\pdfoutput=1
\documentclass[Afour,sageh,times]{sagej}
\usepackage{moreverb,url}\usepackage[colorlinks,bookmarksopen,bookmarksnumbered,citecolor=red,urlcolor=red]{hyperref}

\usepackage{longtable}
\usepackage{supertabular,booktabs}
\usepackage{comment}
\usepackage{amsmath}

\usepackage{comment}
\usepackage{siunitx}
\usepackage{relsize}
\usepackage{ifthen}

\usepackage[caption=false]{subfig}

\usepackage[vlined,ruled,linesnumbered]{algorithm2e}
\usepackage{graphics} 
\usepackage{rotating}
\usepackage{color}
\usepackage{enumerate}
\usepackage[T1]{fontenc}
\usepackage{psfrag}
\usepackage{epsfig} 
\usepackage{booktabs}
\usepackage{graphicx,url}
\usepackage{multirow}
\usepackage{array}
\usepackage{latexsym}
\usepackage{amsfonts}
\usepackage{amsmath}
\usepackage{amssymb}
\usepackage{xstring}
\usepackage[noend]{algorithmic}
\usepackage{multirow}
\usepackage{xcolor}
\usepackage{prettyref}
\usepackage{flexisym}
\usepackage{bigdelim}
\usepackage{breqn} 
\usepackage{listings}
\let\labelindent\relax
\usepackage{enumitem}
\usepackage{xspace}
\usepackage{bm}
\graphicspath{{./figures/}}
\usepackage{tikz}
\usetikzlibrary{matrix,calc}

\usepackage{mdwlist}

\let\itemize\undefined
\makecompactlist{itemize}{stditemize}

\newrefformat{prob}{Problem\,\ref{#1}}
\newrefformat{def}{Definition\,\ref{#1}}
\newrefformat{sec}{Section\,\ref{#1}}
\newrefformat{sub}{Section\,\ref{#1}}
\newrefformat{prop}{Proposition\,\ref{#1}}
\newrefformat{app}{Appendix\,\ref{#1}}
\newrefformat{alg}{Algorithm\,\ref{#1}}
\newrefformat{cor}{Corollary\,\ref{#1}}
\newrefformat{thm}{Theorem\,\ref{#1}}
\newrefformat{lem}{Lemma\,\ref{#1}}
\newrefformat{fig}{Fig.\,\ref{#1}}
\newrefformat{tab}{Table\,\ref{#1}}

\newtheorem{theorem}{Theorem}
\newtheorem{problem}{Problem}

\newtheorem{proposition}[theorem]{Proposition}
\newtheorem{remark}[theorem]{Remark}

\newcommand{\bdmath}{\begin{dmath}}
\newcommand{\edmath}{\end{dmath}}
\newcommand{\beq}{\begin{equation}}
\newcommand{\eeq}{\end{equation}}
\newcommand{\bdm}{\begin{displaymath}}
\newcommand{\edm}{\end{displaymath}}
\newcommand{\bea}{\begin{eqnarray}}
\newcommand{\eea}{\end{eqnarray}}
\newcommand{\beal}{\beq \begin{array}{ll}}
\newcommand{\eeal}{\end{array} \eeq}
\newcommand{\beas}{\begin{eqnarray*}}
\newcommand{\eeas}{\end{eqnarray*}}
\newcommand{\ba}{\begin{array}}
\newcommand{\ea}{\end{array}}
\newcommand{\bit}{\begin{itemize}}
\newcommand{\eit}{\end{itemize}}
\newcommand{\ben}{\begin{enumerate}}
\newcommand{\een}{\end{enumerate}}

\newcommand{\calE}{{\cal E}}

\newcommand{\calL}{{\cal L}}

\newcommand{\calX}{{\cal X}}

\newcommand{\M}[1]{{\bm #1}} 
\renewcommand{\boldsymbol}[1]{{\bm #1}}

\newcommand{\hide}[1]{}

\newcommand{\hiddenText}{{\color{gray} hidden text.}}
\newcommand{\hideWithText}[1]{\hiddenText}

\DeclareMathOperator*{\argmax}{arg\,max}

\newcommand{\normsq}[2]{\left\|#1\right\|^2_{#2}}

\newcommand{\tran}{^{\mathsf{T}}}

\newcommand{\diag}[1]{\mathrm{diag}\left(#1\right)}

\newcommand{\inv}{^{-1}}

\newcommand{\eye}{{\mathbf I}}

\newcommand{\Real}[1]{ { {\mathbb R}^{#1} } }

\newcommand{\at}[1]{^{(#1)}}

\newcommand{\SEthree}{\ensuremath{\mathrm{SE}(3)}\xspace}

\newcommand{\SOthree}{\ensuremath{\mathrm{SO}(3)}\xspace}

\newcommand{\intexpmap}[1]{\mathrm{Exp}\left(#1\right)}

\newcommand{\expmap}[1]{\intexpmap{#1}}

\newcommand{\MA}{\M{A}}

\newcommand{\MD}{\M{D}}

\newcommand{\MM}{\M{M}}

\newcommand{\MQ}{\M{Q}}

\newcommand{\MR}{\M{R}}
\newcommand{\MS}{\M{S}}

\newcommand{\MH}{\M{H}}
\newcommand{\ML}{\M{L}}

\newcommand{\MOmega}{\M{\Omega}}

\newcommand{\vb}{\boldsymbol{b}}

\newcommand{\vg}{\boldsymbol{g}}

\newcommand{\vo}{\boldsymbol{o}}
\newcommand{\vp}{\boldsymbol{p}}

\newcommand{\vr}{\boldsymbol{r}}
\newcommand{\vs}{\boldsymbol{s}}

\newcommand{\vt}{\boldsymbol{t}}
\newcommand{\vxx}{\boldsymbol{x}} 
\newcommand{\vy}{\boldsymbol{y}}

\newcommand{\vtheta}{\boldsymbol{\theta}}

\newcommand{\gtsam}{{\smaller\sf gtsam}\xspace}

\newcommand{\vz}{\boldsymbol{z}}

\lstdefinelanguage{mine} 
{morekeywords={while,True,if,break,=,return,function,for,until,in,input,output,assumptions,assumption,invariant,loop,variant,end,invariant,precondition,variables}, 
sensitive=false, 
morecomment=[l]{\#}, 
morecomment=[il]\%{.}, 
morecomment=[s]{/*}{*/}, 
morestring=[b]", 
} 

\usepackage{calc}
\newlength\listingnumberwidth
\newcommand{\frob}{{\tt F}}

\renewcommand{\ML}{ML\xspace}

\def\volumeyear{2017}
\def\journalname{The International Journal of Robotics Research}
\begin{document}

\runninghead{Choudhary et al.}


  \title{\LARGE{Distributed Mapping with Privacy and Communication Constraints: 
  Lightweight Algorithms and Object-based Models 
 }}


\author{Siddharth Choudhary\affilnum{1}, Luca Carlone\affilnum{2}, Carlos Nieto\affilnum{1}, John Rogers\affilnum{3},  \\Henrik I. Christensen\affilnum{4}, Frank Dellaert\affilnum{1}}

\affiliation{\affilnum{1}College of Computing, Georgia Institute of Technology, USA\\
\affilnum{2}Laboratory for Information \& Decision Systems, Massachusetts Institute of Technology, USA\\
\affilnum{3}U.S. Army Research Laboratory (ARL), USA\\
\affilnum{4}Institute of Contextual Robotics, UC San Diego, USA}

\corrauth{Siddharth Choudhary,
School of Interactive Computing,
College of Computing,
Georgia Institute of Technology,
Atlanta, GA, 
USA
\email{siddharth.choudhary@gatech.edu}
}

\newcommand{\submapping}{submapping\xspace}
\newcommand{\distributed}{decentralized\xspace}
\newcommand{\red}[1]{ {\color{red} #1} }
\newcommand{\myparagraph}[1]{ {\bf #1.}\xspace}
\newcommand{\mysubparagraph}[1]{ \emph{#1.}\xspace}
\newcommand{\myparagraphred}[1]{ {\color{red}{\bf #1.}}\xspace}
\newcommand{\robot}{r\xspace}
\newcommand{\missing}{\red{???}\xspace}
\newcommand{\nrRobots}{n\xspace}
\newcommand{\nrPoses}{N\xspace}
\newcommand{\MRbar}{\bar{\MR}}
\newcommand{\MQbar}{\bar{\MQ}}
\newcommand{\vtbar}{\bar{\vt}}
\newcommand{\vzbar}{\bar{\vz}}
\newcommand{\myEndRemark}{$\blacksquare$}

\newcommand{\forjournal}[1]{\xspace}

\renewcommand{\vs}{\xspace{\tt{vs}}\xspace}
\newcommand{\robots}{ \M{\Omega}\xspace}
\newcommand{\veta}{ \M{\eta}\xspace}
\newcommand{\omegat}{{\tiny \omega_t^2}}
\newcommand{\omegaR}{{\tiny \omega_R^2}}

\newcommand{\ii}{_{ii}}
\newcommand{\inner}{I}
\newcommand{\sep}{S}
\newcommand{\missingRef}{\red{[?]}}

\newcommand{\GN}{GN\xspace}
\renewcommand{\DJ}{DJ\xspace}
\newcommand{\DGS}{DGS\xspace}
\newcommand{\JOR}{JOR\xspace}
\newcommand{\SOR}{SOR\xspace}
\newcommand{\DDFSAM}{DDF-SAM\xspace}

\newcommand{\chairs}{{\tt 25 Chairs}\xspace}
\newcommand{\house}{{\tt House}\xspace}

\newcommand{\stadium}{{\tt stadium}\xspace}
\newcommand{\stadiuma}{{\tt stadium-1}\xspace}
\newcommand{\stadiumb}{{\tt stadium-2}\xspace}
\newcommand{\housea}{{\tt house}\xspace}

\newcommand{\Fig}{Fig.~}
\newcommand{\Figs}{Figs.~}
\newcommand{\maybe}[1]{}
\newcommand{\suba}{a\xspace}
\newcommand{\subb}{b\xspace}
\newcommand{\subc}{c\xspace}
\newcommand{\subd}{d\xspace}

\newcommand{\link}[4]{^{#1_{#2}}_{#3_{#4}}}
\renewcommand{\of}[2]{_{#1_{#2}}}
\newcommand{\SC}[1]{{\color{blue} \textbf{SC}: #1}}
\newcommand{\blue}[1]{{\color{blue} #1}}

\begin{abstract}
We consider the following problem: a team of robots is deployed in an unknown environment and 
it has to collaboratively build a map of the area without a reliable infrastructure for communication. 
The backbone for modern mapping techniques is \emph{pose graph optimization}, which estimates the 
trajectory of the robots, from which the map can be easily built. The first contribution of this paper 
is a set of distributed 
algorithms for pose graph optimization: rather than sending all sensor data to a remote sensor fusion server, 
the robots exchange very partial and noisy information to reach an agreement on the pose graph configuration.
Our approach can be considered as a distributed implementation of the two-stage approach of~\cite{Carlone15icra-init3D}, where 
we use the \emph{Successive Over-Relaxation} (SOR) and the \emph{Jacobi Over-Relaxation} (JOR) 
as workhorses to split the computation among the robots.
We also provide conditions under which
the proposed distributed protocols converge to the solution of the centralized two-stage approach.
As a second contribution, we extend 
the proposed distributed 
algorithms to work with object-based 
map models. The use of object-based models avoids 
the exchange of raw sensor measurements (e.g., point clouds or RGB-D data) further reducing the communication 
burden.
Our third contribution is an extensive experimental evaluation of the proposed 
techniques, 
including tests in realistic Gazebo simulations and field experiments in a military test facility. 
Abundant experimental evidence suggests that 
one of the proposed algorithms (the \emph{Distributed Gauss-Seidel method} or \DGS) has excellent 
performance. The \DGS 
 requires minimal information exchange, has an anytime flavor, 
scales well to large teams (we demonstrate mapping with a team of 50 robots), 
 is robust to noise, and is easy to implement.
Our field tests show that 
the combined use of our distributed algorithms and object-based models
reduces the communication requirements by several orders of magnitude 
and enables distributed mapping with large teams of robots in real-world problems.
\end{abstract}

\maketitle


\section{Introduction}
\label{sec:intro}

The deployment of large teams of cooperative autonomous robots has the potential to enable fast information gathering, 
and more efficient coverage and monitoring of vast areas. 
For military applications such as surveillance, reconnaissance, and battle damage assessment, 
multi-robot systems promise more efficient operation and improved robustness in contested spaces.
In civil applications (e.g., pollution monitoring, precision agriculture, search and rescue, disaster response), 
the use of several inexpensive, heterogeneous, agile platforms 
is an appealing alternative to monolithic single robot systems.

The deployment of multi robot systems in the real world poses many technical challenges, 
ranging from coordination and formation control, to task allocation and 
distributed sensor fusion. In this paper we tackle a specific instance of the sensor fusion problem.
We consider the case in which a team of robots explores an unknown environment 
and each robot has to estimate its trajectory from its own sensor data and by leveraging 
information exchange with the teammates.
Trajectory estimation, also called \emph{pose graph optimization}, is relevant as it constitutes the backbone for many 
estimation and control tasks (e.g., geo-tagging sensor data, mapping, position-aware task allocation, formation control). 
Indeed, in our application, trajectory estimation enables distributed 3D mapping and 
localization (\Fig\ref{fig:3Dclouds}).

We consider a realistic scenario, in which
the robots can only communicate when they are within a given distance.
Moreover, also during a rendezvous (i.e., when the robots are close enough to communicate) 
they cannot exchange a large amount of information, due to \emph{bandwidth constraints}.
Our goal is to design a technique that allows each robot 
to estimate its own trajectory, while asking for minimal knowledge of the trajectory 
of the teammates. This ``privacy constraint'' has a clear motivation in a military application: 
in case one robot is captured, it cannot provide sensitive information 
about the areas covered by the other robots in the team. Similarly, in civilian applications, 
one may want to improve the localization of a device (e.g., a smart 
phone) by exchanging information with other devices, while respecting users' privacy.
Ideally, we want our distributed mapping approach to scale to very large teams of robots. 
Our ultimate vision is to deploy a swarm of agile robots (e.g., micro aerial vehicles) that 
can coordinate by sharing minimal information and using on-board sensing and computation. 
The present paper takes a step in this direction and presents distributed mapping 
techniques that are shown to be extremely effective in large simulations (with up to 50 robots) 
and in real-world problem (with up to 4 robots).

\newcommand{\cwid}{4.2cm}
\begin{figure}[t]
\begin{minipage}{\columnwidth}
\hspace{-0.4cm}
\begin{tabular}{cc}%
\begin{minipage}{\cwid}%
\centering%
\includegraphics[scale=0.14, trim= 0cm 0cm 0cm 0cm, clip]{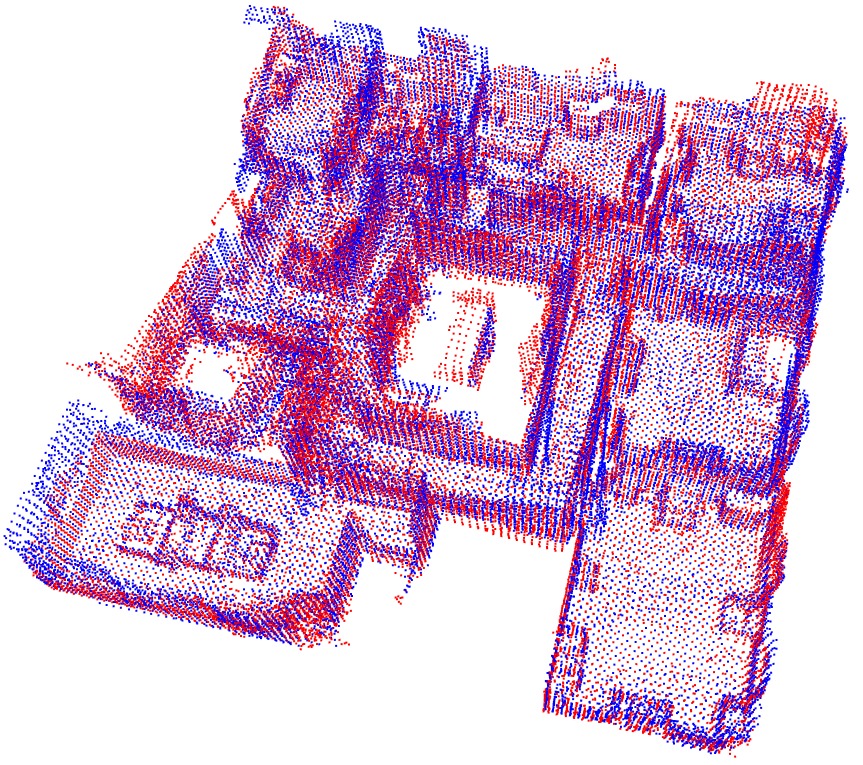} 
\end{minipage}
\hspace{-0.4cm}
&
\begin{minipage}{\cwid}%
\centering%
\includegraphics[scale=0.14, trim= 0cm 0cm 0cm 0cm, clip]{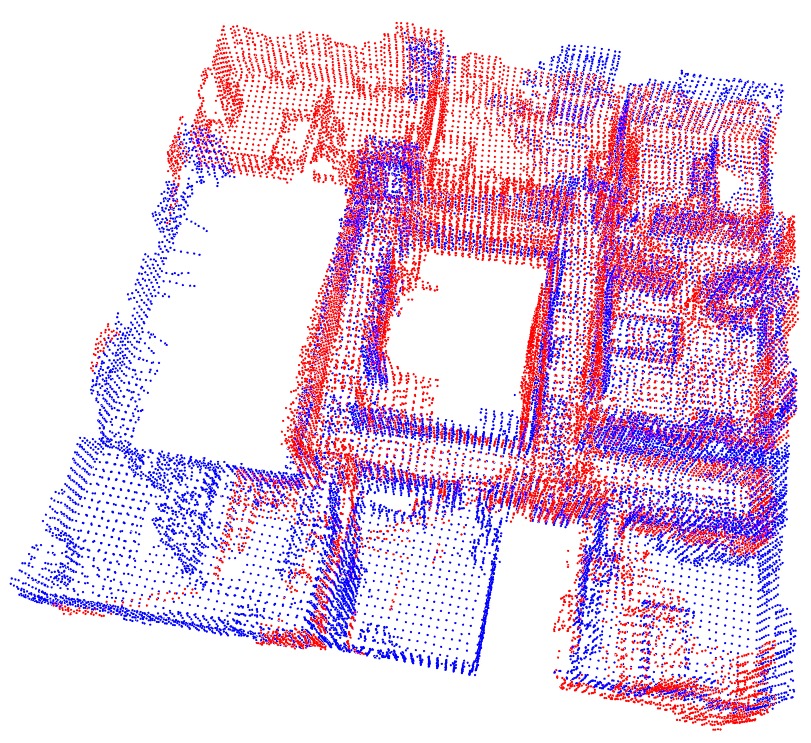} 
\end{minipage}
\vspace{0.2cm}
\\
\begin{minipage}{\cwid}%
\centering%
\includegraphics[scale=0.14, trim= 3.5cm 0cm 4.5cm 0cm, clip, angle=-90]{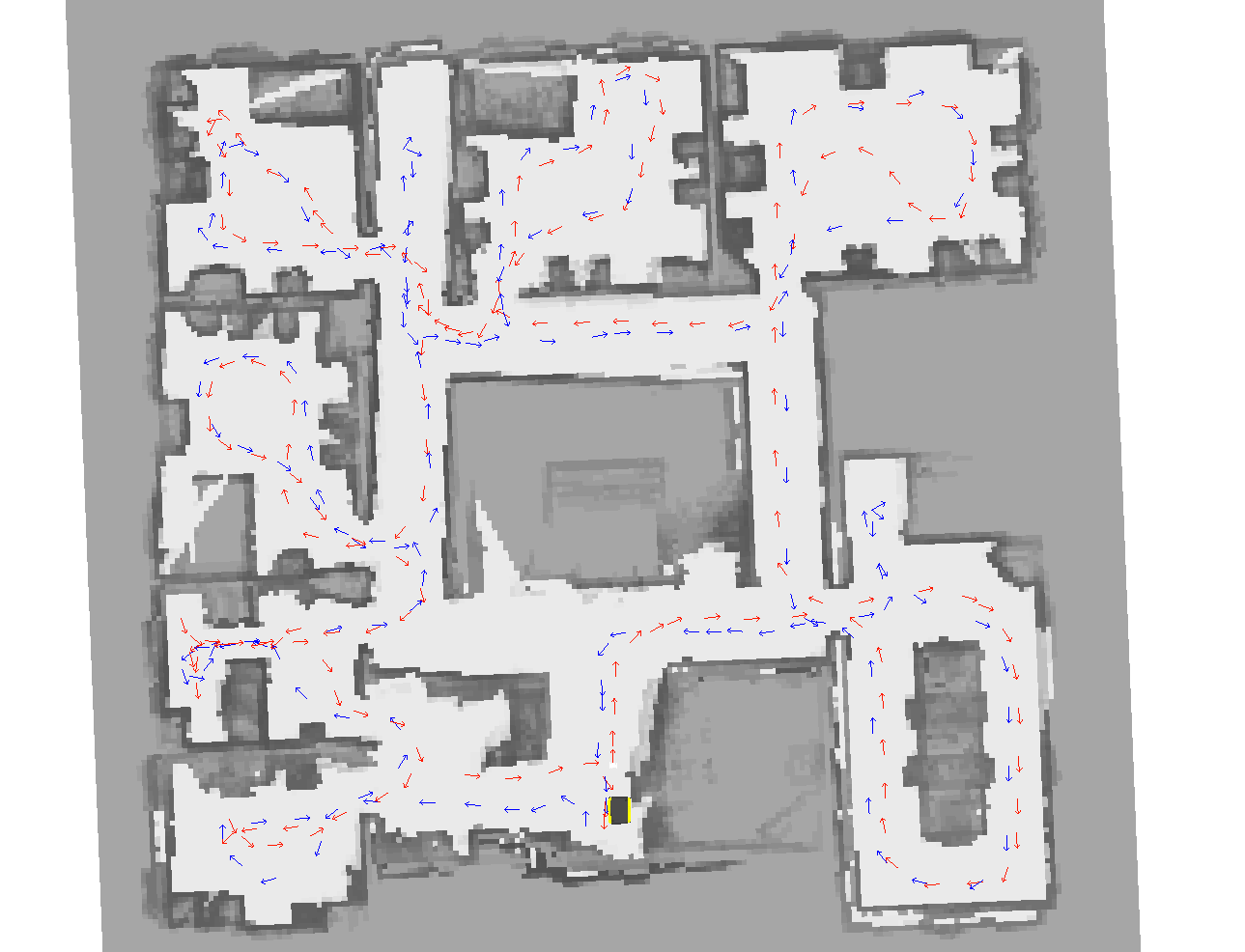} 
\end{minipage}
&
\hspace{-0.4cm}
\begin{minipage}{\cwid}%
\centering%
\includegraphics[scale=0.14, trim= 2.5cm 0cm 3.7cm 0cm, clip]{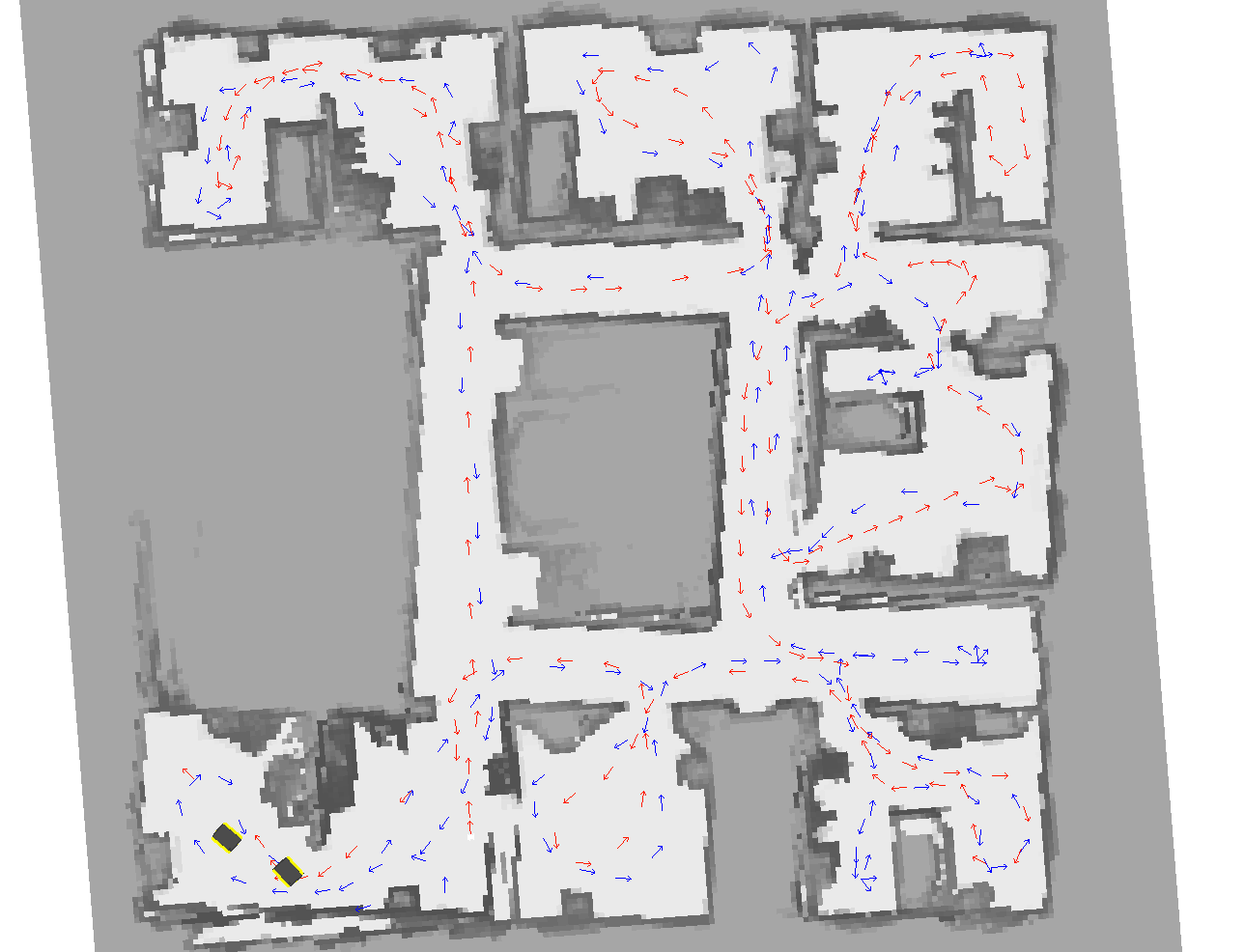}
\end{minipage}
\end{tabular}
\end{minipage}
\caption{In our field experiments, distributed trajectory estimation enables 
3D reconstruction of an entire building using two robots (red, blue). Each column of the figure shows the 
reconstructed point cloud of a floor (top), and the estimated trajectories overlaid on an occupancy grid map (bottom).\label{fig:3Dclouds}
}
\end{figure}

\vspace{0.1cm}
\myparagraph{Contribution} 
We consider a distributed maximum-likelihood (\ML) trajectory estimation problem in which the robots have to collaboratively estimate 
their trajectories while minimizing the amount of exchanged information. 
We focus on a fully 3D case, 
as this setup is of great interest in many robotics applications (e.g., navigation on uneven terrain, 
underwater and  aerial vehicles).
We also consider a fully distributed setup, in which the robots communicate and acquire relative 
measurements only during rendezvous.

We present two key contributions to solve the distributed mapping problem. 
\emph{The first contribution are a set of distributed algorithms that 
enable distributed inference at the estimation back-end}.
This contribution is presented 
in Section~\ref{sec:distributedAlgorithms}.
Our approach can be understood as a distributed implementation of the 
\emph{chordal initialization} 
discussed in~\cite{Carlone15icra-init3D}. The chordal initialization
(recalled in Section~\ref{sec:centralized}) consists in approximating the \ML trajectory estimate 
by solving two quadratic optimization subproblems. The insight of the present work is that 
these quadratic subproblems can be solved in a distributed fashion, leveraging distributed linear system solvers.
In particular, we investigate distributed implementations of the 
Jacobi Over-Relaxation 
and the Successive Over-Relaxation.
These distributed solvers imply a communication burden that is linear in the 
number of rendezvous among the robots. Moreover, they do not rely on the availability of an 
accurate initial guess as in related work (see Section~\ref{sec:relatedWork}).  
In Section~\ref{sec:distributed} we discuss conditions under which the distributed algorithms 
converge to the same estimate of the chordal initialization \citep{Carlone15icra-init3D}, 
which has been extensively shown to be accurate and resilient to measurement noise.

\emph{The second contribution is the use of high-level object-based models at the 
estimation front-end and as map representation}. 
This contribution is presented 
in Section~\ref{sec:objects}.
Traditional approach for multi robot mapping rely on feature-based maps which are composed of low level primitives like points and lines \citep{Davison07tpami}.
These maps become memory-intensive for long-term operation,
contain a lot of redundant information (e.g., it is unnecessary
 to represent a planar surface with thousands of points), and lack the semantic information necessary for
performing wider range of tasks (e.g., manipulation tasks, human-robot interaction). To solve these issues, 
we present an approach for multi robot SLAM which uses
object landmarks \citep{Moreno13cvpr} in a multi robot mapping setup. We show that
this approach further reduces the information exchange among robots, results in compact human-understandable maps,
and has lower computational complexity as compared to low-level feature-based mapping.

\emph{The third contribution is an extensive experimental evaluation including realistic 
simulations in Gazebo and field tests in a military facility.}
This contribution is presented 
in Section~\ref{sec:experiments}.
The experiments demonstrate that one of the proposed algorithms, namely the 
\emph{Distributed Gauss-Seidel method}, provides accurate trajectory estimates, 
is more parsimonious, communication-wise, than 
related techniques, scales to large tea, and is robust to noise.
Finally, our field tests show that 
the combined use of our distribute algorithms and object-based models
reduces the communication requirements by several orders of magnitude 
and enables distributed mapping with large teams of robots. 

Section~\ref{sec:conclusion} concludes the paper and discusses current and future work towards 
 real-world robust mapping with large swarms of flying robots. 

\section{Related Work}
\label{sec:relatedWork}

\myparagraph{Multi Robot Localization and Mapping}
Distributed estimation in multi robot systems is currently an active field of research, 
with special attention being paid to communication constraints~\citep{Paull15icra}, 
heterogeneous teams \citep{Bailey11icra,Indelman12ijrr}, estimation consistency \citep{Bahr09icra}, 
 and robust data association \citep{Indelman14icra,Dong15icra}. 
Robotic literature offers distributed implementations of different estimation techniques, 
including Extended Kalman filters~\citep{Roumeliotis02tra,Zhou06iros}, 
information filters~\citep{Thrun03isrr}, and
particle filters~\citep{Howard06ijrr,Carlone11jirs}.
%
%
More recently, the community reached a large consensus on the use of 
maximum likelihood (\ML) estimation (maximum a-posteriori, in presence of priors), which, applied 
to trajectory estimation, is often referred to as~\emph{pose graph optimization} or \emph{pose-based SLAM}. 
\ML estimators circumvent well-known issues of Gaussian filters (e.g., build-up of linearizion errors)
 and particle filters (e.g., particle depletion), and frame the estimation problem in terms of 
 nonlinear optimization.
In multi robot systems, \ML trajectory estimation can be 
performed by collecting all measurements at a centralized inference engine, which 
performs the optimization~\citep{Andersson08icra,Kim10icra,Bailey11icra}. 
Variants of these techniques invoke partial exchange of raw or preprocessed sensor data~\citep{Lazaro11icra,Indelman14icra}. 

In many applications, however, it is not practical to collect all measurements at a single inference engine.
When operating in hostile environment, a single attack to the centralized inference engine (e.g., one of the 
robot) may threaten the operation of the entire team. Moreover, the centralized approach requires massive communication 
and large bandwidth. Furthermore, solving trajectory estimation over a large team of robots can be too demanding for a 
single computational unit. Finally, the centralized approach poses privacy concerns as it 
requires to collect all information at a single robot; if an enemy robot is able to 
deceive the other robots and convince them that it is part of the team, it can easily gather sensitive 
information (e.g., trajectory covered and places observed by every robot).
These reasons triggered interest towards \emph{distributed trajectory estimation}, in which the robots 
only exploit local communication, in order to reach a consensus on the trajectory estimate.
\cite{Nerurkar09icra} propose an algorithm for cooperative localization
 based on distributed conjugate gradient. 
 \cite{Franceschelli10icra} propose a gossip-based algorithm for distributed 
pose estimation and investigate its convergence in a noiseless setup.  
\cite{Aragues11icra} use a distributed 
 Jacobi approach to estimate a set of 2D poses, or the  centroid of a formation ~\citep{Aragues11scl}. 
 \cite{Aragues12tro} investigate consensus-based approaches for map merging.
\cite{Knuth13icra} estimate 3D poses using distributed gradient descent.
\cite{Cunningham10iros} use Gaussian elimination, and develop an approach, called \DDFSAM, in which 
each robot exchange a Gaussian marginal over the~\emph{separators} (i.e., the variables shared by 
multiple robots); the approach is further extended in~\cite{Cunningham13icra}, to avoid 
storage of redundant data. 

Another related body of work is the literature on 
 parallel and hierarchical approaches for mapping. Also in this case, Gaussian elimination and Schur complement 
  have been used as a key component for
 hierarchical approaches for large-scale mapping~\citep{Ni10iros,Grisetti10icra,Suger14icra}.
 \emph{Decoupled stochastic mapping} was one of the earliest approach for
submapping proposed by  \cite{Leonard01joe}. \cite{Leonard03ijcai} propose a constant-time
SLAM solution which achieves near-optimal results under the assumption that the
robot makes repeated visits to all regions of the environment. \cite{Frese05tro} use
 multi-level relaxations resulting
in a linear time update. \cite{Frese06ar2} propose the TreeMap
algorithm which 
divides the environment 
according to a balanced binary tree. 
%
\cite{Estrada05tro} present a hierarchical SLAM
approach which consist of a set of local maps
and enforces loop consistency when calculating the optimal estimate
at the global level.
\cite{Ni07icra} present an \emph{exact} submapping
approach within a \ML framework, and propose to
cache the factorization of the submaps to speed-up computation.
\cite{Grisetti10icra} propose hierarchical updates on the map:
whenever
an observation is acquired, the highest level of the hierarchy is modified and only
the areas which are substantially modified are changed at lower levels.
\cite{Ni10iros} extend their previous approach \citep{Ni07icra}
to include multiple levels and use nested dissection to minimize the dependence
between two subtrees. \cite{Grisetti12iros} compute a good initial estimate for global alignment
through a submapping approach.
\cite{Zhao13iros} propose an approximation for
large-scale SLAM by solving for a sequence of submaps and joining them
in a divide-and-conquer manner using linear least squares. \cite{Suger14icra} present an approximate SLAM approach based on hierarchical
decomposition to reduce the memory consumption for pose graph optimization.
%
%

While Gaussian elimination has become a popular approach 
it has two major shortcomings. First, the marginals to be exchanged among the robots are dense, and 
the communication cost is quadratic in the number of separators. 
This motivated the use of sparsification techniques to reduce the communication cost~\citep{Paull15icra}.
The second reason is that Gaussian elimination is performed on a linearized version of the
 problem, hence these approaches require good linearization points and 
complex bookkeeping to ensure consistency of the linearization points across the robots~\citep{Cunningham13icra}. 
The need of a linearization point also characterizes gradient-based techniques~\citep{Knuth13icra}. 
In many practical problems, however, no initial guess is available, 
and one has to develop ad-hoc initialization techniques, e.g., \citep{Indelman14icra}.

\myparagraph{Related Work in Other Communities}
Distributed position and orientation estimation is a fertile research area in other communities, 
including sensor networks, computer vision, and multi agent systems.
In these works, the goal is to estimate the state (e.g. position or orientation) of an agent 
(e.g., a sensor or a camera) from relative measurements among the agents. A large body of literature 
deals with distributed localization from distance measurements, 
see~\cite{Anderson10siam,Calafiore12tsmca,Simonetto14tsp,Wei15sensors} and 
the references therein. The case of position estimation from linear measurements is considered 
in~\cite{Barooah05icisip,Barooah07csm,Russell11tsp,Carron14tcns,Todescato15ecc,Freris15cdc};
 the related problem of \emph{centroid estimation} is tackled in~\cite{Aragues11scl}.  
Distributed rotation estimation has been studied in the context of 
 attitude synchronization~\citep{Thunberg11cdc,Hatanaka10cdc,Olfati-Saber06cdc}, 
camera network calibration~\citep{Tron09cdc,Tron12cdc}, sensor network 
localization~\citep{Piovan13automatica}, and distributed consensus on 
manifolds~\citep{Sarlette09sicon,Tron12tac}.

\myparagraph{High-Level Map Representations} \emph{Semantic mapping} using high-level 
object-based representation has gathered a large amount of interest from
the robotics community. \cite{Kuipers2000ai} model the
environment as a spatial semantic hierarchy, where each level expresses
states of partial knowledge corresponding to different level of representations.
 \cite{Ranganathan07rss} present a 3D generative
model for representing places using objects. The object models are
learned in a supervised manner. \cite{Civera11iros} propose a semantic SLAM 
algorithm that annotates the low-level 3D point based maps with precomputed object models. 
\cite{Rogers11iros} recognize door signs and
read their text labels (e.g., room numbers) which are used as landmarks
in SLAM. \cite{Trevor12icra} use planar surfaces corresponding
to walls and tables as landmarks in a mapping system.
\cite{Bao12cvpr} model semantic structure from motion as a joint inference
problem where they jointly recognize and estimate the location of high-level
semantic scene components such as regions and objects in 3D.
SLAM++, proposed by \cite{Moreno13cvpr},
train domain-specific object detectors corresponding to repeated
objects like tables and chairs. The learned detectors are integrated
inside the SLAM framework to recognize and track those objects resulting
in a semantic map. Similarly, \cite{Kim12siggraphAsia} use
learned object models to reconstruct dense 3D models from single scan
of the indoor scene. \cite{Choudhary14iros} proposed an approach for for online 
object discovery and object modeling, and extend a SLAM system to utilize these discovered 
and modeled objects as landmarks to help localize the robot in an online manner. 
\cite{Pillai15rss} develop a SLAM-aware object recognition system which result in a 
 considerably stronger recognition performance as compared to related techniques. \cite{Lopez16ras} present a
 real-time monocular object-based SLAM using a large database of 500 3D objects and 
 show exploiting object rigidity both improve the map and find its real scale. Another body of related work is in the area of dense semantic mapping
where the goal is to categorize each voxel or 3D point with a category label. 
Related work in dense semantic mapping include \cite{Nuchter08ras,Koppula11nips,Pronobis12icra,Finman13ecmr,Kundu14eccv,Vineet15icra,Valentin15tog,McCormac16arxiv} and the references therein.


\section{Dealing with Bandwidth Constraints I: Distributed Algorithms}
\label{sec:distributedAlgorithms}

The first contribution of this paper is to devise distributed algorithms that 
the robots can implement to reach consensus on a globally optimal trajectory estimate 
using minimal communication.
Section~\ref{sec:revisitedPGO} introduces the mathematical notation and formalizes the problem. 
Section~\ref{sec:centralized} presents a centralized algorithm, while 
Section~\ref{sec:distributed} presents the corresponding distributed implementations.

\begin{figure}[t]
\centering%
\includegraphics[width=0.8\columnwidth]{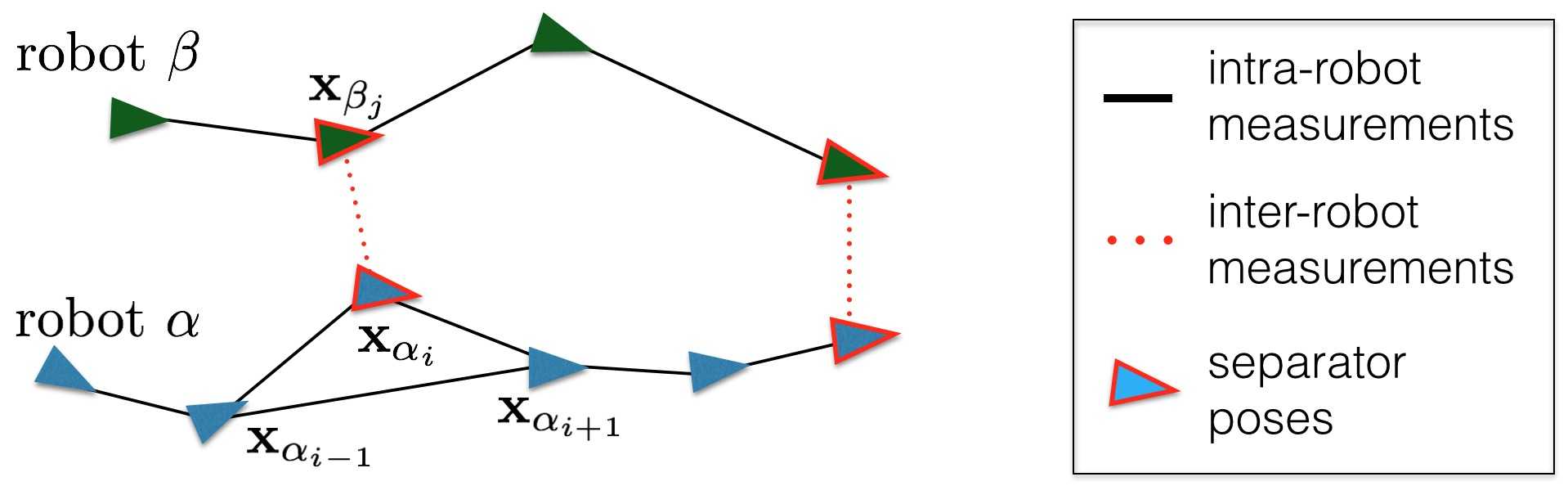} \\
\caption{\label{fig:multiRobotPGO}
An instance of multi robot trajectory estimation: two robots ($\alpha$ in blue, and $\beta$ in dark green) 
traverse and unknown environment, collecting intra-robot measurements (solid black lines). 
During rendezvous, 
each robot can observe the pose of the other robot (dotted red lines). These are called inter-robot measurements and 
relate two \emph{separators} (e.g., $\vxx\of{\alpha}{i}, \vxx\of{\beta}{j}$). 
The goal of the two robots is to compute the \ML estimate of their trajectories.
}
\end{figure}

\subsection{Problem Formulation: Distributed Pose Graph Optimization}
\label{sec:revisitedPGO}

We consider a multi robot system and we denote  
each robot with a Greek letter, such that  the 
set of robots is $\robots =\{\alpha, \beta, \gamma, \ldots\}$. 
The goal of each robot is to estimate 
its own trajectory using the available  measurements, and 
leveraging occasional 
communication with other robots. 
The trajectory estimation problem and the nature of the available measurements 
are made formal in the rest of this section.

We model each trajectory  as a finite set of poses (triangles in Fig.~\ref{fig:multiRobotPGO}); 
the pose assumed by robot $\alpha$ at time $i$ is denoted with $\vxx\of{\alpha}{i}$ 
(we use Roman letters to denote time indices). We are interested in a 3D setup, i.e., $\vxx\of{\alpha}{i} \in \SEthree$, 
where $\SEthree$ is the Special Euclidean group of 3D rigid transformations; when convenient, we write $\vxx\of{\alpha}{i} = (\MR\of{\alpha}{i}, \vt\of{\alpha}{i})$,  making explicit that each pose includes a 
rotation $\MR\of{\alpha}{i} \in \SOthree$, and a position $\vt\of{\alpha}{i} \in \Real{3}$.
 The trajectory of robot $\alpha$ is then denoted as $\vxx\of{\alpha}{} = [\vxx\of{\alpha}{1},\vxx\of{\alpha}{2}, \ldots]$.

\myparagraph{Measurements} We assume that each robot acquires relative pose measurements. In practice these are obtained 
by post-processing raw sensor data (e.g., scan matching on 3D laser scans). 
We consider two types of measurements: intra-robot  
and inter-robot measurements. The \emph{intra-robot measurements} involve the poses of a single robot 
at different time instants; common examples of intra-robot measurements are odometry measurements
 (which constrain consecutive robot poses, e.g., $\vxx\of{\alpha}{i}$ and $\vxx\of{\alpha}{i+1}$ in Fig.~\ref{fig:multiRobotPGO}) 
 or loop closures (which constrain non-consecutive poses, e.g., $\vxx\of{\alpha}{i-1}$ and $\vxx\of{\alpha}{i+1}$ in Fig.~\ref{fig:multiRobotPGO}).
The \emph{inter-robot measurements} are the ones relating the poses of different robots.
For instance, during a rendezvous, 
 robot $\alpha$ (whose local time is $i$), observes a second robot $\beta$ (whose local time is $j$)
and uses on-board sensors to measure the relative pose of the observed robot in its own reference frame.
Therefore, robot $\alpha$ acquires an inter-robot measurement, describing the relative pose between $\vxx\of{\alpha}{i}$ and $\vxx\of{\beta}{j}$ (red links in Fig.~\ref{fig:multiRobotPGO}). We use the term 
\emph{separators} to refer to the poses involved in an inter-robot measurement.

While our classification of the measurements (inter \vs intra) is based on the robots involved in the 
measurement process, all relative measurements can be framed within the same measurement model.
Since all measurements correspond to noisy observation of the relative pose between a pair of poses, say 
 $\vxx\of{\alpha}{i}$ and $\vxx\of{\beta}{j}$, a general measurement model is:
\newcommand{\subs}{\link{\alpha}{i}{\beta}{j}}
\beq
\label{eq:measurementModel}
\begin{array}{c}
\vzbar\subs \doteq (\MRbar\subs, \vtbar\subs), \quad \text{with:} 
\left\{
\begin{array}{l}
\MRbar\subs = (\MR\of{\alpha}{i})\tran \MR\of{\beta}{j} \MR_\epsilon    \\
\vtbar\subs = (\MR\of{\alpha}{i})\tran ( \vt\of{\beta}{j} \!-\! \vt\of{\alpha}{i}  ) \!+\! \vt_{\epsilon}
\end{array} 
\right.                      
\end{array}
\eeq
where the relative pose measurement $\vzbar\subs$ includes the relative rotation measurements
$\MRbar\subs$, which describes the attitude $\MR\of{\beta}{j}$ with respect to the 
  reference frame of robot $\alpha$ at time $i$,
``plus'' a random rotation $\MR_\epsilon$ (measurement noise), 
and the relative position measurement $\vtbar\subs$, which describes the position 
$\vt\of{\beta}{j}$ in the reference frame of robot $\alpha$ at time $i$, plus random
 noise $\vt_{\epsilon}$. 
According to our previous definition, 
intra robot measurements are in the form $\vzbar\link{\alpha}{i}{\alpha}{k}$, for some robot $\alpha$ 
and for two time instants $i \neq k$; inter-robot measurements, instead, are in the form  
 $\vzbar\subs$ for two robots $\alpha \neq \beta$.

 In the following, we denote with $\calE^{\alpha}_{\inner}$ the set of intra-robot 
 measurements for robot $\alpha$, while we call $\calE_{\inner}$ the set of intra-robot 
 measurements for all robots in the team, i.e., $\calE_{\inner} = \cup_{\alpha \in \robots} \calE^{\alpha}_{\inner}$.
 The set of inter-robot measurements involving robot $\alpha$ is denoted with $\calE^{\alpha}_{\sep}$
   ($\sep$ is the mnemonic for ``separator''). 
 The set of all inter-robot measurements is denoted with $\calE_{\sep}$.
 The set of all available measurements  is then $\calE = \calE_{\inner} \cup \calE_{\sep}$.
 Note that each robot only has access to its own intra and inter-robot measurements 
 $\calE^{\alpha}_{\inner}$ and $\calE^{\alpha}_{\sep}$.



\myparagraph{\ML trajectory estimation}
Let us collect all robot trajectories in a single 
(to-be-estimated) set of poses $\vxx = [\vxx\of{\alpha}{}, \vxx\of{\beta}{}, \vxx\of{\gamma}{}, \ldots ]$. 
%
The \ML estimate for $\vxx$ is defined as the maximum of the measurement likelihood:
\beq
\label{eq:maxLike}
\vxx^\star = 
\argmax_{\vxx} \prod_{(\alpha_i,\beta_j) \in \calE} \calL( \vzbar\subs  \;|\; \vxx)
\eeq
where we took the standard assumption of independent measurements. 
The expression of the likelihood function depends on the distribution 
of the measurements noise, i.e., $\MR_\epsilon, \vt_{\epsilon}$ in~\eqref{eq:measurementModel}.
%
We follow the path of~\cite{Carlone15iros-duality3D} and 
 assume that translation noise is distributed according to a zero-mean Gaussian with information matrix
 $\omegat \eye_3$, while the rotation noise follows a Von-Mises distribution with 
concentration parameter $\omegaR$. 

Under these assumptions, it is possible to demonstrate~\citep{Carlone15iros-duality3D}
that the \ML estimate $\vxx \doteq \{(\MR\of{\alpha}{i},\vt\of{\alpha}{i}), \forall \alpha \in \Omega, \forall i\}$ 
can be computed as solution of the following optimization problem:
%
\bea
\label{eq:MRPGO}
\!\!\!\!
\min_{
\substack{
\vt\of{\alpha}{i} \in \Real{3}, 
\MR\of{\alpha}{i} \in \SOthree \\
\forall \alpha \in \robots, \forall i
}} 
 \sum_{(\alpha_i,\beta_j) \in \calE}  & 
 \omegat \normsq{ \vt\of{\beta}{j} \!\!-\! \vt\of{\alpha}{i} \!\!-\! \MR\of{\alpha}{i} \vtbar\subs }{}  
 \!\!+ \nonumber
 \\ 
 &  \frac{\omegaR}{2} \normsq{ \MR\of{\beta}{j} \!\!-\! \MR\of{\alpha}{i} \MRbar\subs}{\frob}  
\eea
The peculiarity of~\eqref{eq:MRPGO} is the use of the \emph{chordal distance} 
$\textstyle \| \MR\of{\beta}{j} \!\!-\! \MR\of{\alpha}{i} \MRbar\subs\|_{\frob}$ to quantify 
rotation errors, while the majority of related works in robotics uses 
 the \emph{geodesic distance}, see~\citep{Carlone15icra-init3D} for details.

A centralized approach to solve the multi robot pose graph optimization problem~\eqref{eq:MRPGO} 
works as follows.
A robot collects all measurements $\calE$.
Then, the optimization problem~\eqref{eq:MRPGO} is solved using 
 iterative optimization on manifold~\citep{Dellaert12tr}, fast approximations~\citep{Carlone15icra-init3D}, 
 or convex relaxations~\citep{Rosen16wafr}.

In this paper we consider the more interesting case in which it is not possible to 
collect all measurements at a centralized estimator, and the problem has to be solved in a distributed 
fashion. More formally, the problem we solve is the following.

\begin{problem}[Distributed Trajectory Estimation]
\label{prob:statement}
Design an algorithm that each robot $\alpha$ can 
execute 
during a rendezvous with a subset of other robots $\Omega_r \subseteq \Omega \setminus \{\alpha\}$, and that
\bit
\item takes as input: (i) the intra-robot measurements $\calE^{\alpha}_{\inner}$ 
and (ii) the subset of inter-robot measurements $\calE^{\alpha}_{\sep}$, 
(iii) partial estimates of the trajectory of robots $\beta \in \Omega_r$; 
\item returns as output: the \ML estimate $\vxx^\star_\alpha$, which is such that 
$\vxx^\star = [\vxx^\star\of{\alpha}{}, \vxx^\star \of{\beta}{}, \vxx^\star\of{\gamma}{}, \ldots ]$ 
is a minimizer of~\eqref{eq:MRPGO}.
\eit
\end{problem}

While the measurements $\calE^{\alpha}_{\inner}$ and $\calE^{\alpha}_{\sep}$ 
are known by robot $\alpha$, gathering the estimates from robots $\beta \in \Omega_r$ requires 
communication, hence we want our distributed algorithm to exchange a very small portion of the trajectory estimates. 

 The next sections present our solution to Problem~\ref{prob:statement}.   
To help readability, we start with a centralized description of the approach, which is an adaptation 
of the chordal initialization of~\cite{Carlone15icra-init3D} to the multi robot case. Then we tailor the discussion to 
the distributed setup in Section~\ref{sec:distributed}.

\subsection{Two-Stage Pose Graph Optimization: Centralized Description}
\label{sec:centralized}


The present work is based on two key observations.
The first one is that the optimization problem~\eqref{eq:MRPGO} has a quadratic objective; 
what makes~\eqref{eq:MRPGO} hard is the presence of non-convex constraints,
i.e., $\MR\of{\alpha}{i} \in \SOthree$.
 Therefore, as already proposed in~\cite{Carlone15icra-init3D} (for the single robot, centralized case), 
 we use a two-stage approach: we first solve a relaxed version of~\eqref{eq:MRPGO} and get an estimate for the rotations 
 $\MR\of{\alpha}{i}$ of all robots, and then we recover the full poses and top-off the result with a Gauss-Newton (\GN) iteration. 
The second key observation is that each of the two stages can be solved in distributed fashion, 
 exploiting existing distributed linear system solvers. 
In the rest of this section we review the two-stage approach of~\cite{Carlone15icra-init3D}, 
while we discuss the use of distributed 
 solvers in Section~\ref{sec:distributed}. 

The two-stage approach of~\cite{Carlone15icra-init3D} 
first solves for the unknown rotations, and then recovers the full poses via a 
single \GN iteration. The two stages are detailed in the following.

\myparagraph{Stage 1: rotation initialization via relaxation and projection} 
The first stage computes a good estimate of the rotations of all robots by solving the following rotation subproblem:
\beq
\label{eq:MRPGO-R}
\min_{
\substack{
\MR\of{\alpha}{i} \in \SOthree \\
\forall \alpha \in \robots, \forall i
}} \sum_{(\alpha_i,\beta_j) \in \calE} \omegaR \normsq{ \MR\of{\beta}{j} \!\!-\! \MR\of{\alpha}{i} \MRbar\subs}{\frob}  
\eeq
which amounts to estimating the rotations of all robots in the team by considering only the 
relative rotation measurements (the second summand in~\eqref{eq:MRPGO}). 

While problem~\eqref{eq:MRPGO-R} is nonconvex (due to the nonconvex constraints $\MR\of{\alpha}{i} \in \SOthree$), 
many algorithms to approximate its solution are available in literature.
Here we use the approach proposed in~\cite{Martinec07cvpr} and reviewed in~\cite{Carlone15icra-init3D}. 
The approach first solves the quadratic relaxation obtained by 
dropping the constraints $\MR\of{\alpha}{i} \in \SOthree$, and then projects the relaxed solution to $\SOthree$. 
In formulas, the quadratic relaxation is:
\beq
\label{eq:MRPGO-R-relax}
\min_{
\substack{
\MR\of{\alpha}{i}, 
\forall \alpha \in \robots, \forall i
}} \sum_{(\alpha_i,\beta_j) \in \calE} \omegaR \normsq{ \MR\of{\beta}{j} \!\!-\! \MR\of{\alpha}{i} \MRbar\subs}{\frob}  
\eeq
which simply rewrites~\eqref{eq:MRPGO-R} without the constraints.
Since~\eqref{eq:MRPGO-R-relax} is quadratic in the unknown 
rotations $\MR\of{\alpha}{i}, \forall \alpha \in \robots, \forall i$, we
 can rewrite it as:
\beq
\label{eq:MRPGO-R-relax2}
\min_\vr
\|\MA_r \vr - \vb_r \|^2  
\eeq
where we stacked all the entries of the unknown rotation matrices 
$\MR\of{\alpha}{i}, \forall \alpha \in \robots, \forall i$ into a single
vector $\vr$, and we built the (known) matrix $\MA_r$ and 
(known) vector $\vb_r$ accordingly (the presence of a nonzero vector $\vb_r$ 
 follows from setting one of the rotations to be the reference frame, e.g., $\MR\of{\alpha}{1} = \eye_3$).

Since~\eqref{eq:MRPGO-R-relax} is a linear least-squares problem, its solution can be found by 
solving the normal equations: 
\beq
\label{eq:normEq-R}
(\MA_r\tran \MA_r) \vr = \MA_r\tran \vb_r
\eeq
Let us denote with $\breve{\vr}$ the solution of~\eqref{eq:normEq-R}.
Rewriting $\breve{\vr}$ in matrix form, we obtain the matrices  $\breve{\MR}\of{\alpha}{i}$, 
$\forall \alpha \in \robots, \forall i$.
 Since these rotations were obtained from a relaxation of~\eqref{eq:MRPGO-R}, they 
 are not guaranteed to satisfy the constraints $\MR\of{\alpha}{i} \in \SOthree$; therefore
 the approach~\citep{Martinec07cvpr}
  projects them to $\SOthree$, and gets the rotation estimate $\hat{\MR}\of{\alpha}{i} = 
 {\tt project}(\breve{\MR}\of{\alpha}{i})$, $\forall \alpha \in \robots, \forall i$.
 The projection only requires to perform 
 an SVD of $\breve{\MR}\of{\alpha}{i}$ and can be performed independently for each rotation~\citep{Carlone15icra-init3D}.

\myparagraph{Stage 2: full pose recovery via single \GN iteration} 
In the previous stage we obtained an estimate for the rotations $\hat{\MR}\of{\alpha}{i}, 
\forall \alpha \in \robots, \forall i$. In this stage we use this estimate to reparametrize  
problem~\eqref{eq:MRPGO}. In particular, we rewrite each unknown rotation $\MR\of{\alpha}{i}$ 
as the known estimate $\hat{\MR}\of{\alpha}{i}$ ``plus'' an unknown perturbation; 
in formulas, we rewrite each rotation as $\MR\of{\alpha}{i} = \hat{\MR}\of{\alpha}{i} \expmap{\vtheta\of{\alpha}{i}}$, 
where $\expmap{\cdot}$ is the exponential map for $\SOthree$, and $\vtheta\of{\alpha}{i} \in \Real{3}$ 
(this is our new parametrization for the rotations). 
With this parametrization, eq.~\eqref{eq:MRPGO} becomes: 
\bea
\label{eq:MRPGO-reparametrized}
\min_{
\substack{
\vt\of{\alpha}{i}, \vtheta\of{\alpha}{i} \\
\forall \alpha \in \robots, \forall i
}} 
 \;\;\; \sum_{(\alpha_i,\beta_j) \in \calE} \!\!
 \omegat \normsq{ \vt\of{\beta}{j} \!\!-\! \vt\of{\alpha}{i} \!\!-\! \hat{\MR}\of{\alpha}{i} \expmap{\vtheta\of{\alpha}{i}} \vtbar\subs }{} \\ 
 \!\!+ \frac{\omegaR}{2} \normsq{ \hat{\MR}\of{\beta}{j} \expmap{\vtheta\of{\beta}{j}} 
 \!\!-\! \hat{\MR}\of{\alpha}{i} \expmap{\vtheta\of{\alpha}{i}} \MRbar\subs}{\frob}    \nonumber
\eea
Note that the reparametrization allowed to drop the constraints (we are now trying to estimate 
 vectors in $\Real{3}$), but moved the nonconvexity to the objective ($\expmap{\cdot}$ is nonlinear in its argument).
In order to solve~\eqref{eq:MRPGO-reparametrized}, we take a quadratic approximation of the cost function.
For this purpose we use the following first-order approximation of the exponential map:
\beq
\label{eq:firstOrderExpmap}
\expmap{\vtheta\of{\alpha}{i}} \simeq 
 \eye_3 + \MS(\vtheta\of{\alpha}{i}) 
\eeq
where $\MS(\vtheta\of{\alpha}{i})$ is a skew symmetric matrix whose entries are 
defined by the vector $\vtheta\of{\alpha}{i}$.
Substituting~\eqref{eq:firstOrderExpmap} into~\eqref{eq:MRPGO-reparametrized} we get the 
desired quadratic approximation: 
\bea
\label{eq:MRPGO-approx}
\min_{
\substack{
\vt\of{\alpha}{i}, \vtheta\of{\alpha}{i} \in \Real{3} \\
\forall \alpha \in \robots, \forall i
}} 
 \!\! \sum_{(\alpha_i,\beta_j) \in \calE} \!\!
\omegat
 \normsq{ \vt\of{\beta}{j} \!\!-\! \vt\of{\alpha}{i} \!\!-\! \hat{\MR}\of{\alpha}{i} \vtbar\subs
 \!\!-\! \hat{\MR}\of{\alpha}{i} \MS(\vtheta\of{\alpha}{i}) \vtbar\subs }{} \\ 
 \!\!+ \frac{\omegaR}{2} \normsq{ 
 \hat{\MR}\of{\beta}{j}  
 \!-\! 
 \hat{\MR}\of{\alpha}{i}  \MRbar\subs 
  + \hat{\MR}\of{\beta}{j} \MS(\vtheta\of{\beta}{j}) 
 \!-\! 
 \hat{\MR}\of{\alpha}{i} \MS(\vtheta\of{\alpha}{i}) \MRbar\subs 
 }{\frob}   \nonumber
\eea
%
Rearranging the unknown $\vt\of{\alpha}{i}, \vtheta\of{\alpha}{i}$ of all robots into a single vector $\vp$, 
we rewrite~\eqref{eq:MRPGO-approx} as a linear least-squares problem:
\bea
\label{eq:MRPGO-leastsquares}
\min_{\vp} \;\; \|\MA_p \; \vp - \vb_p \|^2
\eea
whose solution can be found by solving the linear system:
\bea
\label{eq:MRPGO-linSys}
(\MA_p\tran \MA_p) \vp = \MA_p\tran \vb_p
\eea
From the solution of~\eqref{eq:MRPGO-linSys} we can build our trajectory 
estimate: the entries of $\vp$ directly define the 
positions $\vt\of{\alpha}{i}$, $\forall \alpha \in \robots, \forall i$; moreover, $\vp$ includes the rotational corrections
 $\vtheta\of{\alpha}{i}$, from which we get our rotation estimate as:
$\MR\of{\alpha}{i} = \hat{\MR}\of{\alpha}{i} \expmap{\vtheta\of{\alpha}{i}}$.

\begin{remark}[{\bf Advantage of Centralized Two-Stage Approach}] The approach reviewed in this section
 has three advantages. First, as shown in~\cite{Carlone15icra-init3D}, in common problem instances 
  (i.e., for reasonable levels of measurement noise) it 
 returns a solution that is very close to the \ML estimate.
 Second, the approach only requires to solve two linear 
 systems (the cost of projecting the rotations is negligible), hence it is computationally efficient.
 Finally, the approach does not require an initial guess, therefore, it is able to converge even when the 
 initial trajectory estimate is inaccurate (in those instances, iterative optimization 
 tends to fail~\citep{Carlone15icra-init3D}) or is unavailable. 
 \myEndRemark
\end{remark}


\subsection{Distributed Pose Graph Optimization}
\label{sec:distributed}

In this section we show that the two-stage approach described in Section~\ref{sec:centralized}
can be implemented in a distributed fashion. Since the approach only requires solving 
two linear systems, every distributed linear system solver can be used as workhorse to 
split the computation among the robots. For instance, one could adapt the Gaussian elimination 
approach of~\cite{Cunningham10iros} to solve the linear systems~\eqref{eq:normEq-R} and \eqref{eq:MRPGO-linSys}.
In this section we propose alternative approaches, based on the Distributed Jacobi Over-Relaxation and 
the Distributed Successive Over-Relaxation algorithms, 
 and we discuss their advantages.

 In both~\eqref{eq:normEq-R} and \eqref{eq:MRPGO-linSys} we need to solve a linear system where 
 the unknown vector can be partitioned into subvectors, such that each subvector contains the 
 variables associated to a single robot in the team. For instance, we can partition the vector $\vr$ in~\eqref{eq:normEq-R}, 
  as $\vr = [\vr_\alpha, \vr_\beta, \ldots]$, such that $\vr_\alpha$ describes the rotations of robot $\alpha$.
  Similarly, we can partition  $\vp = [\vp_\alpha, \vp_\beta, \ldots]$ in~\eqref{eq:MRPGO-linSys}, such 
  that $\vp_\alpha$ describes the trajectory of robot $\alpha$.
  Therefore,~\eqref{eq:normEq-R} and \eqref{eq:MRPGO-linSys} can be framed in the general form:
\beq
\label{eq:partition}
\MH \vy = \vg \Leftrightarrow
\left[
\begin{array}{ccc}
\MH_{\alpha\alpha} & \MH_{\alpha\beta} & \ldots \\
\MH_{\beta\alpha} & \MH_{\beta\beta} & \ldots \\
\vdots & \vdots & \ddots
\end{array}
\right]
\left[
\begin{array}{c}
\vy_\alpha \\ \vy_\beta \\ \vdots 
\end{array}
\right]
=
\left[
\begin{array}{c}
\vg_\alpha \\ \vg_\beta \\ \vdots 
\end{array}
\right]
\eeq
where we want to compute the vector $\vy = [\vy_\alpha, \vy_\beta, \ldots]$ given 
the (known) block matrix $\MH$ and the (known) block vector $\vg$; on the right of~\eqref{eq:partition} 
we partitioned the square matrix $\MH$ and the vector $\vg$ according to the block-structure of $\vy$.

\newcommand{\indTwo}{\delta}

In order to introduce the distributed  algorithms, we first observe that 
 the linear system~\eqref{eq:partition} can be rewritten as:
 \[
\sum_{\indTwo \in \MOmega } \MH_{\alpha\indTwo} \vy_\indTwo = \vg_\alpha \qquad \forall \alpha \in \MOmega
 \] 
Taking the contribution of 
$\vy_\alpha$ out of the sum, we get:
\beq
\label{eq:distPreamble}
\MH_{\alpha\alpha} \vy_\alpha = - \!\!\!\sum_{\indTwo \in \MOmega \setminus \{\alpha\} } \!\!\! \MH_{\alpha\indTwo} 
\vy_\indTwo + \vg_\alpha
\qquad \forall \alpha \in \MOmega
\eeq
The set of equations~\eqref{eq:distPreamble} is the same as the original system~\eqref{eq:partition}, 
but clearly exposes the contribution of the variables associated to each robot.
The equations~\eqref{eq:distPreamble} constitute the basis for 
 the \emph{Successive Over-Relaxation} (\SOR) and the \emph{Jacobi Over-Relaxation} (\JOR) 
methods that we describe in the following sections. 

\subsubsection{Distributed Jacobi Over-Relaxation (\JOR):\label{sec:jor}} The distributed \JOR 
algorithm~\citep{Bertsekas89book} starts at an arbitrary initial estimate 
$\vy\at{0} = [\vy_\alpha\at{0}, \vy_\beta\at{0}, \ldots]$ and
solves the linear system~\eqref{eq:partition} by repeating the following iterations:
\begin{multline}
\label{eq:jorIterations}
 \vy\at{k+1}_\alpha = (1-\gamma)\vy\at{k}_\alpha  \\ +
 (\gamma)\MH_{\alpha\alpha}\inv \left(- \sum_{\indTwo \in \MOmega \setminus \{\alpha\} }
 \MH_{\alpha\indTwo} \vy\at{k}_\indTwo + \vg_\alpha \right)
 \quad  \forall \alpha \in \MOmega
\end{multline}
where $\gamma$ is the \emph{relaxation factor}. 
Intuitively, at each iteration robot $\alpha$ attempts to solve eq.~\eqref{eq:distPreamble} 
(the second 
summand in~\eqref{eq:jorIterations} is the solution of~\eqref{eq:distPreamble} 
with the estimates of the other robots kept fixed), while remaining close to the previous estimate 
$\vy\at{k}_\alpha$ (first summand in~\eqref{eq:jorIterations}).
If the iterations~\eqref{eq:jorIterations} converge to a fixed point, 
say $\vy_\alpha$ $\forall \alpha$, then the resulting estimate solves 
 the linear system~\eqref{eq:distPreamble} exactly~\cite[page 131]{Bertsekas89book}.
To prove this fact we only need to rewrite~\eqref{eq:jorIterations} after convergence:
\begin{multline}
 \vy_\alpha = (1-\gamma)\vy_\alpha  +
 (\gamma)\MH_{\alpha\alpha}\inv \left(- \sum_{\indTwo \in \MOmega \setminus \{\alpha\} }
 \MH_{\alpha\indTwo} \vy_\indTwo + \vg_\alpha \right) \nonumber
\end{multline}
which can be easily seen to be identical to~\eqref{eq:distPreamble}. 

In our multi robot problem, the distributed \JOR algorithm can be understood in a simple way: at each iteration, each robot  
estimates its own variables ($\vy\at{k+1}_\alpha$) by assuming that the ones of the other robots are constant 
($\vy\at{k}_\indTwo$); iterating this procedure, the robots reach an agreement on the estimates, 
and converge to the solution of eq.~\eqref{eq:partition}. Using the distributed \JOR approach, the 
robots solve~\eqref{eq:normEq-R} and \eqref{eq:MRPGO-linSys} in a distributed manner. When $\gamma=1$, the 
distributed \JOR method is also known as the \emph{distributed Jacobi} (\DJ) method. 

We already mentioned that when the iterations~\eqref{eq:jorIterations} converge, then they return the exact solution of the linear system. So a natural question is: \emph{when do the Jacobi iteration converge?}
A general answer is given by the following proposition.

\begin{proposition}[{\bf Convergence of \JOR}~\citep{Bertsekas89book}]
\label{prop:convergenceJOR}
Consider the linear systems~\eqref{eq:partition}
and 
 define the block diagonal matrix $\MD \doteq \diag{\MH_{\alpha\alpha}, \MH_{\beta\beta}, \ldots}$.
Moreover, define the matrix:
\beq
\MM = (1-\gamma) \eye - \gamma \MD\inv (\MH - \MD)
\eeq
where $\eye$ is the identity matrix of suitable size.
Then, the \JOR iterations~\eqref{eq:jorIterations} converge from any initial estimate 
if and only if $\rho(\MM)<1$, where $\rho(\cdot)$ denotes the spectral radius 
(maximum of absolute value of the eigenvalues) of a matrix.
\end{proposition} 

The proposition is the same of Proposition 6.1 in~\citep{Bertsekas89book} 
(the condition that $\eye - \MM$ is invertible is guaranteed to hold as 
noted in the footnote on page 144 of~\citep{Bertsekas89book}).

It is non-trivial to establish whether our 
linear systems~\eqref{eq:normEq-R} and \eqref{eq:MRPGO-linSys} satisfy the condition
of Proposition~\ref{prop:convergenceJOR}.
In the experimental section, we empirically observe that the Jacobi iterations indeed converge 
whenever $\gamma \leq 1$. For the \SOR algorithm, presented in the next section, instead, 
we can provide stronger theoretical convergence guarantees.

\subsubsection{Distributed Successive Over-Relaxation (\SOR)}
\label{sec:sor}

The distributed \SOR algorithm~\citep{Bertsekas89book} starts at an arbitrary initial estimate
$\vy\at{0} = [\vy_\alpha\at{0}, \vy_\beta\at{0}, \ldots]$ and, at iteration $k$,
applies the following update rule, for each $\alpha \in \MOmega$:
\begin{multline}
\label{eq:sorIterations}
 \vy\at{k+1}_\alpha = (1-\gamma)\vy\at{k}_\alpha  \\ + (\gamma)\MH_{\alpha\alpha}\inv \left(
  - \sum_{\indTwo \in \MOmega^+_\alpha } 
 \MH_{\alpha\indTwo} \vy\at{k+1}_\indTwo
 - \sum_{\indTwo \in \MOmega^-_\alpha } 
 \MH_{\alpha\indTwo} \vy\at{k}_\indTwo  
 + \vg_\alpha \right)
\end{multline}
where $\gamma$ is the \emph{relaxation factor}, $\MOmega^+_\alpha$ is the set of robots that already computed the $(k+1)$-th estimate, while
$\MOmega^-_\alpha$ is the set of robots that still have to perform the update~\eqref{eq:sorIterations},
excluding node $\alpha$ (intuitively: each robot uses the latest estimate).
As for the \JOR algorithm, by comparing~\eqref{eq:sorIterations} and \eqref{eq:distPreamble}, we see that
if the sequence produced by the iterations~\eqref{eq:sorIterations} converges to a fixed point,
then such point  satisfies~\eqref{eq:distPreamble}, and indeed solves the original linear
system~\eqref{eq:partition}. When $\gamma=1$, the distributed \SOR method is known
as the \emph{distributed Gauss-Seidel} (\DGS) method. 

The following proposition, whose proof trivially follows from~\cite[Proposition 6.10, p. 154]{Bertsekas89book} 
(and the fact that the involved matrices are positive definite), 
establishes when the distributed \SOR algorithm converges to the desired solution.

\begin{proposition}[{\bf Convergence of \SOR}]
\label{prop:convergenceSOR}
The \SOR iterations~\eqref{eq:sorIterations} applied to the linear systems~\eqref{eq:normEq-R} and \eqref{eq:MRPGO-linSys} 
converge to the solution of the corresponding linear system   
 (from any initial estimate) whenever $\gamma \in (0,2)$, while the iterations do no 
 converge to the correct solution whenever $\gamma \notin (0,2)$.  
\end{proposition} 

According to~\cite[Proposition 6.10, p. 154]{Bertsekas89book}, 
for $\gamma \notin (0,2)$, 
the \SOR iterations~\eqref{eq:sorIterations} do not converge to the 
solution of the linear system in general, hence also in practice, we restrict 
the choice of $\gamma$ in the open interval $(0,2)$. In the experimental section, we 
show that the choice $\gamma = 1$ ensures the fastest convergence.

\subsubsection{Communication Requirements for \JOR and \SOR}

\begin{figure}[h!]
\centering%
\includegraphics[width=0.8\columnwidth]{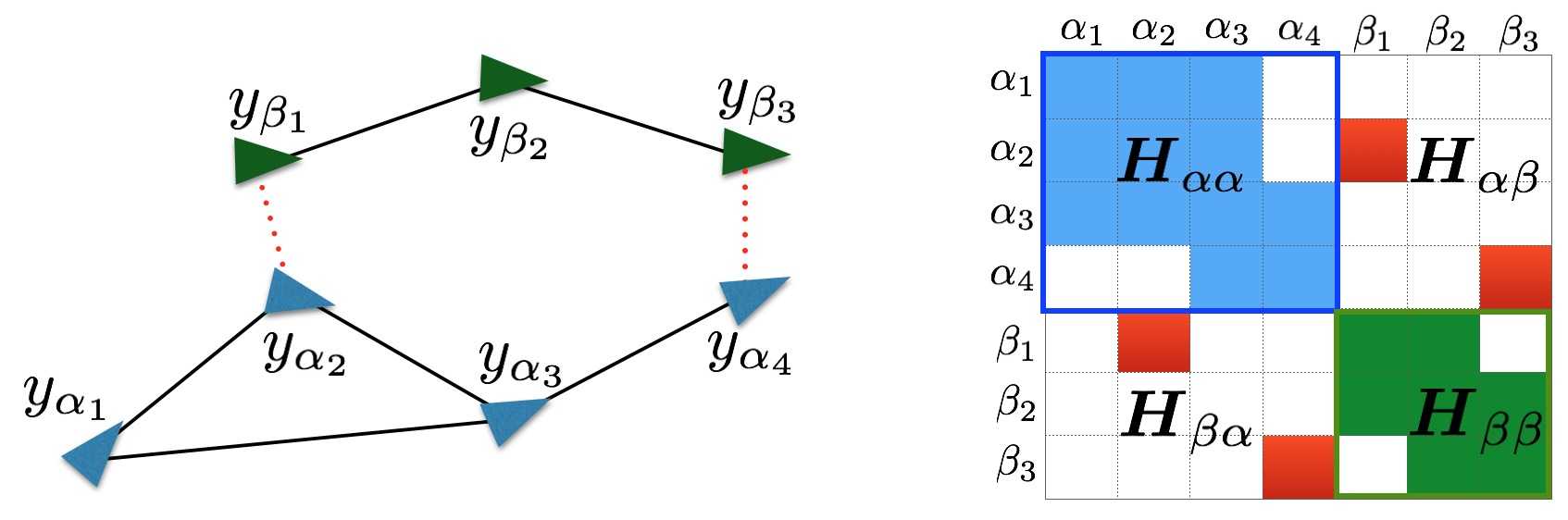}\\
\caption{Example: (left) trajectory estimation problem and (right) 
 corresponding block structure of the matrix $\MH$.\label{fig:exampleStructure}}
\end{figure}

In this section we observe that to execute the \JOR and \SOR iterations~\eqref{eq:jorIterations}\eqref{eq:sorIterations}, 
robot $\alpha$ only needs its intra and inter-robot measurements $\calE^\alpha_\inner$ and $\calE^\alpha_\sep$, 
and an estimate of the separators, involved in the inter-robot measurements in  $\calE^\alpha_\sep$. 
For instance, in the graph of Fig.~\ref{fig:exampleStructure} robot $\alpha$ only needs the 
 estimates of $\vy_{\beta_1}$ and $\vy_{\beta_3}$, while does not require any 
 knowledge about the other poses of $\beta$.  


To understand this fact, we note that both~\eqref{eq:normEq-R} and \eqref{eq:MRPGO-linSys} model 
an estimation problem from pairwise relative measurements. It is well known that the 
matrix $\MH$ (sometimes called the \emph{Hessian}~\citep{Dellaert05rss}) underlying these problems has a block structure defined by the Laplacian matrix of the 
underlying graph~\citep{Barooah07csm}. For instance, Fig.~\ref{fig:exampleStructure} (right) shows the 
block sparsity of the matrix $\MH$ describing the graph on the left: off-diagonal block-elements 
in position $(\alpha_i,\beta_j)$ are non zero if and only if there is an edge (i.e., a measurement) between $\alpha_i$ 
and $\beta_j$. 

By exploiting the block sparsity of $\MH$, we can further simplify the \JOR ~\eqref{eq:jorIterations} iterations
as:
\begin{multline}
\label{eq:jorIterationsSparsity}
\hspace{-0.2cm}
 \vy\at{k+1}_\alpha = (1-\gamma)\vy\at{k}_\alpha  \\ + (\gamma)\MH_{\alpha\alpha}\inv \left(-
 \sum_{(\alpha_i,\indTwo_j) \in \calE^\alpha_\sep} 
 \MH_{\alpha_i\indTwo_j} \vy\at{k}_{\indTwo_j} + \vg_\alpha \right), \;\;  \forall \alpha \in \MOmega
\end{multline}
where we simply removed the contributions of the zero blocks from the sum in~\eqref{eq:jorIterations}. 

Similarly we can simplify the \SOR~\eqref{eq:sorIterations} iterations as:

\begin{multline}
\label{eq:sorIterationsSparsity}
\hspace{-0.2cm}
 \vy\at{k+1}_\alpha \!\! = (1-\gamma)\vy\at{k}_\alpha  \\ + (\gamma) \MH_{\alpha\alpha}\inv \!\! \left( \,
 \; - \!\!\!\!\!\!\!\!\!\! \sum_{\;\;(\alpha_i,\indTwo_j) \in {\calE^\alpha_\sep}^+ } \!\!\!\!\!\!\!\!\! \MH_{\alpha_i\indTwo_j} \vy\at{k+1}_{\indTwo_j} 
 \! - \!\!\!\!\!\!\!\!\! \sum_{(\alpha_i,\indTwo_j) \in {\calE^\alpha_\sep}^- } \!\!\!\!\!\!\!\!\! \MH_{\alpha_i\indTwo_j} \vy\at{k}_{\indTwo_j} 
 \!\!+ \vg_\alpha \right) 
\end{multline}
where we  removed the contributions of the zero blocks from the sum in~\eqref{eq:sorIterations}; 
the sets ${\calE^\alpha_\sep}^+$ and ${\calE^\alpha_\sep}^-$ satisfy ${\calE^\alpha_\sep}^+ \cup {\calE^\alpha_\sep}^- = {\calE^\alpha_\sep}$, 
and are such that ${\calE^\alpha_\sep}^+$ includes the inter-robot measurements involving  
robots which already performed the $(k+1)$-th iteration, while ${\calE^\alpha_\sep}^-$ 
is the set of measurements involving robots that have not performed the iteration yet (as before: 
each robot simply uses its latest estimate).

Eqs.~\eqref{eq:jorIterationsSparsity} and~\eqref{eq:sorIterationsSparsity} highlight that the \JOR and \SOR iterations (at robot $\alpha$) only require the estimates 
for poses involved in its inter-robot measurements $\calE^\alpha_\sep$. 
Therefore both \JOR and \SOR involve almost no ``privacy violation'': every other robot 
$\beta$ in the team does not need to communicate any other information about its own 
trajectory, but only sends an estimate of its rendezvous poses.
\subsubsection{Flagged Initialization}
\label{sec:flaggedInit}

As we will see in the experimental section and according to 
 Proposition~\ref{prop:convergenceSOR}, the \JOR and \SOR approaches converge
from any initial condition when $\gamma$ is chosen appropriately. However, starting from a ``good'' initial  
condition can reduce the number of iterations to converge, and in turns reduces the communication 
burden (each iteration~\eqref{eq:jorIterationsSparsity} or \eqref{eq:sorIterationsSparsity} requires the robots to exchange 
their estimate of the separators). 

In this work, we follow the path of~\cite{Barooah05icisip} and adopt a \emph{flagged initialization}. 
A flagged initialization scheme only alters the first \JOR or \SOR iteration as follows.
Before the first iteration, all robots are marked as ``uninitialized''. 
Robot $\alpha$ performs its iteration~\eqref{eq:jorIterationsSparsity} or \eqref{eq:sorIterationsSparsity} without 
considering the inter-robot measurements, i.e., eqs.~\eqref{eq:jorIterationsSparsity}-\eqref{eq:sorIterationsSparsity} 
become $\vy\at{k+1}_\alpha = \MH_{\alpha\alpha}\inv \vg_\alpha$; then the robot $\alpha$ 
marks itself as ``initialized''. When the robot $\beta$ performs its iteration, it includes 
only the separators from the robots that are initialized; after performing the \JOR or \SOR iteration, 
also $\beta$ marks itself as initialized. Repeating this procedure, all robots 
become initialized after performing the first iteration. The following iterations then proceed 
according to the standard \JOR~\eqref{eq:jorIterationsSparsity} or \SOR~\eqref{eq:sorIterationsSparsity} update. 
\cite{Barooah05icisip} show a significant improvement in convergence 
 using flagged initialization. As discussed in the experiments, flagged initialization is 
 also advantageous in our distributed pose graph optimization problem.


\section{Dealing With Bandwidth Constraints II: Compressing Sensor Data via Object-based Representations}
\label{sec:objects}

The second contribution of this paper is the use of high-level object-based models at the 
estimation front-end and as a map representation for multi robot SLAM.
Object-based abstractions are crucial to further reduce the memory storage and the
 information exchange among the robots. 

Previous approaches for multi robot mapping rely on feature-based maps which become memory-intensive for long-term operation, contain a 
large amount of redundant information, and lack the semantic understanding necessary to
perform a wider range of tasks (e.g., manipulation, human-robot interaction). To solve these issues, 
we present an approach for multi robot SLAM which uses
object landmarks \citep{Moreno13cvpr} in a multi robot mapping setup. 

Section~\ref{sec:problemFormulationObjectSLAM} introduces the additional mathematical notation and formalizes the problem of distributed object-based SLAM. Section~\ref{sec:objectBasedSLAM} presents the implementation details of our distributed object-based SLAM system.

\begin{figure}[h]
\centering
\includegraphics[width=0.7\columnwidth, trim=0cm 0cm 0cm 0cm,clip]{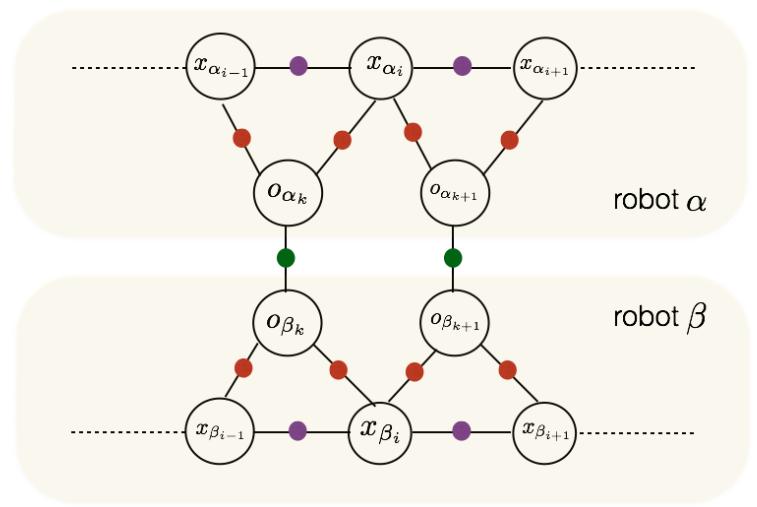} 
\caption{\label{fig:objectSLAMFactorGraph} Factor graph representation of Multi-Robot Object based SLAM. $\vxx\of{\alpha}{i}$ and $\vxx\of{\beta}{i}$
denote the poses assumed by robot $\alpha$ and $\beta$ at time $i$ respectively. The 
pose of the $k^{th}$ object in the coordinate frame of 
robot $\alpha$ and $\beta$ is denoted with $\vo\of{\alpha}{k}$ and $\vo\of{\beta}{k}$ respectively. Green dots shows inter-robot factors 
whereas orange and purple dots shows intra-robot factors.}
\end{figure}

\subsection{Distributed Object-based SLAM}
\label{sec:problemFormulationObjectSLAM}

We consider a multi robot system as defined in Section \ref{sec:revisitedPGO}.
Each robot, in addition to estimating its own trajectory using local measurements and occasional communication with other robots, also estimates the pose of a set of objects in the environment.  
We model each trajectory  as a finite set of poses; the trajectory of robot $\alpha$ is  $\vxx\of{\alpha}{} = [\vxx\of{\alpha}{1},\vxx\of{\alpha}{2}, \ldots]$.
  In addition,  we denote with $\vo\of{\alpha}{k}\in \SEthree$ the pose of the $k^{th}$ object in the coordinate frame of robot $\alpha$ (Fig. \ref{fig:objectSLAMFactorGraph}).

\myparagraph{Measurements} Similar to distributed pose graph optimization (Section \ref{sec:revisitedPGO}), we assume that each robot acquires two types of relative pose measurements: intra-robot and inter-robot measurements.  The \emph{intra-robot measurements} consist of the odometry measurements, which constrain consecutive robot poses (e.g., $\vxx\of{\alpha}{i}$ and $\vxx\of{\alpha}{i+1}$ in Fig.~\ref{fig:objectSLAMFactorGraph}), and object measurements which constrains robot poses with the corresponding visible object landmarks (e.g., $\vxx\of{\alpha}{i}$ and $\vo\of{\alpha}{k}$ in Fig.~\ref{fig:objectSLAMFactorGraph}). 
Contrarily to Section~\ref{sec:revisitedPGO}, 
the \emph{inter-robot measurements} relate the object poses observed by different robots. 
During a rendezvous between robot $\alpha$ and robot $\beta$, each robot shares the label and pose of detected object landmarks with the other robot. Then, for each object observed by both robots, 
the teammates add an inter-robot measurements, 
enforcing the object pose estimate to be consistent across the teammates. 
For instance, if $\vo\of{\beta}{k}$ and $\vo\of{\alpha}{k}$ in Fig.~\ref{fig:objectSLAMFactorGraph} 
 model the pose of the same object, then the two poses should be identical. For this reason, intra-robot measurement between a pair of associated object poses is zero.  
 
 The intra-robot object measurements follow the same measurements model of eq.~\eqref{eq:measurementModel}.
 For instance, if the robot $\alpha$ at time $i$ and at pose  $\vxx\of{\alpha}{i}$ 
 observes an object at pose $\vo\of{\alpha}{k}$, then the corresponding measurement 
 $\vzbar^{\vxx_{\alpha_{i}}}_{\vo_{\alpha_{k}}}$ measures the relative pose between $\vxx\of{\alpha}{i}$  and $\vo\of{\alpha}{k}$.
 Consistently with our previous notation, we denote intra-robot object measurement between $\vxx\of{\alpha}{i}$ and $\vo\of{\alpha}{k}$ as  $\vzbar^{\vxx_{\alpha_{i}}}_{\vo_{\alpha_{k}}}$, 
 and inter-robot measurement between object poses $\vo\of{\alpha}{k}$ and $\vo\of{\beta}{k}$ as $\vzbar^{\vo_{\alpha_{k}}}_{\vo_{\beta_{k}}}$. 
 
In the following, we denote with $\calE^{\alpha}_{I}$ the set of intra-robot odometry for robot $\alpha$, while we call $\calE_{I}$ the set of intra-robot odometry measurements for all robots in the team, i.e., $\calE_{I} = \cup_{\alpha \in \robots} \calE^{\alpha}_{I}$. 
Similarly the set of intra-robot object measurements for robot $\alpha$ is denoted as  $\calE^{\alpha}_{o}$, whereas the set of all intra-robot object measurements is denoted as $\calE_{o}$.
 Similar to Section \ref{sec:revisitedPGO}, the set of inter-robot measurements involving robot $\alpha$ is denoted with $\calE^{\alpha}_{\sep}$. 
 The set of all inter-robot measurements is denoted with $\calE_{\sep}$. The set of all available measurements  is then $\calE = \calE_{I} \cup \calE_{O}  \cup \calE_{\sep}$.
Note that each robot only has access to its own intra and inter-robot measurements 
$\calE^{\alpha}_{I}$, $\calE^{\alpha}_{O}$ and $\calE^{\alpha}_{\sep}$.



\myparagraph{\ML trajectory and objects estimation}
Let us collect all robot trajectories and object poses in a 
(to-be-estimated) set of robot poses $\vxx = [\vxx\of{\alpha}{}, \vxx\of{\beta}{}, \vxx\of{\gamma}{}, \ldots ]$ and set of object poses $\vo = [\vo\of{\alpha}{}, \vo\of{\beta}{}, \vo\of{\gamma}{}, \ldots ]$. 
The \ML estimate for $\vxx$ and $\vo$ is defined as the maximum of the measurement likelihood:
%
\begin{multline}
\label{eq:maxLikeObjectSLAM}
\vxx^\star, \vo^\star = 
\argmax_{\vxx, \vo} \prod_{(\vxx_{\alpha_{i}},\vxx_{\alpha_{i+1}}) \in \calE_{I}} 
\calL(\vzbar^{\vxx_{\alpha_{i}}}_{\vxx_{\alpha_{i+1}}} \;|\; \vxx) \\
\prod_{(\vxx_{\alpha_{i}},\vo_{\alpha_{k}}) \in \calE_{O}} \calL(\vzbar^{\vxx_{\alpha_{i}}}_{\vo_{\alpha_{k}}}  \;|\; \vxx, \vo)
\prod_{(\vo_{\alpha_{i}},\vo_{\beta_{j}}) \in \calE_{\sep}} \calL(\vzbar^{\vo_{\alpha_{i}}}_{\vo_{\beta_{j}}}  \;|\; \vxx, \vo)
\end{multline}
%
where we used the same assumptions on measurement noise as in Section \ref{sec:revisitedPGO}.  
Defining $\calX= \vxx \cup \vo$, we rewrite eq.~\eqref{eq:maxLikeObjectSLAM} as:

\beq
\label{eq:maxLikeObjectSLAMSimplified}
\calX^\star =  
\argmax_{\calX} \prod_{(\alpha_i,\beta_j) \in \calE} \calL( \vzbar\subs  \;|\; \calX)
\eeq

Since the optimization problem in eq.~\eqref{eq:maxLikeObjectSLAMSimplified} has the same structure of the one in
 eq.~\eqref{eq:maxLike}, we follow the same steps to solve 
it in a distributed manner using the Distributed Gauss-Seidel method.   

The next section  presents the implementation details of our distributed object-based SLAM system.

\begin{figure}[h]
\centering
\includegraphics[width=0.8\columnwidth, trim=0cm 0cm 0cm 0cm,clip]{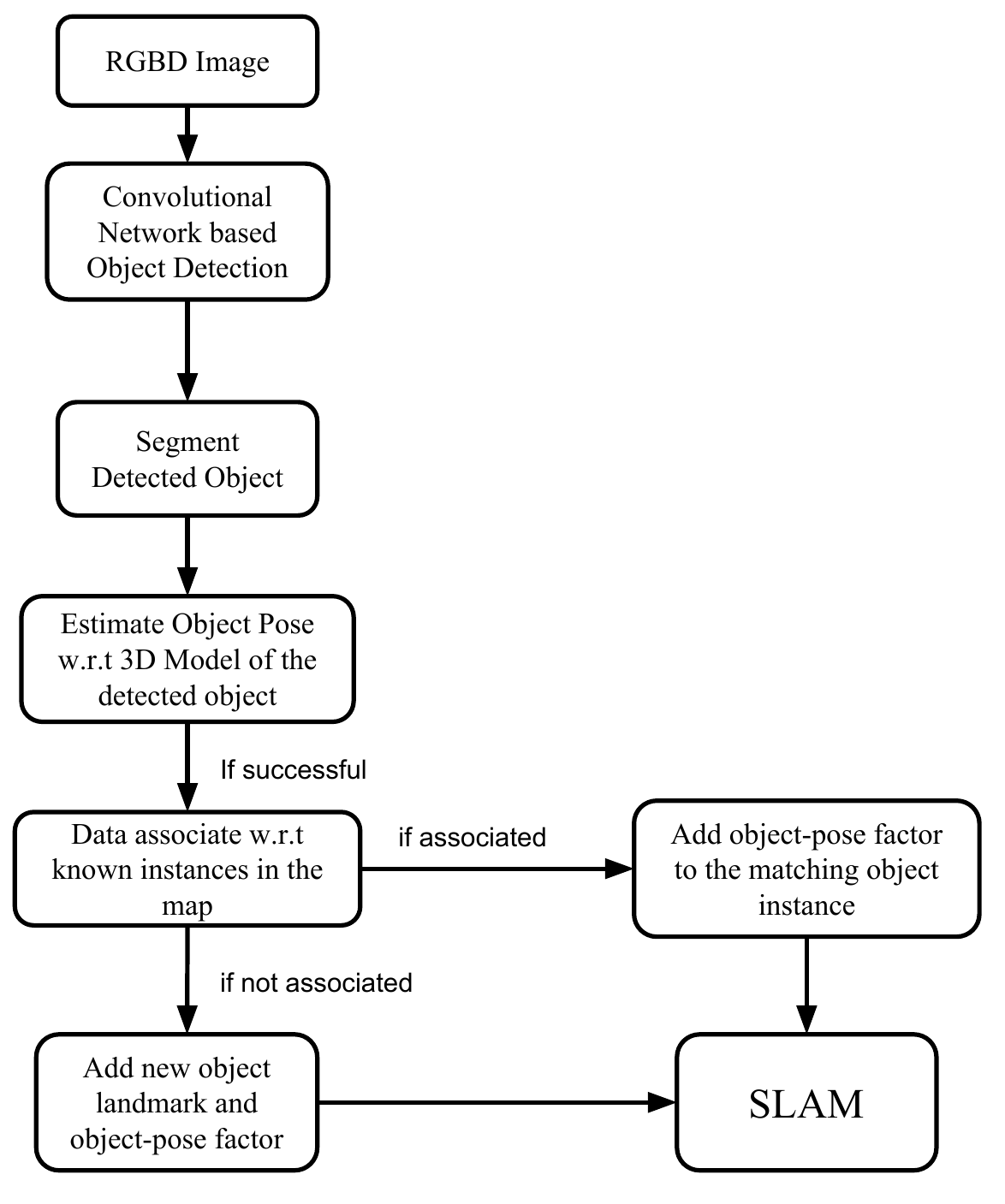} 
\caption{\label{fig:objectSLAMFlowChart}Flowchart of Object based SLAM}
\end{figure}

\subsection{Object-based SLAM Implementation}
\label{sec:objectBasedSLAM}

\myparagraph{Object detection and pose estimation}
Each robot collects RGBD data using a depth camera, and measures its ego-motion 
through wheel odometry. In our approach, each RGB frame (from RGBD) is passed to the YOLO object detector~\citep{Redmon15arxiv},
which detects objects at 45 frames per second.
Compared to object-proposal-based detectors, YOLO is fast, since it avoids the computation burden of extracting object proposals, and is less likely to produce false positives in the background. We fine-tune the YOLO detector on a subset of objects from the \emph{BigBird} dataset (\cite{Singh14icra}). The training dataset contains the object images in a clean background taken from different viewpoints and labeled images of the same objects taken by a robot in an indoor environment. During testing, we use a probability threshold of 0.3 to avoid false detections. 

Each detected object bounding box is  segmented using the \emph{organized point cloud segmentation}~\citep{Trevor13spme}. The segmented object is then matched to the 3D template of the detected object class to estimate its pose. We extract PFHRGB features \citep{Rusu08iros} in the source (object segment) and target (object model) point clouds and register the two point clouds in a Sample Consensus Initial Alignment framework \citep{RusuDoctoralDissertation}. If we have at least 12 inlier correspondences, GICP (generalized iterative closest 
point~\cite{Segal09rss} is performed to further refine the registration and the final transformation is used as the object pose estimate. If less than 12 inlier correspondences are found, the detection is considered to be a false positive and the 
corresponding measurement is discarded. In hindsight, this approach verifies the detection both semantically and geometrically.  

\myparagraph{Data Association}
If object pose estimation is successful, it is data-associated with other instances already present in the map by finding the object landmark having the same category label within $2\sigma$ distance of the newly detected object. If there are multiple objects with the same label within that distance, the newly detected object is matched to the nearest object instance. If there exists no object having the same label, a new object landmark is created. 

Before the first rendezvous event, each robot performs 
standard single-robot SLAM using OmniMapper \citep{Trevor12icra}. 
Both wheel odometry and relative pose measurements to the observed objects 
are fed to the SLAM back-end.
 A flowchart of the approach is given in Fig.~\ref{fig:objectSLAMFlowChart}.

\myparagraph{Robot Communication}
During a rendezvous between robots $\alpha$ and $\beta$, robot  $\alpha$ communicates the category labels (class) and poses (in robot $\alpha$'s frame) of all the detected objects to robot $\beta$. 
We assume that the initial pose of each robot is known to all the robots, hence, 
given the initial pose of robot $\alpha$, robot $\beta$ is able to transform the communicated object 
poses from robot $\alpha$'s frame to its own frame.\footnote{The knowledge of the initial pose is only used to facilitate data association but it is not actually 
used during pose graph optimization. We believe that this assumption can be easily relaxed but 
for space reasons we leave this task to future work.} For each object in the list communicated by 
robot $\alpha$, 
robot $\beta$ finds the nearest object in its map, having the 
same category label and within $2\sigma$ distance. 
If such an object exists, it is added to the list of \emph{shared} objects: this is the set of objects seen by both robots.
The list of shared objects contains pairs $(\vo\of{\alpha}{k},\vo\of{\beta}{l})$ and informs the 
robots that the poses $\vo\of{\alpha}{k}$ and $\vo\of{\beta}{l}$ correspond to the same physical object, 
observed by the two robots.  
For this reason, in the optimization we enforce the relative pose between
 $\vo\of{\alpha}{k}$ and $\vo\of{\beta}{l}$  to be zero.
 
 We remark that, while before the first rendezvous the robots $\alpha$ and $\beta$ have 
different reference frames,  the object-object factors enforce both robots to have a single 
shared reference frame, facilitating future data association. 

Next we show the experimental evaluation which includes realistic Gazebo simulations and field experiments in a military test facility. 



\definecolor{dgreen}{rgb}{0,0.5,0}

\section{Experiments}
\label{sec:experiments}

We evaluate the distributed \JOR and \SOR along with \DJ and \DGS approaches 
(with and without using objects) in large-scale simulations (Section~\ref{sec:simulations} and \ref{sec:simulations_objectSLAM}) 
and field tests (Section~\ref{sec:real} and \ref{sec:real_objectSLAM}). 
The results demonstrate that (i) the \DGS dominates the other algorithms 
considered in this paper in terms of convergence speed,
(ii) the \DGS algorithm is accurate, scalable, and robust to noise,
(iii) the \DGS requires less communication than techniques from related work (i.e., DDF-SAM), 
and (iv) in real applications, the combination of \DGS and object-based mapping reduces the 
communication requirements by several orders of magnitude compared to approaches 
exchanging raw measurements. 

\subsection{Simulation Results: Multi Robot Pose Graph Optimization}
\label{sec:simulations}

\newcommand{\widthCol}{0.3\columnwidth}
\newcommand{\scaleFig}{0.18}
\begin{figure}[t]
\begin{minipage}{\columnwidth}
 \small
\begin{tabular}{ccc}%
\!\!\!\!\!\!
\begin{minipage}{\widthCol}%
\centering%
\includegraphics[scale=0.2, trim=0cm 0cm 0cm 0cm,clip]{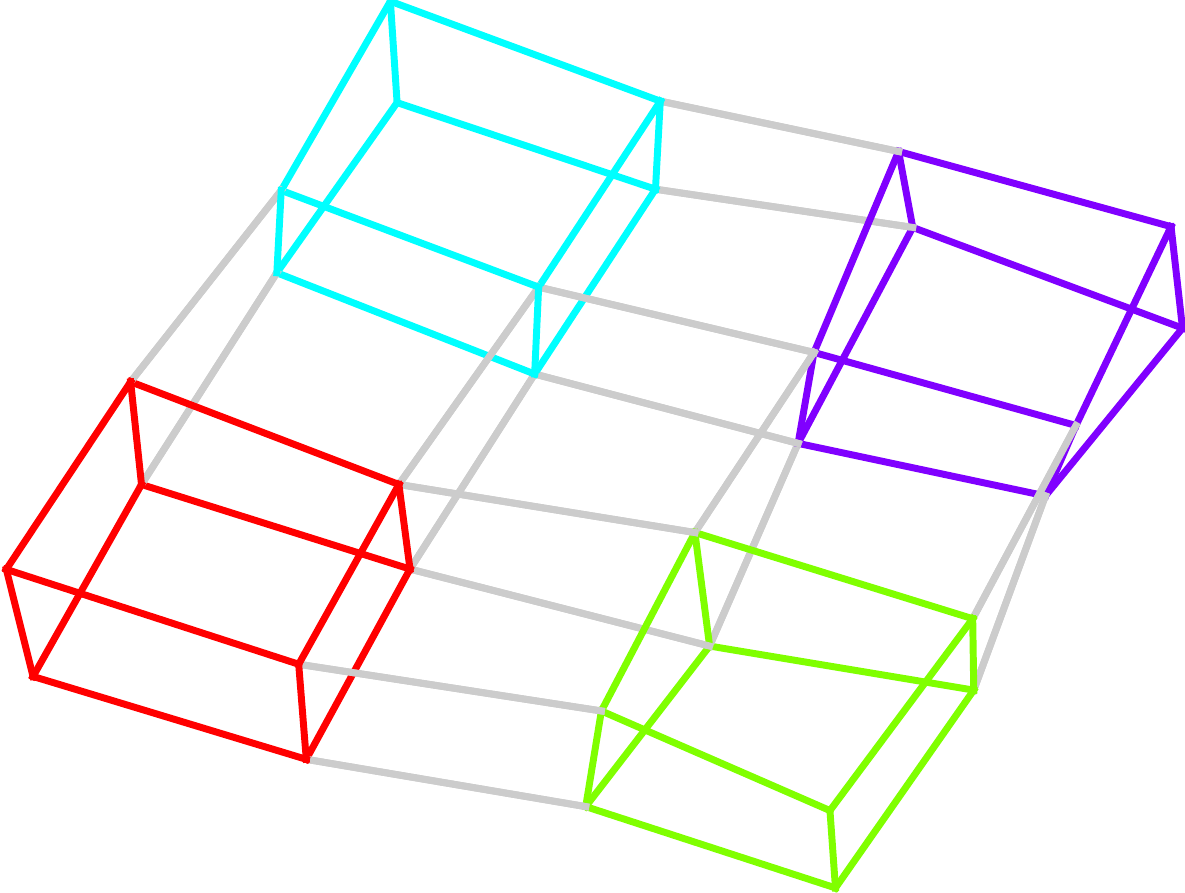} 
\end{minipage}
\!\!\!\! & \!\!\!\!\!\!
\begin{minipage}{\widthCol}%
\centering%
\includegraphics[scale=0.21, trim=0cm 0cm 0cm 0cm,clip]{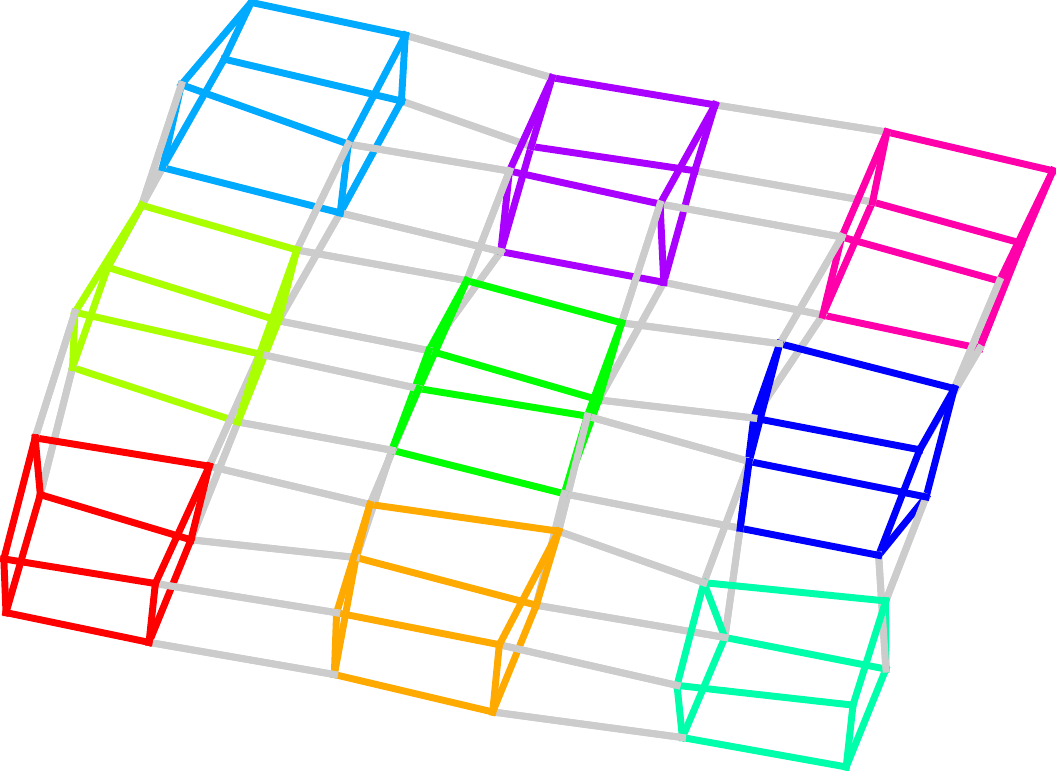} 
\end{minipage}
\!\!\!\! & \!\!\!\!
\begin{minipage}{\widthCol}%
\centering%
\includegraphics[scale=0.23, trim=0cm 0cm 0cm 0cm,clip]{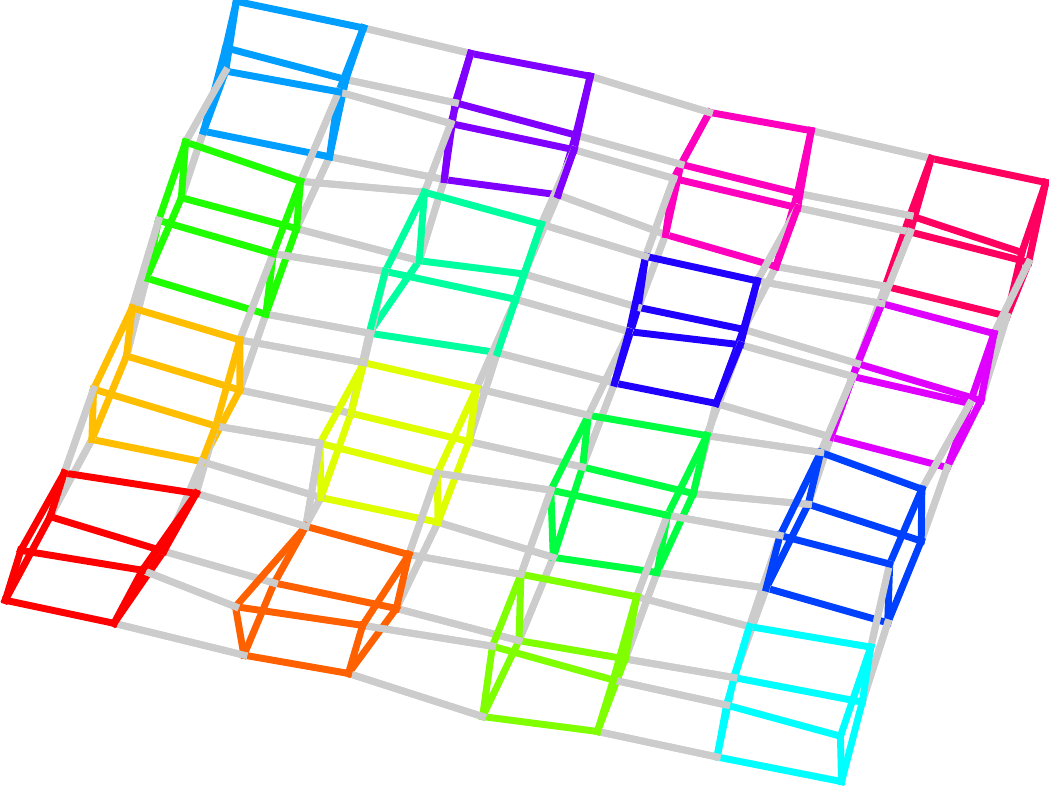} 
\end{minipage}
\\
(a) 4 Robots & (b) 9 Robots  & \hspace{0.3cm}(c) 16 Robots
\end{tabular}
\end{minipage}
\caption{\label{fig:datasetPlot} Simulated 3D datasets with different number of robots. Robots are shown in different colors.
Gray links denote inter-robot measurements.} 
\vspace{-0.3cm}
\end{figure}

In this section, we characterize the performance of the proposed approaches in terms of
convergence, scalability (in the number of robots and separators), 
and sensitivity to noise. 

\myparagraph{Simulation setup and performance metrics}
For our tests, we created simulation datasets in six different configurations 
with increasing number of robots: 4, 9, 16, 25, 36 and 49 robots. The robots are arranged
in a 3D grid with each robot moving on a cube, as shown in Fig.~\ref{fig:datasetPlot}. 
When the robots are at contiguous corners, they can communicate (gray links).
Unless specified otherwise, we generate measurement noise from a zero-mean Gaussian distribution with 
standard deviation $\sigma_R = 5^\circ$ for the rotations and $\sigma_t = 0.2$m for the translations.
Results are averaged over 10 Monte Carlo runs. 

In our problem, \JOR or \SOR are used to sequentially solve two linear systems,~\eqref{eq:normEq-R} 
and \eqref{eq:MRPGO-linSys}, which return the 
minimizers of~\eqref{eq:MRPGO-R-relax2} and~\eqref{eq:MRPGO-leastsquares}, respectively. 
Defining, $m_r \doteq \min_\vr \|\MA_r \vr - \vb_r \|^2$, we use
the following metric, named the \emph{rotation estimation error}, to quantify the error in solving~\eqref{eq:normEq-R}:
\beq
e_r(k) = \| \MA_r \vr\at{k} - \vb_r \|^2 - m_r
\eeq
$e_r(k)$ quantifies how far is the current estimate $\vr\at{k}$ (at the $k$-th iteration) 
from the minimum of the quadratic cost. Similarly, we define the \emph{pose estimation error} as:
\beq
e_p(k) = \| \MA_p \vp\at{k} - \vb_p \|^2 - m_p
\eeq
with $m_p \doteq \min_{\vp} \;\; \|\MA_p \; \vp - \vb_p \|^2$. 
Ideally, we want $e_r(k)$ and $e_p(k)$ to quickly converge to zero for increasing $k$.

Ultimately, the accuracy of the proposed approach depends on the number of iterations, 
hence we need to set a stopping condition for the \JOR or \SOR iterations.
We use the following criterion: we stop the iterations if the change in the estimate is sufficiently small.
More formally, the iterations stop when $\|\vr\at{k+1}-\vr\at{k}\|\leq \eta_r$ (similarly, for the 
second linear system $\|\vp\at{k+1}-\vp\at{k}\|\leq \eta_p$). We use $\eta_r=\eta_p=10^{-1}$ as 
 stopping condition unless specified otherwise. 

\definecolor{dgreen}{rgb}{0,0.5,0}
\begin{figure}[t]
\centering
\begin{minipage}{\columnwidth}
\hspace{-5mm}
\begin{tabular}{cc}%
\begin{minipage}{0.5\columnwidth}%
\centering%
\includegraphics[scale=0.28, trim=0cm 0cm 0cm 0cm,clip]{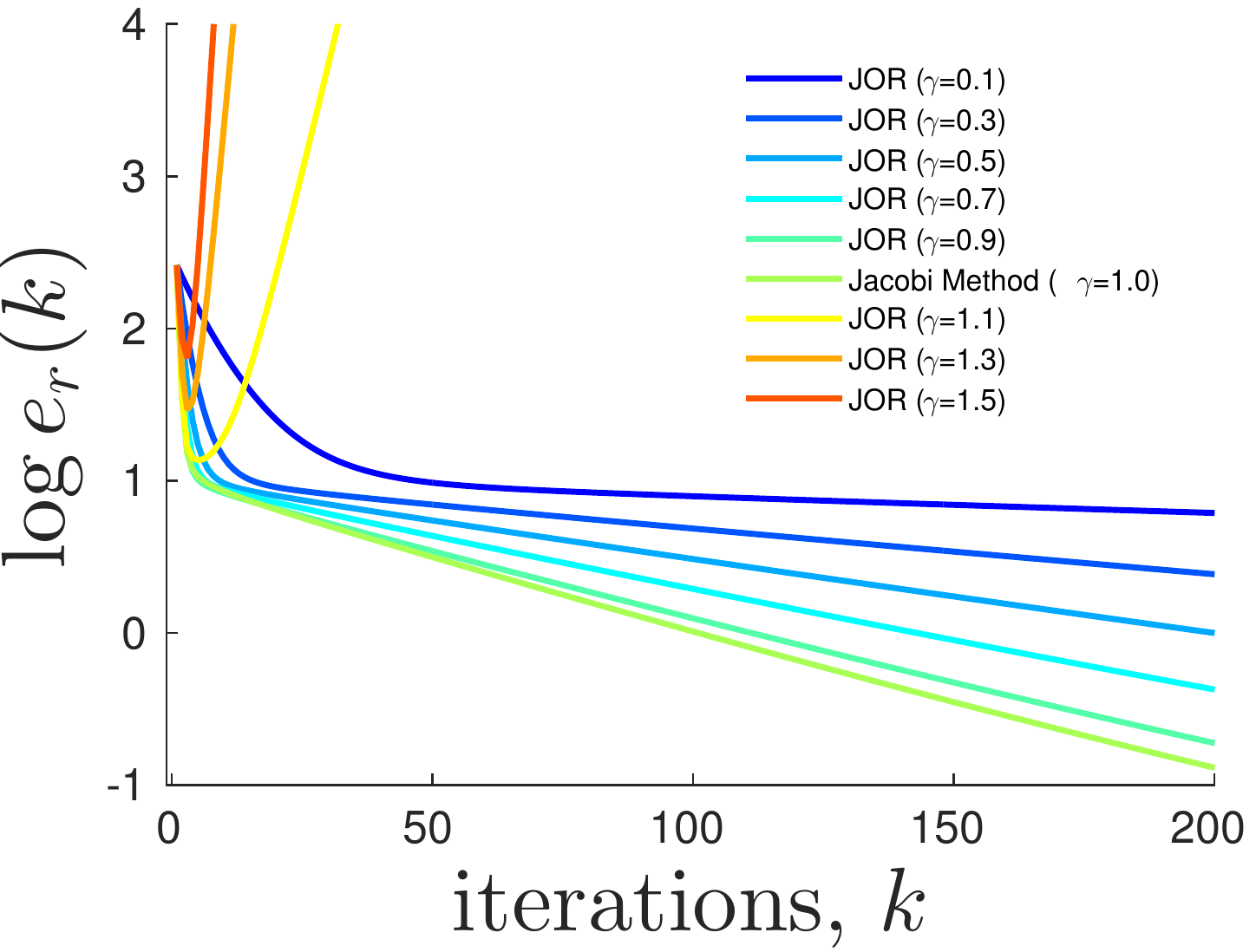} 
\end{minipage}
&\hspace{-1mm}
\begin{minipage}{0.5\columnwidth}%
\centering%
\includegraphics[scale=0.30, trim=0cm 0cm 0cm 0cm,clip]{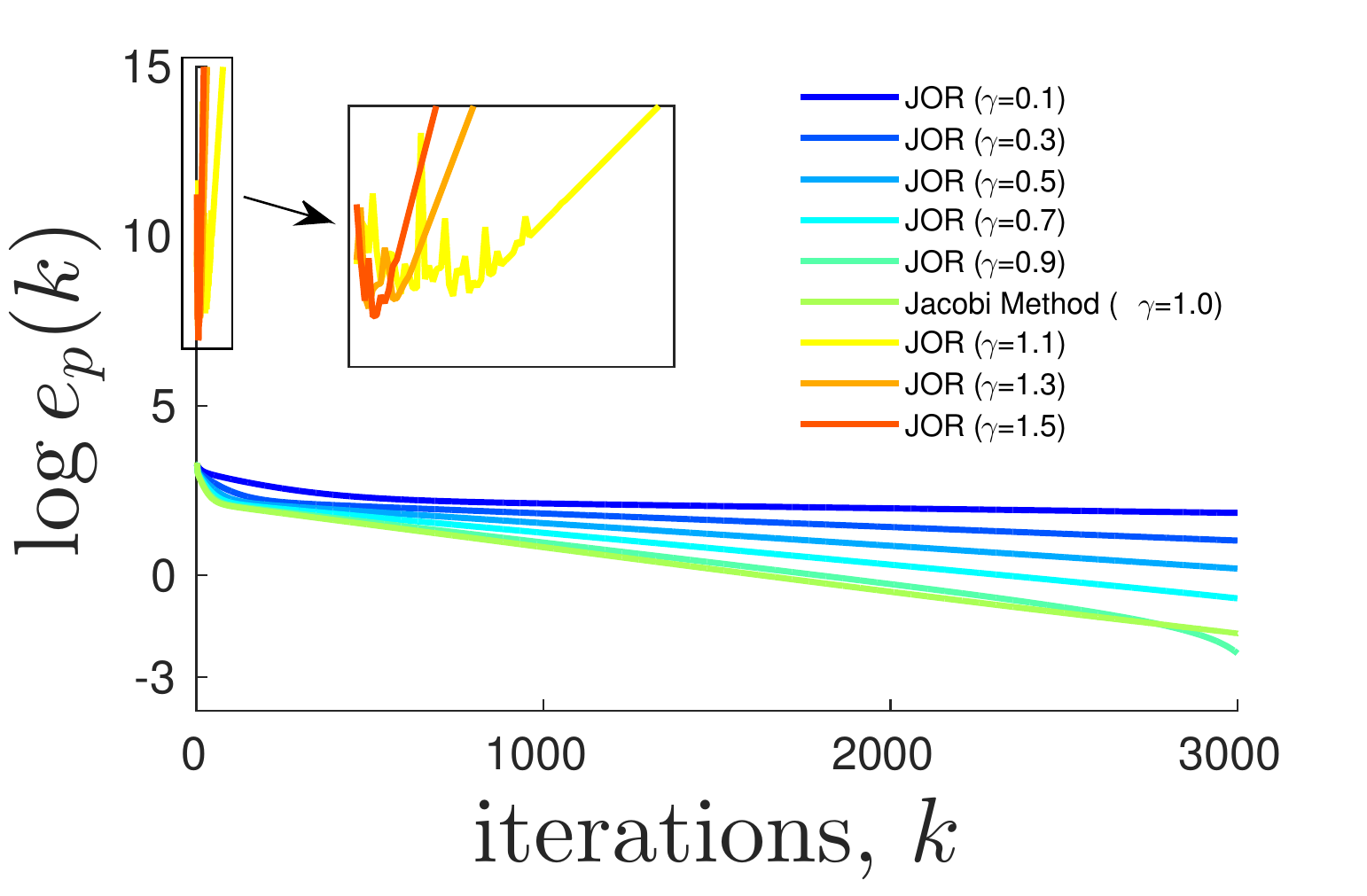} 
\end{minipage}
\vspace{0.1cm}
\\
\hspace{0.2cm}(a) Rotation Estimation  & (b) Pose Estimation
\\
\end{tabular}
\caption{\label{fig:jorConvergenceVSgamma} 
\JOR: convergence of (a) rotation estimation and (b) pose estimation
for different values of $\gamma$ (grid scenario, 49 robots).
In the case of pose estimation, the gap between the initial values
of $\gamma > 1$ and $\gamma \le 1$ is due to the bad initialization provided by the rotation estimation for $\gamma > 1$. 
}
\end{minipage}
\end{figure}
\definecolor{dgreen}{rgb}{0,0.5,0}
\begin{figure}[t]
\centering
\begin{minipage}{\columnwidth}
\hspace{-5mm}
\begin{tabular}{cc}%
\begin{minipage}{0.5\columnwidth}%
\centering%
\includegraphics[scale=0.30, trim=0cm 0cm 0cm 0cm,clip]{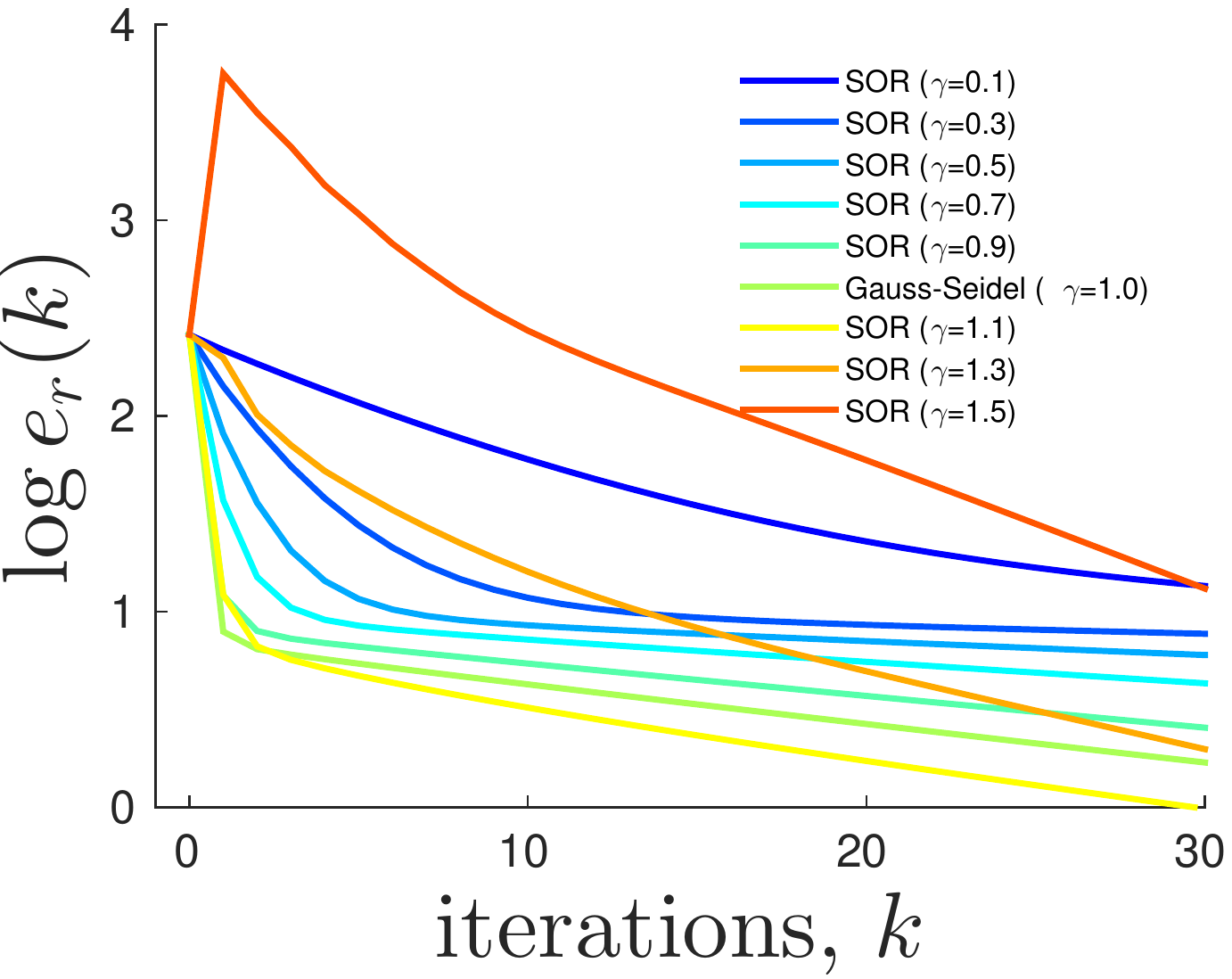} 
\end{minipage}
&\hspace{-3mm}
\begin{minipage}{0.5\columnwidth}%
\centering%
\includegraphics[scale=0.30, trim=0cm 0cm 0cm 0cm,clip]{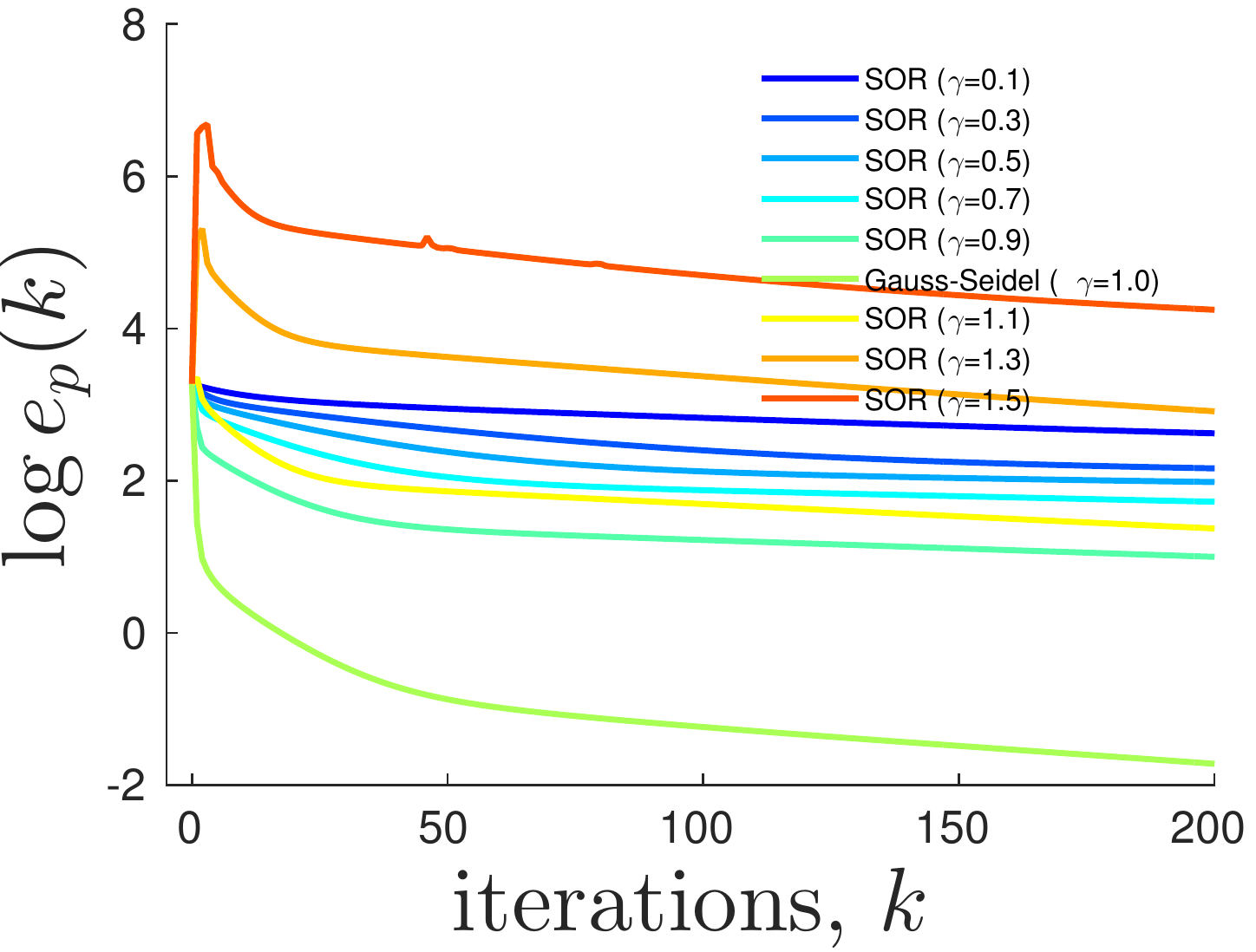} 
\end{minipage}
\\
\hspace{0.2cm}(a) Rotation Error  & (b) Pose Error
\\
\end{tabular}
\caption{\label{fig:sorConvergenceVSgamma} \SOR: convergence of (a) rotation estimation and (b) pose estimation
for different values of $\gamma$ (grid scenario, 49 robots).}
\end{minipage}
\end{figure}

\myparagraph{Comparisons among the distributed algorithms}
In this section we consider the scenario with 49 robots.
We start by studying the convergence properties of the JOR and SOR algorithms in 
isolation. Then we compare the two algorithms in terms of convergence speed.
\Fig\ref{fig:jorConvergenceVSgamma} shows the rotation and the pose error versus the number of iterations 
for different choices of the parameter $\gamma$ for the \JOR algorithm.
\Fig\ref{fig:jorConvergenceVSgamma}\suba confirms the result of Proposition~\ref{prop:convergenceJOR}: 
 \JOR applied to the rotation subproblem converges as long as $\gamma \leq 1$. 
 \Fig\ref{fig:jorConvergenceVSgamma}\suba shows that for any  $\gamma > 1$ the estimate diverges, 
 while the critical value $\gamma = 1$ (corresponding to the \DJ method) ensures the fastest convergence rate.
\Fig\ref{fig:sorConvergenceVSgamma} shows the rotation and the pose error versus the number of iterations 
for different choices of the parameter $\gamma \in (0,2)$ for the \SOR algorithm.
The figure confirms the result of Proposition~\ref{prop:convergenceSOR}: the \SOR algorithm converges 
for any choice of $\gamma \in (0,2)$. 
\Fig\ref{fig:sorConvergenceVSgamma}\suba shows that choices of $\gamma$ close to 1 ensure fast 
convergence rates, while \Fig\ref{fig:sorConvergenceVSgamma}\subb established 
$\gamma = 1$  (corresponding to the \DGS method) as the parameter selection with 
faster convergence.
In summary, both \JOR and \SOR have top performance when $\gamma = 1$.
Later in this section we show that $\gamma = 1$ is the best choice independently 
on the number of robots and the measurement noise.

\definecolor{dgreen}{rgb}{0,0.5,0}
\begin{figure}[t]
\centering
\begin{minipage}{\columnwidth}
\hspace{-5mm}
\begin{tabular}{cc}%
\begin{minipage}{0.5\columnwidth}%
\centering%
\includegraphics[scale=0.26, trim=0cm 0cm 0cm 0cm,clip]{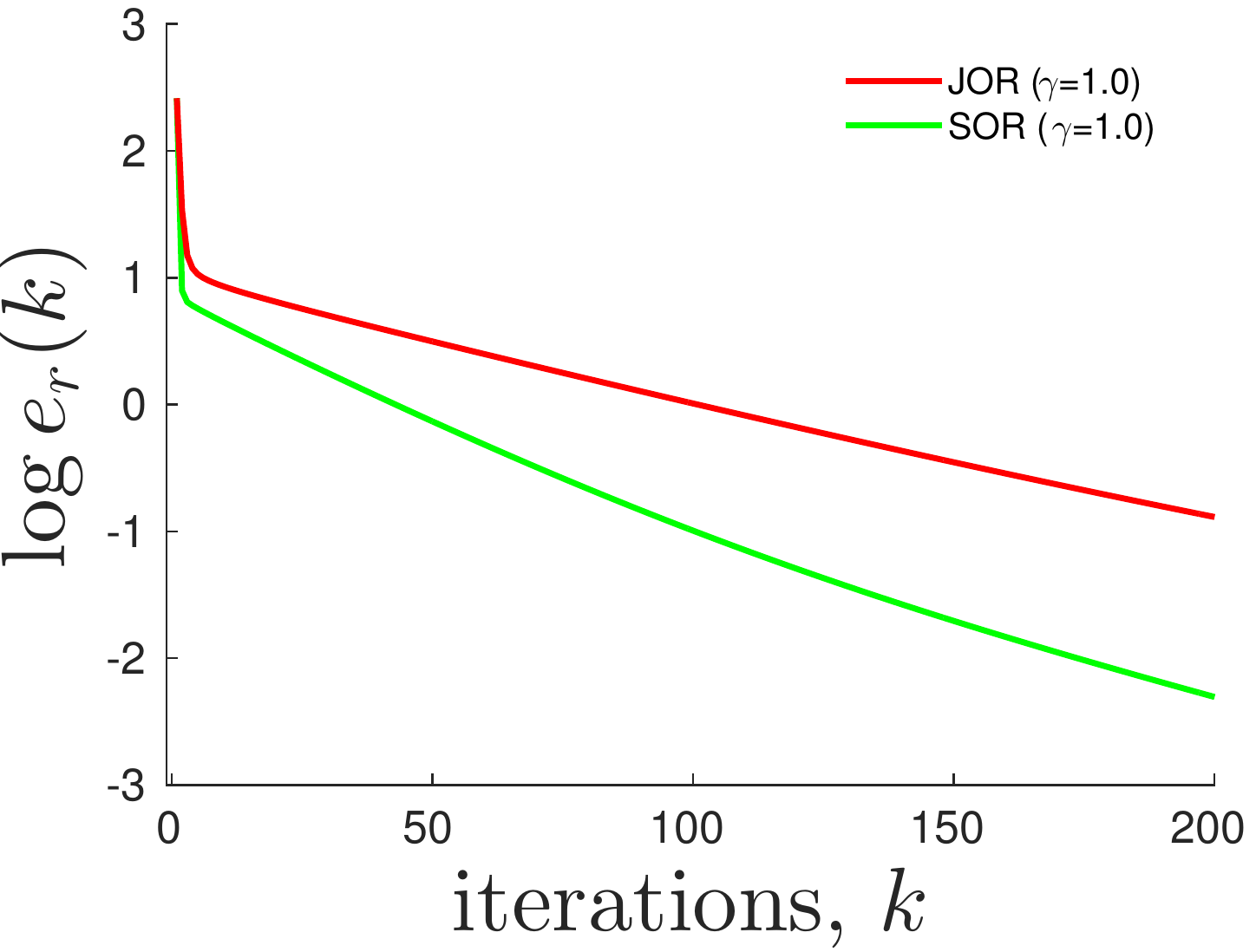} 
\end{minipage}
&\hspace{-2mm}
\begin{minipage}{0.5\columnwidth}%
\centering%
\includegraphics[scale=0.26, trim=0cm 0cm 0cm 0cm,clip]{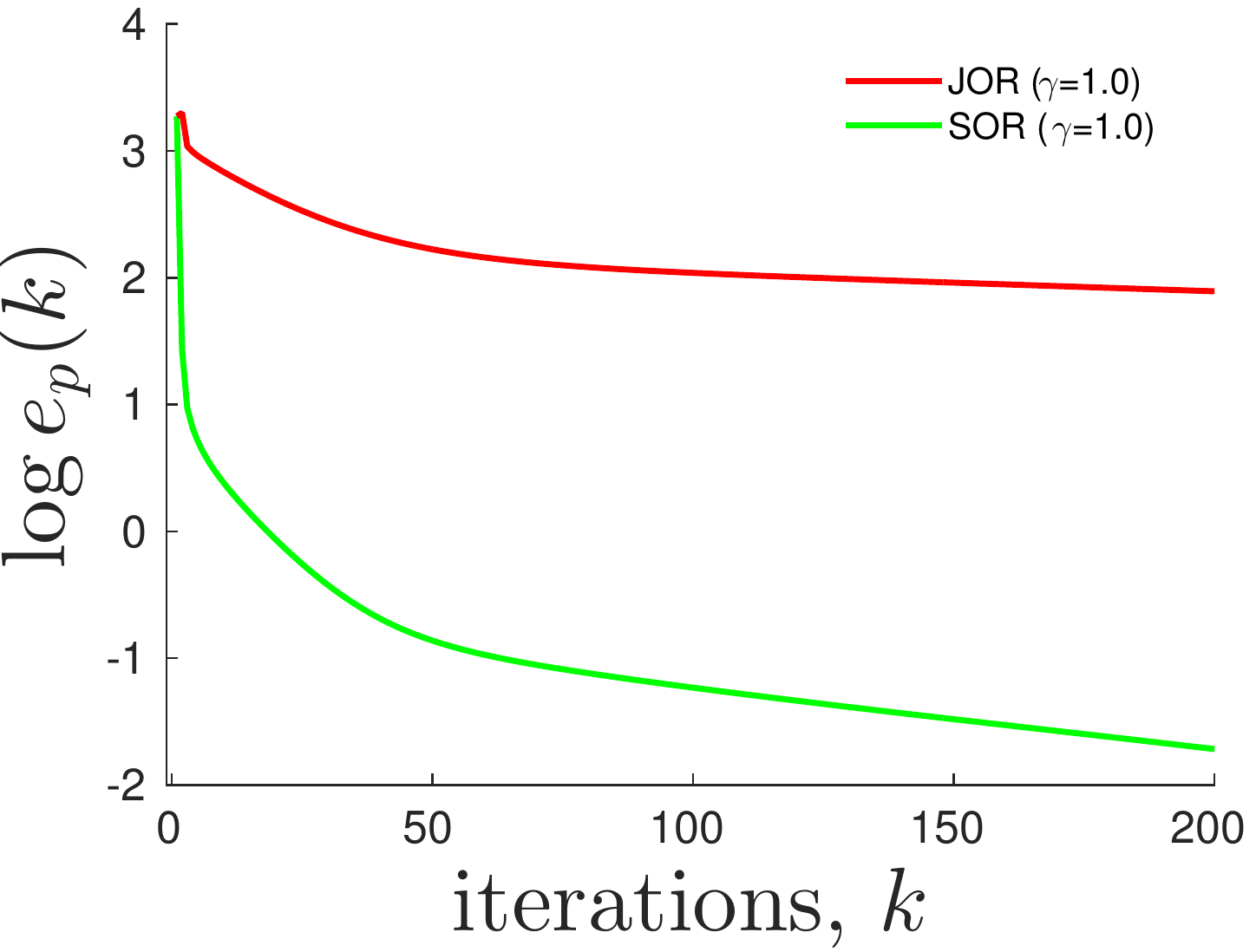} 
\end{minipage}

\\
\hspace{0.2cm}(a) Rotation Error & (b) Pose Error
\\
\end{tabular}
\caption{\label{fig:sorVSjor} \JOR \vs \SOR: 
convergence of (a) rotation estimation and (b) pose estimation for 
the \JOR and \SOR algorithms with $\gamma = 1$ (grid scenario, 49 robots).
}
\end{minipage}
\end{figure}

\begin{figure}[t]
\centering
\begin{minipage}{\columnwidth}
\hspace{-5mm}
\begin{tabular}{cc}%
\begin{minipage}{0.5\columnwidth}%
\centering%
\includegraphics[scale=0.26, trim=0cm 0cm 0cm 0cm,clip]{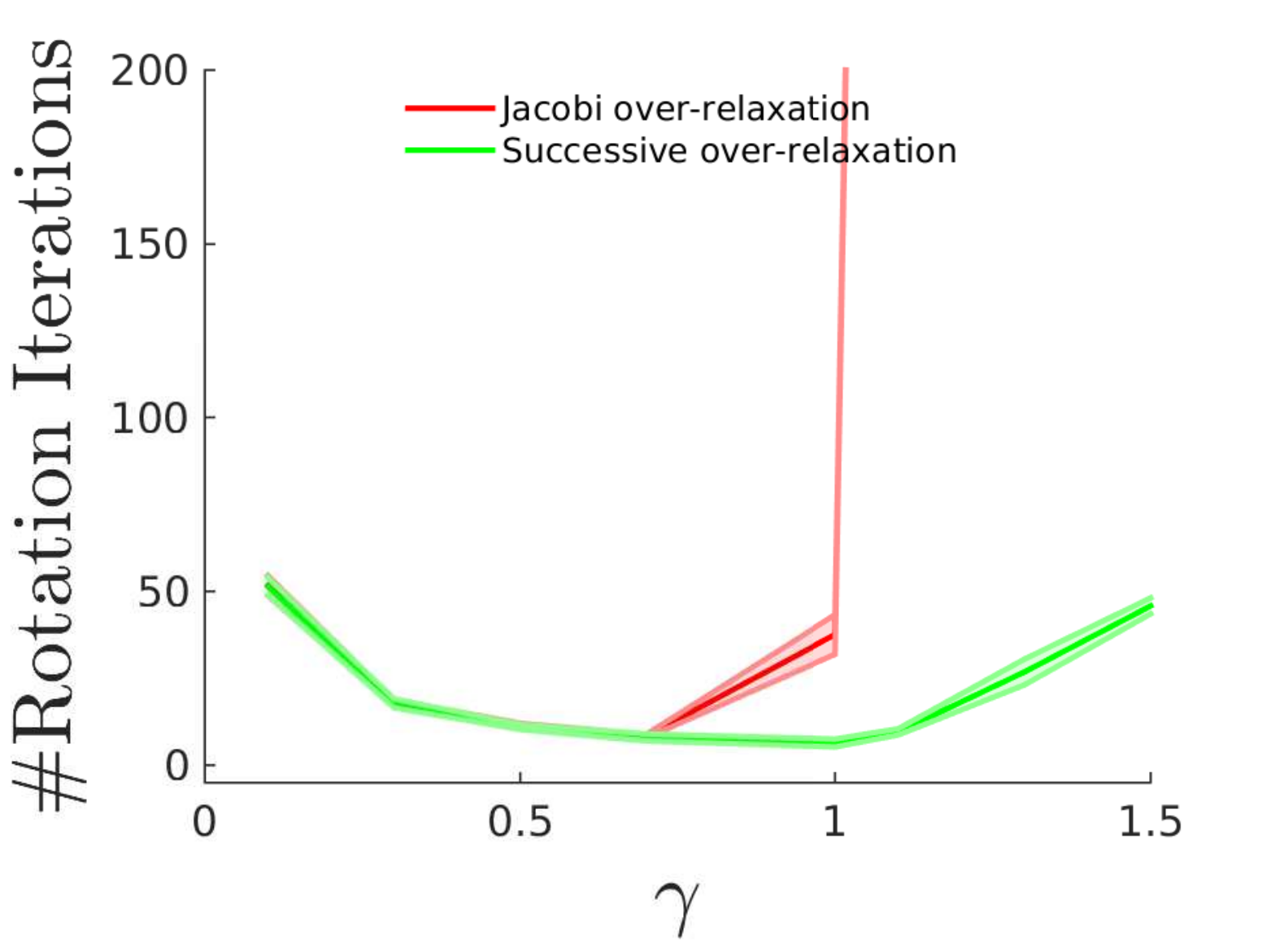} 
\end{minipage}
&\hspace{-1mm}
\begin{minipage}{0.5\columnwidth}%
\centering%
\includegraphics[scale=0.26, trim=0cm 0cm 0cm 0cm,clip]{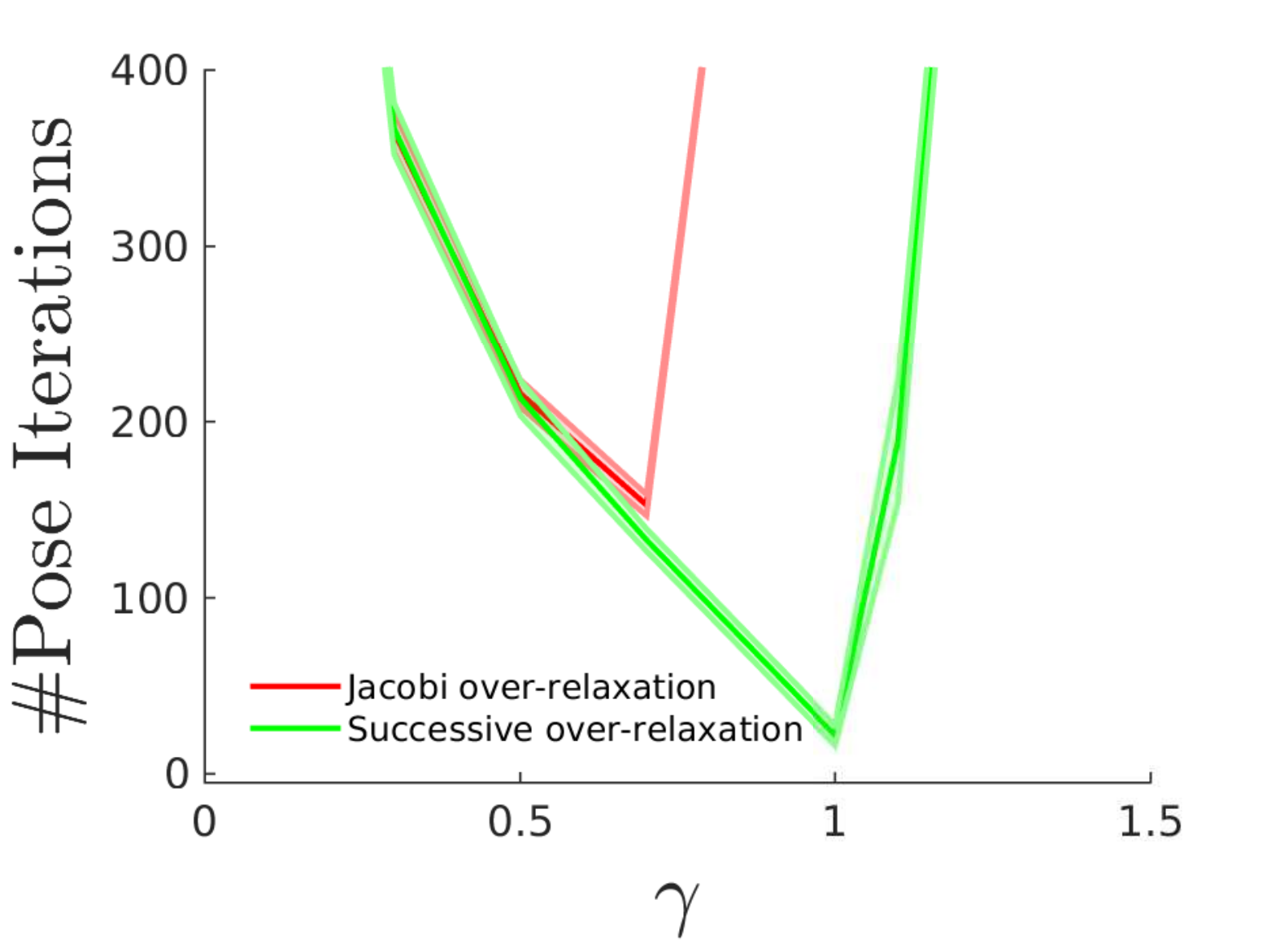} 
\end{minipage}

\\
\hspace{0.2cm}(a) Rotation Estimation & (b) Pose Estimation 
\\
\end{tabular}
\caption{\label{fig:sorVSjorIterations} \JOR \vs \SOR: number of iterations 
required for (a) rotation estimation and (b) pose estimation for 
the \JOR and \SOR algorithms with $\gamma = 1$ (grid scenario, 49 robots).
The average number of iterations is shown as a solid line, while the 1-sigma standard deviation 
is shown as a shaded area.
}
\end{minipage}
\end{figure}

Let us now compare \JOR and \SOR in terms of convergence.
\Fig\ref{fig:sorVSjor} compares the convergence rate of \SOR and \JOR 
for both the rotation subproblem (\Fig\ref{fig:sorVSjor}\suba)
and the pose subproblem (\Fig\ref{fig:sorVSjor}\subb).
 We set $\gamma = 1$ in \JOR and \SOR since we already observed that 
 this choice ensures the best performance. 
 The figure confirms that \SOR dominates \JOR in both subproblems.
\Fig\ref{fig:sorVSjorIterations} shows the number of iterations for 
convergence (according to our stopping conditions) 
and for different choices of the parameter $\gamma$. 
Once again, the figure confirms that the \SOR with $\gamma = 1$ is able to 
converge in the smallest number of iterations, requiring only few tens of iterations 
in both the rotation and the pose subproblem.

\definecolor{dgreen}{rgb}{0,0.5,0}
\begin{figure}[t]
\centering
\begin{minipage}{\columnwidth}
\hspace{-5mm}
\begin{tabular}{cc}%
\begin{minipage}{0.5\columnwidth}%
\centering%
\includegraphics[scale=0.28, trim=0cm 0cm 0cm 0cm,clip]{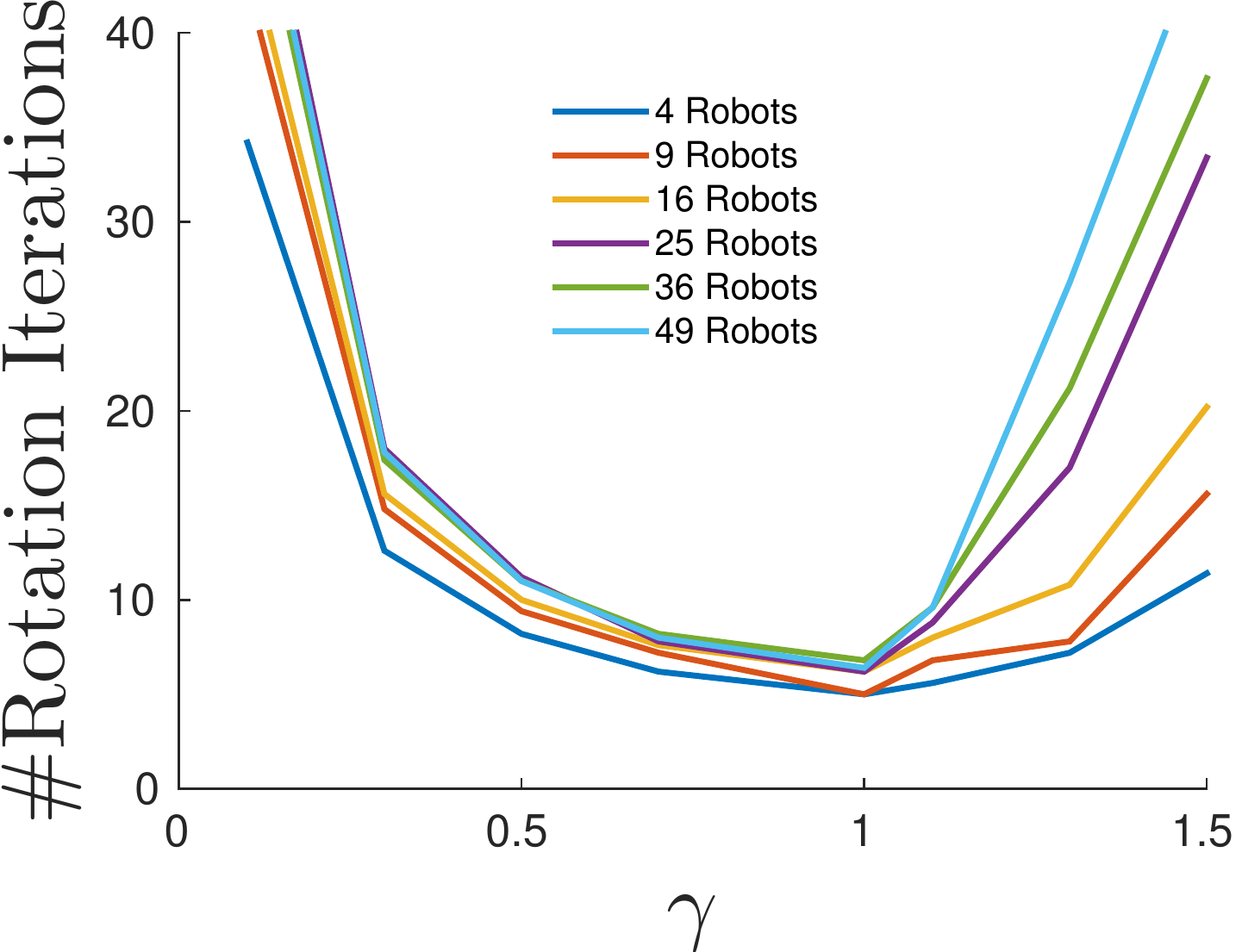} 
\end{minipage}
&\hspace{-3mm}
\begin{minipage}{0.5\columnwidth}%
\centering%
\includegraphics[scale=0.28, trim=0cm 0cm 0cm 0cm,clip]{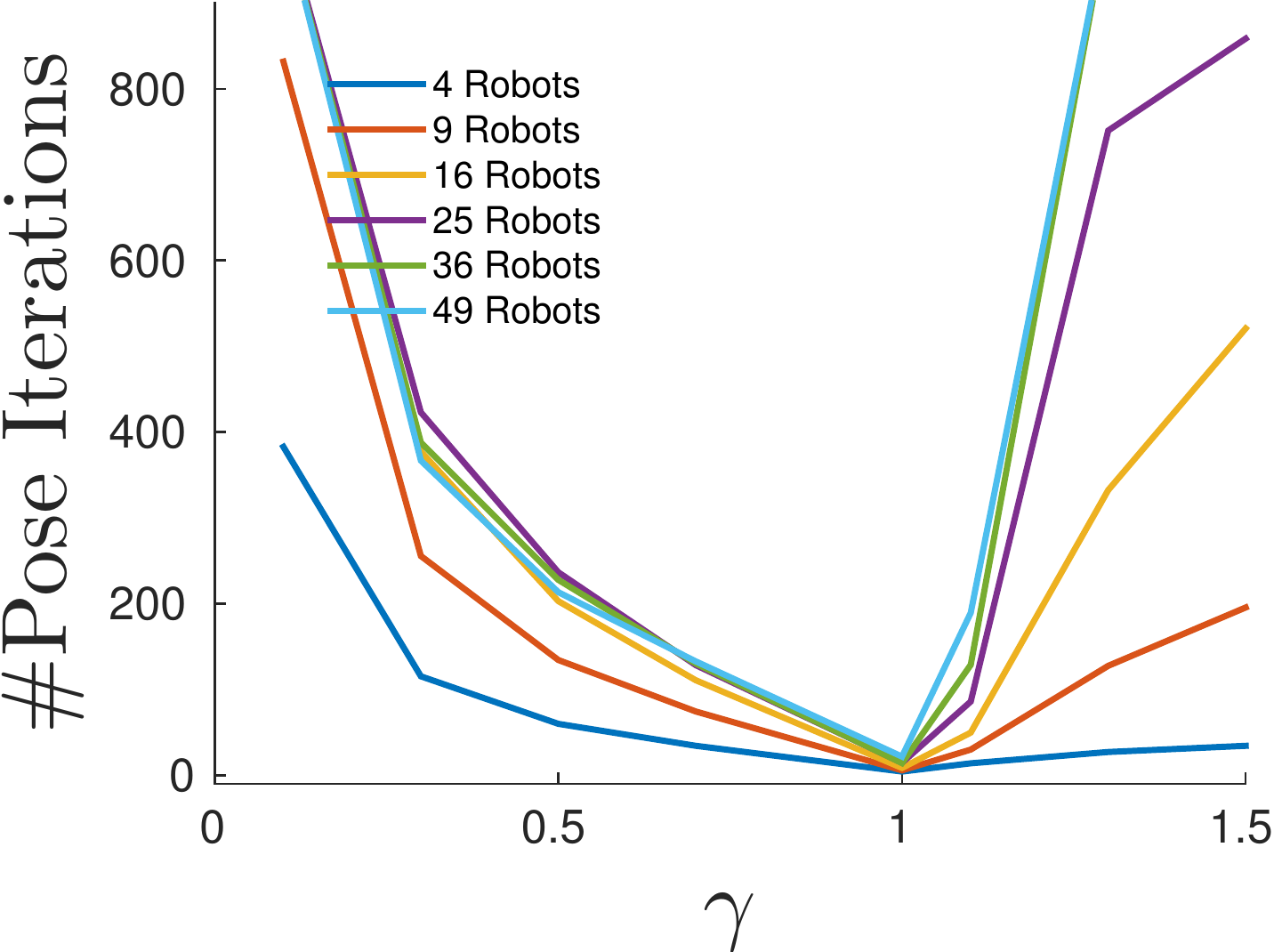} 
\end{minipage}
\\
\hspace{0.2cm}(a) Rotation Error  & (b) Pose Error
\\
\end{tabular}
\caption{\label{fig:sorVariantsNumRobots} 
\SOR: number of iterations required for (a) rotation estimation and (b) pose estimation in the SOR algorithm
for different choices of $\gamma$ and increasing number of robots.
}
\end{minipage}
\end{figure}
\definecolor{dgreen}{rgb}{0,0.5,0}
\begin{figure}[t]
\centering
\begin{minipage}{\columnwidth}
\hspace{-5mm}
\begin{tabular}{cc}%
\begin{minipage}{0.5\columnwidth}%
\centering%
\includegraphics[scale=0.28, trim=0cm 0cm 0cm 0cm,clip]{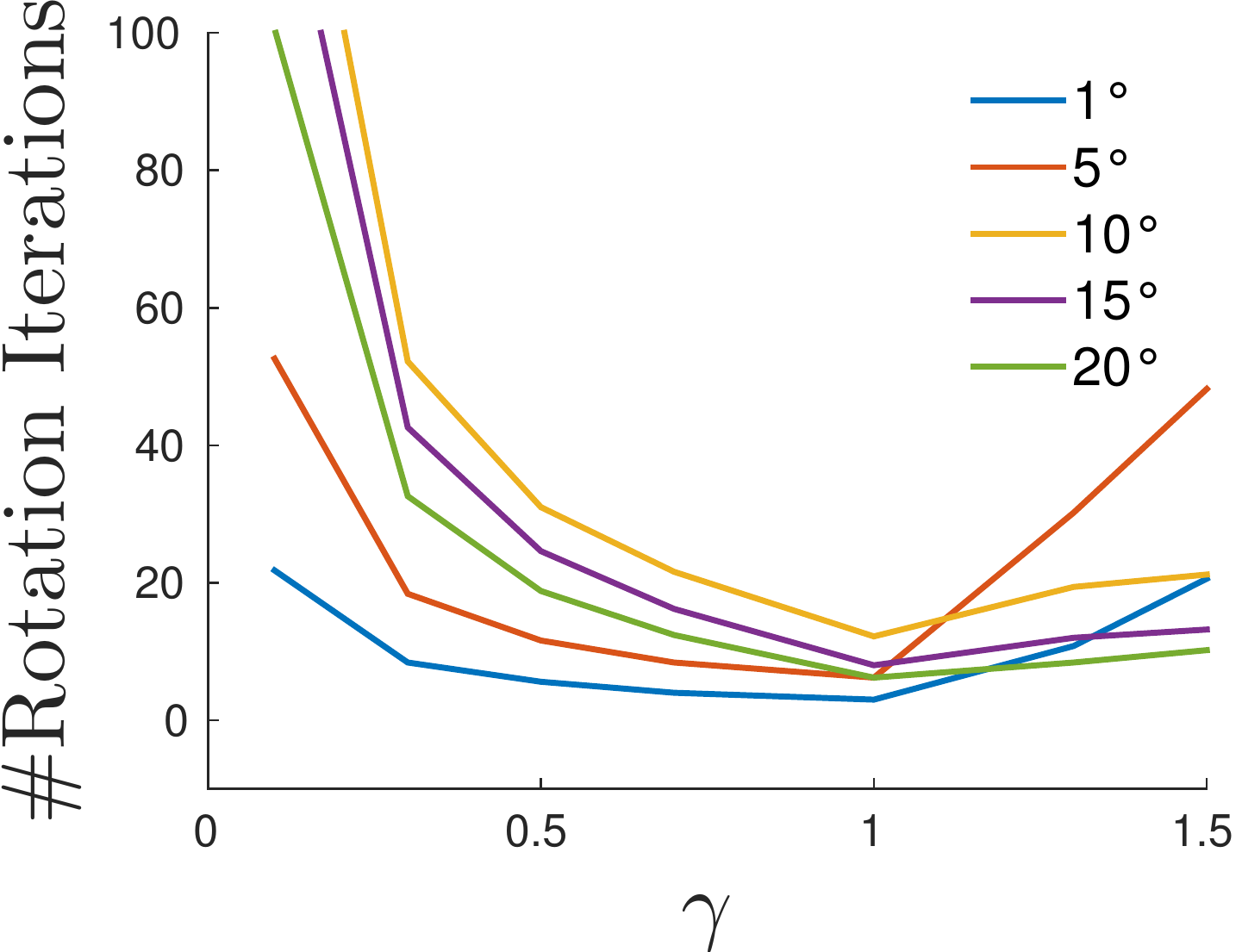} 
\end{minipage}
&\hspace{-2mm}
\begin{minipage}{0.5\columnwidth}%
\centering%
\includegraphics[scale=0.28, trim=0cm 0cm 0cm 0cm,clip]{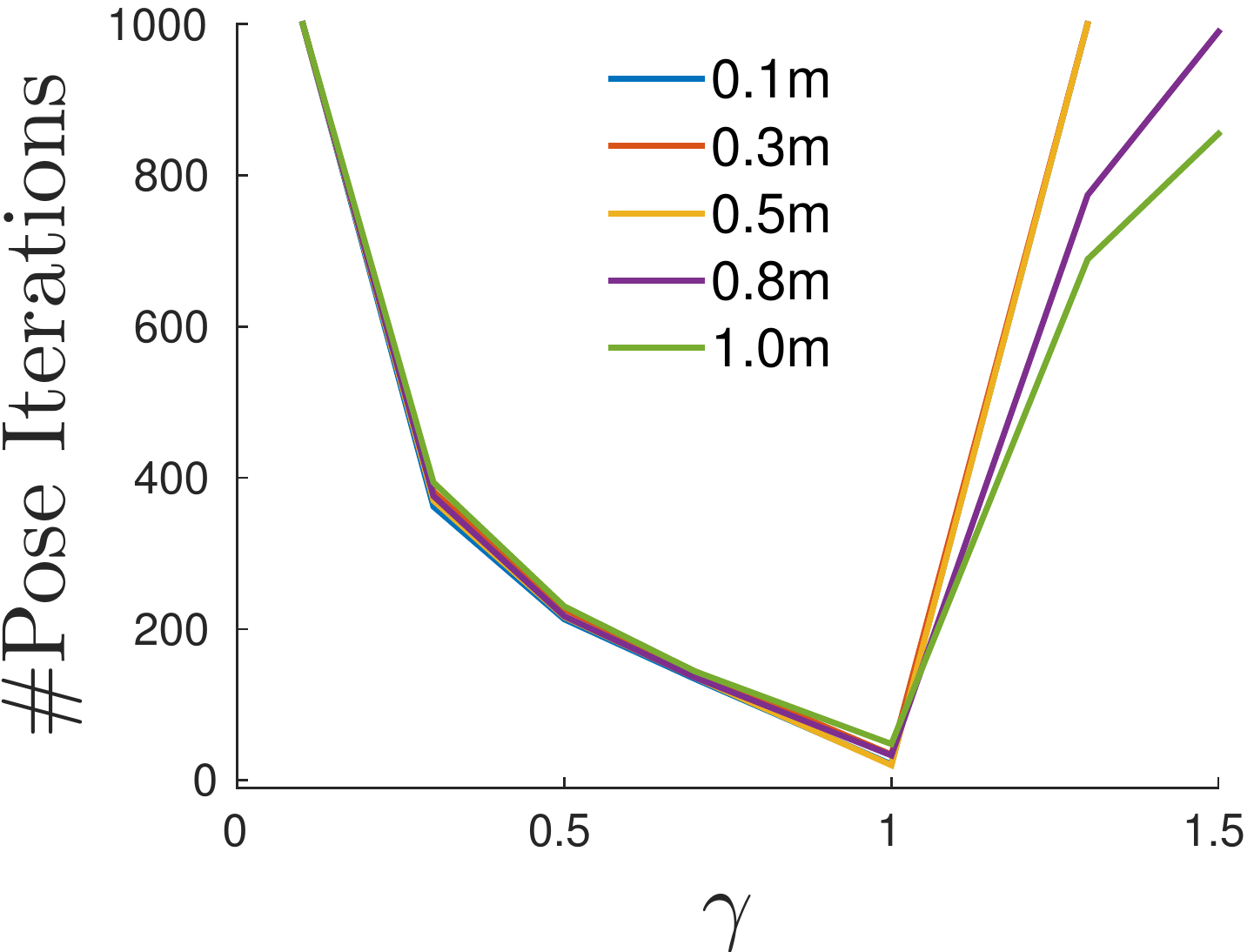} 
\end{minipage}
\\
\hspace{0.2cm}(a) Rotation Error  & (b) Pose Error
\\
\end{tabular}
\caption{\label{fig:sorVariantsSeparatedNoise} 
\SOR: number of iterations required for (a) rotation estimation and (b) pose estimation in the SOR algorithm
for different choices of $\gamma$ and increasing measurement noise.  
}
\end{minipage}
\end{figure}

We conclude this section by showing that setting $\gamma = 1$ in \SOR ensure faster
convergence regardless the number of robots and the measurement noise.
\Fig\ref{fig:sorVariantsNumRobots} compares
the number of iterations required to converge for increasing number of robots for varying $\gamma$ values. Similarly \Fig\ref{fig:sorVariantsSeparatedNoise} compares
the number of iterations required to converge for increasing noise for varying $\gamma$ value. We can see that in both the cases $\gamma=1$ has the fastest convergence (required the least number of iterations) 
irrespective of the number of robots and measurement noise. 
Since \SOR with $\gamma = 1$, i.e., the \DGS method, is the 
top performer in all test conditions, in the rest of the paper we restrict our analysis to this algorithm.

\definecolor{dgreen}{rgb}{0,0.5,0}
\begin{figure}[t]
\centering
\begin{minipage}{\columnwidth}
\hspace{-5mm}
\begin{tabular}{cc}%
\begin{minipage}{0.5\columnwidth}%
\centering%
\includegraphics[scale=0.28, trim=0cm 0cm 0cm 0cm,clip]{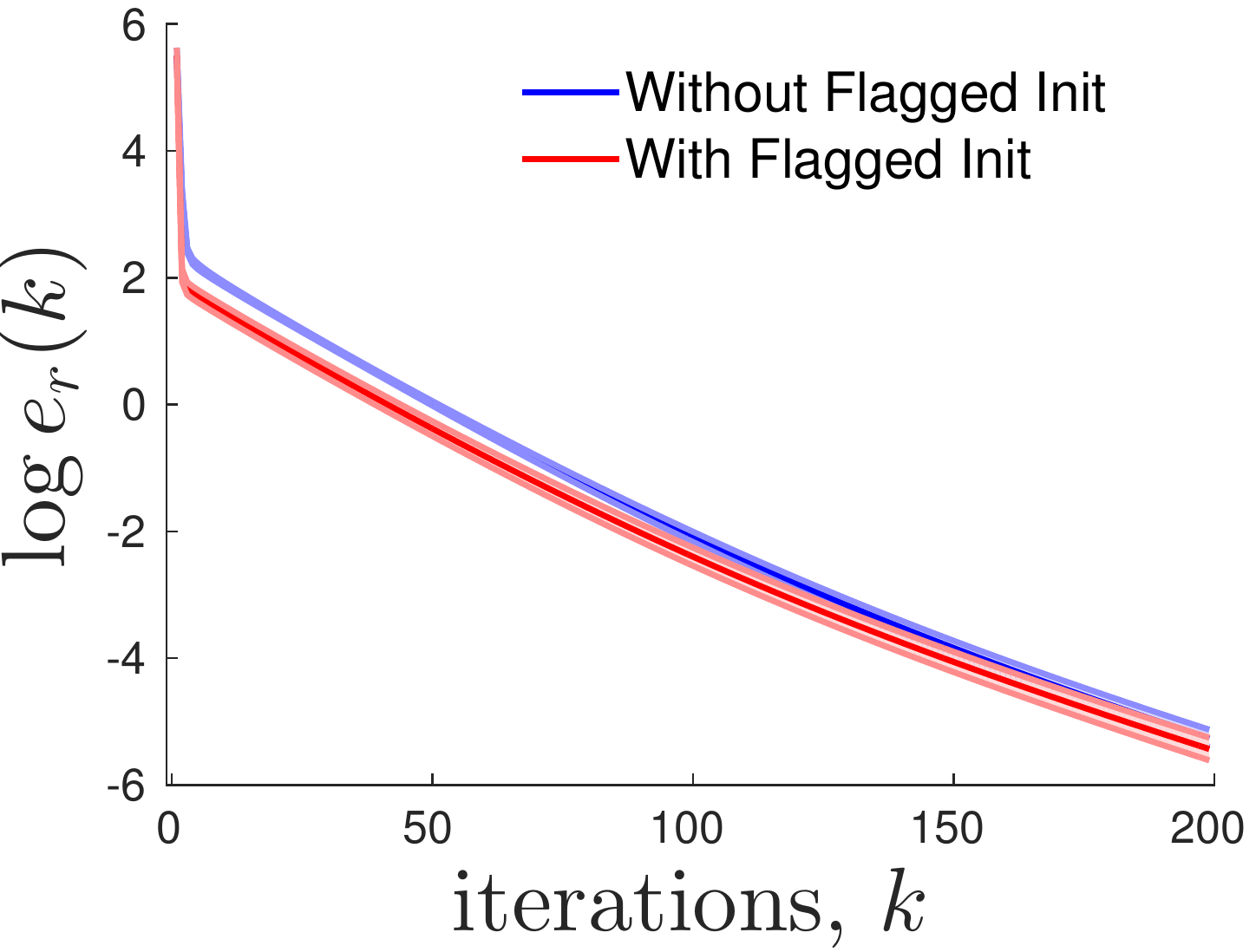} 
\end{minipage}
&\hspace{-3mm}
\begin{minipage}{0.5\columnwidth}%
\centering%
\includegraphics[scale=0.28, trim=0cm 0cm 0cm 0cm,clip]{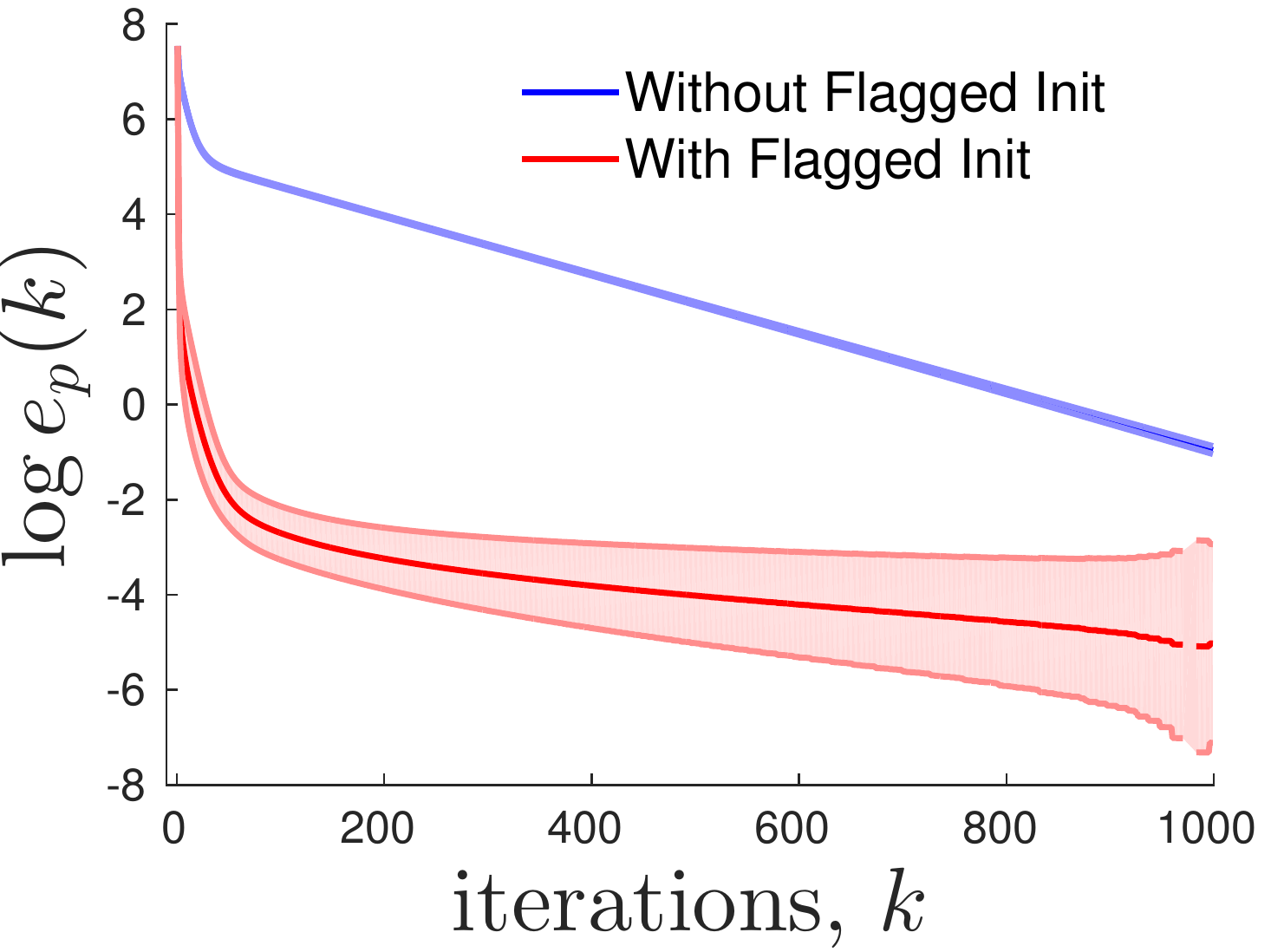} 
\end{minipage}
\vspace{0.1cm}
\\
\hspace{0.2cm}(a) Rotation  Error & (b) Pose  Error
\\
\end{tabular}
\caption{\label{fig:flaggedInit} \DGS: Comparison between flagged and non-flagged initialization 
on the grid scenario with 49 robots. 
Average estimation errors (solid line) and 1-sigma standard deviation (shaded area) are in log scale.
}
\end{minipage}
\vspace{-0.5cm}
\end{figure}

\myparagraph{Flagged initialization} In this paragraph we discuss the advantages of the flagged initialization.
We compare the \DGS method with flagged initialization against a naive initialization in which the variables 
($\vr\at{0}$ and $\vp\at{0}$, respectively) are initialized to zero. The results, for the dataset with 
49 robots, are shown in Fig.~\ref{fig:flaggedInit}.  
In both cases the estimation errors go to zero, but the convergence is faster when using the flagged initialization. 
 The speed-up is significant for the second linear system (Fig.~\ref{fig:flaggedInit}b).
We noticed a similar advantage across all tested scenarios.
Therefore, in the rest of the paper we always adopt the flagged initialization.

\definecolor{dgreen}{rgb}{0,0.5,0}
\begin{figure}[t]
\centering
\begin{minipage}{\columnwidth}
\begin{tabular}{cc}%
\begin{minipage}{0.45\columnwidth}%
\centering%
\includegraphics[width=\columnwidth, trim=0cm 0cm 0cm 0cm,clip]{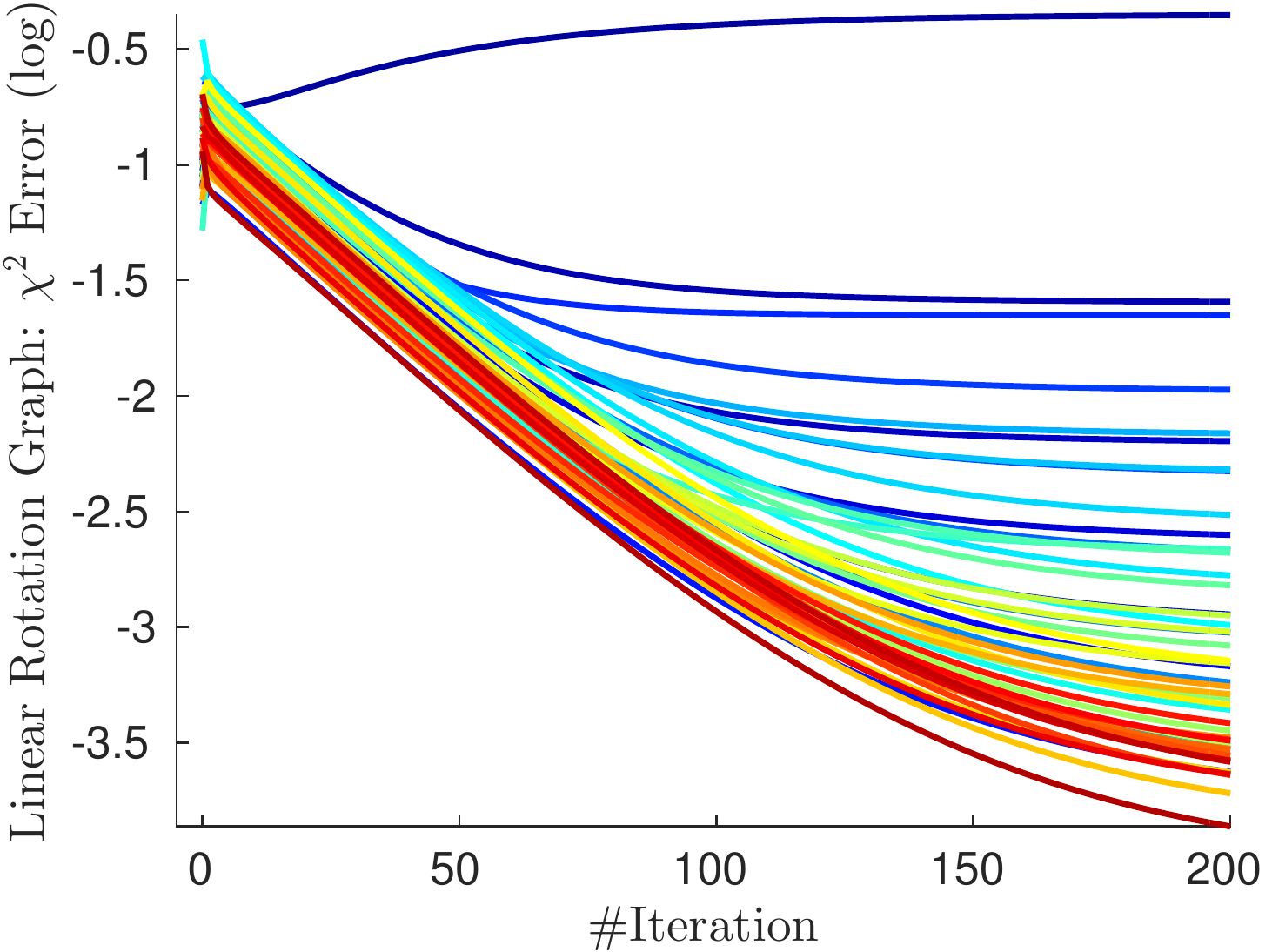} 
\end{minipage}
&
\begin{minipage}{0.45\columnwidth}%
\centering%
\includegraphics[width=\columnwidth, trim=0cm 0cm 0cm 0cm,clip]{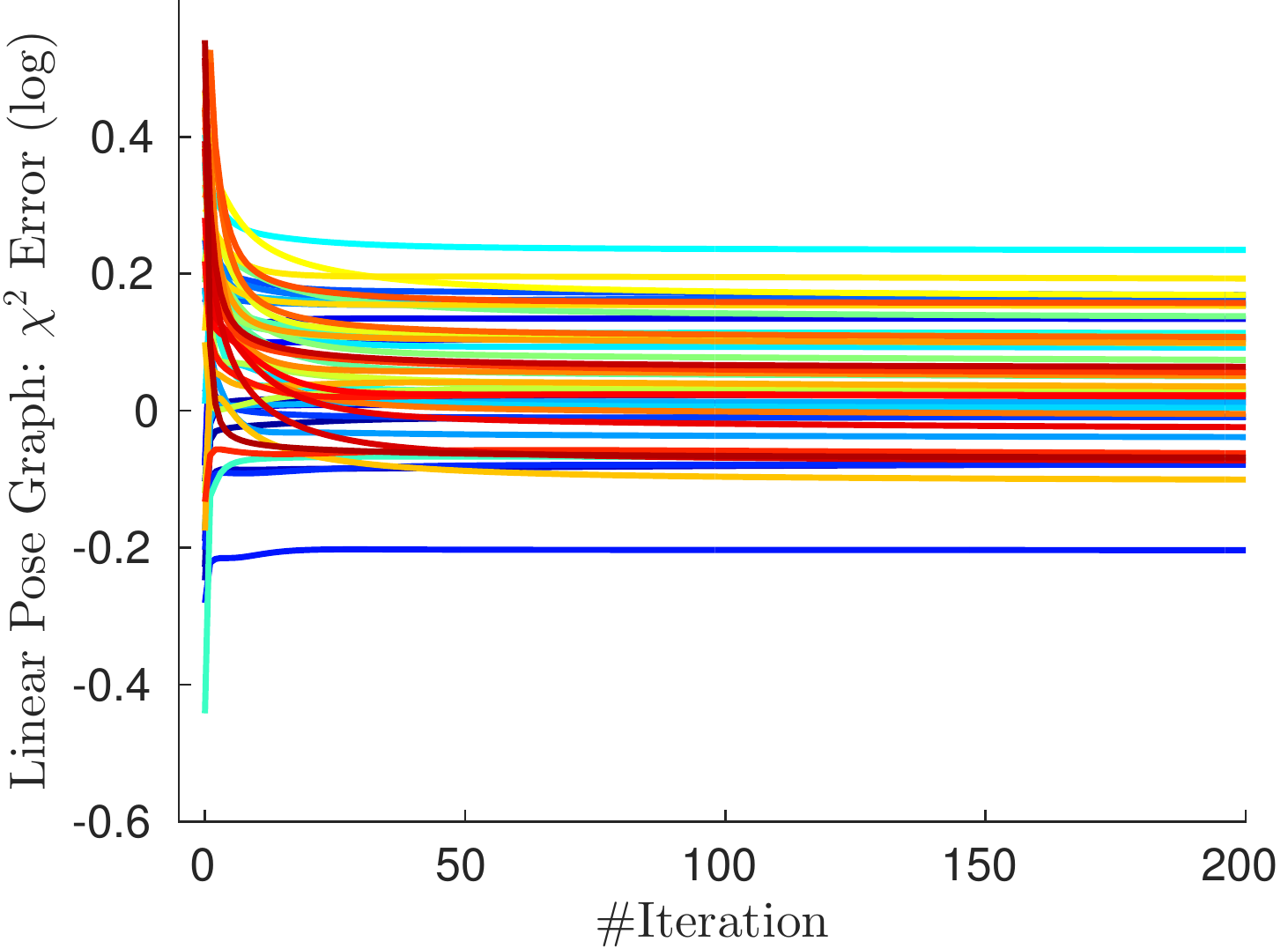} 
\end{minipage}
\\
(a) Rotation Error & (b) Pose Error
\\
\begin{minipage}{0.45\columnwidth}%
\centering%
\includegraphics[width=\columnwidth, trim=0cm 0cm 0cm 0cm,clip]{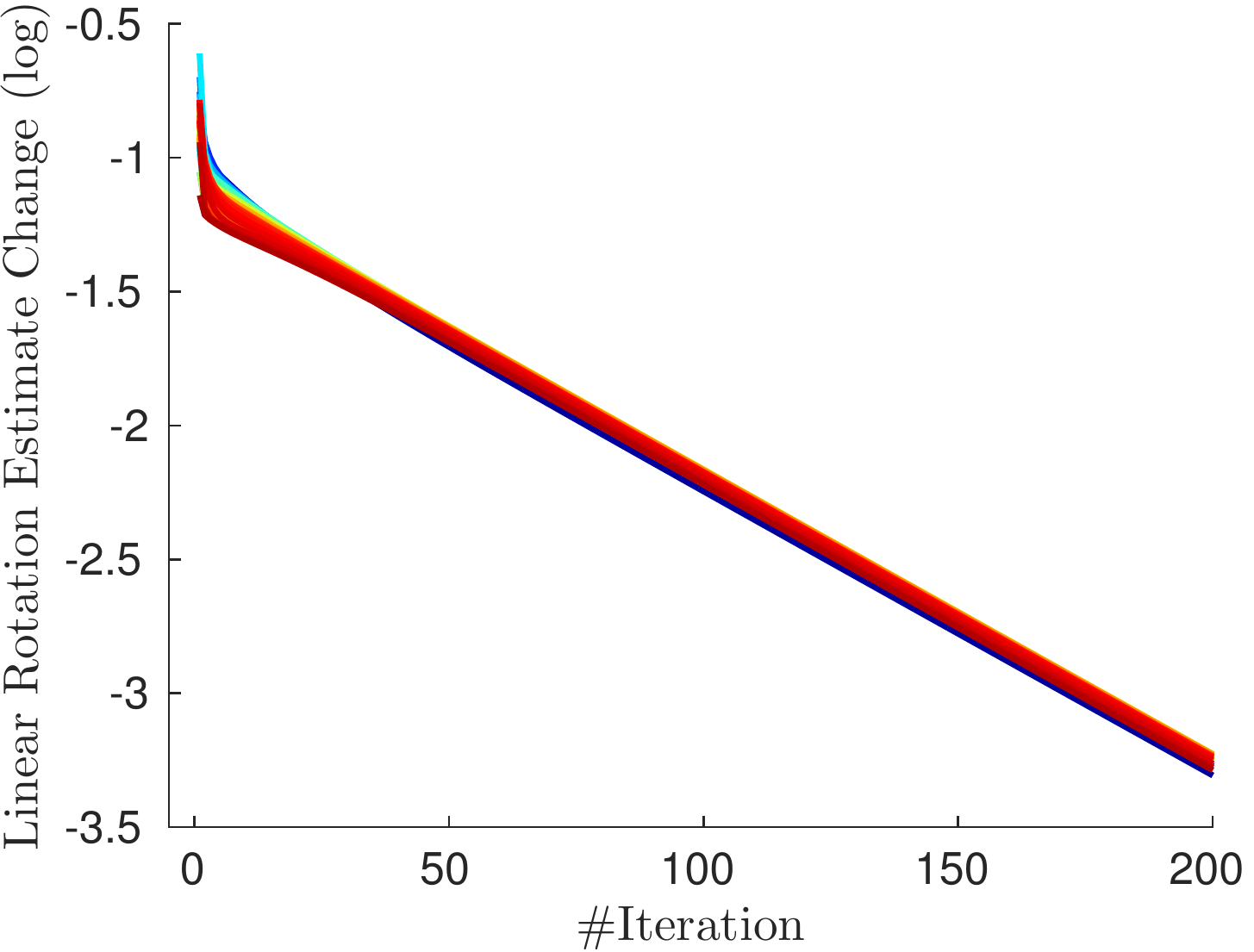} 
\end{minipage}
&
\begin{minipage}{0.45\columnwidth}%
\centering%
\includegraphics[width=\columnwidth, trim=0cm 0cm 0cm 0cm,clip]{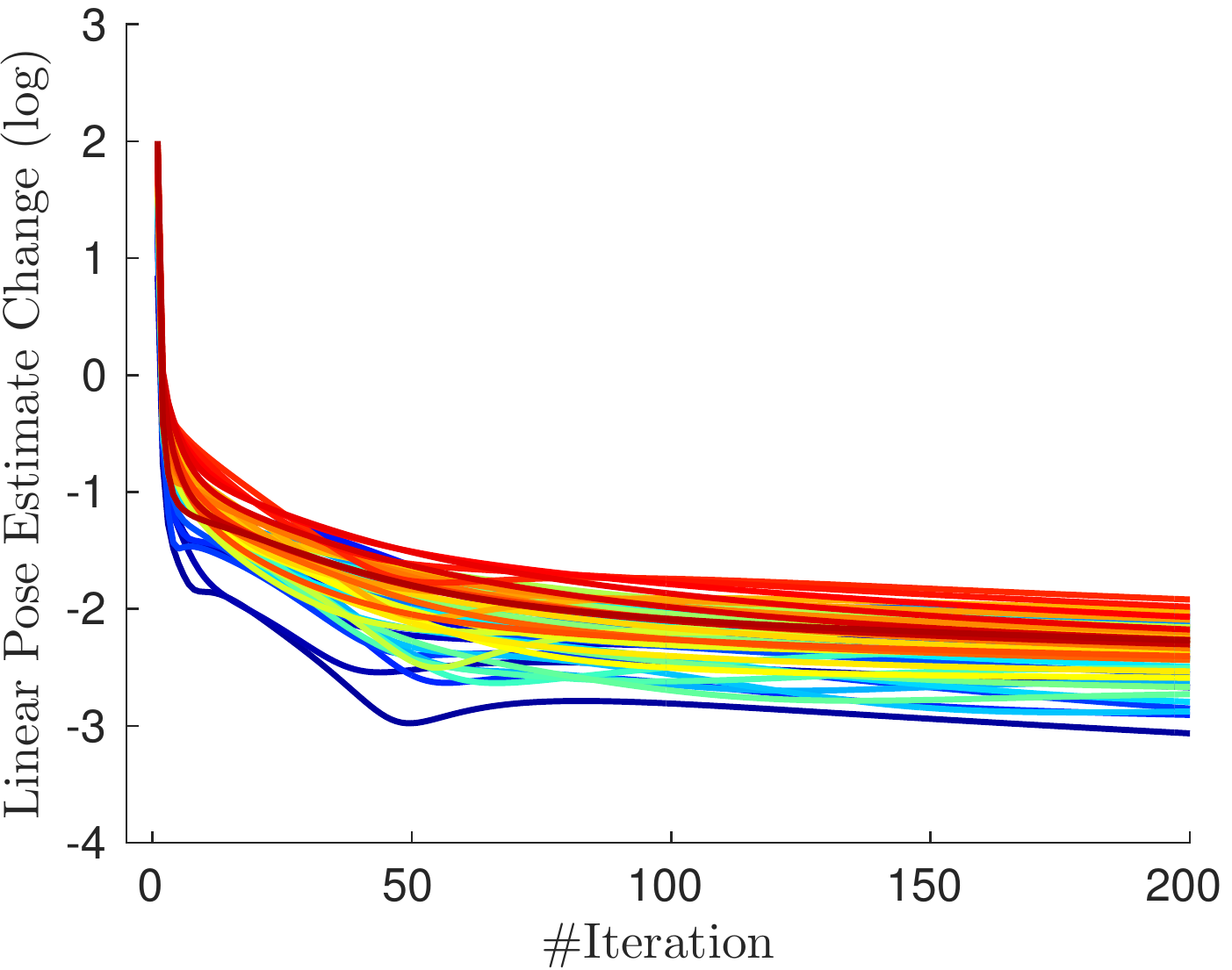} 
\end{minipage}
\\
(c) Rotation Estimate Change & (d) Pose Estimate Change
\\
\end{tabular}%
\caption{\label{fig:convergence} \DGS:  convergence statistics of rotation estimation and pose estimation for each robot (49 Robots). Robots are represented by different color lines.
}
\end{minipage}%
\end{figure}

\myparagraph{Stopping conditions and anytime flavor} 
This section provides extra insights on the convergence of the  \DGS method.
\Fig\ref{fig:convergence}\suba-\subb show the evolution of  the rotation and pose error for \emph{each} robot in the 49-robot grid: 
the error associated to each robot (i.e., to each subgraph corresponding to a robot trajectory) 
is not monotonically decreasing 
and the error for some robot can increase to bring down the overall error. 
 \Fig\ref{fig:convergence}\subc-\subd report the change in the
 rotation and pose estimate for individual robots. 
 Estimate changes become negligible within few tens of iterations. As mentioned at the 
beginning of the section, we stop the \DGS iterations when the estimate change is sufficiently small
(below the thresholds $\eta_r$ and $\eta_p$).

\begin{figure}[h!]
\hspace{-5mm}
\begin{minipage}{\columnwidth}
\begin{tabular}{ccc}%
\begin{minipage}{\widthCol}%
\centering%
\includegraphics[scale=0.22, trim=0cm 0cm 0cm 0cm,clip]{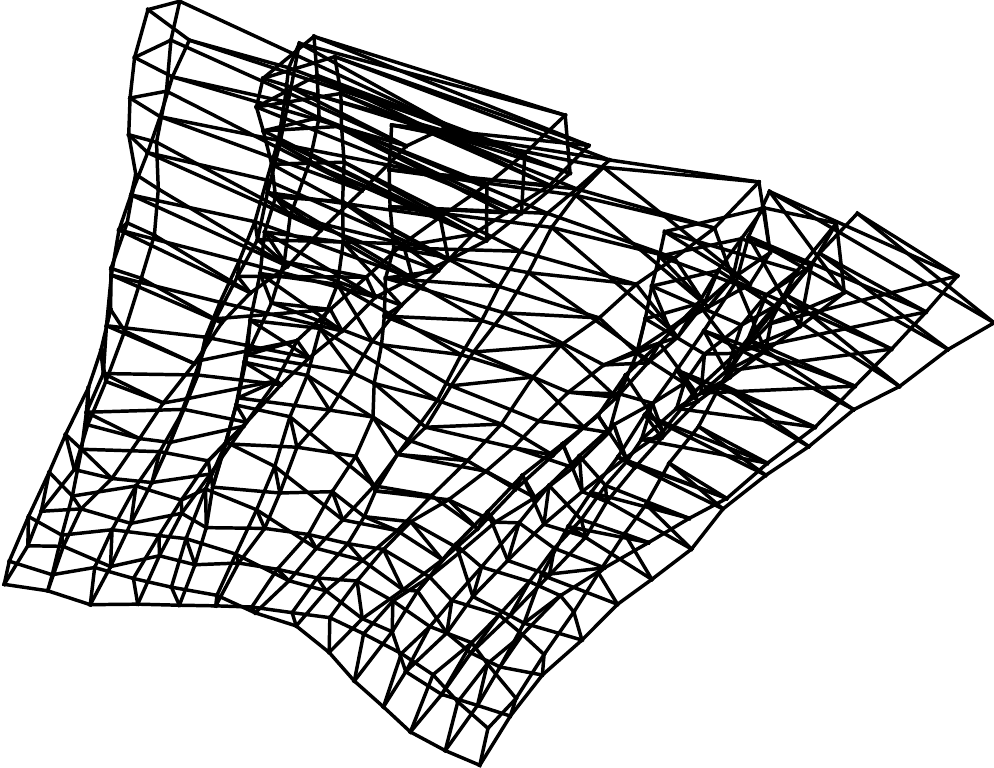} 
\end{minipage}
&
\begin{minipage}{\widthCol}%
\centering%
\includegraphics[scale=0.18, trim=2.8cm 0cm 0cm 1.5cm,clip]{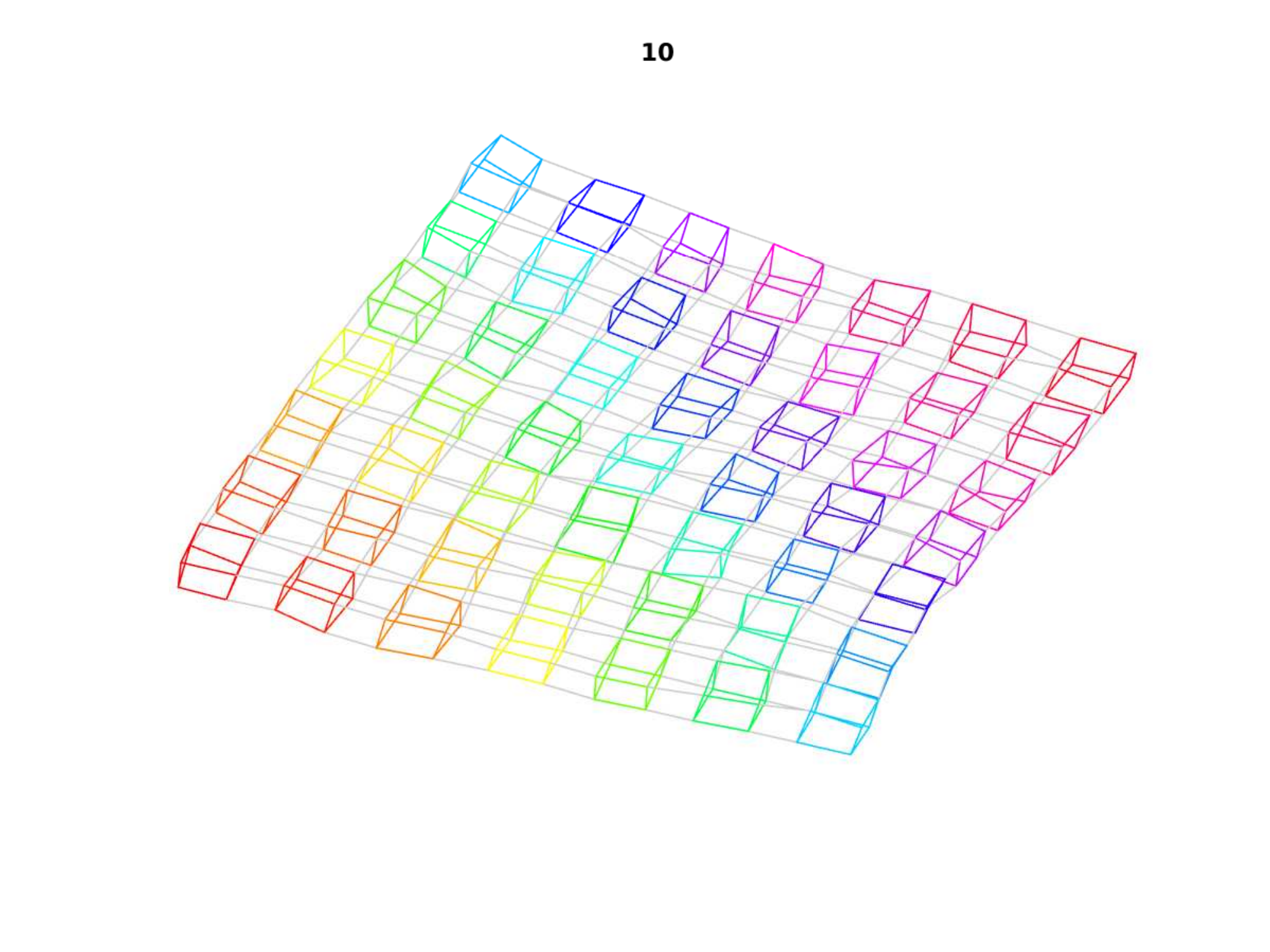} 
\end{minipage}
&
\begin{minipage}{\widthCol}%
\centering%
\includegraphics[scale=0.18, trim=1.5cm 0cm 0cm 1.5cm,clip]{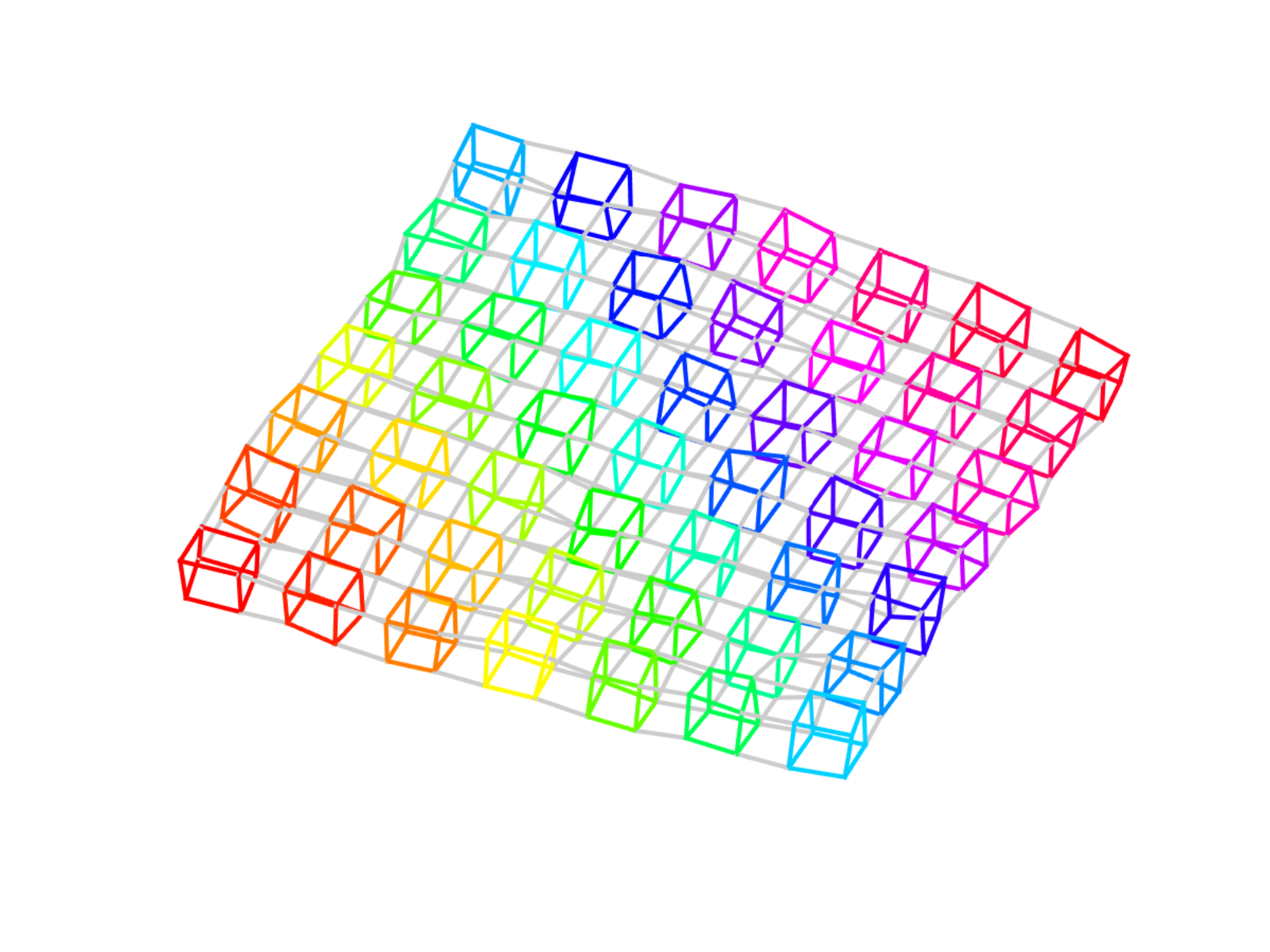} 
\end{minipage} 
\vspace{-0.4cm}
\\
(a) Initial & (b) 10 iterations & (c) 1000 iterations
\\
\end{tabular}
\end{minipage}
\caption{\label{fig:snapshots} \DGS: Trajectory estimates for the scenario with 49 robots. 
(a) Odometric estimate (not used in our approach and only given for visualization purposes), (b)-(c) \DGS estimates after given number of iterations.
}
\end{figure}

\Fig\ref{fig:snapshots} shows the estimated trajectory after 10 and 1000 iterations 
of the \DGS algorithm for the 49-robot grid. The odometric estimate (\Fig\ref{fig:snapshots}\suba) is shown for visualization purposes, while 
it is not used in our algorithm. We can see that 
the estimate after 10 iterations is already visually close to the estimate after 1000 iterations. 
The \DGS algorithm has an any-time flavor: the trajectory estimates are already accurate
after few iterations and asymptotically converge to the centralized estimate.

\definecolor{dgreen}{rgb}{0,0.5,0}

\begin{figure}[t]

\centering
\begin{minipage}{\columnwidth}
\hspace{-4mm}
\begin{tabular}{cc}%
\begin{minipage}{0.48\columnwidth}%
\centering%
\includegraphics[scale=0.28, trim=0cm 0cm 0cm 0cm,clip]{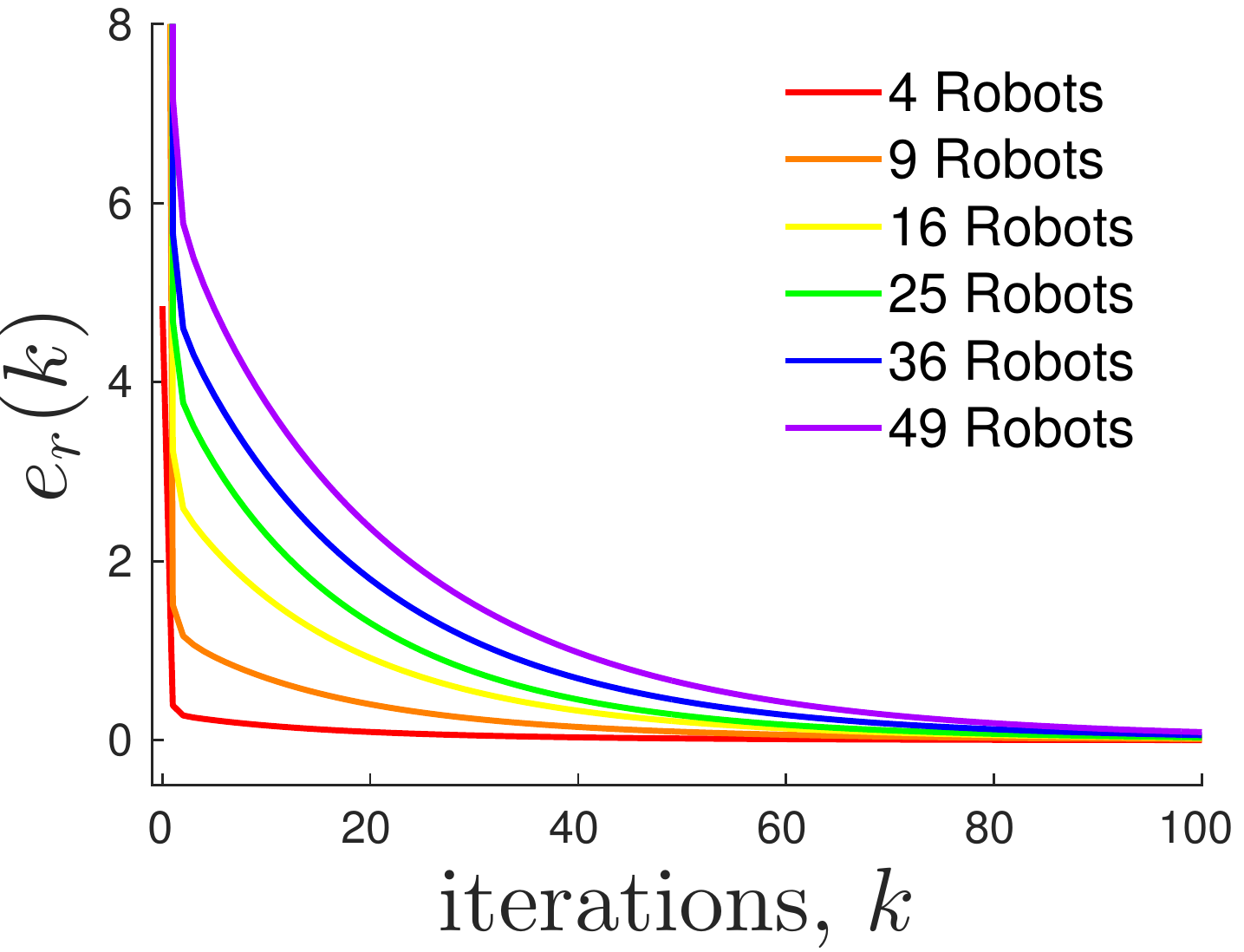} 
\end{minipage}
& \hspace{-4mm}
\begin{minipage}{0.48\columnwidth}%
\centering%
\includegraphics[scale=0.28, trim=0cm 0cm 0cm 0cm,clip]{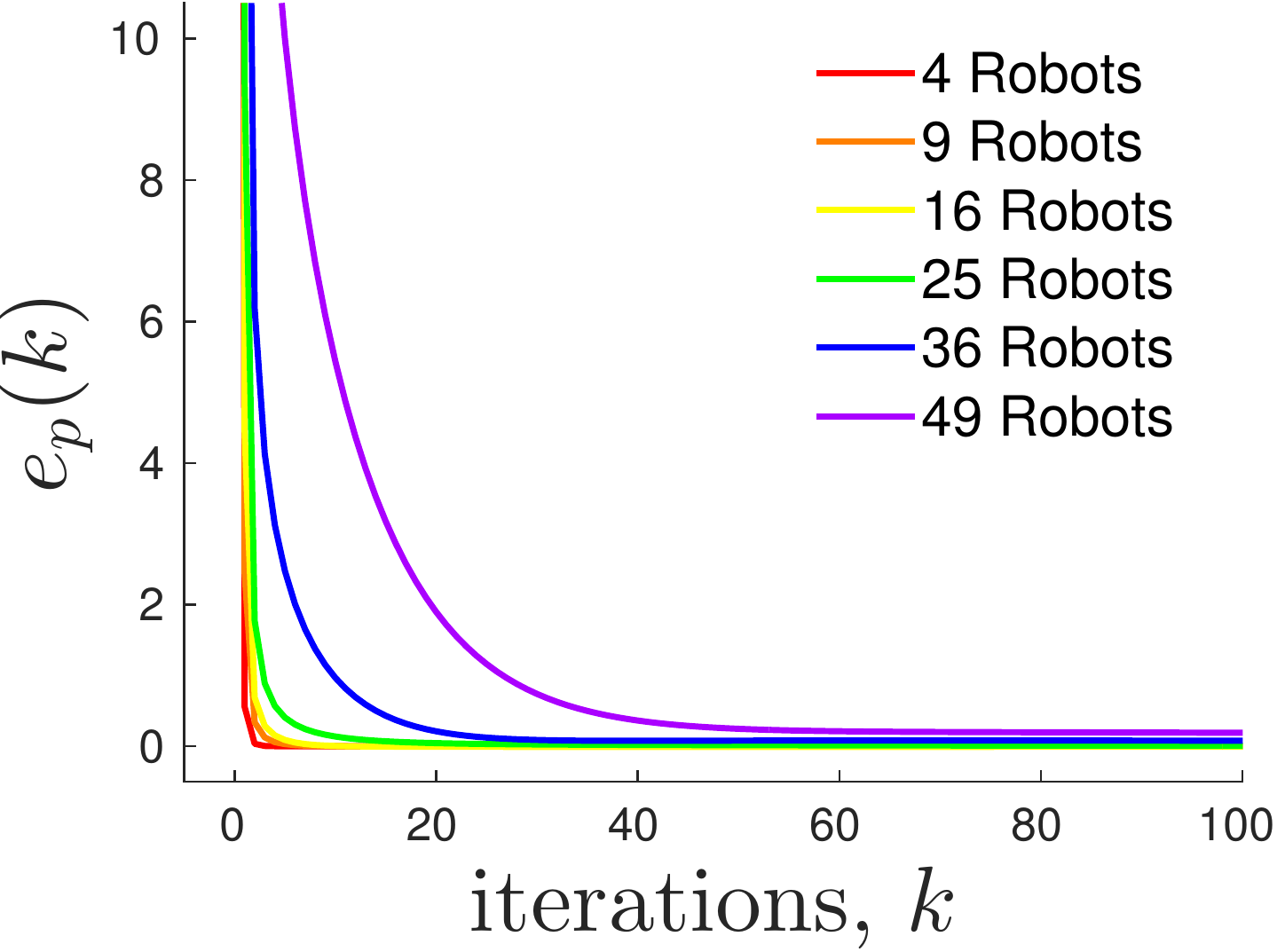} 
\end{minipage}
\\
\hspace{0.2cm}(a) Rotation  Error & (b) Pose  Error 
\\
\end{tabular}%
\caption{\label{fig:convergenceWRTNumberRobots} \DGS: convergence for scenarios with 
increasing number of robots. 
}
\end{minipage}%
\end{figure}

\myparagraph{Scalability in the number of robots}
 \Fig\ref{fig:convergenceWRTNumberRobots} shows the average rotation and pose errors for all the simulated datasets
 (4, 9, 16, 25, 36 and 49 robots). 
In all cases the errors quickly converge to zero. 
For large number or robots the convergence rate becomes slightly slower, while in all cases
the errors is negligible in few tens of iterations. 


\begin{table*}
\centering
{\renewcommand{\arraystretch}{1.2}%
\begin{tabular}{|c|c|c|c|c|c|c|}
\hline 
\multirow{3}{*}{\#Robots} &  \multicolumn{4}{c|}{Distributed Gauss-Seidel} & \multicolumn{2}{c|}{Centralized}\\
\cline{2-7}
& \multicolumn{2}{c|}{$\eta_r =\eta_p = 10^{-1}$} & \multicolumn{2}{c|}{$\eta_r =\eta_p = 10^{-2}$} & Two-Stage & \GN \\
\cline{2-5}
& \#Iter & Cost & \#Iter & Cost & Cost & Cost \\
\hline 
\hline 
4 & 10 & 1.9 & 65 & 1.9 & 1.9 & 1.9 \\ 
 \hline 
9 & 14 & 5.3 & 90 & 5.2 & 5.2 & 5.2 \\ 
 \hline 
16 & 16 & 8.9 & 163 & 8.8 & 8.8 & 8.7 \\ 
 \hline 
25 & 17 & 16.2 & 147 & 16.0 & 16.0 & 15.9 \\ 
 \hline 
36 & 28 & 22.9 & 155 & 22.7 & 22.6 & 22.5 \\ 
 \hline 
49 & 26 & 35.1 & 337 & 32.9 & 32.7 & 32.5 \\ 
 \hline 
 \end{tabular}}
 \vspace{0.1cm}
\caption{\label{tab:distributedCentralizedComparison} 
Number of iterations and cost attained in problem~\eqref{eq:MRPGO} by the \DGS algorithm 
(for two choices of the stopping conditions), versus a centralized two-stage approach and a \GN method.
Results are shown for scenarios with increasing number of robots.
}
\end{table*}

While so far we considered the errors for each subproblem ($e_r(k)$ and $e_p(k)$), 
we now investigate the overall accuracy of the \DGS algorithm to solve our original problem~\eqref{eq:MRPGO}. 
We compare the proposed approach against the centralized two-stage approach of~\cite{Carlone15icra-init3D} 
and against a standard (centralized) Gauss-Newton (\GN) method, 
available in \gtsam~(\cite{Dellaert12tr}). We use the cost 
attained in problem~\eqref{eq:MRPGO}  by each technique as accuracy metric (the lower the better).
Table~\ref{tab:distributedCentralizedComparison} reports the number of iterations and the cost 
attained in problem~\eqref{eq:MRPGO}, for the compared techniques. 
The number of iterations is the sum of the number of iterations required to solve~\eqref{eq:normEq-R} 
and \eqref{eq:MRPGO-linSys}.
The cost of the \DGS approach is given for two choices of the thresholds $\eta_r$ and $\eta_p$.
As already reported in~\cite{Carlone15icra-init3D}, the last two columns of the table confirm that 
the centralized two-stage approach is practically as accurate as a \GN method.
When using a strict stopping condition ($\eta_r = \eta_p = 10^{-2}$), the \DGS approach 
 produces the same error as the centralized counterpart (difference smaller than $1\%$). 
Relaxing the stopping conditions to $\eta_r = \eta_p = 10^{-1}$ implies a consistent reduction 
in the number of iterations, at a small loss in accuracy (cost increase 
is only significant for the scenario with 49 robots). In summary, 
the \DGS algorithm (with $\eta_r = \eta_p = 10^{-1}$) ensures accurate estimation within 
few iterations, even for large teams. 

\definecolor{dgreen}{rgb}{0,0.5,0}
\begin{figure}[t]
\hspace{-4mm}
\centering
\begin{minipage}{\columnwidth}
\centering
\begin{tabular}{cc}%
\begin{minipage}{0.48\columnwidth}%
\centering%
\includegraphics[scale=0.28, trim=0cm 0cm 0cm 0cm,clip]{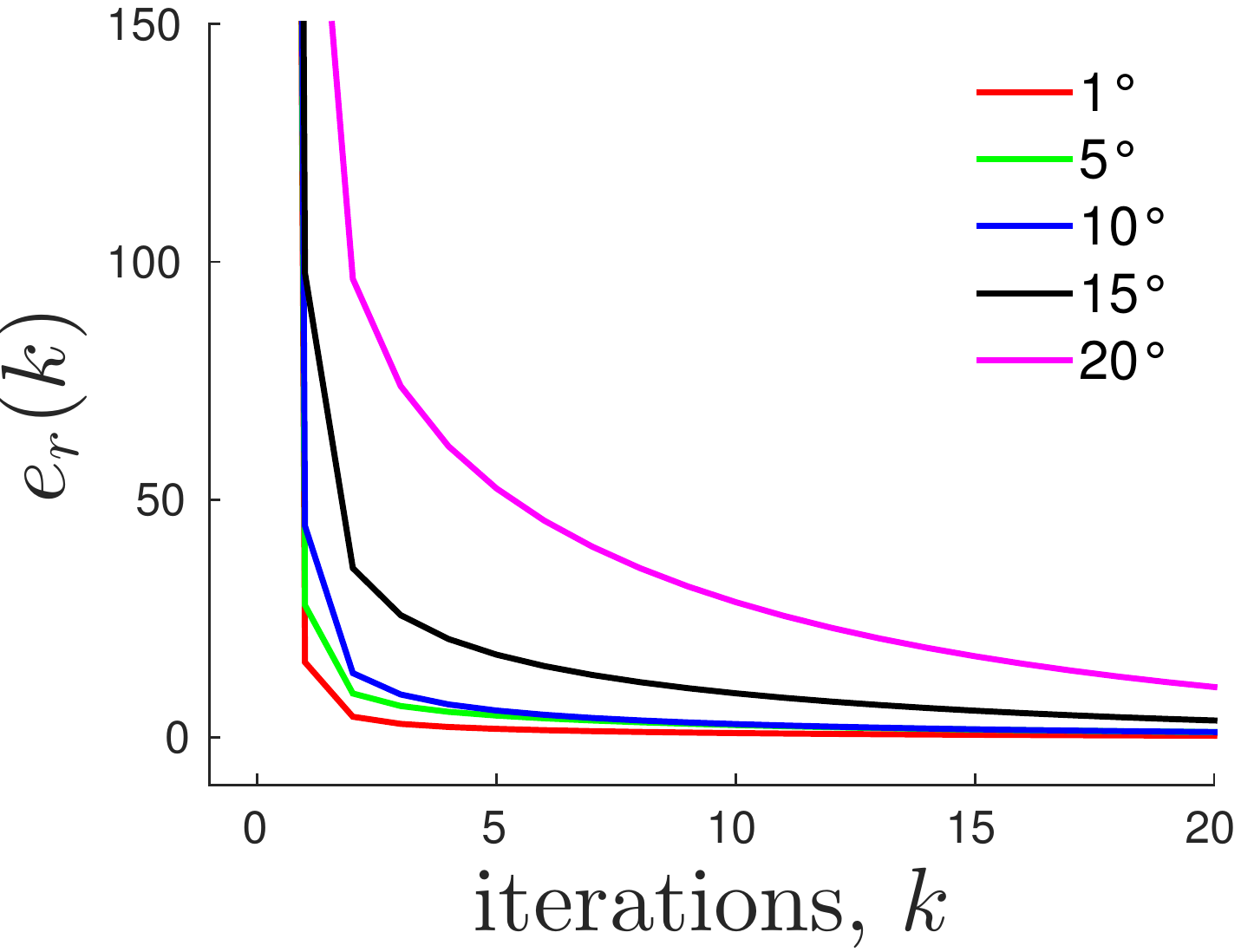} 
\end{minipage}
& \hspace{-4mm}
\begin{minipage}{0.48\columnwidth}%
\centering%
\includegraphics[scale=0.28, trim=0cm 0cm 0cm 0cm,clip]{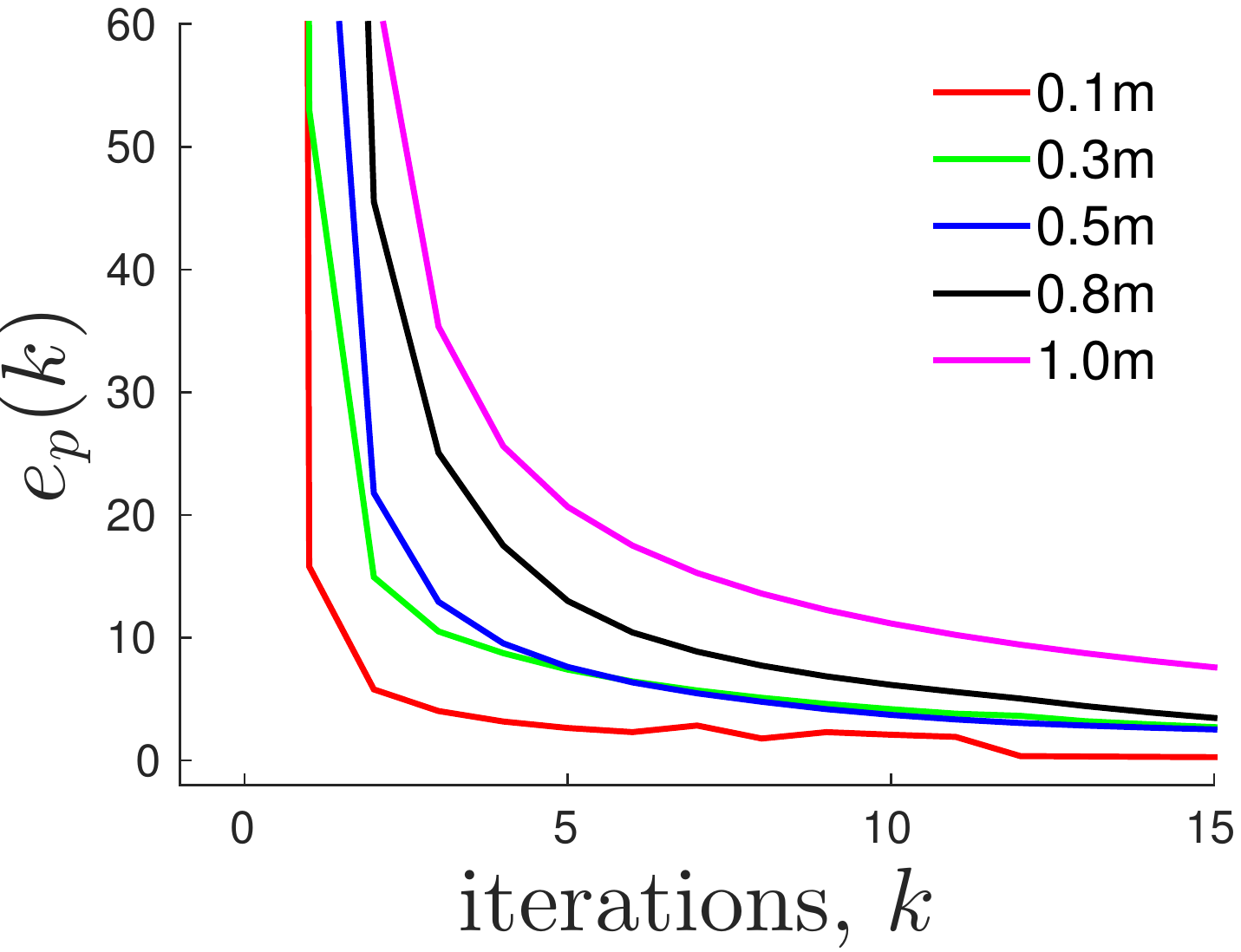} 
\end{minipage}
\vspace{0.1cm}
\\
\hspace{0cm}(a) Rotation Noise & (b) Translation Noise  \\
\end{tabular}
\caption{\label{fig:sensitivityAnalysis} \DGS: convergence for increasing levels of noise 
(scenario with 49 Robots). (a) Average rotation error for 
$\sigma_R = \{1, 5, 10, 15, 20\}^\circ$. (b) Average pose error for $\sigma_t = \{0.1, 0.3, 0.5, 0.8, 1.0\}$m.
}
\end{minipage}
\vspace{-0.3cm}
\end{figure}

\myparagraph{Sensitivity to measurement noise} 
Fig.~\ref{fig:sensitivityAnalysis} shows the average rotation and pose errors
for increasing levels of noise 
in the scenario with 49 robots. 
Also in this case, while larger noise seems to imply longer convergence tails, 
the error becomes sufficiently small after few  tens of iterations. 

Table~\ref{tab:sensitivityAnalysisTable} evaluates the performance of the \DGS method in 
solving problem~\eqref{eq:MRPGO} for increasing levels of noise, comparing it against  
the centralized two-stage approach of~\cite{Carlone15icra-init3D} 
and the Gauss-Newton method.
The \DGS approach is 
able to replicate the accuracy of the centralized two-stage approach, regardless the 
noise level, while the choice of thresholds $\eta_r = \eta_p = 10^{-1}$ ensures accurate estimation within 
few tens of iterations. 



\begin{table*}
\centering
{\renewcommand{\arraystretch}{1.2}%
\begin{tabular}{|cc|c|c|c|c|c|c|}
\hline 
\multicolumn{2}{|c|}{Measurement} &  \multicolumn{4}{c|}{Distributed Gauss-Seidel} & \multicolumn{2}{c|}{Centralized}\\
\cline{3-8}
\multicolumn{2}{|c|}{noise} & \multicolumn{2}{c|}{$\eta_r\!=\!\eta_p\!=\!10^{-1}$} & \multicolumn{2}{c|}{$\eta_r\!=\!\eta_p\!=\!10^{-2}$} & \!\!Two-Stage\!\! & \GN \\
\cline{3-6}
$\!\!\sigma_r (^{\circ})\!\!$ & $\!\!\sigma_t (\text{m})\!\!$ & \#Iter & Cost & \#Iter & Cost & Cost & Cost \\
\hline 
\hline 
 $\!\!1\!\!$ & $\!\!0.05\!\!$ & 8.5 & 2.1 & 51.0 & 1.8 & \!\!1.8\!\! & 1.8 \\ 
 \hline 
$\!\!5\!\!$ & $\!\!0.1\!\!$  & 21.8 & 14.8 & 197.8 & 14.0 & \!\!14.0\!\! & 13.9 \\ 
 \hline 
$\!\!10\!\!$ & $\!\!0.2\!\!$ & 35.6 & 58.4 & 277.7 & 56.6 & \!\!56.6\!\! & 56.0 \\  
 \hline 
$\!\!15\!\!$ & $\!\!0.3\!\!$ & 39.8 & 130.5 & 236.8 & 128.4 & \!\!129.3\!\! & \!\!126.0\!\! \\ 
 \hline 
 \end{tabular}}
 \vspace{0.1cm}
\caption{\label{tab:sensitivityAnalysisTable} 
Number of iterations and cost attained in problem~\eqref{eq:MRPGO} by the \DGS algorithm 
(for two choices of the stopping conditions), versus a centralized two-stage approach and a \GN method.
Results are shown for increasing measurement noise.
}
\end{table*}

\definecolor{dgreen}{rgb}{0,0.5,0}
\renewcommand{\scaleFig}{}
\begin{figure}[t]
\centering
\begin{minipage}{\columnwidth}
\begin{tabular}{cc}%
\hspace{-0.1cm}
\begin{minipage}{0.45\columnwidth}%
\centering%
\includegraphics[scale=0.28, trim=0cm 0cm 0cm 0cm,clip]{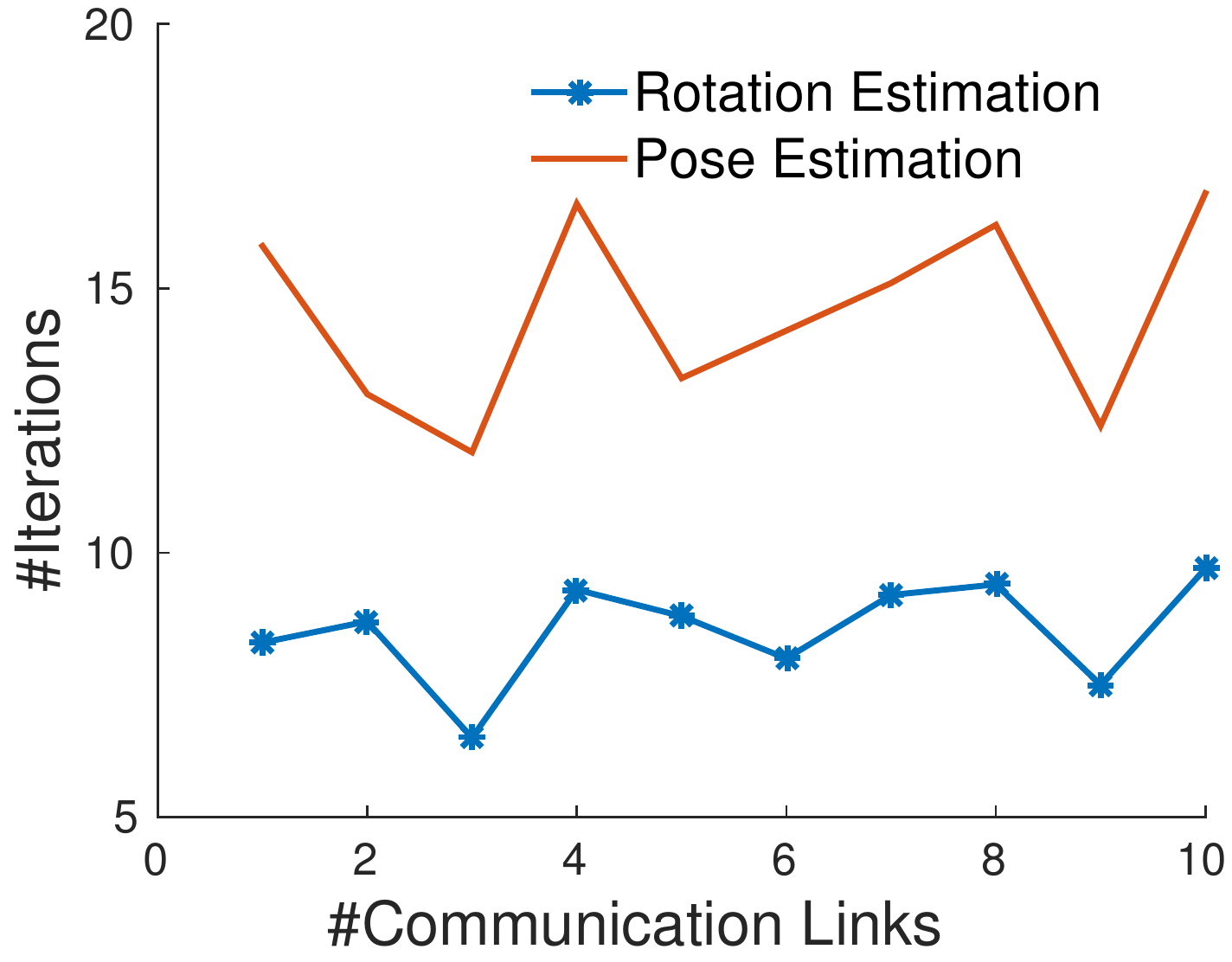} 
\end{minipage}
&
\begin{minipage}{0.45\columnwidth}%
\centering%
\includegraphics[scale=0.28, trim=0cm 0cm 0cm 0cm,clip]{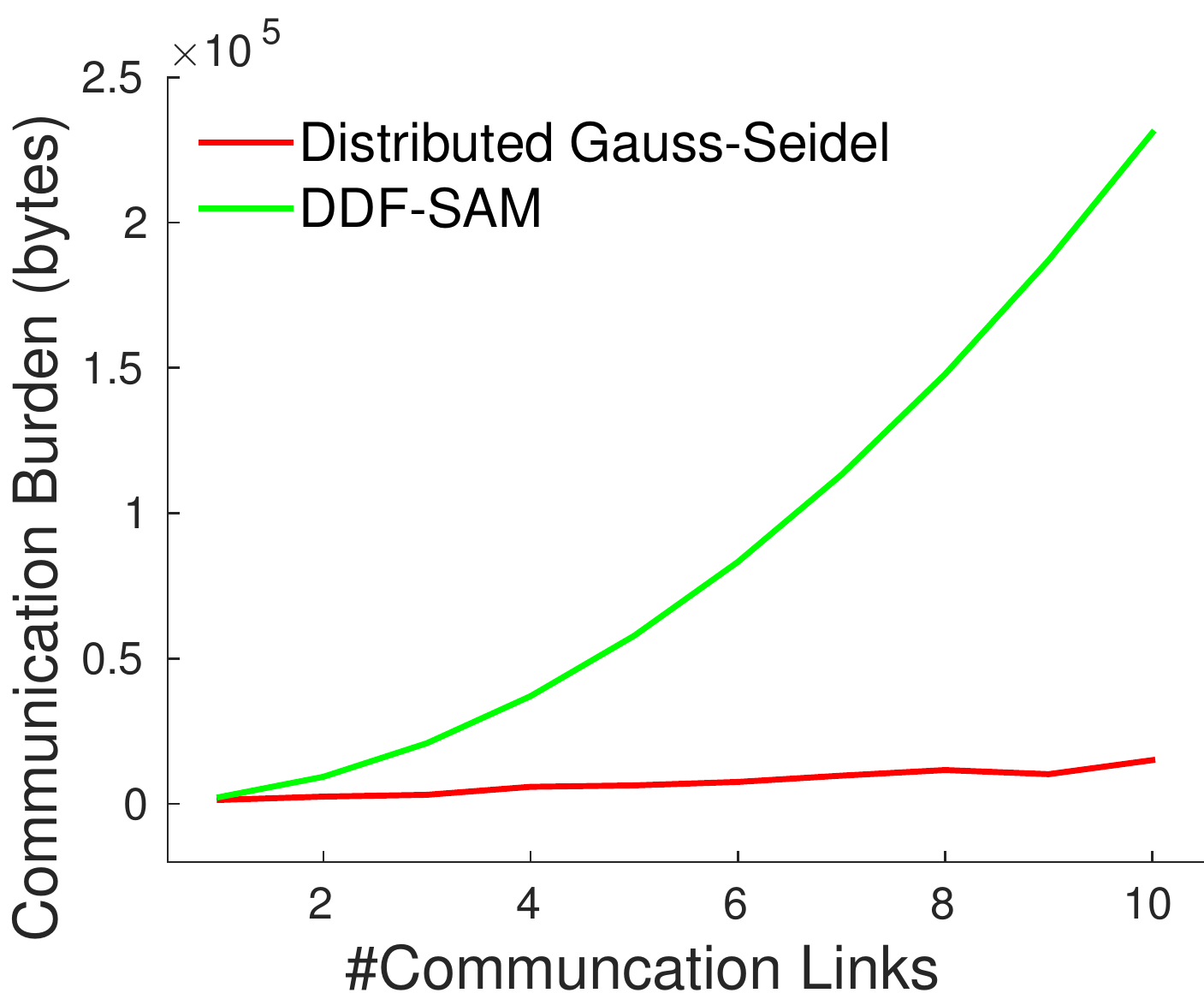} 
\end{minipage}
\vspace{0.1cm}
\\
\hspace{0.2cm}(a)  & (b) 
\\
\end{tabular}
\caption{\label{fig:convergenceWRTNumberCommunicationLinks} 
\DGS \vs \DDFSAM:
(a) average number of iterations versus number of separators for the \DGS algorithm.
(b) communication burden (bytes of exchanged information) for \DGS and \DDFSAM, for increasing 
number of separators.}
\end{minipage}%
\vspace{-0.1cm} 
\end{figure}

\myparagraph{Scalability in the number of separators}
In order to evaluate the impact of the number of separators on convergence, 
we simulated two robots moving along parallel tracks for 10 time stamps.
The number of communication links were varied from 1 (single communication) to 10 (communication at every time), 
 hence the number of separators (for each robot) ranges from 1 to 10. 
Fig.~\ref{fig:convergenceWRTNumberCommunicationLinks}a shows the number of iterations 
required by the \DGS algorithm ($\eta_r = \eta_p = 10^{-1}$), for increasing 
number of communication links. 
The number of iterations is fairly insensitive to the number of communication links.
\newcommand{\pro}{\,} 

Fig.~\ref{fig:convergenceWRTNumberCommunicationLinks}b compares the 
information exchanged in the \DJ algorithm against a state-of-the-art algorithm, 
\DDFSAM~(\cite{Cunningham10iros}). In \DDFSAM, each robot sends 
$K_{\GN} \, \left[s \pro B_p +\left(s \pro B_p\right)^{2}\right]$ bytes,
where $K_{\GN}$ is the number of iterations required by a \GN method applied to problem~\eqref{eq:MRPGO} 
(we consider the best case $K_{\GN} = 1$), 
$s$ is the number of separators and $B_p$ is the size of a pose in bytes.
In the \DGS algorithm, each robots sends
$K^r_{\text{DGS}}\pro\left(s \pro B_r\right) + K^p_{\text{DGS}}\pro\left(s \pro B_p\right)$ 
bytes,
where $K^r_{\text{DGS}}$ and $K^p_{\text{DGS}}$ are the number of iterations required by the 
\DGS algorithm to solve the linear systems~\eqref{eq:normEq-R} 
and \eqref{eq:MRPGO-linSys}, respectively, and $B_r$ is the size of a rotation (in bytes). 
We assume $B_r = 9$ doubles (72 bytes)\footnote{In the linear system~\eqref{eq:normEq-R} we relax the orthogonality constraints
hence we cannot parametrize the rotations with a minimal 3-parameter representation.}  and $B_p= 6$ doubles (48 bytes).
The number of iterations $K^r_{\text{DGS}}$ and $K^p_{\text{DGS}}$ are the one plotted in 
Fig.~\ref{fig:convergenceWRTNumberCommunicationLinks}\suba. 
From Fig.~\ref{fig:convergenceWRTNumberCommunicationLinks}\subb we see that 
the communication burden of \DDFSAM quickly becomes unsustainable, while the 
linear increase in communication of the  \DGS algorithm implies large communication saving.

%

\definecolor{dgreen}{rgb}{0,0.5,0}
\renewcommand{\scaleFig}{0.25}

\begin{figure*}[t]
\begin{minipage}{0.9\columnwidth}
\begin{tabular}{cc|cc}%
\begin{minipage}{0.5\columnwidth}%
\centering%
\includegraphics[width=\columnwidth, trim=0cm 0cm 0cm 0cm,clip]{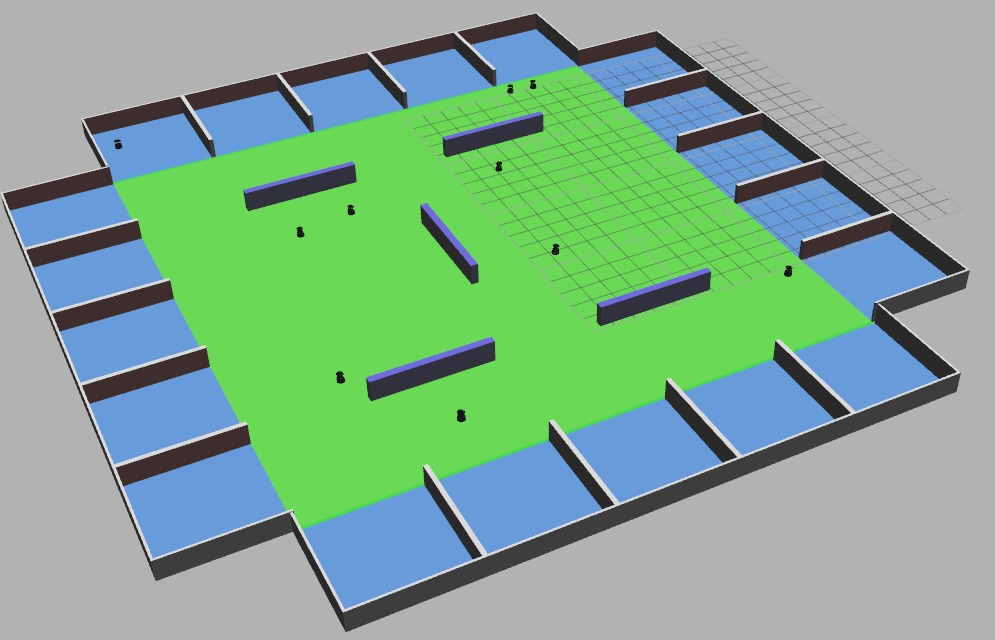} 
\end{minipage}
&
\begin{minipage}{0.5\columnwidth}%
\centering%
\includegraphics[width=\columnwidth, trim=0cm 0cm 0cm 0cm,clip]{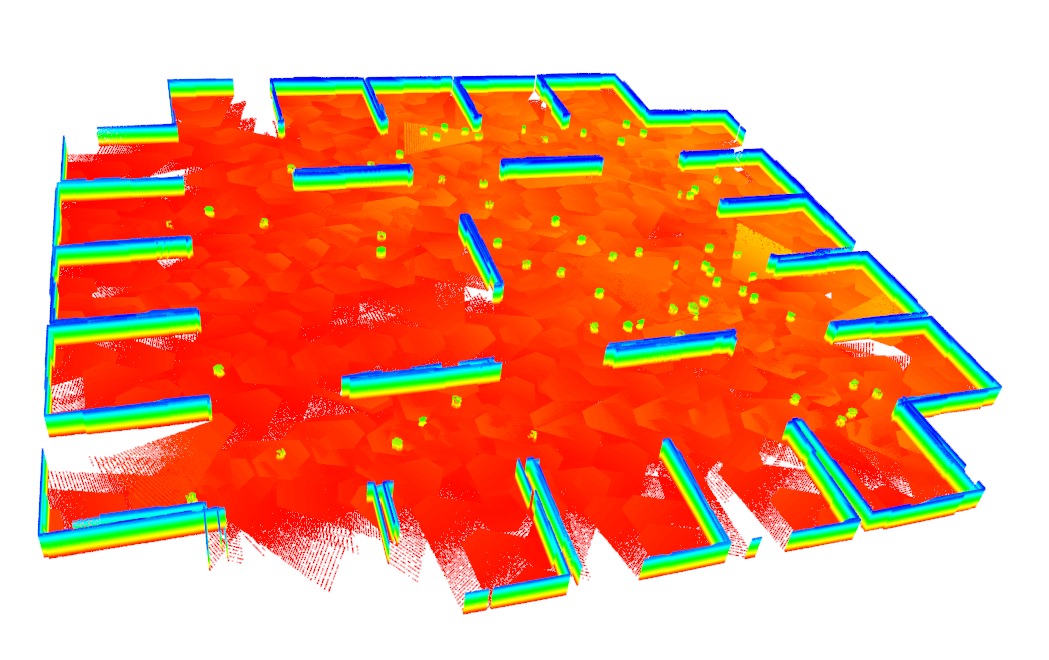} 
\end{minipage}
&
\begin{minipage}{0.5\columnwidth}%
\centering%
\includegraphics[width=\columnwidth, trim=0cm 0cm 0cm 0cm,clip]{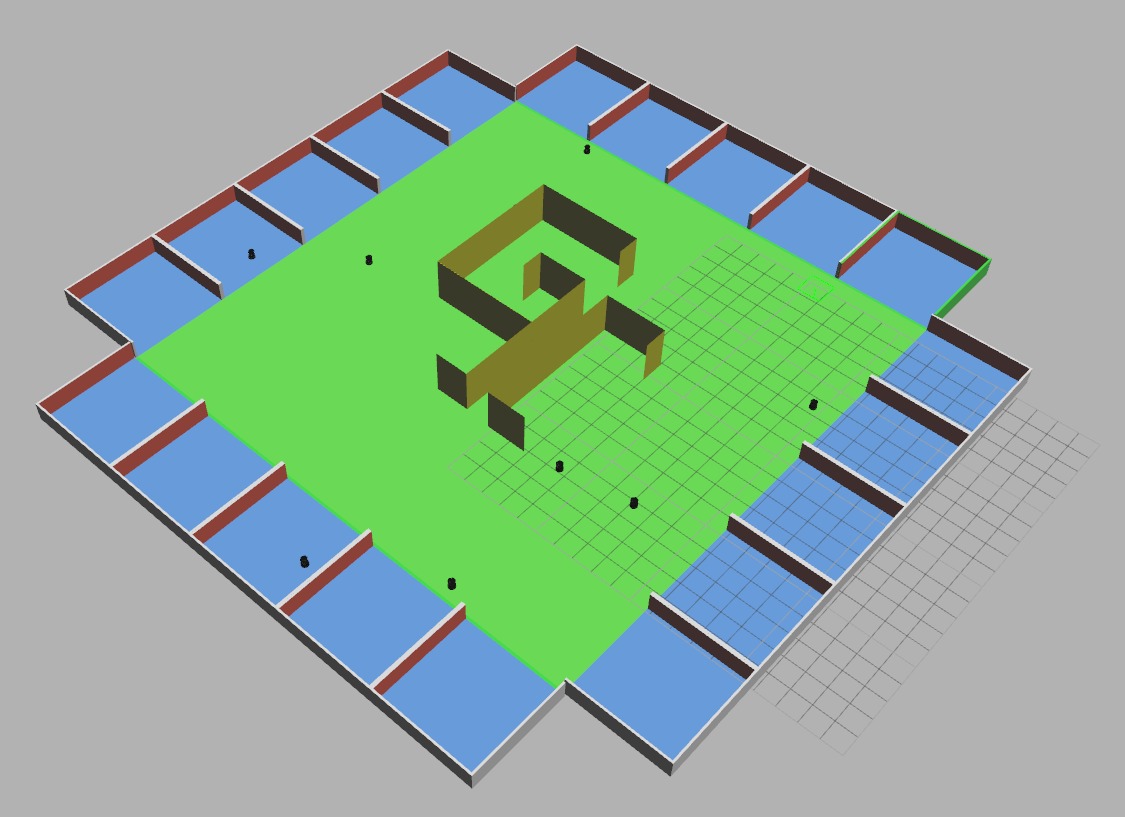} 
\end{minipage}
&
\begin{minipage}{0.5\columnwidth}%
\centering%
\includegraphics[width=\columnwidth, trim=0cm 0cm 0cm 0cm,clip]{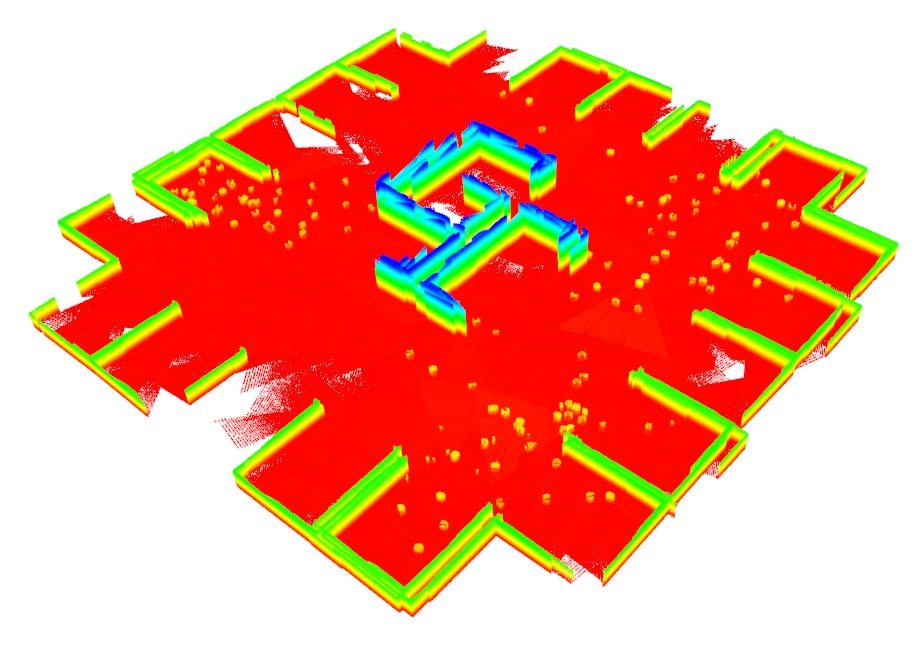} 
\end{minipage}
\vspace{0.1cm}
\\
\hspace{0.2cm} GroundTruth & Estimate &  GroundTruth  & Estimate
\\
\end{tabular}
\end{minipage}
\caption{\label{fig:gazeboExperiment} Gazebo tests: ground truth environments and aggregated point clouds
corresponding to the \DGS estimate.
}
\end{figure*}



\definecolor{dgreen}{rgb}{0,0.5,0}

\begin{figure}[t]

\centering
\begin{minipage}{\columnwidth}
\hspace{-4mm}
\centering%
\begin{tabular}{cc}%
\begin{minipage}{0.48\columnwidth}%
\centering%
\includegraphics[scale=0.28, trim=0cm 0cm 0cm 0cm,clip]{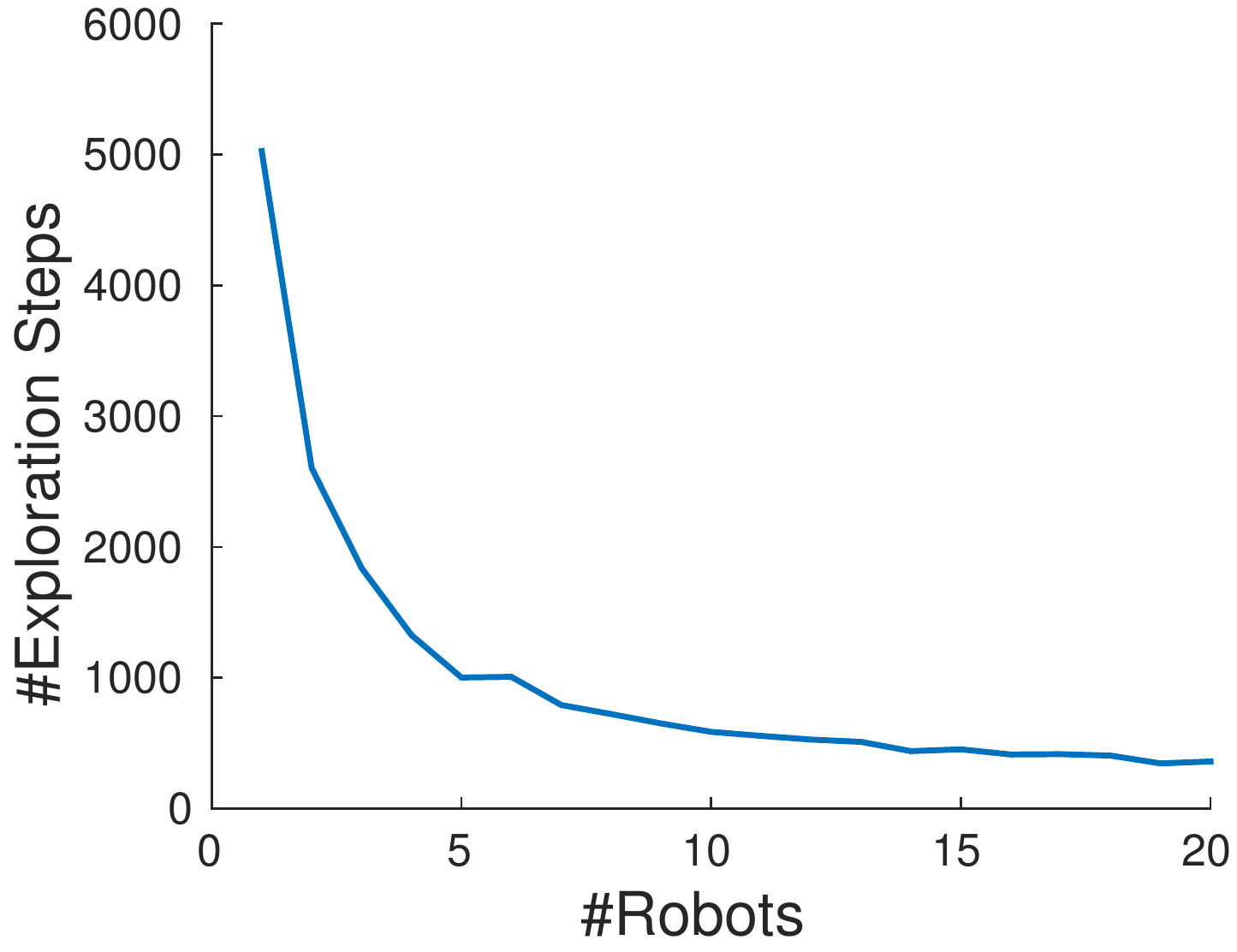} 
\end{minipage}
& 
\begin{minipage}{0.48\columnwidth}%
\centering%
\includegraphics[scale=0.28, trim=0cm 0cm 0cm 0cm,clip]{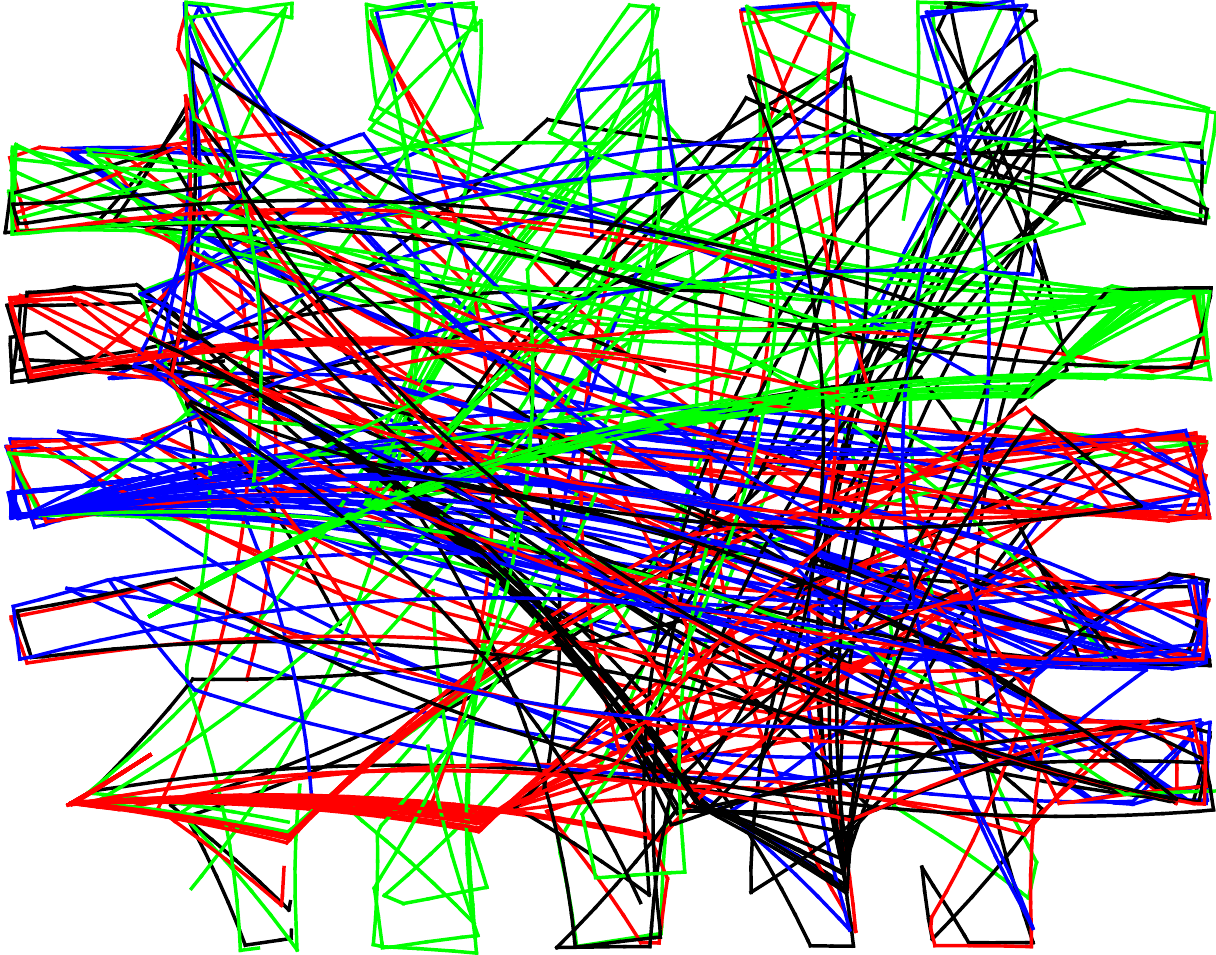} 
\end{minipage}
\\
\hspace{0.2cm}(a) Exploration steps & (b) Monte Carlo Runs 
\\
\end{tabular}%
\caption{\label{fig:robotsVsExplorationTime} (a) Number of exploration steps required to explore a fixed-sized grid with  increasing number of robots. (b) Samples of robot trajectories from our Gazebo-based Monte Carlo experiments.}
\end{minipage}%
\end{figure}
\maybe{
\definecolor{dgreen}{rgb}{0,0.5,0}
\begin{figure}[t]
\hspace{-4mm}
\centering
\begin{minipage}{\columnwidth}
\centering
\begin{tabular}{cc}%
\begin{minipage}{0.48\columnwidth}%
\centering%
\includegraphics[scale=0.14, trim=0cm 0cm 0cm 0cm,clip]{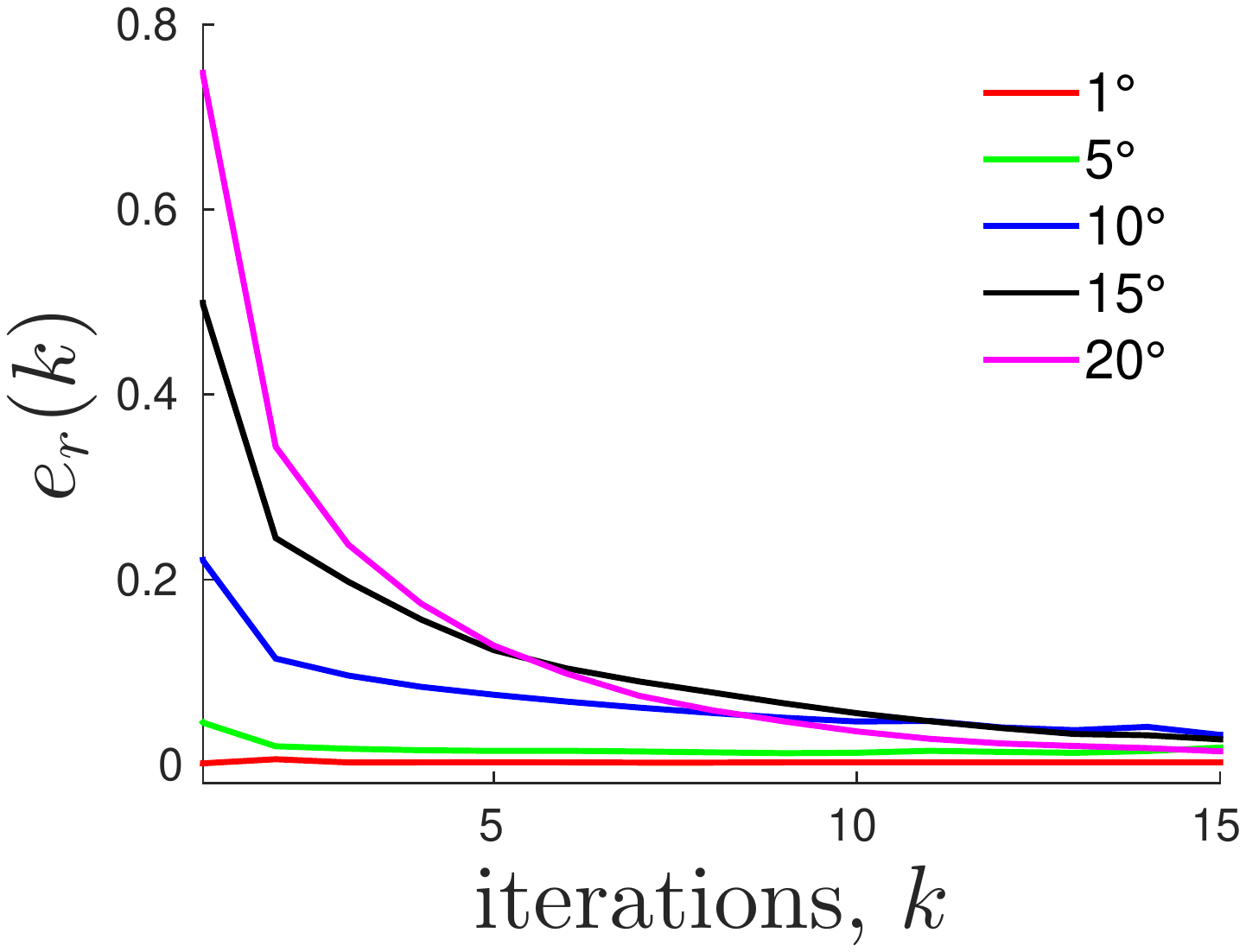} 
\end{minipage}
& \hspace{-4mm}
\begin{minipage}{0.48\columnwidth}%
\centering%
\includegraphics[scale=0.14, trim=0cm 0cm 0cm 0cm,clip]{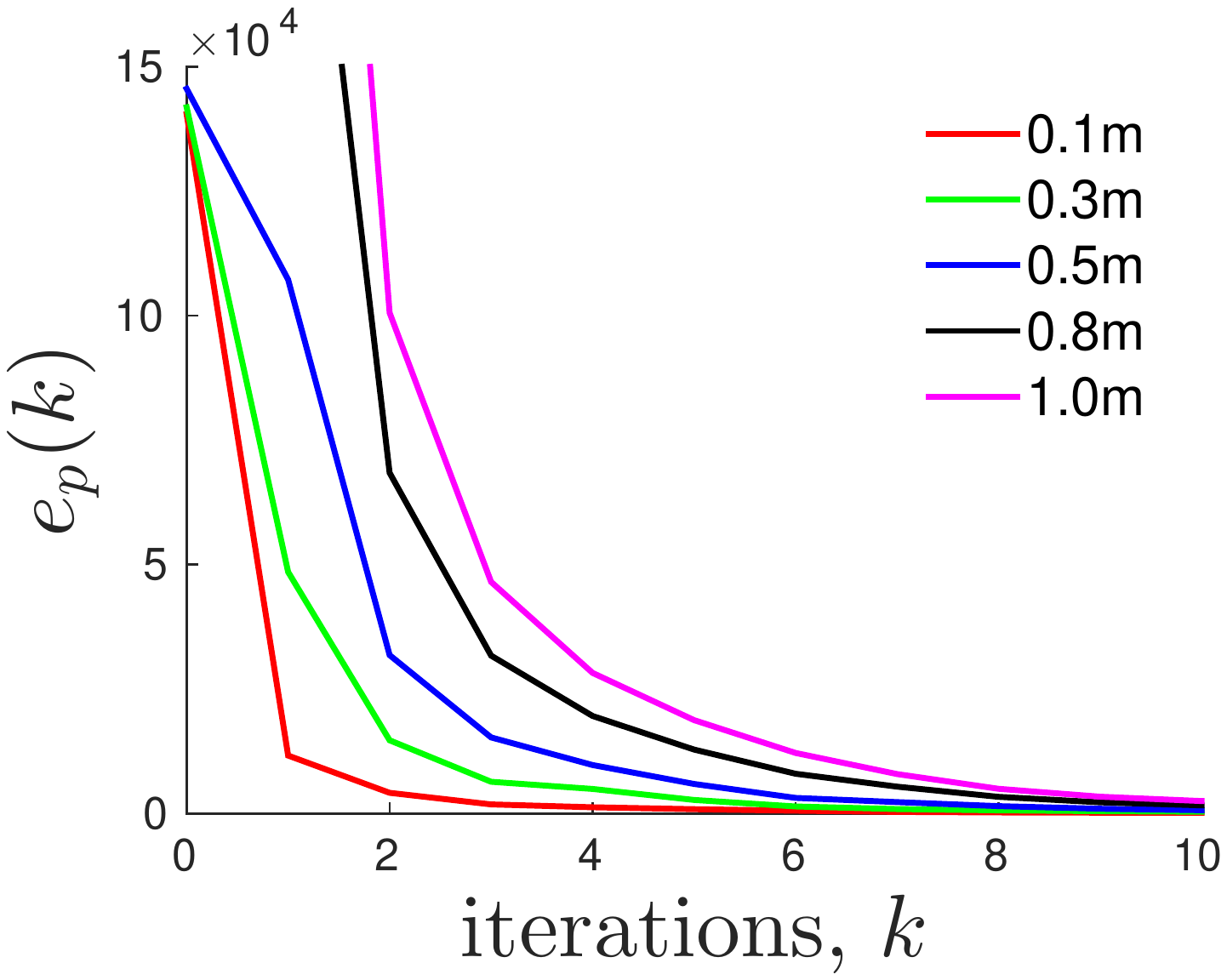} 
\end{minipage}
\vspace{0.1cm}
\\
\hspace{0cm}(a) Rotation Noise & (b) Translation Noise  \\
\end{tabular}
\caption{\label{fig:sensitivityAnalysisGazebo} Convergence for increasing levels of noise 
(scenario with 2 Robots in Gazebo). (a) Average rotation estimation error for 
$\sigma_R = \{1, 5, 10, 15, 20\}^\circ$. (b) Average pose estimation error for $\sigma_t = \{0.1, 0.3, 0.5, 0.8, 1.0\}$m.
}
\end{minipage}
\end{figure}

\begin{figure}[t]
\hspace{-4mm}
\centering
\begin{minipage}{\columnwidth}
\centering
\begin{tabular}{cc}%
\begin{minipage}{0.48\columnwidth}%
\centering%
\includegraphics[scale=0.14, trim=0cm 0cm 0cm 0cm,clip]{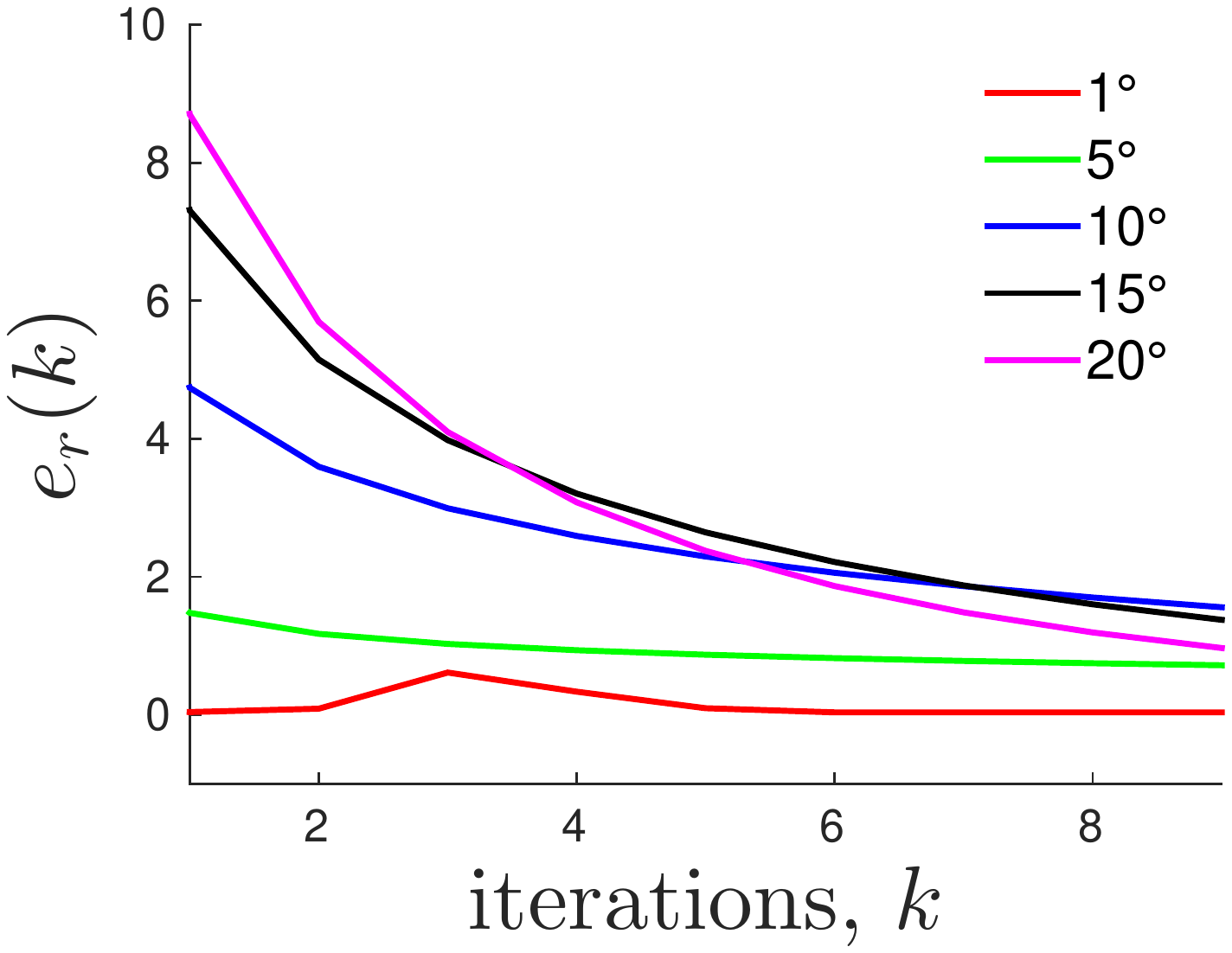} 
\end{minipage}
& \hspace{-4mm}
\begin{minipage}{0.48\columnwidth}%
\centering%
\includegraphics[scale=0.14, trim=0cm 0cm 0cm 0cm,clip]{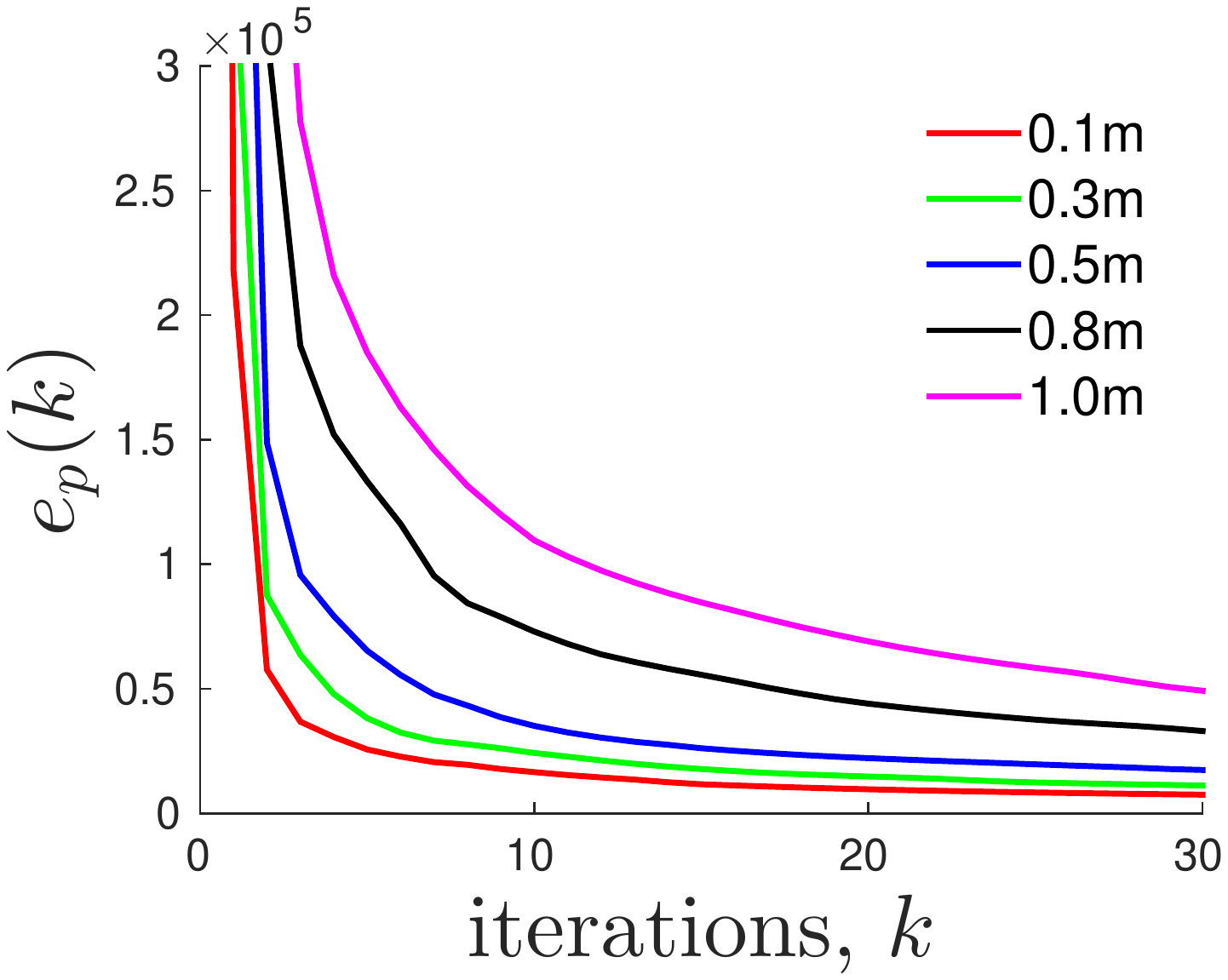} 
\end{minipage}
\vspace{0.1cm}
\\
\hspace{0cm}(a) Rotation Noise & (b) Translation Noise  \\
\end{tabular}
\caption{\label{fig:sensitivityAnalysisGazebo4Robots} Convergence for increasing levels of noise 
(scenario with 4 Robots in Gazebo). (a) Average rotation estimation error for 
$\sigma_R = \{1, 5, 10, 15, 20\}^\circ$. (b) Average pose estimation error for $\sigma_t = \{0.1, 0.3, 0.5, 0.8, 1.0\}$m.
}
\end{minipage}
\end{figure}

}

\myparagraph{Realistic simulations in Gazebo}
We tested our \DGS-based approach in two scenarios in Gazebo simulations as shown in \Fig\ref{fig:gazeboExperiment}.
The robots start at fixed locations  and explore the environment by moving according to a random walk. 
Each robot is equipped with a 3D laser range finder, which is used to intra-robot and inter-robot measurements 
via scan matching.
In both  scenarios, two robots communicate only when they are within close proximity of each other ($0.5$m in our tests).
Results are average over 100 Monte-Carlo runs. 

\Fig\ref{fig:gazeboExperiment} shows the aggregated point cloud corresponding to the \DGS trajectory estimate, for one of the runs. The point cloud closely resembles the ground truth environment shown in the same figure. \Fig\ref{fig:robotsVsExplorationTime}\suba shows that number of steps required to explore the whole environment quickly decreases with  increasing number of robots. This intuitive observation motivates our interest towards mapping techniques that can  
scale to large teams of robots. 
 \Fig\ref{fig:robotsVsExplorationTime}\subb reports trajectory samples 
for different robots in our Monte Carlo analysis.
\maybe{
  \Figs\ref{fig:sensitivityAnalysisGazebo} and~\ref{fig:sensitivityAnalysisGazebo4Robots} show the average errors $e_r(k)$ and $e_p(k)$ for  increasing levels of noise 
and for the scenario with 2 and 4 robots respectively. 
}

\subsection{Simulation Results: Multi Robot Object based SLAM}
\label{sec:simulations_objectSLAM}

In this section we characterize the performance of the \DGS algorithms associated with our object-based model 
described in Section~\ref{sec:objects}.
We test the resulting
multi robot object-based SLAM approach in terms of scalability in the number of robots and sensitivity to noise.

\begin{figure}[t]
\centering
\begin{minipage}{1.0\columnwidth}
\begin{tabular}{c}%

 25 Chairs Scene	 \\

\begin{minipage}{0.9\columnwidth}%
\centering%
\includegraphics[width=\columnwidth, trim=0cm 0cm 0cm 0cm,clip]{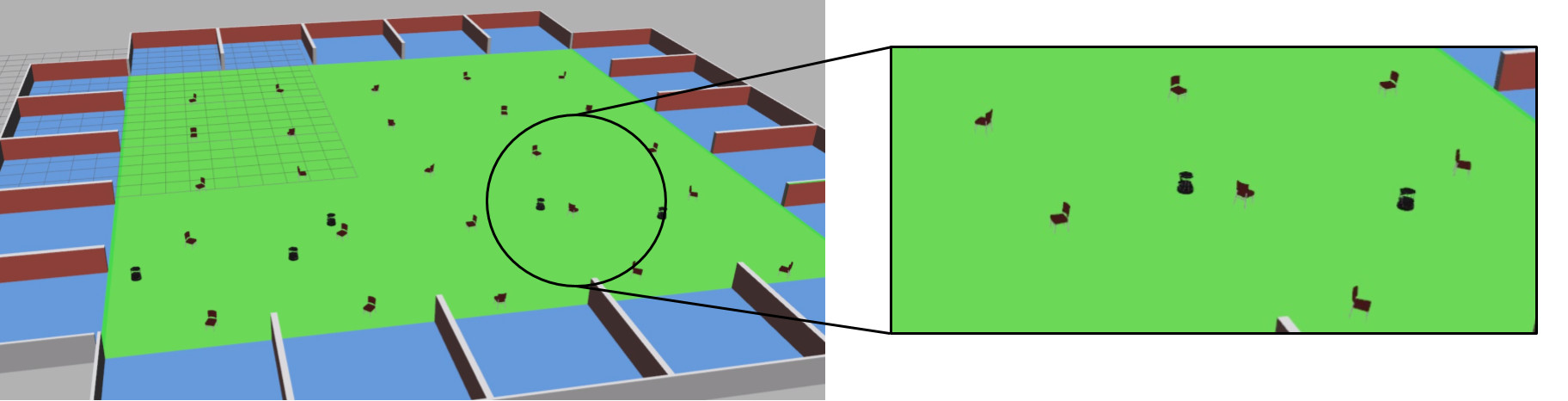} 
\end{minipage}\vspace{0.2cm}
\\

House Scene \\
\begin{minipage}{0.9\columnwidth}%
\centering%
\includegraphics[width=\columnwidth, trim=0cm 0cm 0cm 0cm,clip]{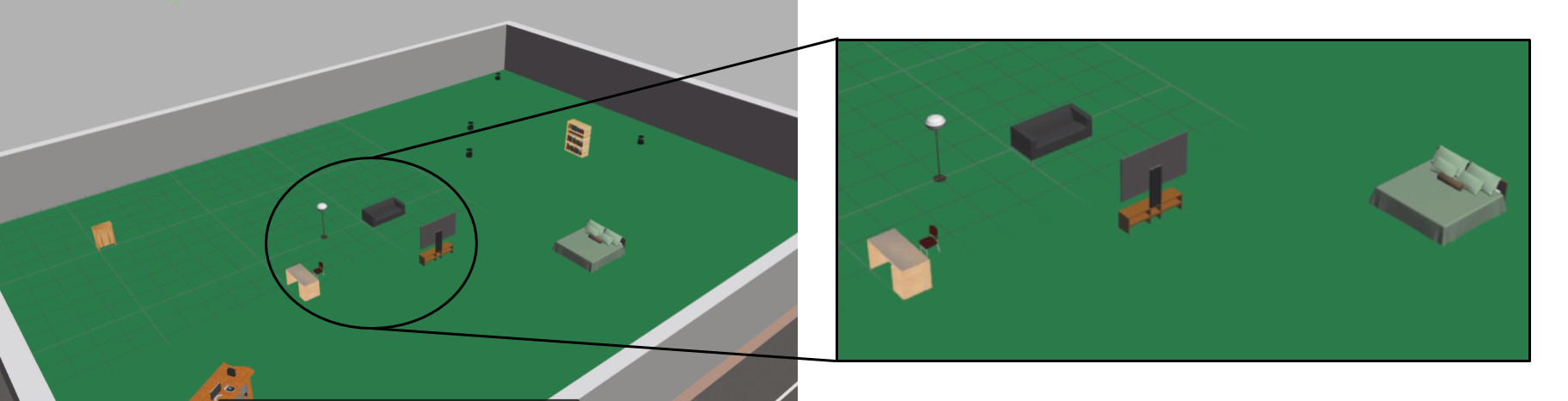} 
\end{minipage}
\\

\end{tabular}
  \end{minipage}
  \caption{\label{fig:scenario} Multi robot object-based SLAM in Gazebo: the \chairs and \house scenarios simulated in Gazebo. 
 }

\end{figure}

\myparagraph{Simulation setup and performance metrics}
We consider two scenarios, the \chairs  and the \house, which we simulated in Gazebo. In the \chairs scenario, we placed 25 chairs as objects on a grid, with each chair placed at a random angle. In  the \house scenario, we placed furniture as objects in order to simulate a living room environment. \Fig\ref{fig:scenario} shows the two scenarios.
Unless specified otherwise, we generate measurement noise from a zero-mean Gaussian distribution with standard deviation $\sigma_R = 5^\circ$ for the rotations and $\sigma_t = 0.2$m for the translations. Six robots are used by default. 
Results are averaged over 10 Monte Carlo runs. 

We use the \emph{absolute translation error} (ATE*) and \emph{absolute rotation error} (ARE*) of the robot and landmark poses with respect to the centralized estimate as error metric. 
These two metrics are formally defined below.

\mysubparagraph{Absolute Translation Error (ATE*)} Similar to the formulation by Sturm et al. \cite{Sturm12iros}, the average translation error measures the absolute distance between the trajectory and object poses  estimated by our approach versus the centralized \GN method. 
The ATE* is defined as: 
\begin{equation}
ATE^*=
\left( \frac{1}{\sum_{\alpha \in \Omega} n_\alpha} \sum_{\alpha \in \Omega} \sum_{i=1}^{n_\alpha}\| \vt\of{\alpha}{i} - \vt\of{\alpha}{i}^* \|^{2}
\right)^{\frac{1}{2}}
\label{eq:ate}
\end{equation}
where $\vt\of{\alpha}{i}$ is the position estimate for robot $\alpha$ at time $i$, 
$\vt\of{\alpha}{i}^*$  is the corresponding estimate from GN, and $n_\alpha$ is the number of 
poses in the trajectory of $\alpha$. A similar definition holds for the object positions.
 
\mysubparagraph{Absolute Rotation Error (ARE*)} The average rotation error is computed by 
evaluating the angular mismatch between the (trajectory and objects) rotations produced by the proposed
 approach versus a centralized GN method:
\begin{equation}
ARE^*=
 \left( \frac{1}{\sum_{\alpha \in \Omega} n_\alpha} \sum_{\alpha \in \Omega} \sum_{i=1}^{n_\alpha} 
 \| \text{Log}  \left( (\MR\of{\alpha}{i}^*)\tran \MR\of{\alpha}{i} \right) \|^{2}
\right)^{\frac{1}{2}}
\label{eq:are}
\end{equation}
where $\MR\of{\alpha}{i}$ is the rotation estimate for robot $\alpha$ at time $i$, 
$\MR\of{\alpha}{i}^*$  is the corresponding estimate from GN. 
A similar definition holds for the object rotations.

\begin{figure}[t]
\centering
\begin{minipage}{\columnwidth}\hspace{-5mm}
\begin{tabular}{cc}%
 Centralized & Distributed \\
\begin{minipage}{0.5\columnwidth}%
\centering%
\includegraphics[width=0.75\columnwidth, trim=0cm 0cm 0cm 0cm,clip]{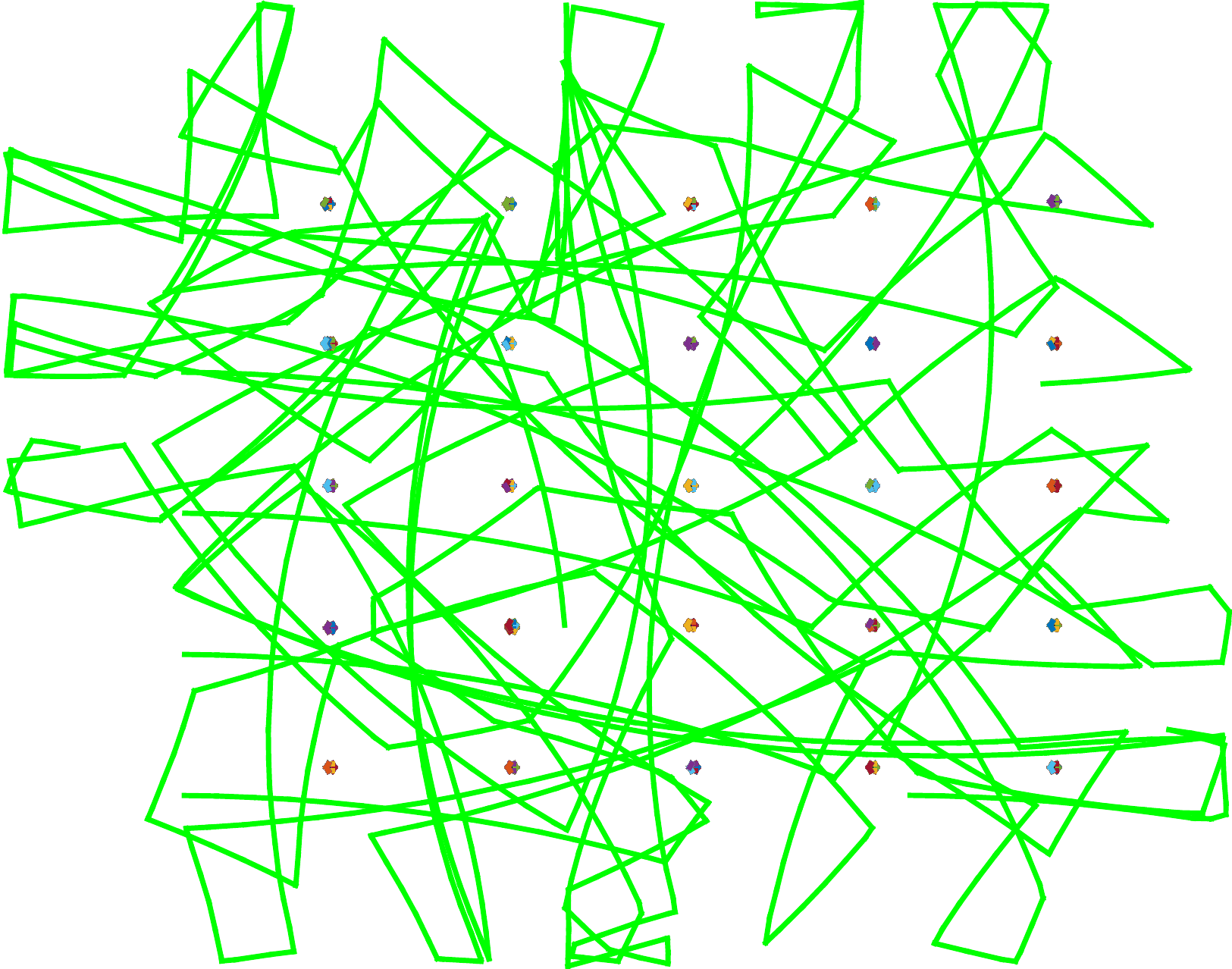} 
\end{minipage}
&
\begin{minipage}{0.5\columnwidth}%
\centering%
\includegraphics[width=0.75\columnwidth, trim= 0cm 0cm 0cm 0cm, clip]{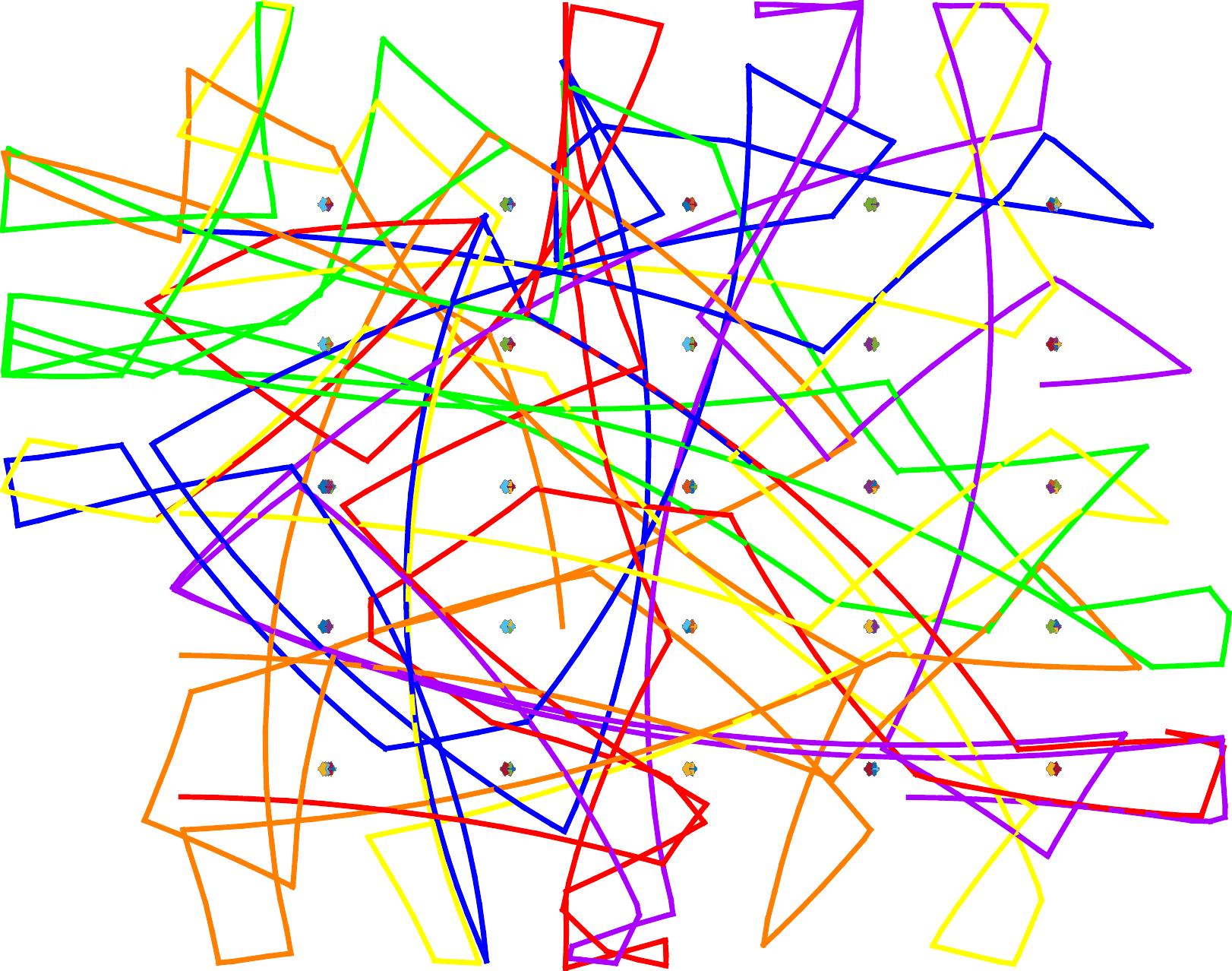} \vspace{0.2cm} \end{minipage}\\

\begin{minipage}{0.5\columnwidth}%
\centering%
\includegraphics[width=0.75\columnwidth, trim=0cm 0cm 0cm 0cm,clip]{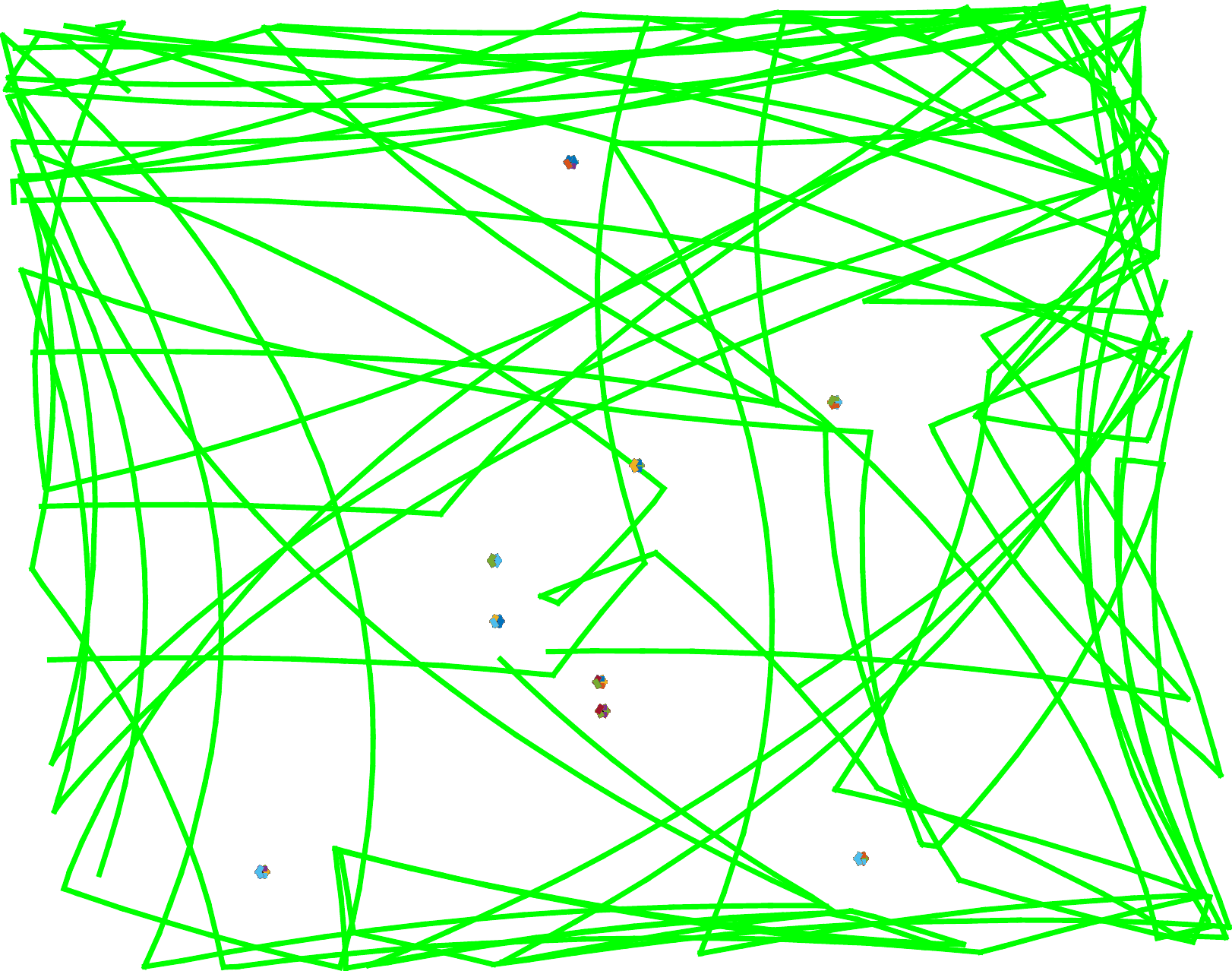} 
\end{minipage}
&
\begin{minipage}{0.5\columnwidth}%
\centering%
\includegraphics[width=0.75\columnwidth, trim= 0cm 0cm 0cm 0cm, clip]{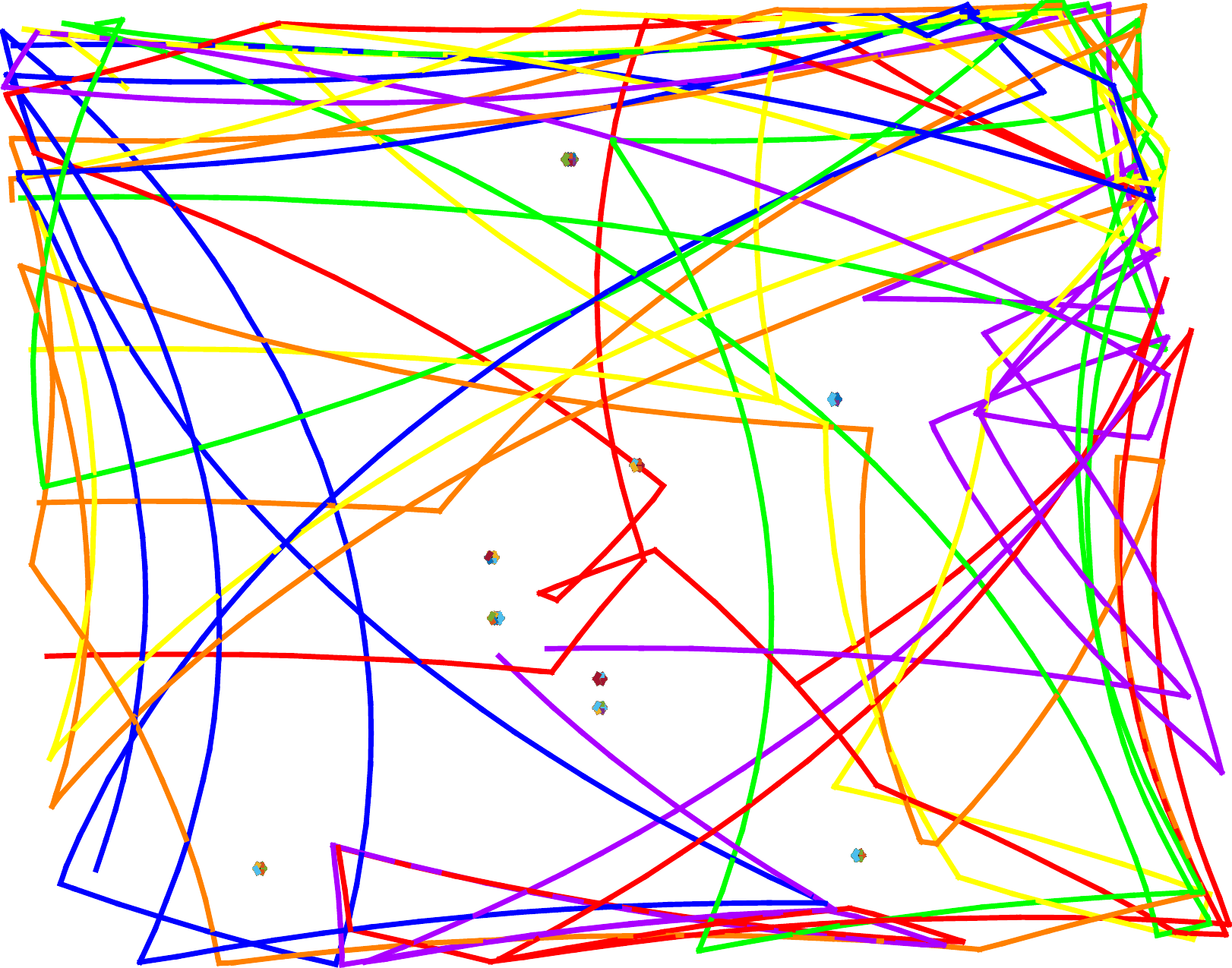} \end{minipage}\\

\end{tabular}
  \end{minipage}
  \caption{\label{fig:simulationExperiments}  
Trajectories of the six robots and object locations (shows as red dots)  estimated using the centralized \GN method and 
the proposed \DGS method for the \chairs(top) and \house scenarios (bottom).
\vspace{-5mm}
  }
\end{figure}


\begin{table*}[h!]
\centering
{\renewcommand{\arraystretch}{1.5}%
\begin{tabular}{|c|c|c|c|c|c|c|c|c|c|}
\hline 
\multirow{3}{*}{\#Robots} &  \multicolumn{4}{c|}{Distributed Gauss-Seidel} & Centralized & \multicolumn{2}{c|}{ATE* (m)} & \multicolumn{2}{c|}{ARE* (deg)}\\
\cline{2-10}
& \multicolumn{2}{c|}{$\eta\!=\!10^{-1}$} & \multicolumn{2}{c|}{$\eta\!=\!10^{-2}$} & GN & \multirow{2}{*}{Poses} & \multirow{2}{*}{Lmrks.} & \multirow{2}{*}{Poses} & \multirow{2}{*}{Lmrks.} \\
\cline{2-5}
& \#Iter & Cost & \#Iter & Cost & Cost  & & & &\\
\hline 
\hline 

2 & 5.0 & 56.1 & 9.0 & 56.0 & 54.7 & 1.5e-03 & 8.0e-04 & 2.1e-01 & 2.8e-01\\
 \hline
4 & 5.0 & 118.0 & 8.0 & 117.9 & 113.8 & 9.7e-04 & 7.5e-04 & 2.0e-01 & 2.8e-01\\
 \hline
6 & 5.0 & 166.6 & 7.0 & 166.5 & 160.9 & 3.1e-03 & 2.1e-03 & 3.3e-01 & 4.0e-01\\
 \hline
  \end{tabular}}
 \vspace{0.1cm}
\caption{\label{tab:numRobotsAnalysis_ObjectSLAM} 
Number of iterations, cost, ATE* and ARE* of our approach compared to the centralized Gauss-Newton method
for increasing number of robots. ATE* and ARE* are measured using $\eta\!=\!10^{-1}$ as stopping condition. 
}
\end{table*}


\begin{table*}[h!]
\centering
{\renewcommand{\arraystretch}{1.5}%
\begin{tabular}{|cc|c|c|c|c|c|c|c|c|c|}
\hline 
\multicolumn{2}{|c|}{Measurement} &  \multicolumn{4}{c|}{Distributed Gauss-Seidel} & Centralized & \multicolumn{2}{c|}{ATE* (m)} & \multicolumn{2}{c|}{ARE* (deg)}\\
\cline{3-11}
\multicolumn{2}{|c|}{noise} & \multicolumn{2}{c|}{$\eta\!=\!10^{-1}$} & \multicolumn{2}{c|}{$\eta\!=\!10^{-2}$} & GN &\multirow{2}{*}{Poses} & \multirow{2}{*}{Lmrks.} & \multirow{2}{*}{Poses} & \multirow{2}{*}{Lmrks.}\\
\cline{3-6}
$\sigma_r (^{\circ})$ & $\sigma_t (\text{m})$ & \#Iter & Cost & \#Iter & Cost & Cost  & & & &\\
\hline 
\hline 

1 & 0.1 & 5.0 & 12.7 & 6.0 & 12.7 & 12.5 & 1.8e-04 & 1.3e-04 & 7.5e-02 & 9.0e-02\\
 \hline
5 & 0.1 & 5.0 & 166.6 & 7.0 & 166.5 & 160.9 & 3.1e-03 & 2.1e-03 & 3.3e-01 & 4.0e-01\\
 \hline
10 & 0.2 & 5.0 & 666.2 & 8.0 & 665.9 & 643.4 & 1.3e-02 & 8.8e-03 & 6.7e-01 & 8.1e-01\\
 \hline
15 & 0.3 & 6.0 & 1498.3 & 10.0 & 1497.8 & 1447.2 & 3.0e-02 & 2.1e-02 & 1.0e+00 & 1.2e+00\\
 \hline
   \end{tabular}}
 \vspace{0.1cm}
\caption{\label{tab:sensitivityAnalysisTable_ObjectSLAM} 
Number of iterations, cost, ATE* and ARE* of our approach compared to a centralized Gauss-Newton method for increasing measurement noise. ATE* and ARE* are measured using $\eta\!=\!10^{-1}$ as stopping condition.}
\end{table*}

\myparagraph{Accuracy in the number of robots} 
\Fig\ref{fig:simulationExperiments} compares the object locations and trajectories estimated using multi-robot mapping and centralized mapping for the two scenarios. 
Videos showing the map building for the two scenarios are available at: \url{https://youtu.be/nXJamypPvVY} and \url{https://youtu.be/nYm2sSHuGjo}.

Table \ref{tab:numRobotsAnalysis_ObjectSLAM} reports the number of iterations and our accuracy metrics
(cost, ATE*, ARE*) for increasing number of robots.  The table confirms that the distributed approach is nearly as accurate as the centralized Gauss-Newton method and the number of iterations does not increase with increasing number of robots, making our approach scalable to large teams. Usually, few tens of iterations suffice to reach an accurate estimate. Note that even when the cost of the \DGS method is slightly higher than \GN, the actual mismatch in the 
pose estimates is negligible (in the order of millimeters for positions and less than a degree for rotations). 

\myparagraph{Sensitivity to measurement noise}
We further test the
accuracy of our approach by evaluating the
number of iterations, the cost, the ATE* and ARE* for increasing levels of noise. Table \ref{tab:sensitivityAnalysisTable_ObjectSLAM} 
shows that our approach is able to replicate the accuracy of the
centralized Gauss-Newton method, regardless of the noise level.


\begin{figure}[t]
\includegraphics[width=1.01\columnwidth, trim= 0cm 0cm 0cm 0cm, clip]{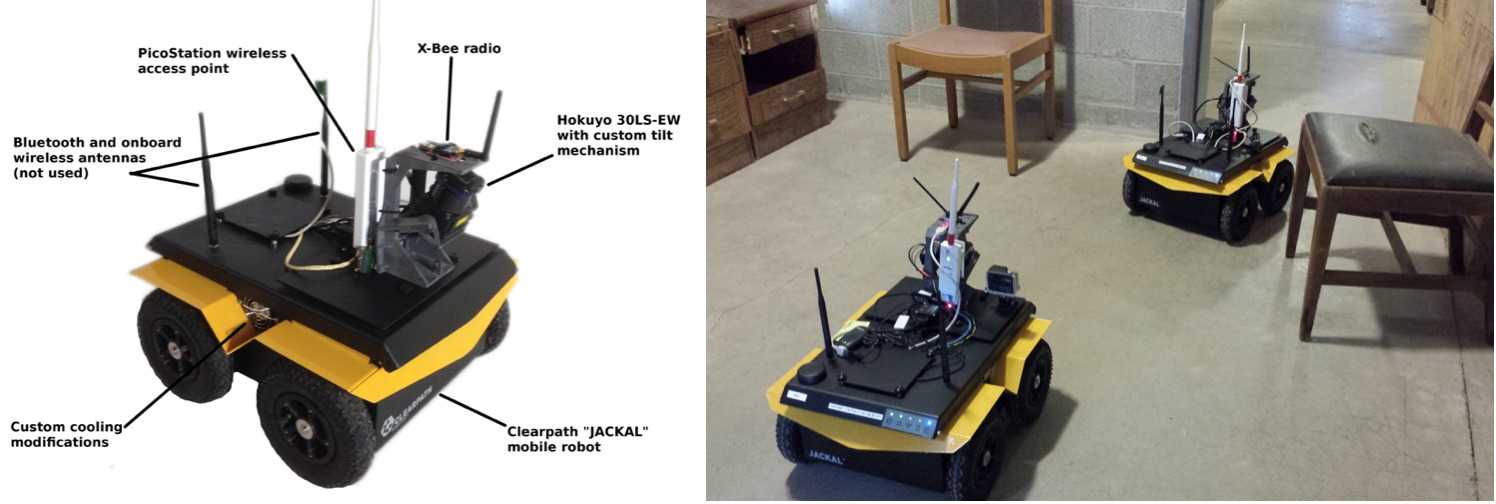}  \\
\caption{
(Left) Clearpath Jackal robot used for the field tests: platform and sensor layout; (right) snapshot of the 
test facility with the two Jackal robots. \label{fig:jackalRobot}
}
\end{figure}

\begin{figure}[t]
\includegraphics[width=1.01\columnwidth, trim= 0cm 0cm 0cm 0cm, clip]{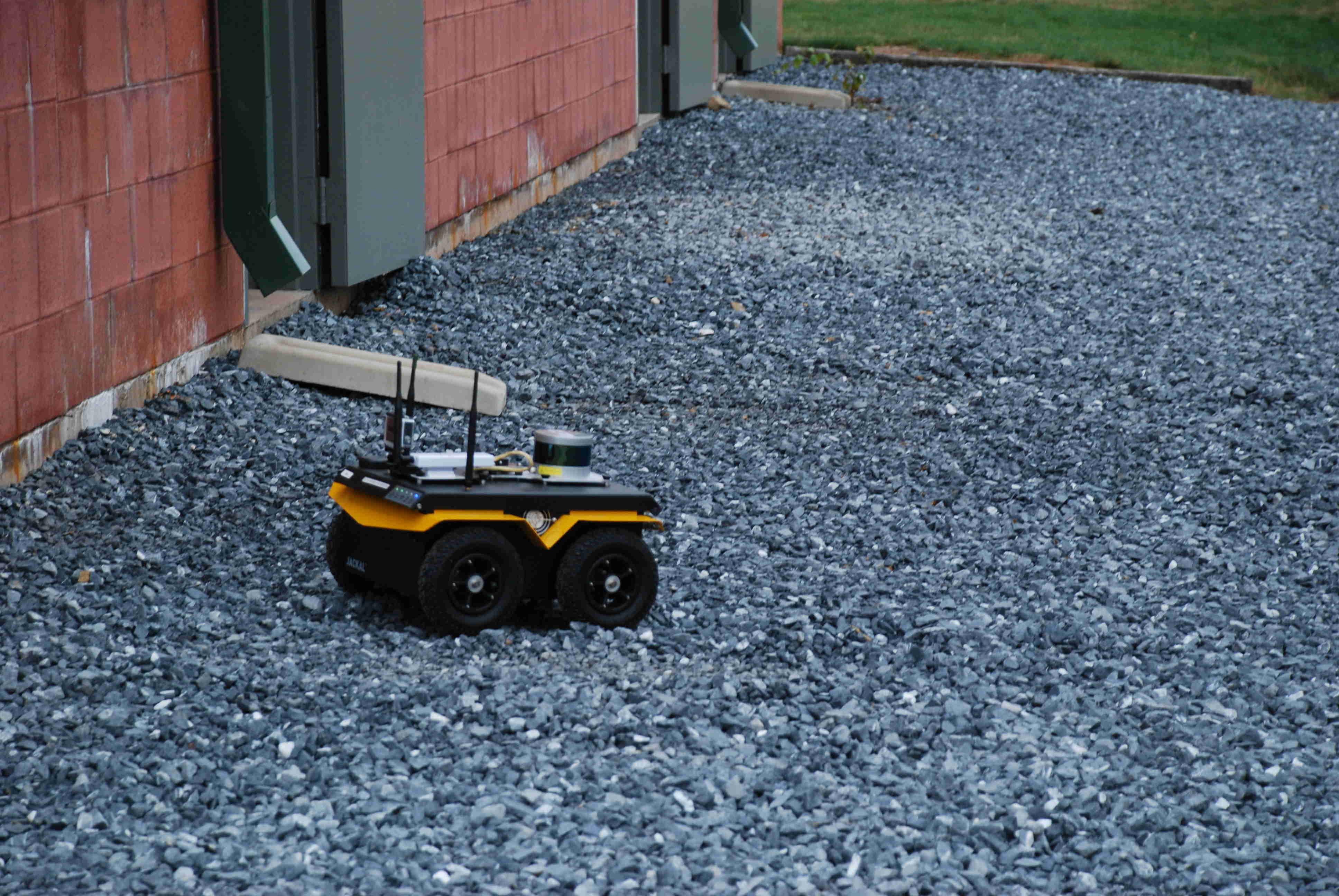}  \\
\caption{
Clearpath Jackal robot moving on gravel. \label{fig:jackalRobotOnGravel}
}
\end{figure}



\definecolor{dgreen}{rgb}{0,0.5,0}
\renewcommand{\scaleFig}{0.25}

\begin{figure*}[t]
\begin{minipage}{0.7\columnwidth}
\begin{tabular}{c|ccc|c}%

 Point Cloud & \DGS &  DDF-SAM  & Centralized  & Occupancy Grid \\

\begin{minipage}{0.5\columnwidth}%
\centering%
\includegraphics[width=\columnwidth, trim=15cm 0cm 20cm 0cm,clip]{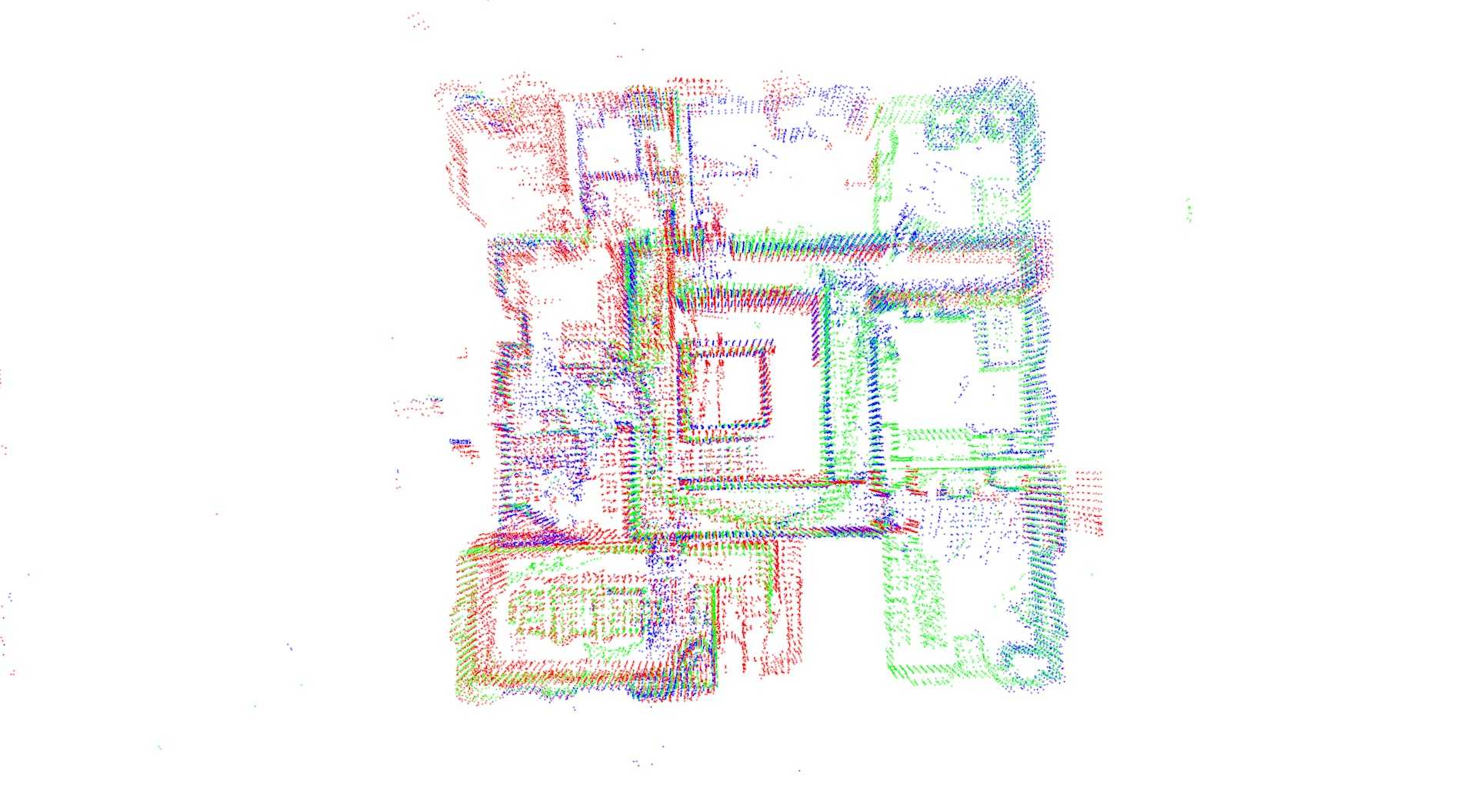} 
\end{minipage}
&
\begin{minipage}{0.5\columnwidth}%
\centering%
\includegraphics[width=\columnwidth, trim=0cm 0cm 0cm 0cm,clip]{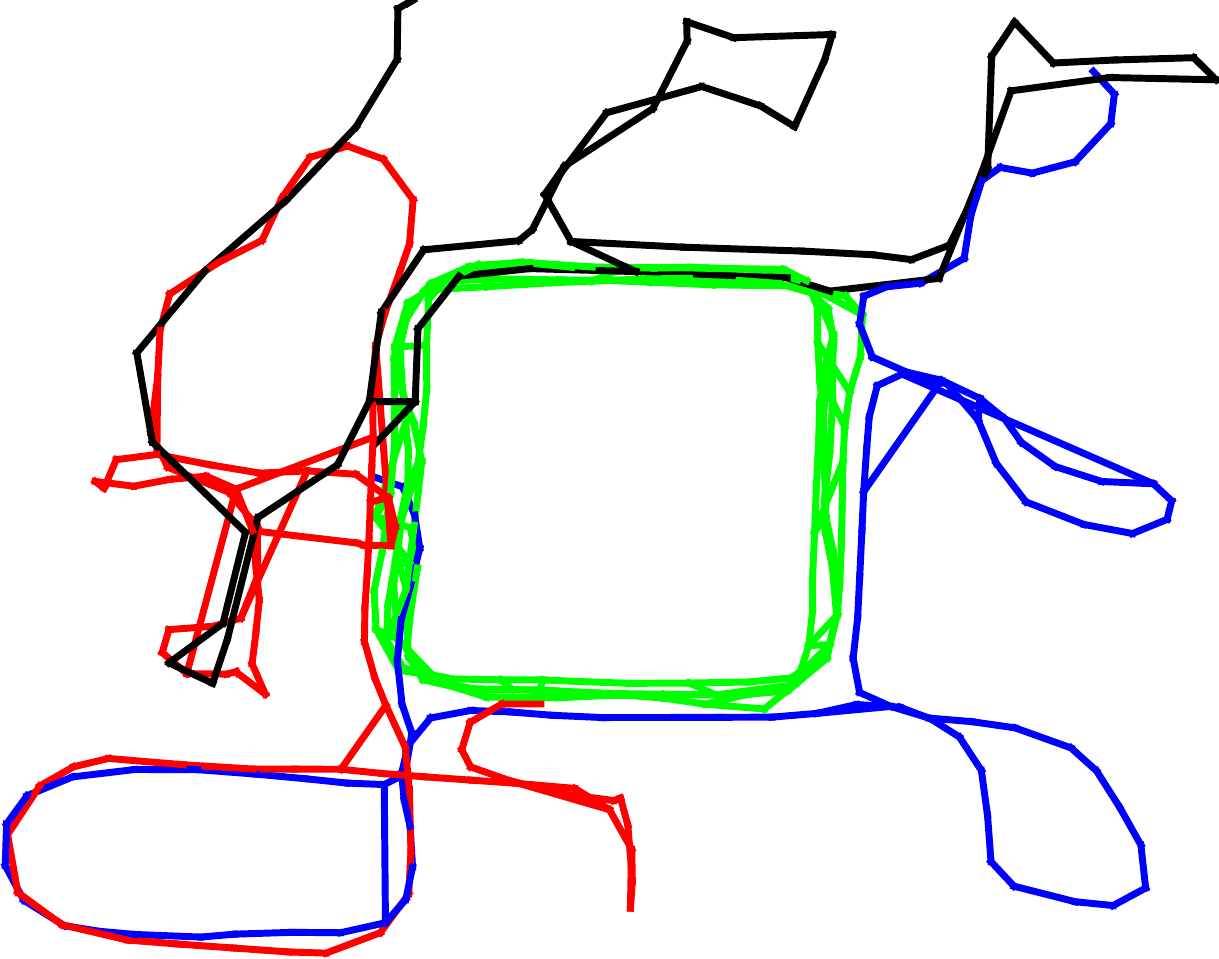} 
\end{minipage}
&
\begin{minipage}{0.5\columnwidth}%
\centering%
\includegraphics[width=\columnwidth, trim=0cm 0cm 0cm 0cm,clip]{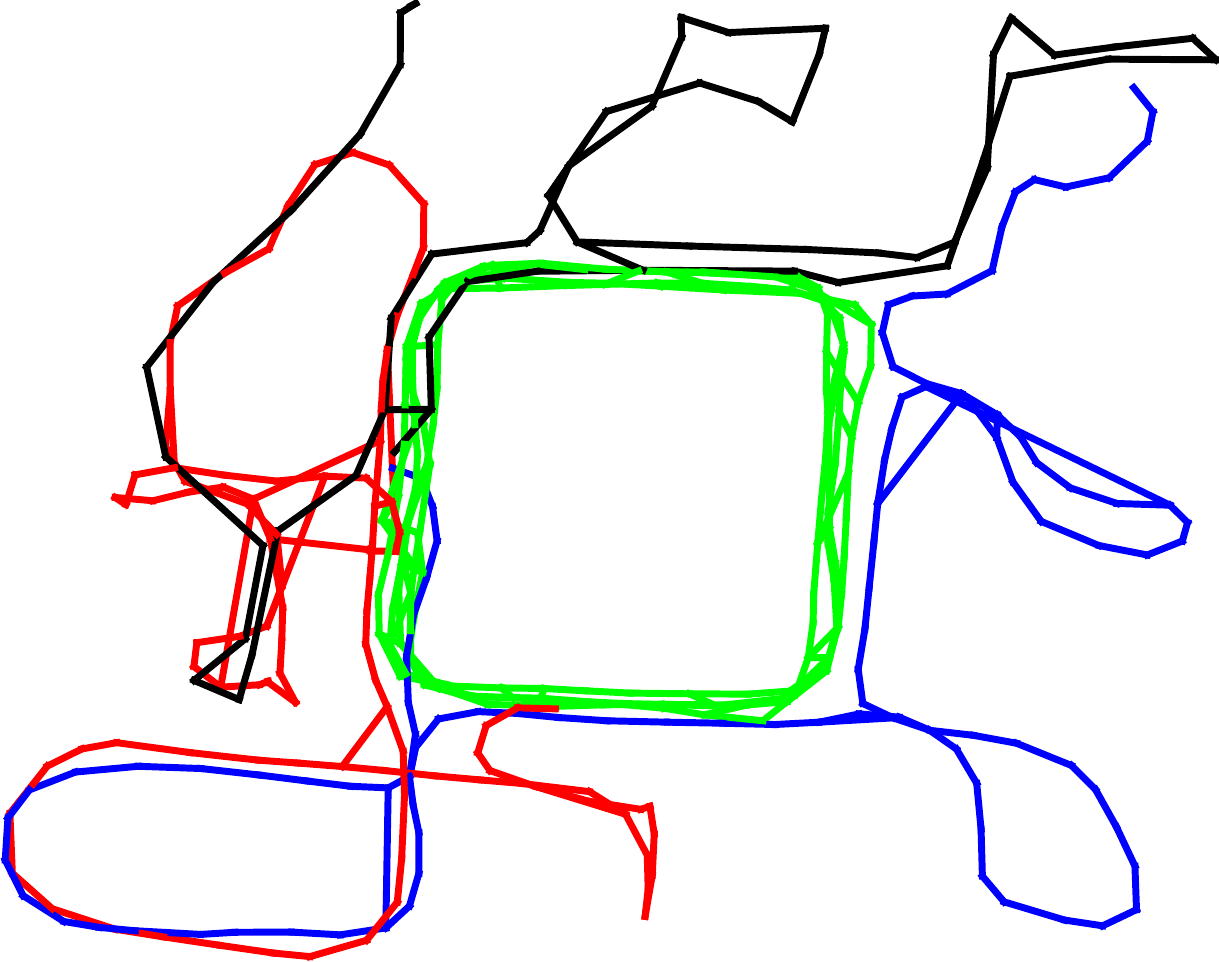} 
\end{minipage}

&
\begin{minipage}{0.5\columnwidth}%
\centering%
\includegraphics[width=\columnwidth, trim=0cm 0cm 0cm 0cm,clip]{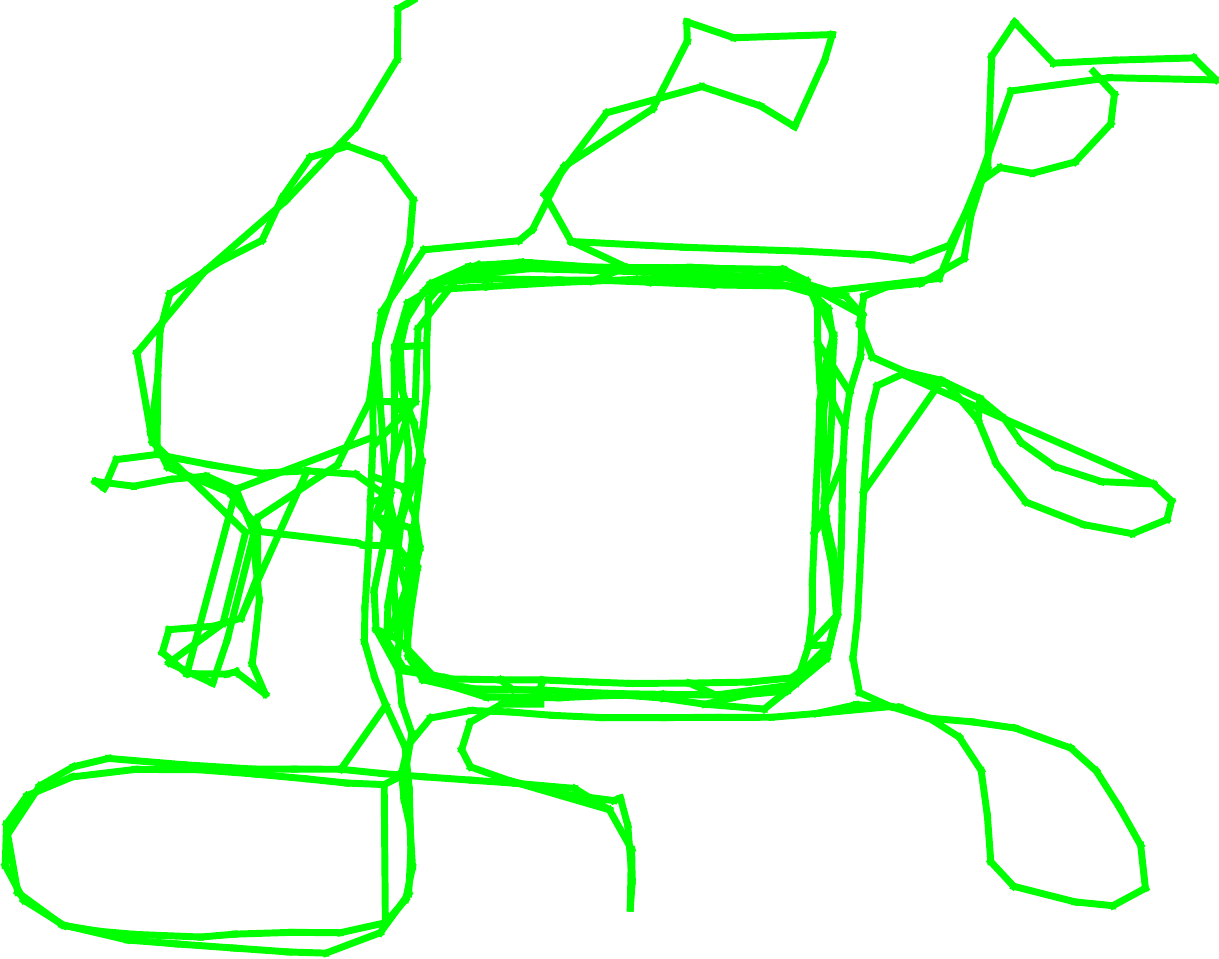} 
\end{minipage}
&
\begin{minipage}{0.5\columnwidth}%
\centering%
\includegraphics[width=\columnwidth, trim= 3.5cm 0cm 4.5cm 0cm, clip]{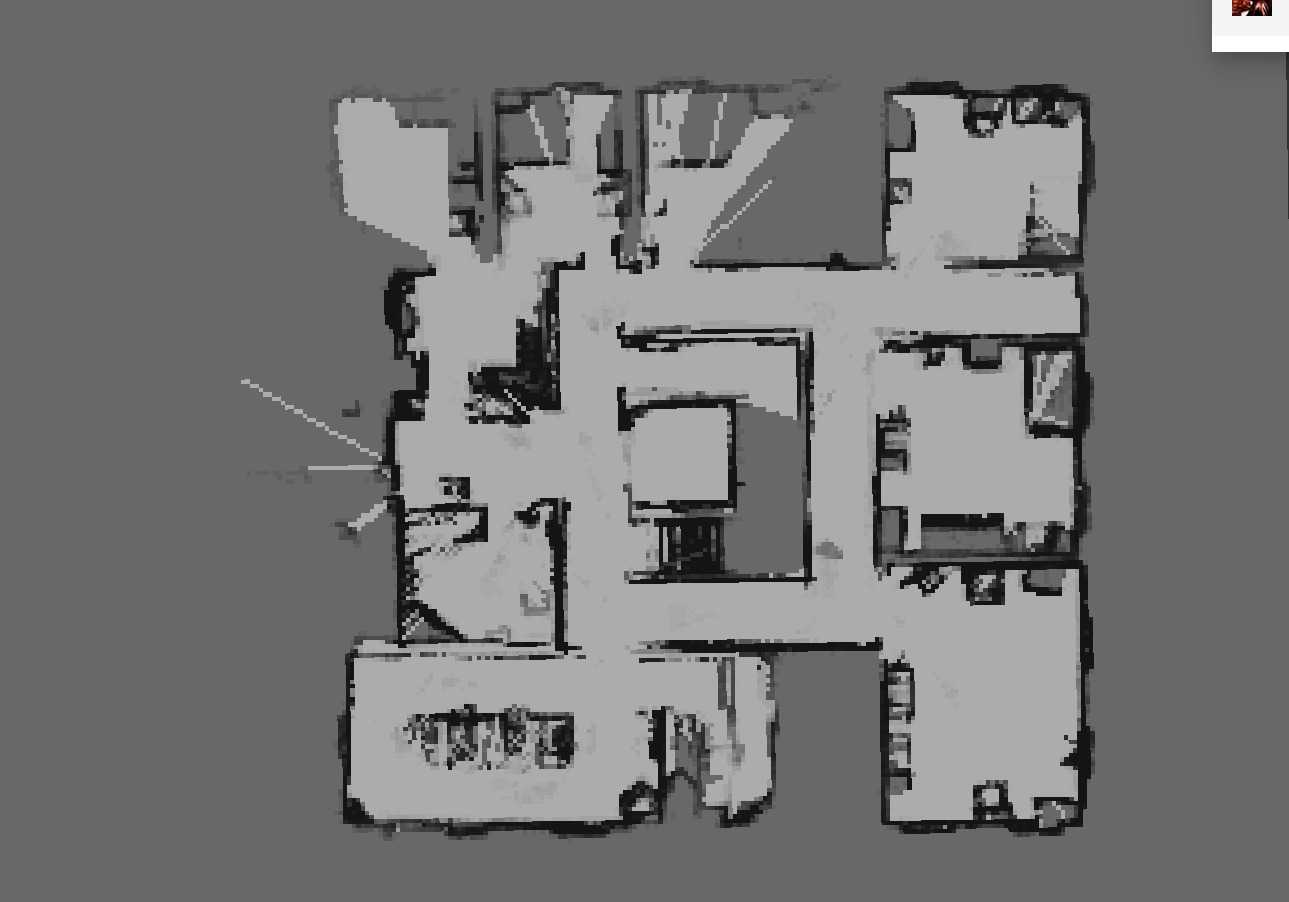} 
\end{minipage}\\
\vspace{0.2cm}

\begin{minipage}{0.5\columnwidth}%
\centering%
\includegraphics[width=\columnwidth, trim=15cm 0cm 20cm 0cm,clip]{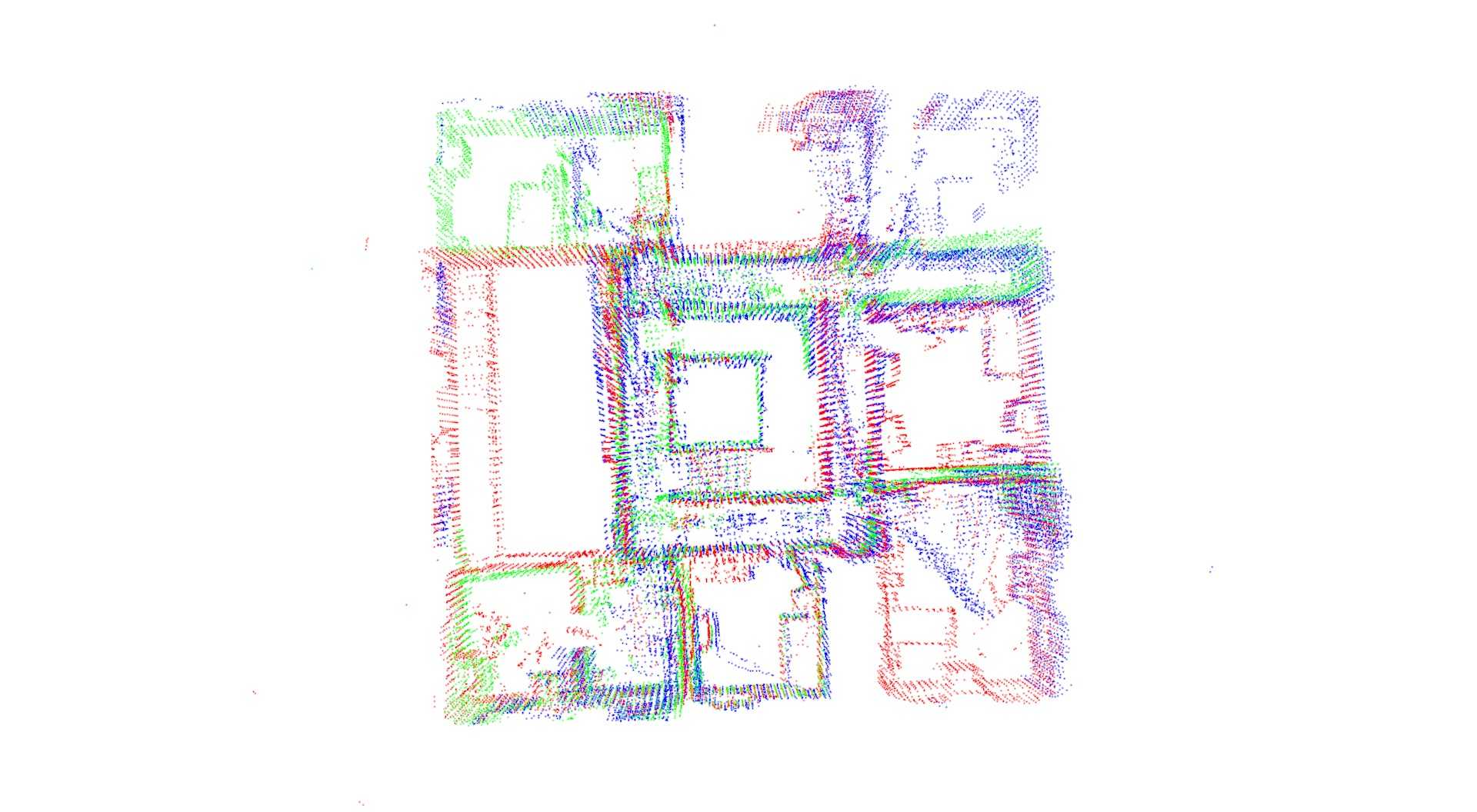} 
\end{minipage}
&
\begin{minipage}{0.5\columnwidth}%
\centering%
\includegraphics[width=\columnwidth, trim=0cm 0cm 0cm 0cm,clip]{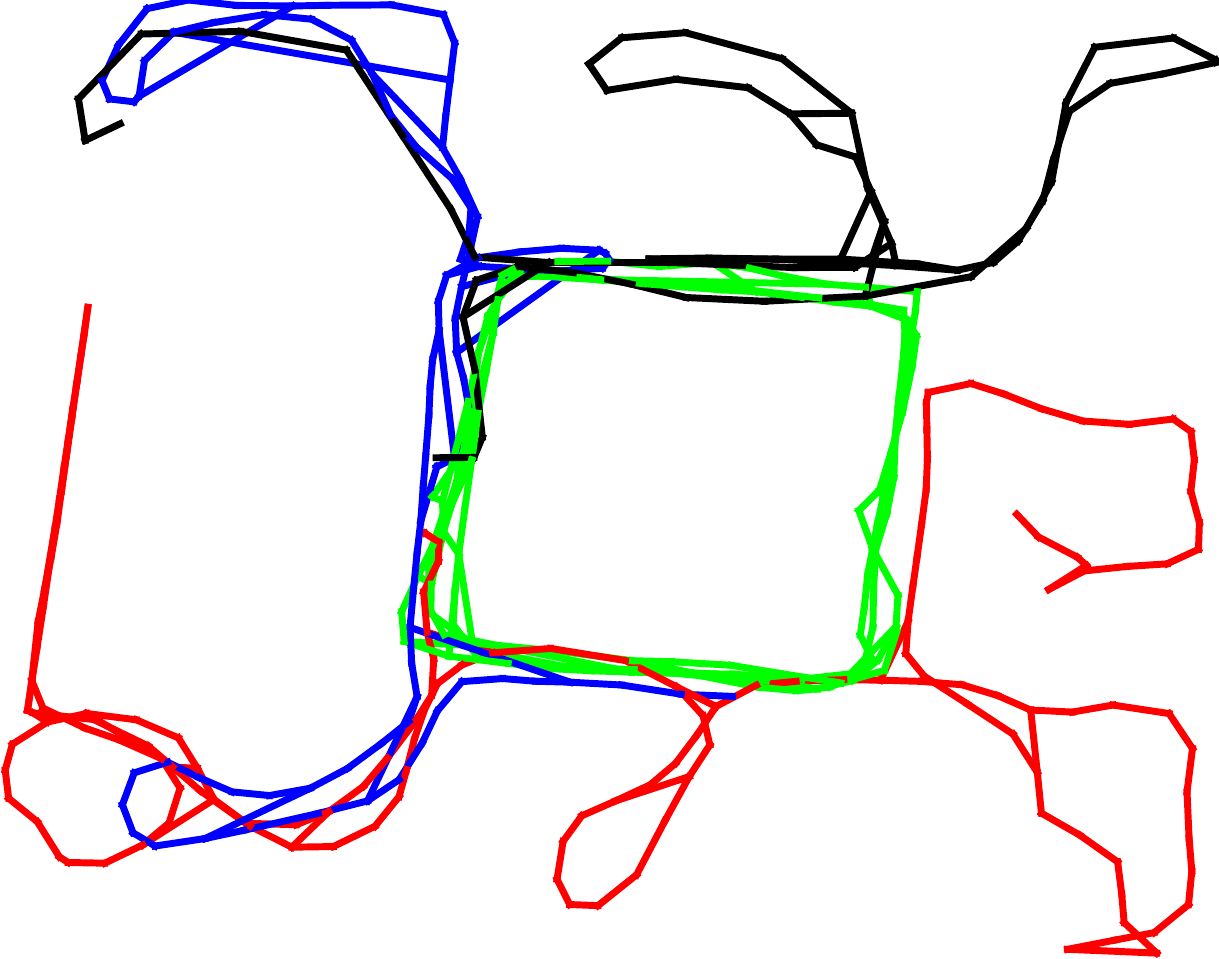} 
\end{minipage}
&
\begin{minipage}{0.5\columnwidth}%
\centering%
\includegraphics[width=\columnwidth, trim=0cm 0cm 0cm 0cm,clip]{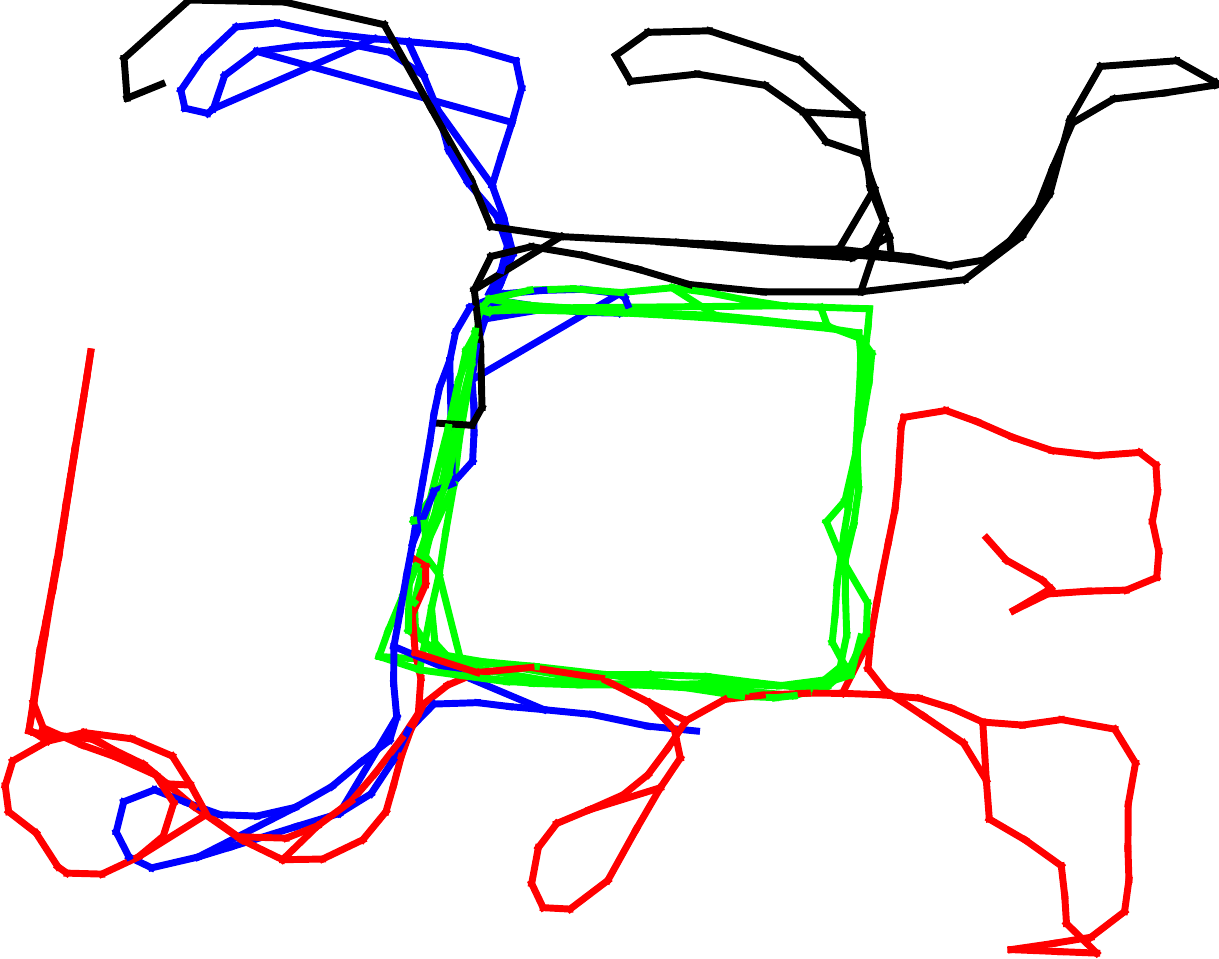} 
\end{minipage}

&
\begin{minipage}{0.5\columnwidth}%
\centering%
\includegraphics[width=\columnwidth, trim=0cm 0cm 0cm 0cm,clip]{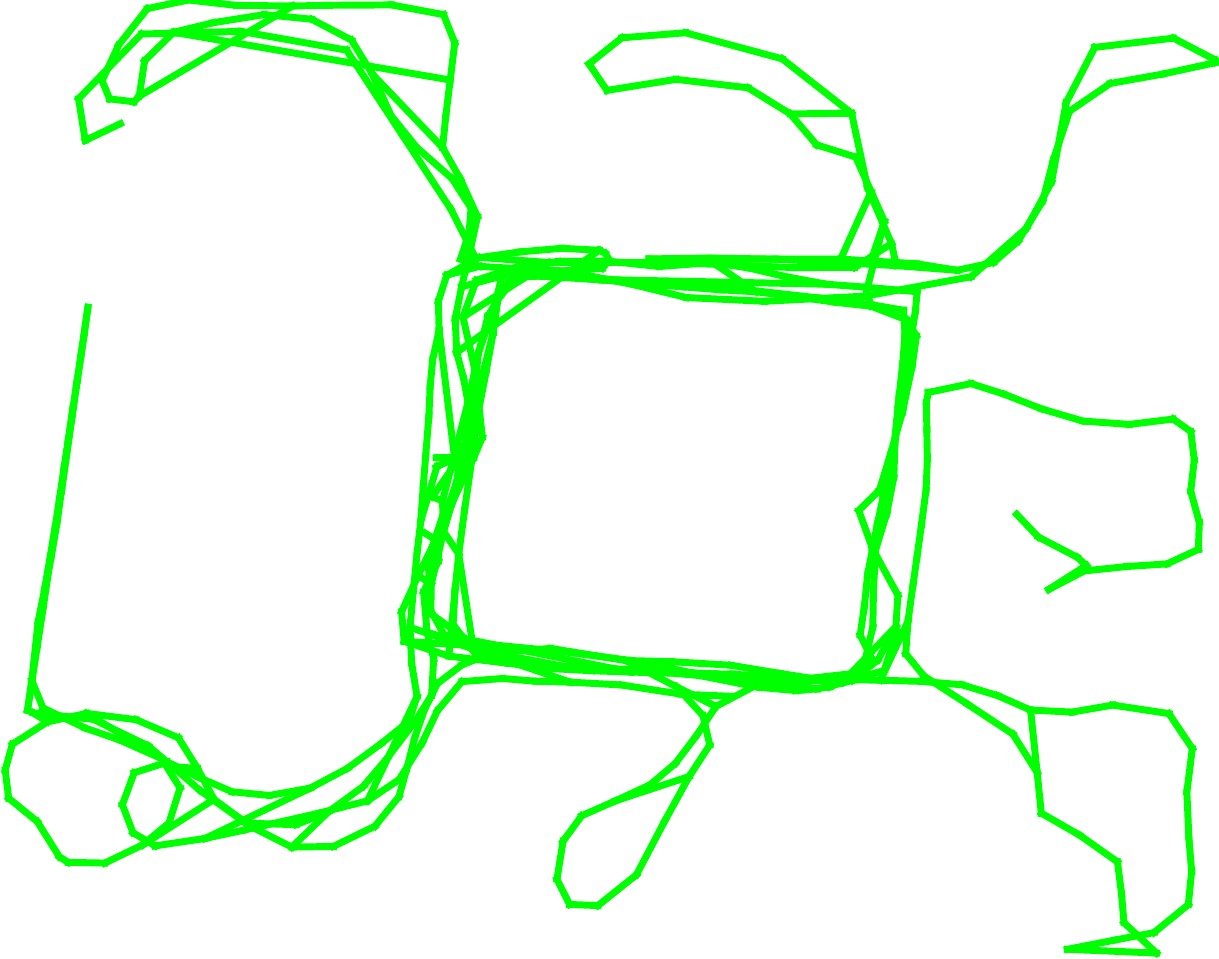} 
\end{minipage}
&
\begin{minipage}{0.5\columnwidth}%
\centering%
\includegraphics[width=\columnwidth, trim= 3.5cm 0cm 4.5cm 0cm, clip]{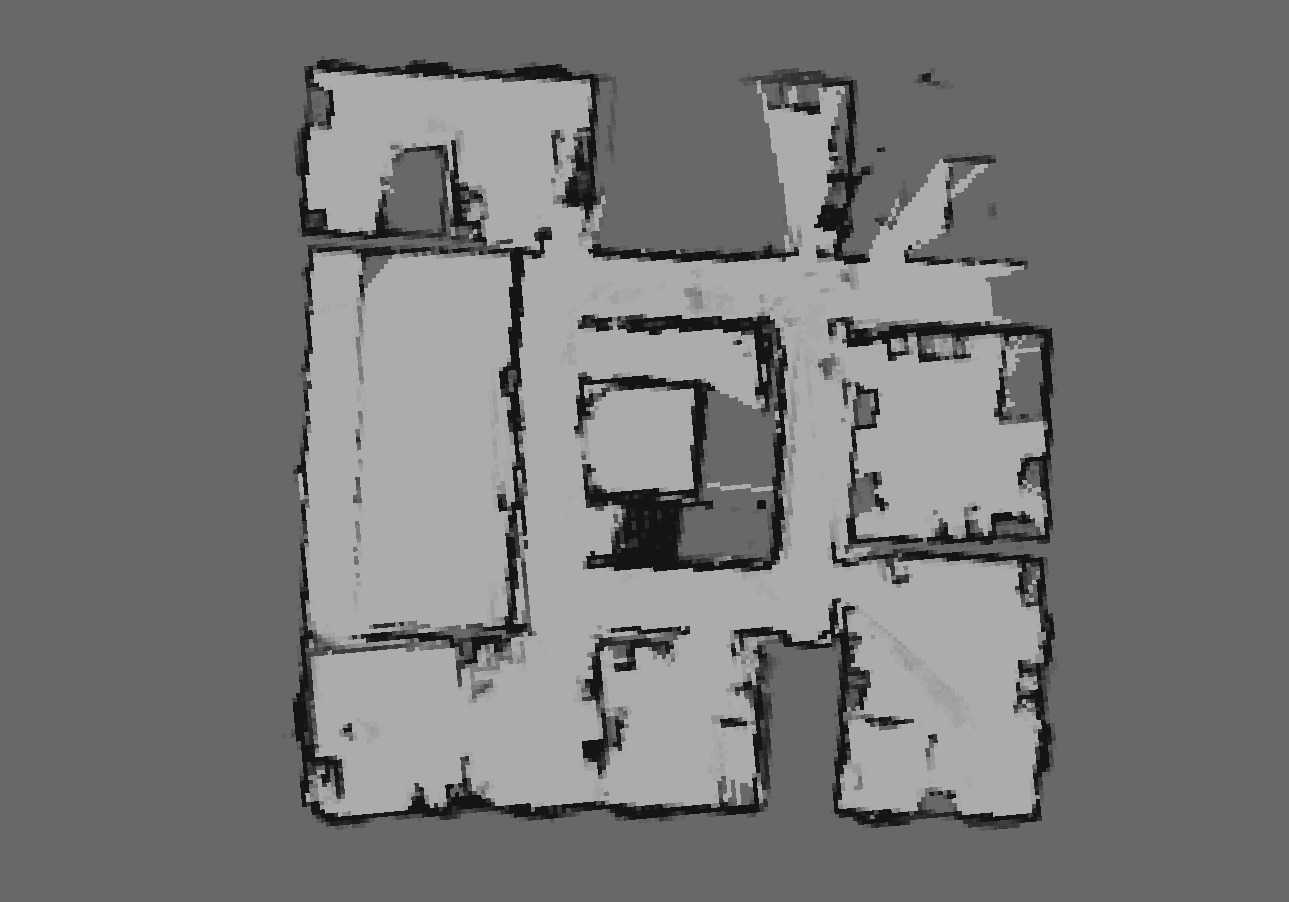} 
\end{minipage}\\
\vspace{0.2cm}

\begin{minipage}{0.5\columnwidth}%
\centering%
\includegraphics[width=\columnwidth, trim=15cm 0cm 20cm 0cm,clip]{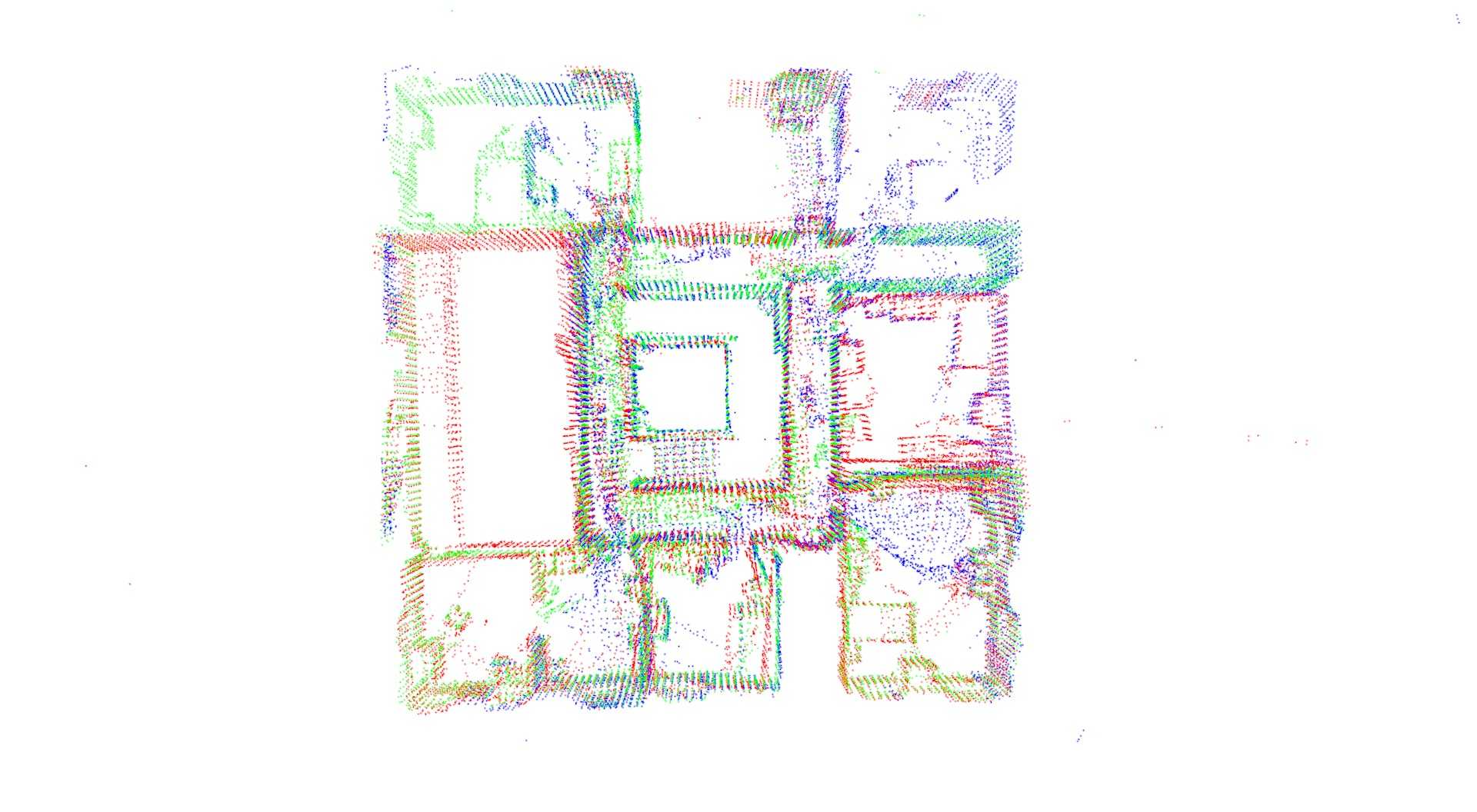} 
\end{minipage}
&
\begin{minipage}{0.5\columnwidth}%
\centering%
\includegraphics[width=\columnwidth, trim=0cm 0cm 0cm 0cm,clip]{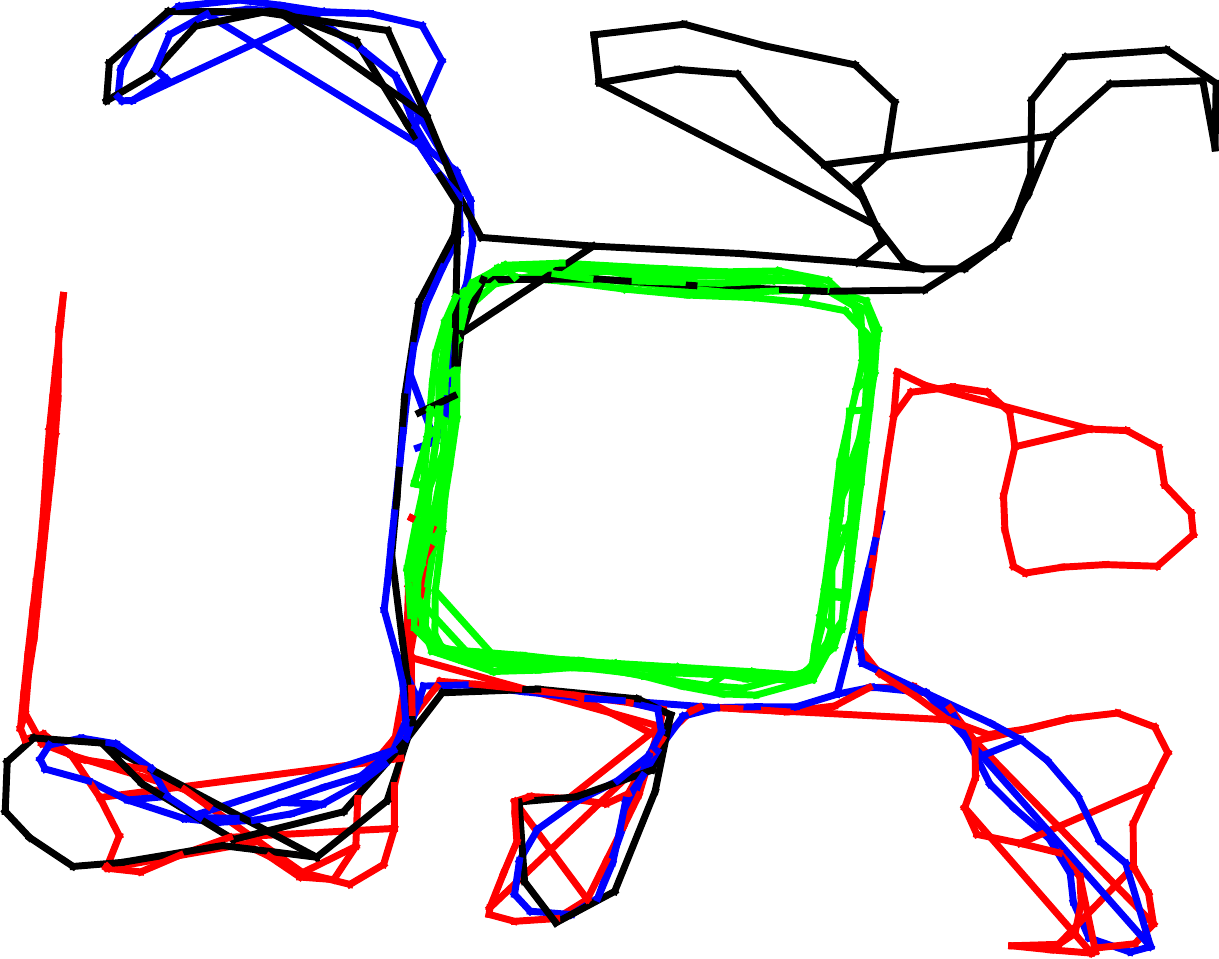} 
\end{minipage}
&
\begin{minipage}{0.5\columnwidth}%
\centering%
\includegraphics[width=\columnwidth, trim=0cm 0cm 0cm 0cm,clip]{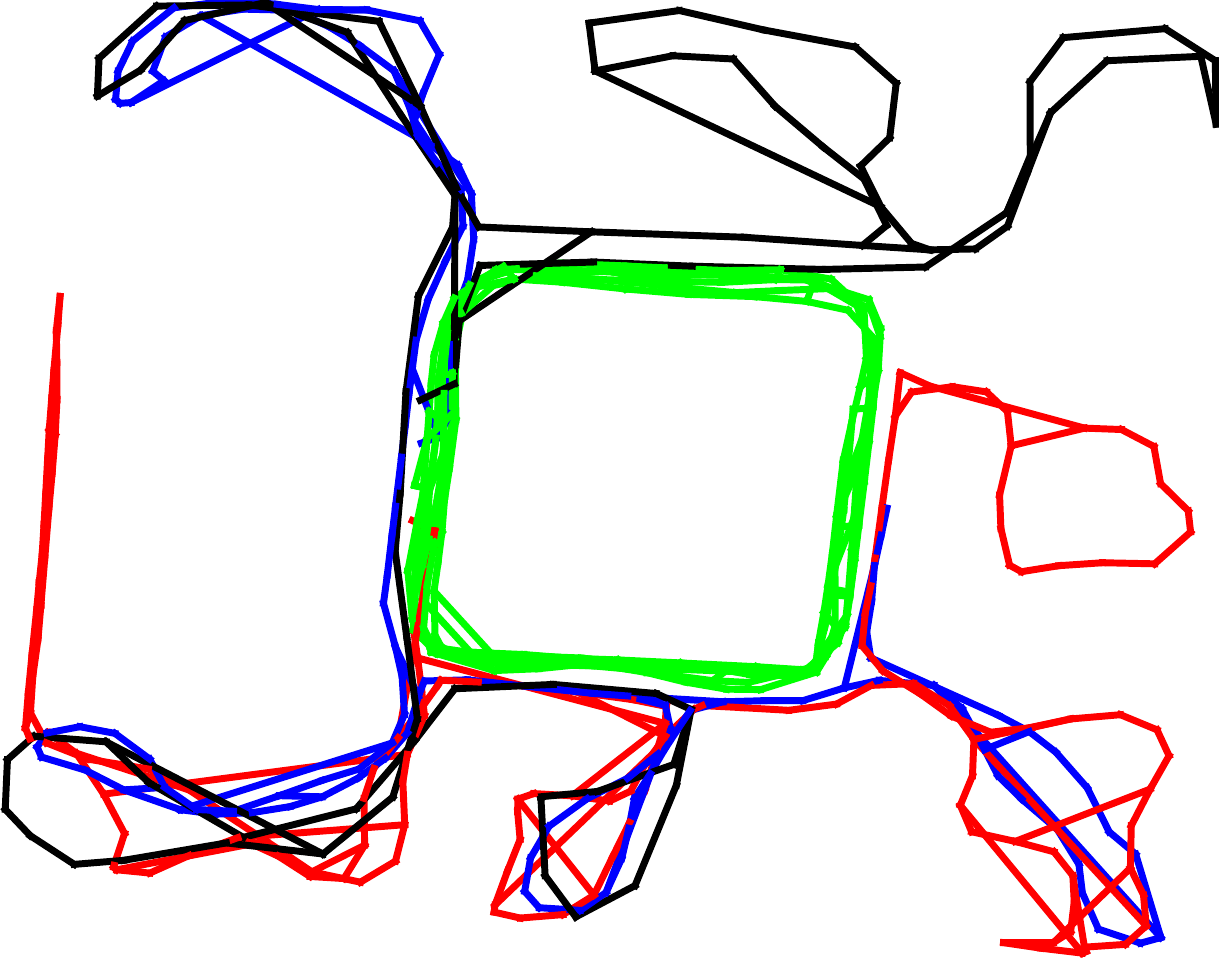} 
\end{minipage}

&
\begin{minipage}{0.5\columnwidth}%
\centering%
\includegraphics[width=\columnwidth, trim=0cm 0cm 0cm 0cm,clip]{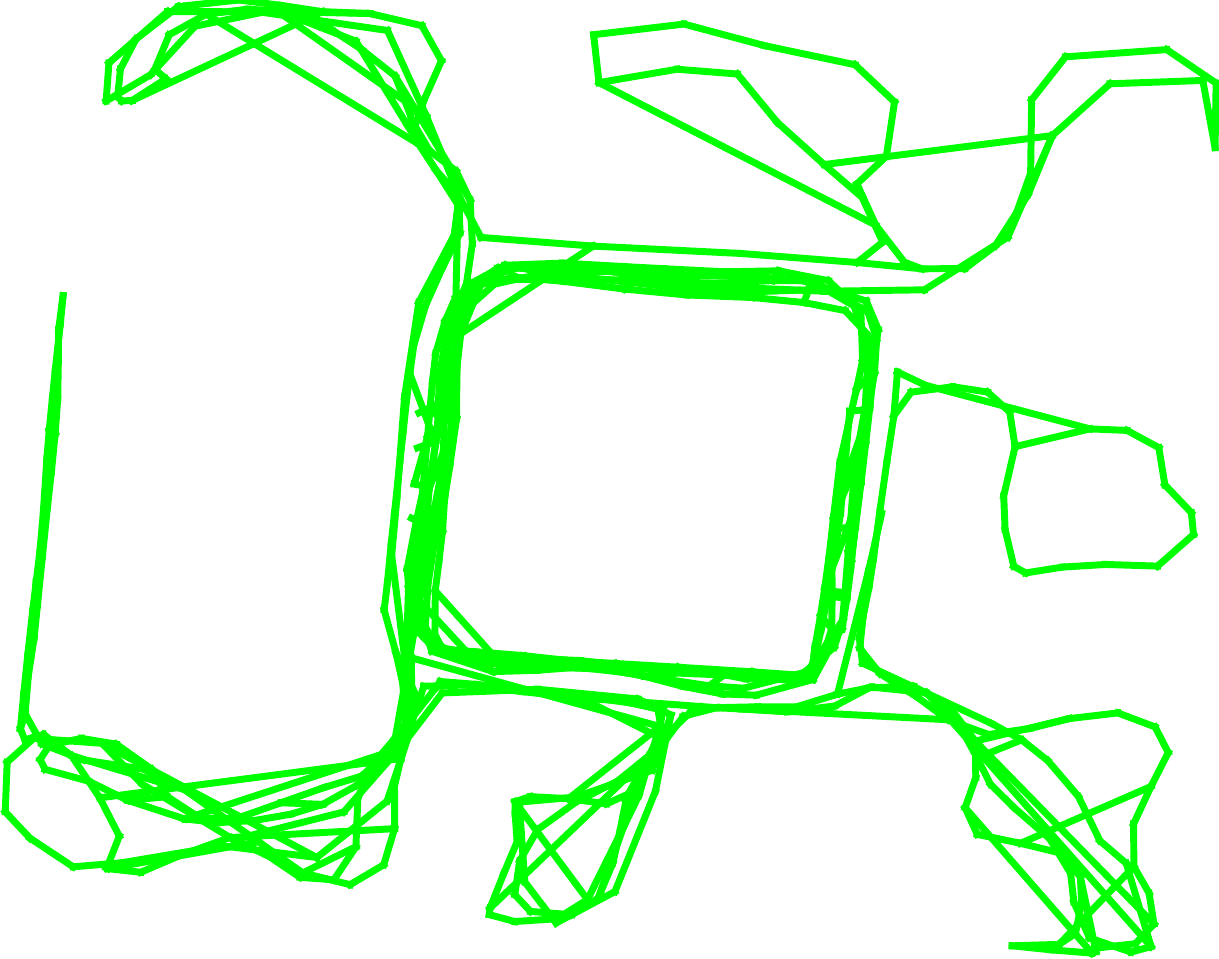} 
\end{minipage}
&
\begin{minipage}{0.5\columnwidth}%
\centering%
\includegraphics[width=\columnwidth, trim= 3.5cm 0cm 4.5cm 0cm, clip]{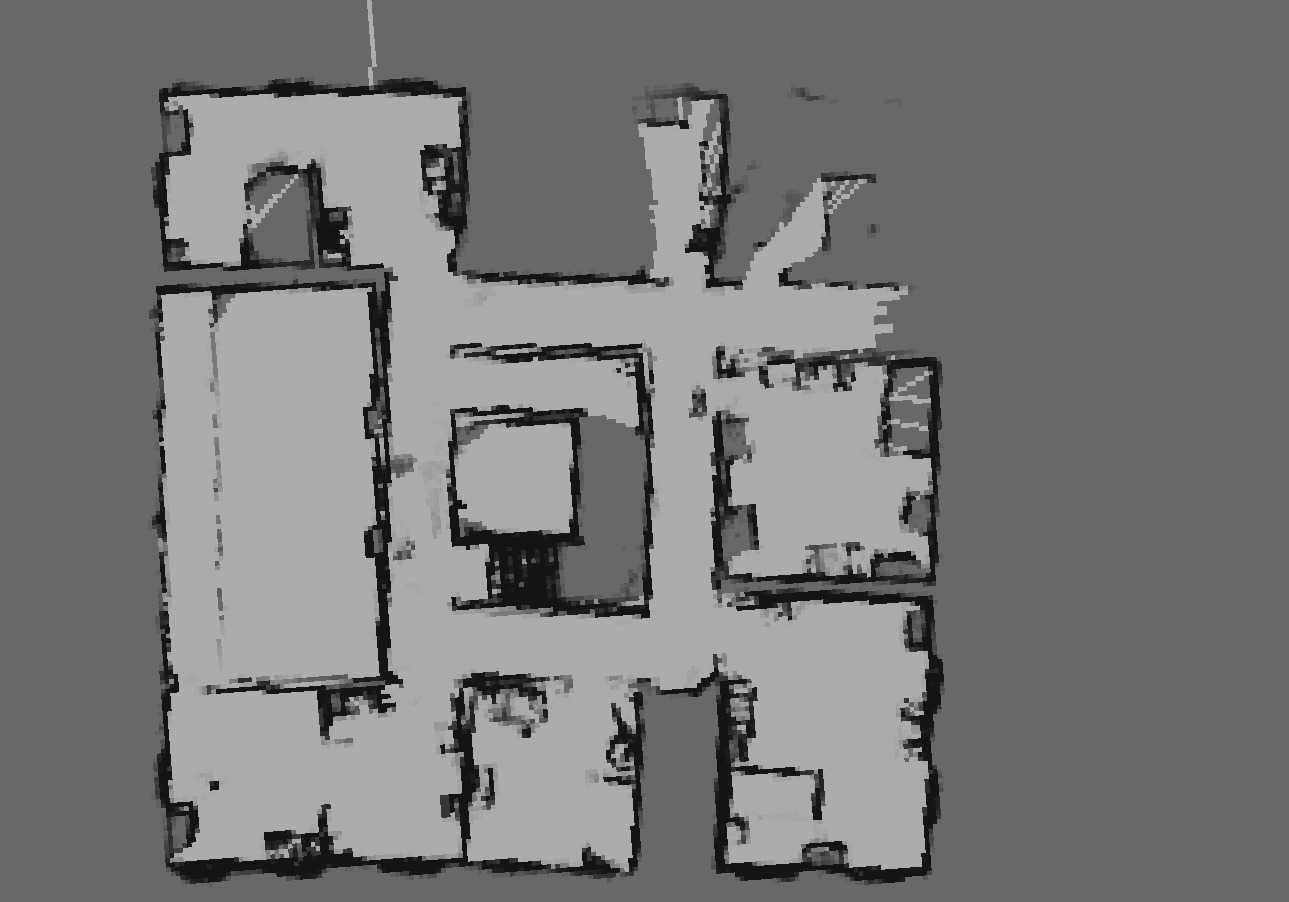} \end{minipage}\\
\vspace{0.2cm}

\begin{minipage}{0.5\columnwidth}%
\centering%
\includegraphics[width=\columnwidth, trim=15cm 0cm 20cm 0cm,clip]{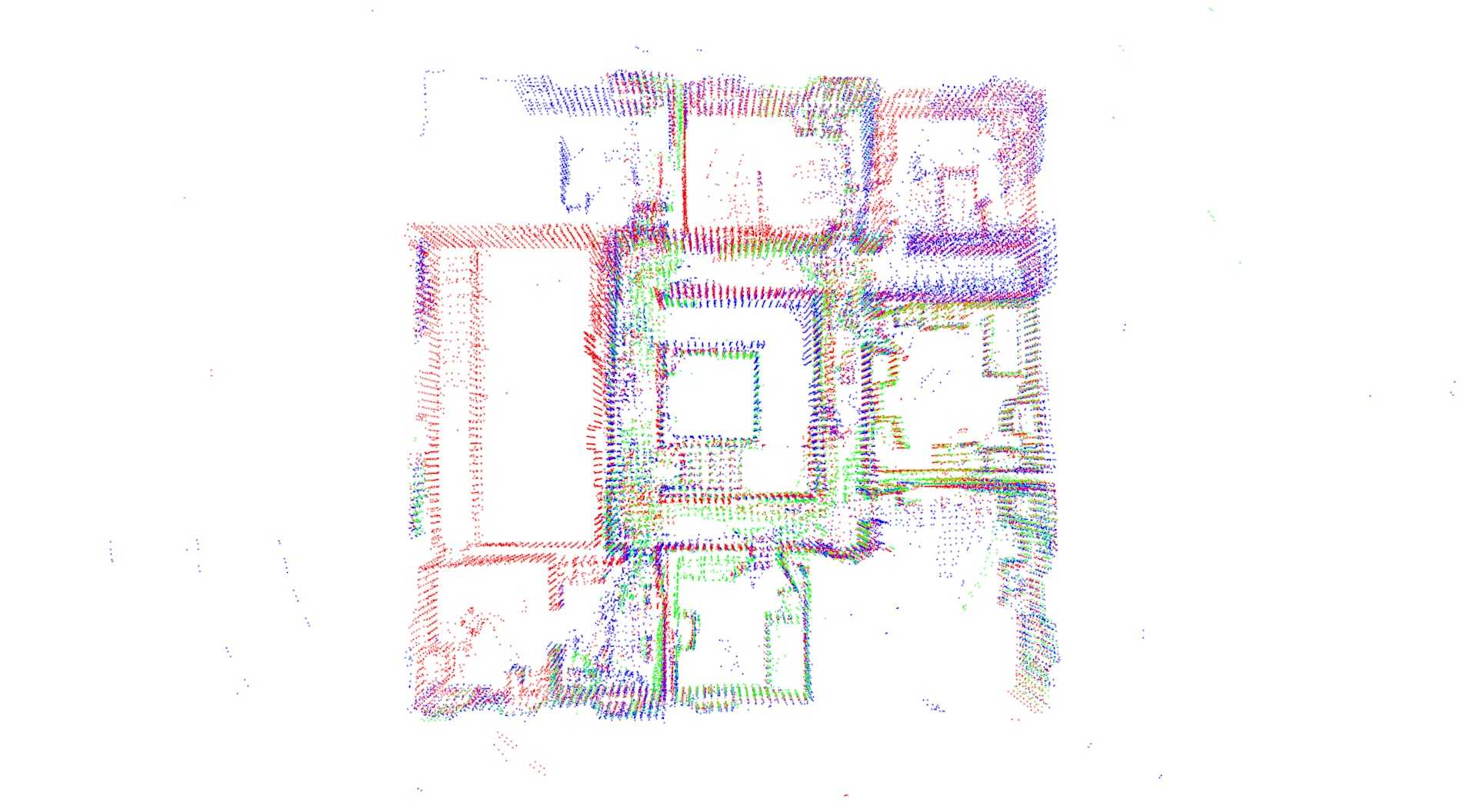} 
\end{minipage}
&
\begin{minipage}{0.5\columnwidth}%
\centering%
\includegraphics[width=\columnwidth, trim=0cm 0cm 0cm 0cm,clip]{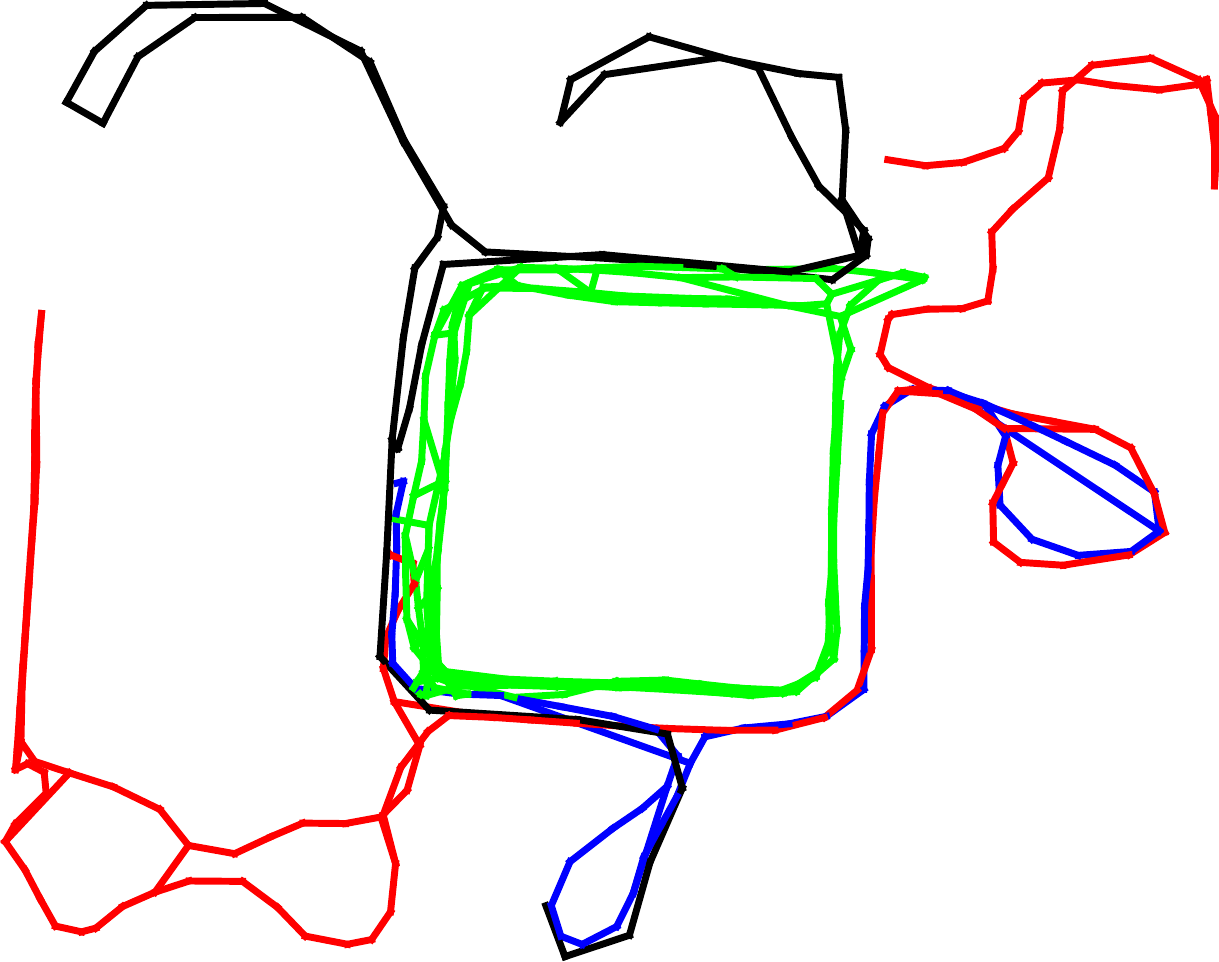} 
\end{minipage}
&
\begin{minipage}{0.5\columnwidth}%
\centering%
\includegraphics[width=\columnwidth, trim=0cm 0cm 0cm 0cm,clip]{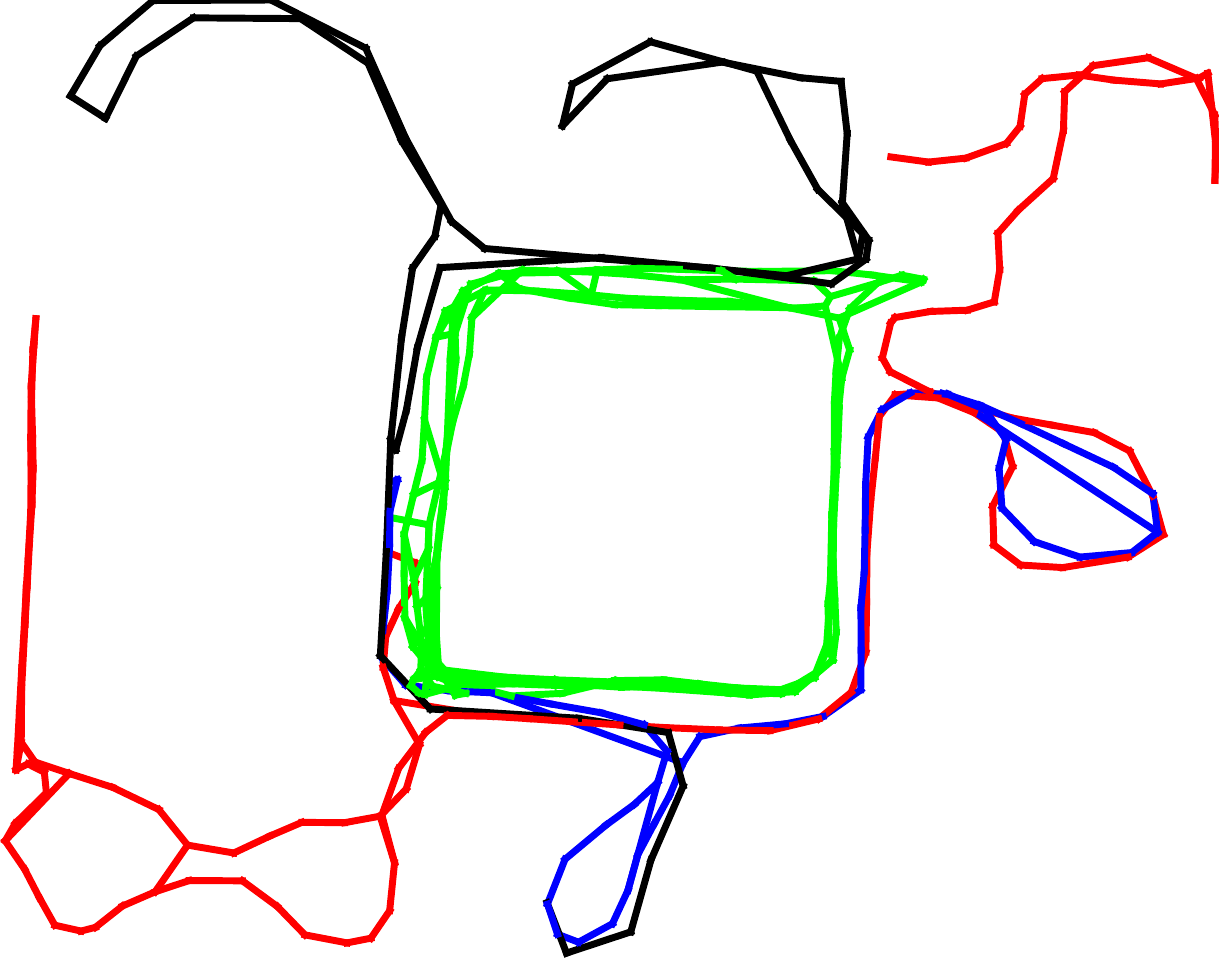} 
\end{minipage}

&
\begin{minipage}{0.5\columnwidth}%
\centering%
\includegraphics[width=\columnwidth, trim=0cm 0cm 0cm 0cm,clip]{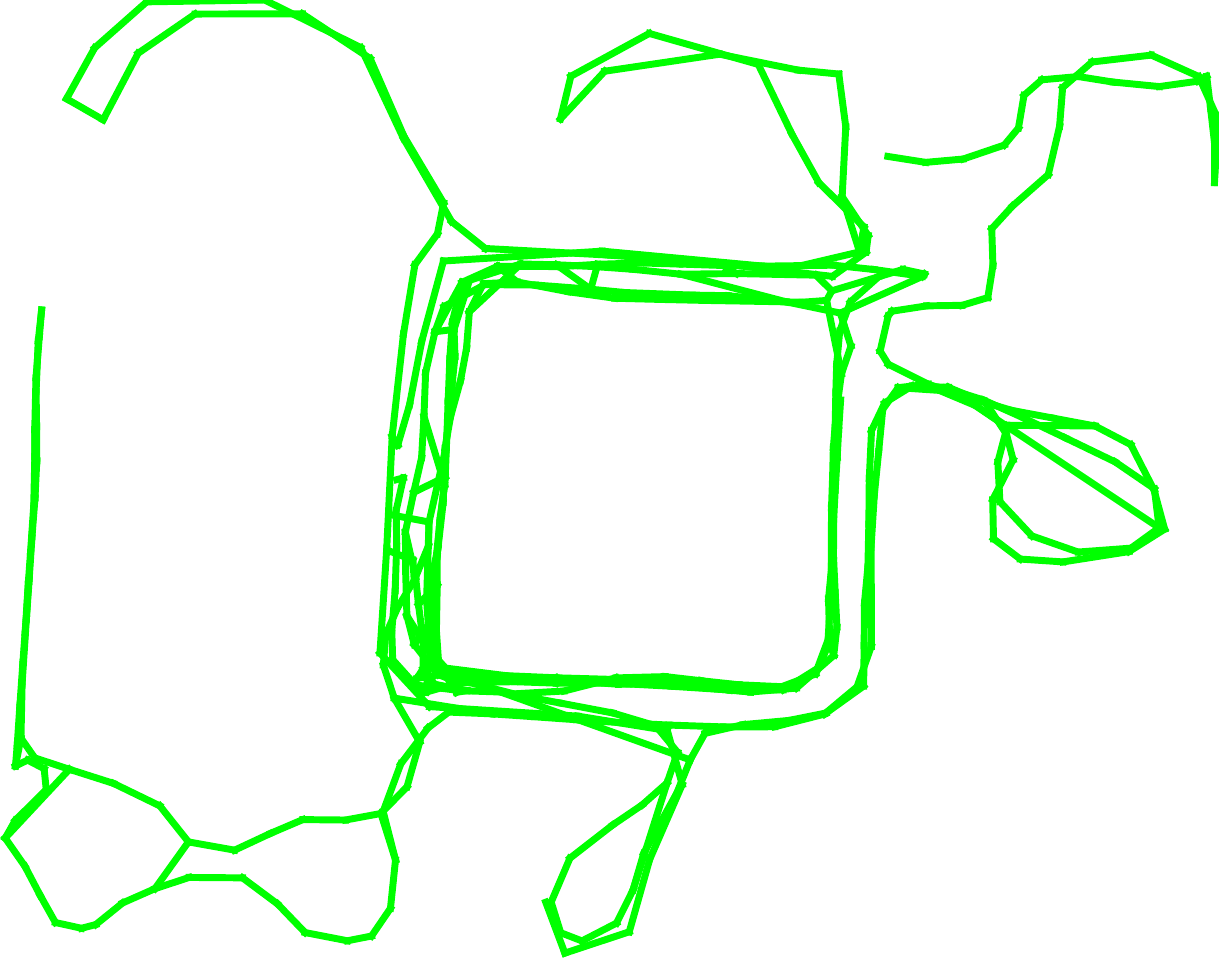} 
\end{minipage}
&
\begin{minipage}{0.5\columnwidth}%
\centering%
\includegraphics[width=\columnwidth, trim= 3.5cm 0cm 4.5cm 0cm, clip]{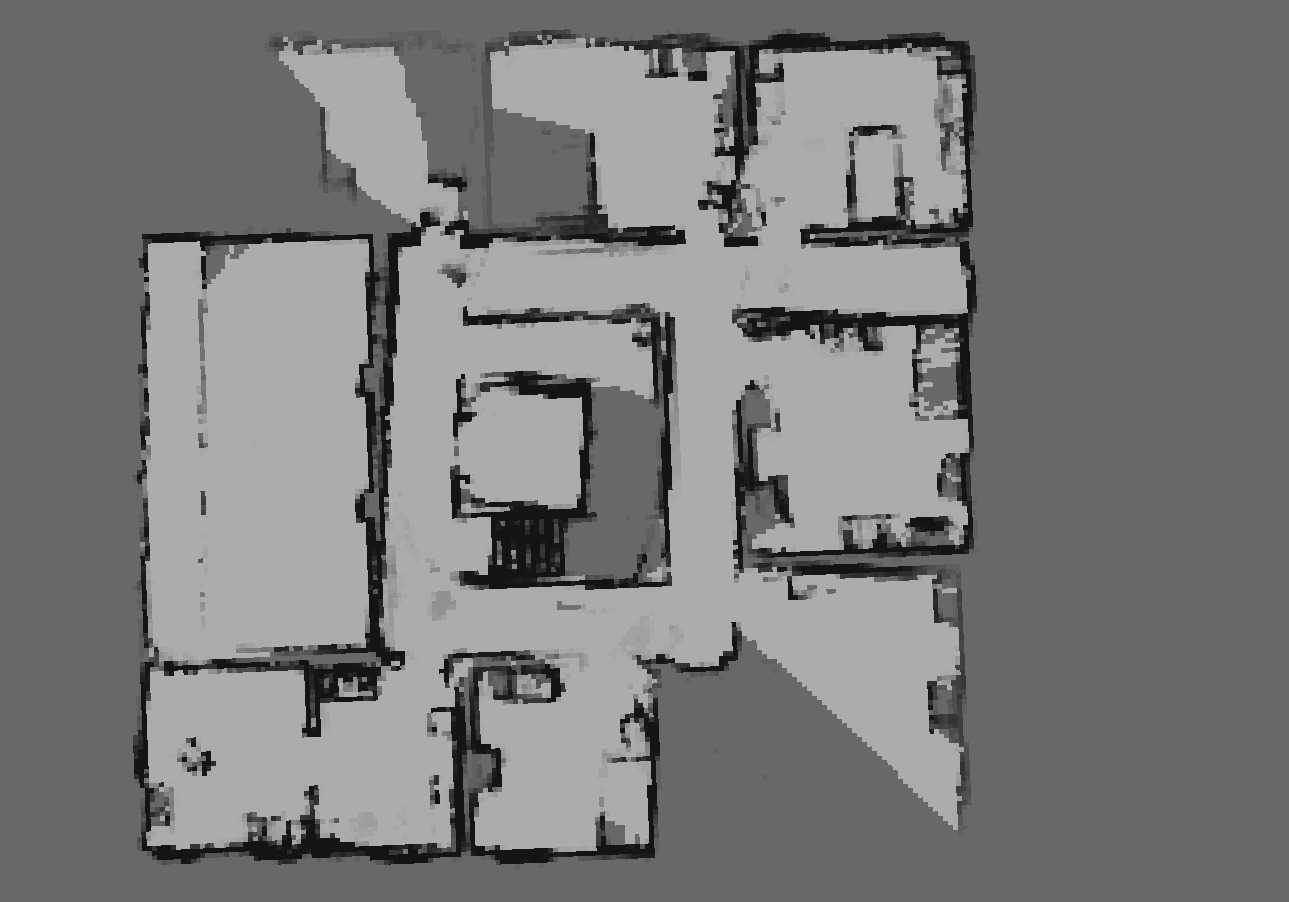} \end{minipage}\\
\vspace{0.2cm}

\begin{minipage}{0.5\columnwidth}%
\centering%
\includegraphics[width=\columnwidth, trim=15cm 0cm 20cm 0cm,clip]{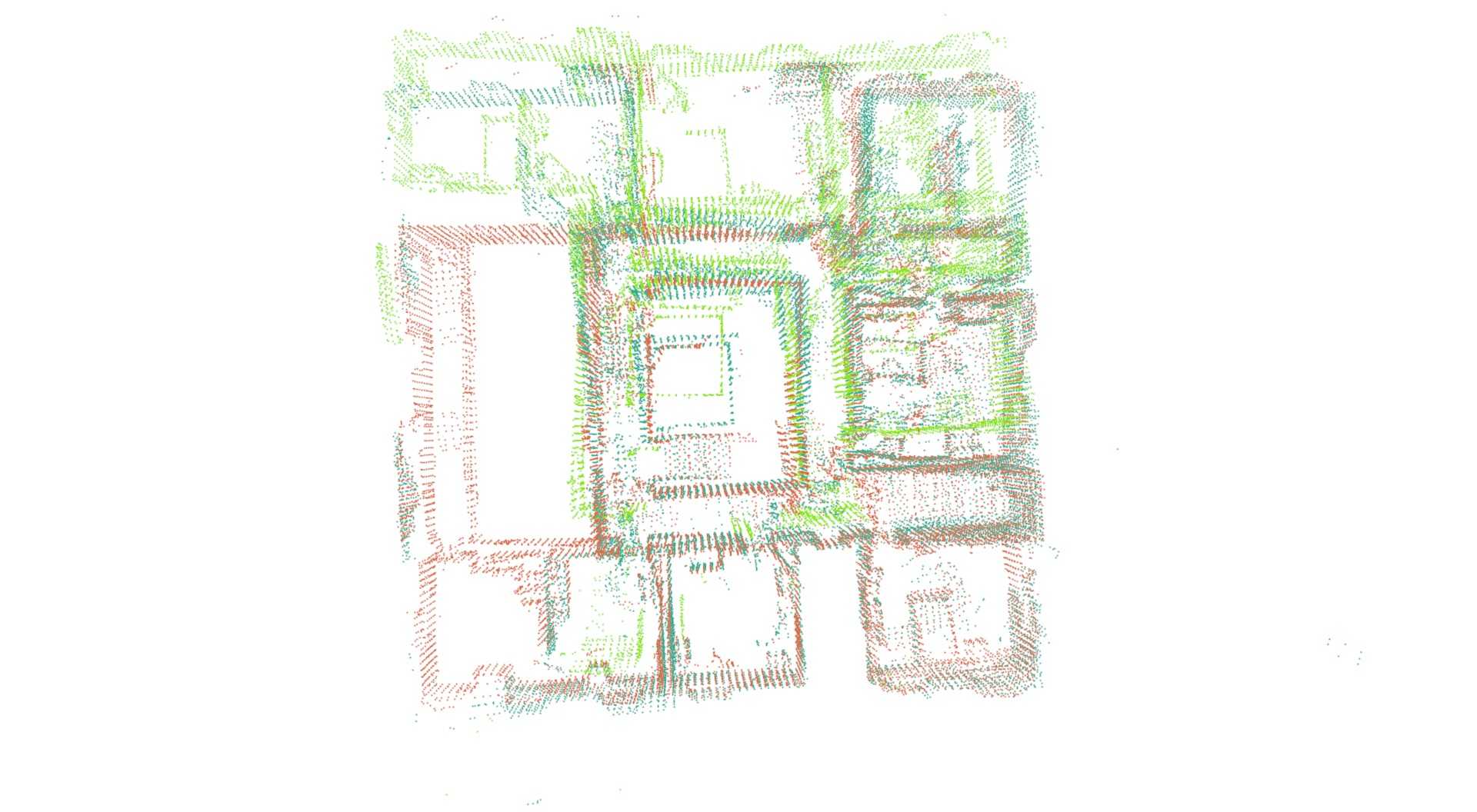} 
\end{minipage}
&
\begin{minipage}{0.5\columnwidth}%
\centering%
\includegraphics[width=\columnwidth, trim=0cm 0cm 0cm 0cm,clip]{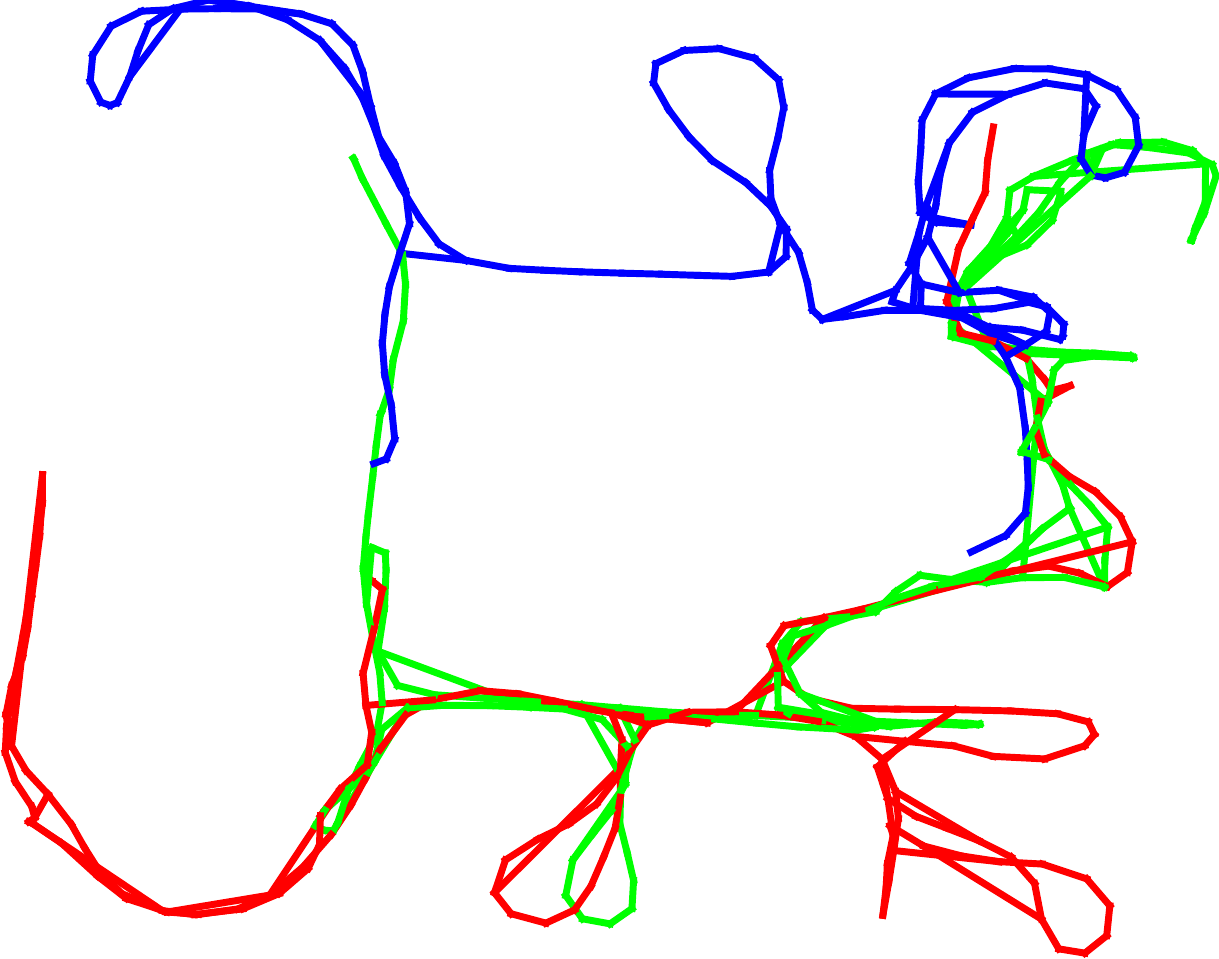} 
\end{minipage}
&
\begin{minipage}{0.5\columnwidth}%
\centering%
\includegraphics[width=\columnwidth, trim=0cm 0cm 0cm 0cm,clip]{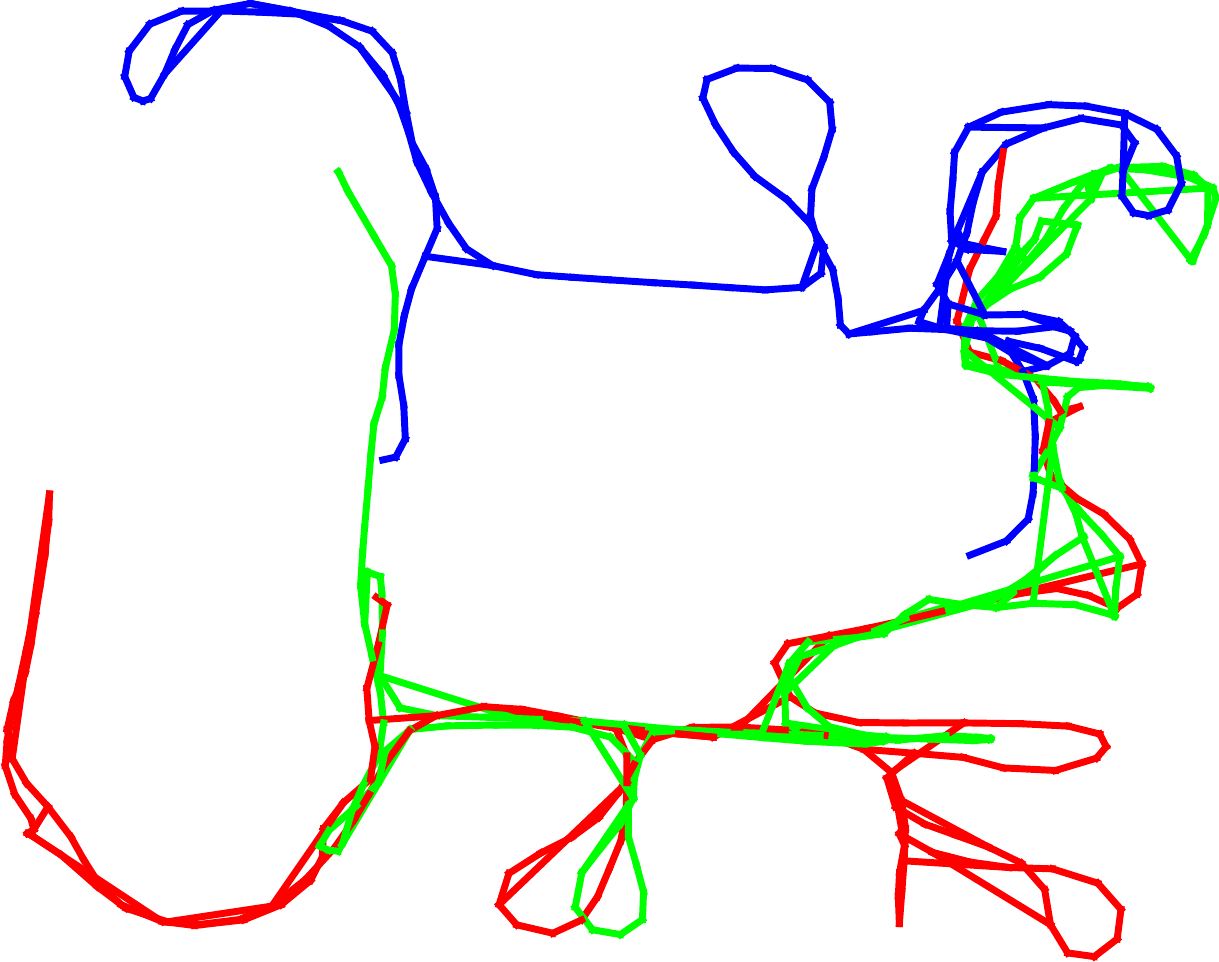} 
\end{minipage}
&
\begin{minipage}{0.5\columnwidth}%
\centering%
\includegraphics[width=\columnwidth, trim=0cm 0cm 0cm 0cm,clip]{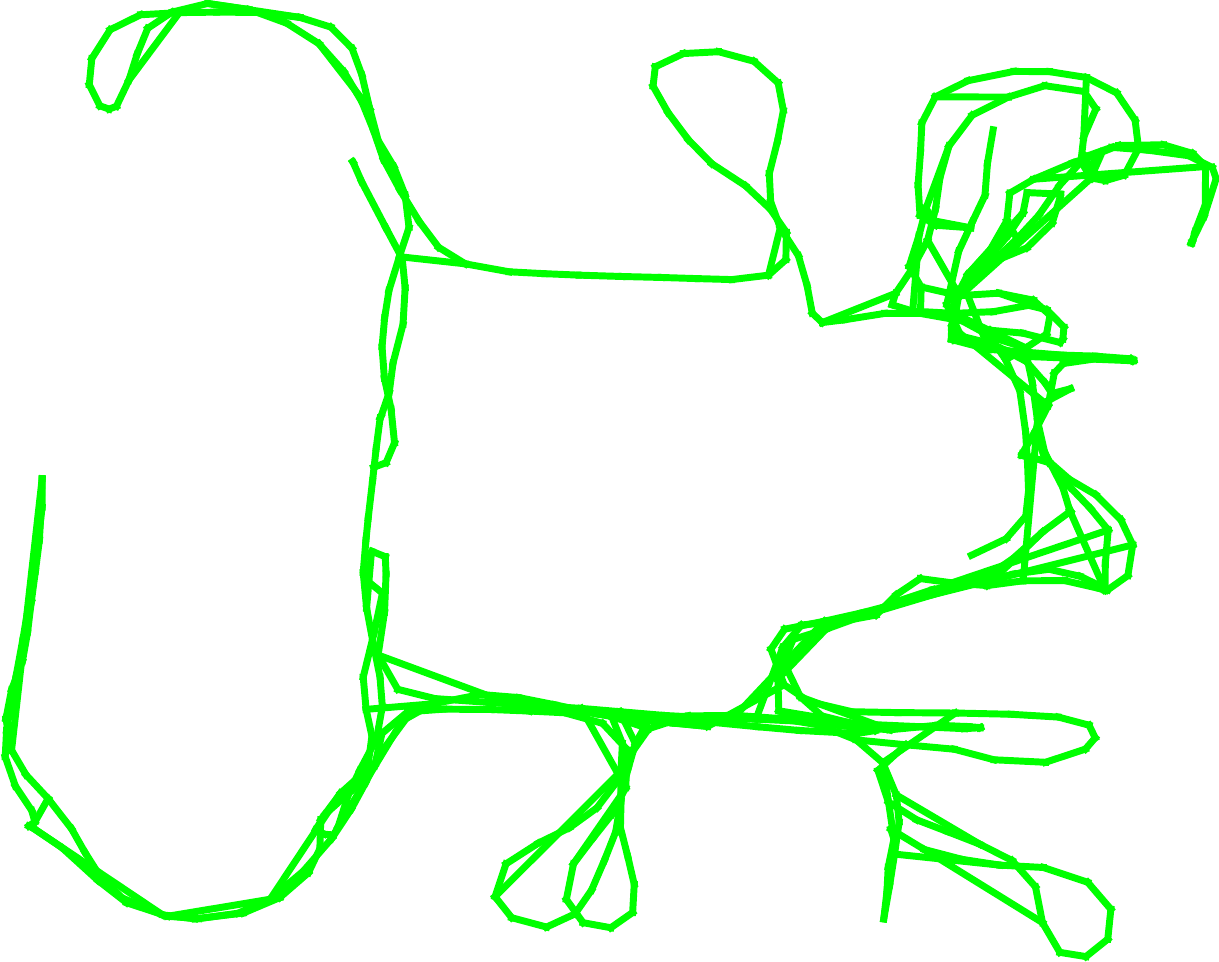} 
\end{minipage}
&
\begin{minipage}{0.5\columnwidth}%
\centering%
\includegraphics[width=\columnwidth, trim= 3.5cm 0cm 4.5cm 0cm, clip]{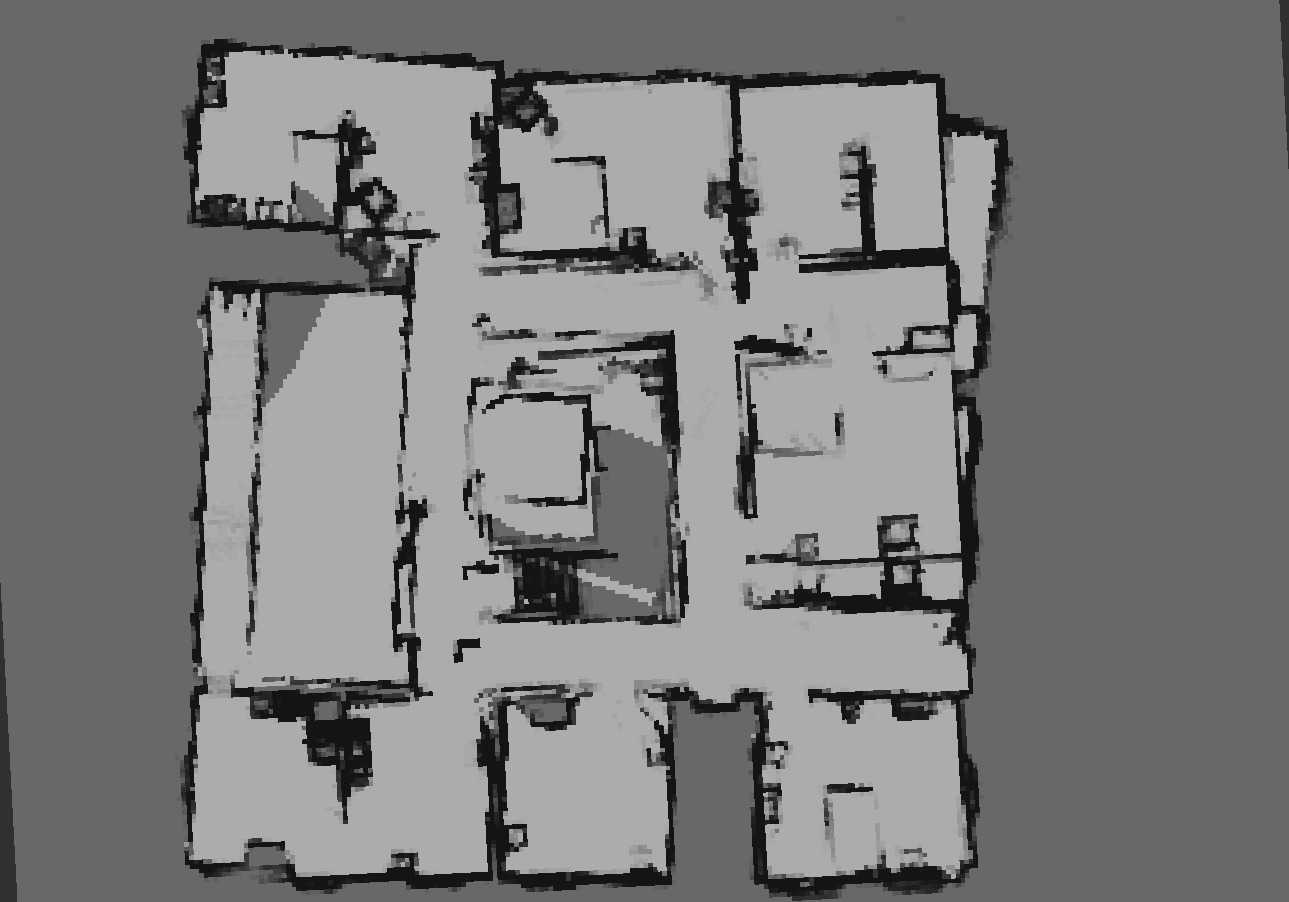} \end{minipage}\\
\vspace{0.2cm}

\begin{minipage}{0.5\columnwidth}%
\centering%
\includegraphics[width=\columnwidth, trim=15cm 0cm 15cm 0cm,clip]{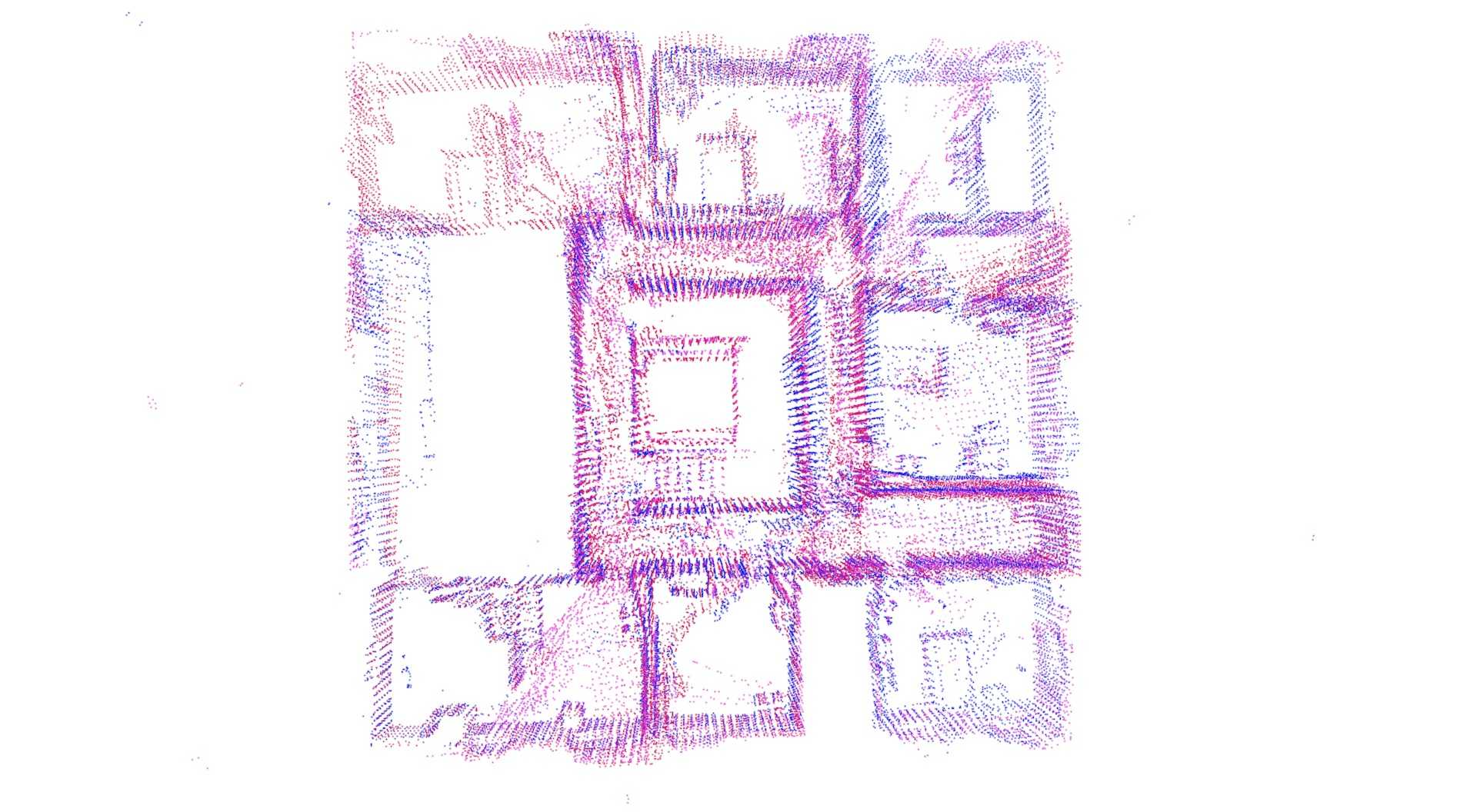} 
\end{minipage}
&
\begin{minipage}{0.5\columnwidth}%
\centering%
\includegraphics[width=\columnwidth, trim=0cm 0cm 0cm 0cm,clip]{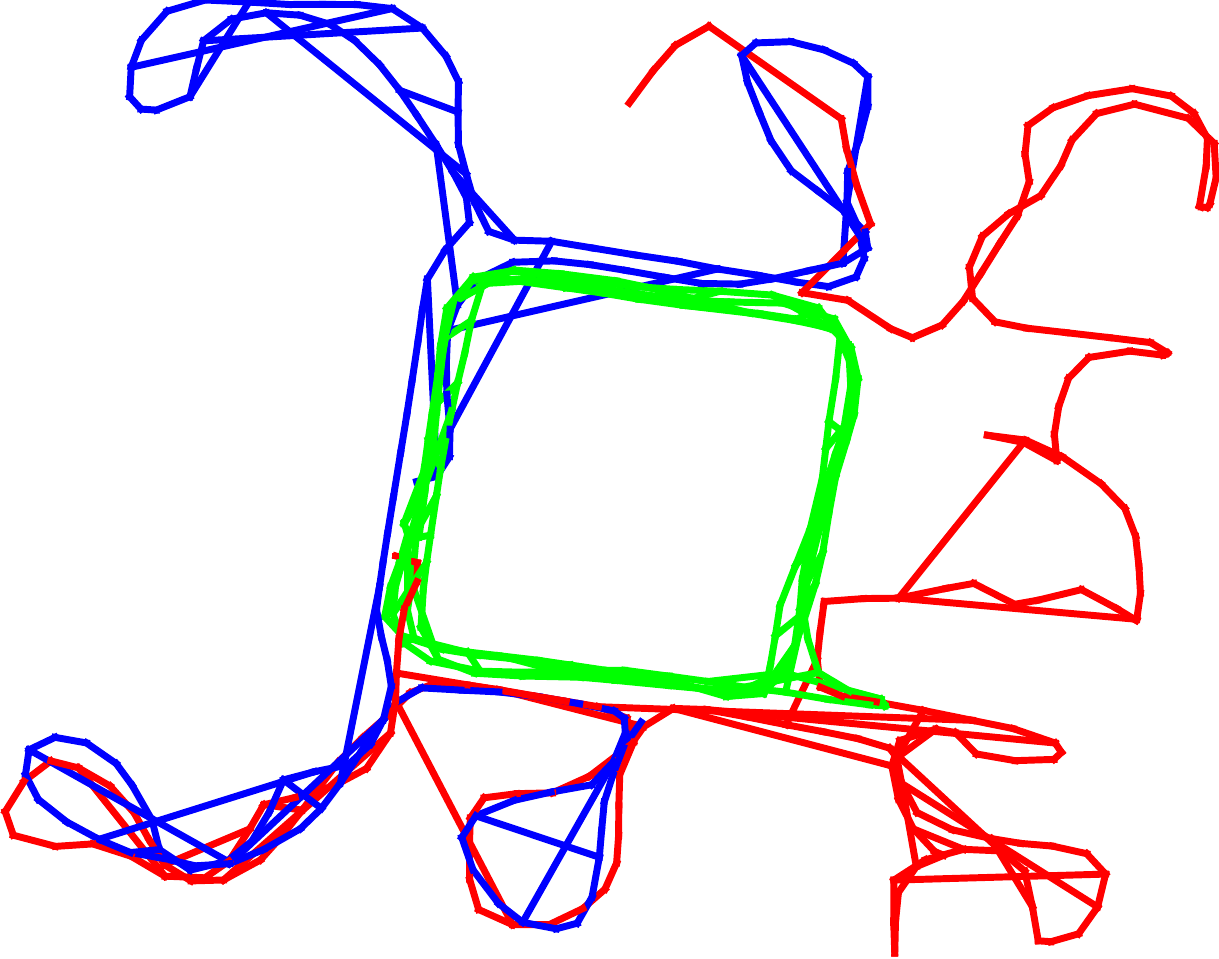} 
\end{minipage}
&
\begin{minipage}{0.5\columnwidth}%
\centering%
\includegraphics[width=\columnwidth, trim=0cm 0cm 0cm 0cm,clip]{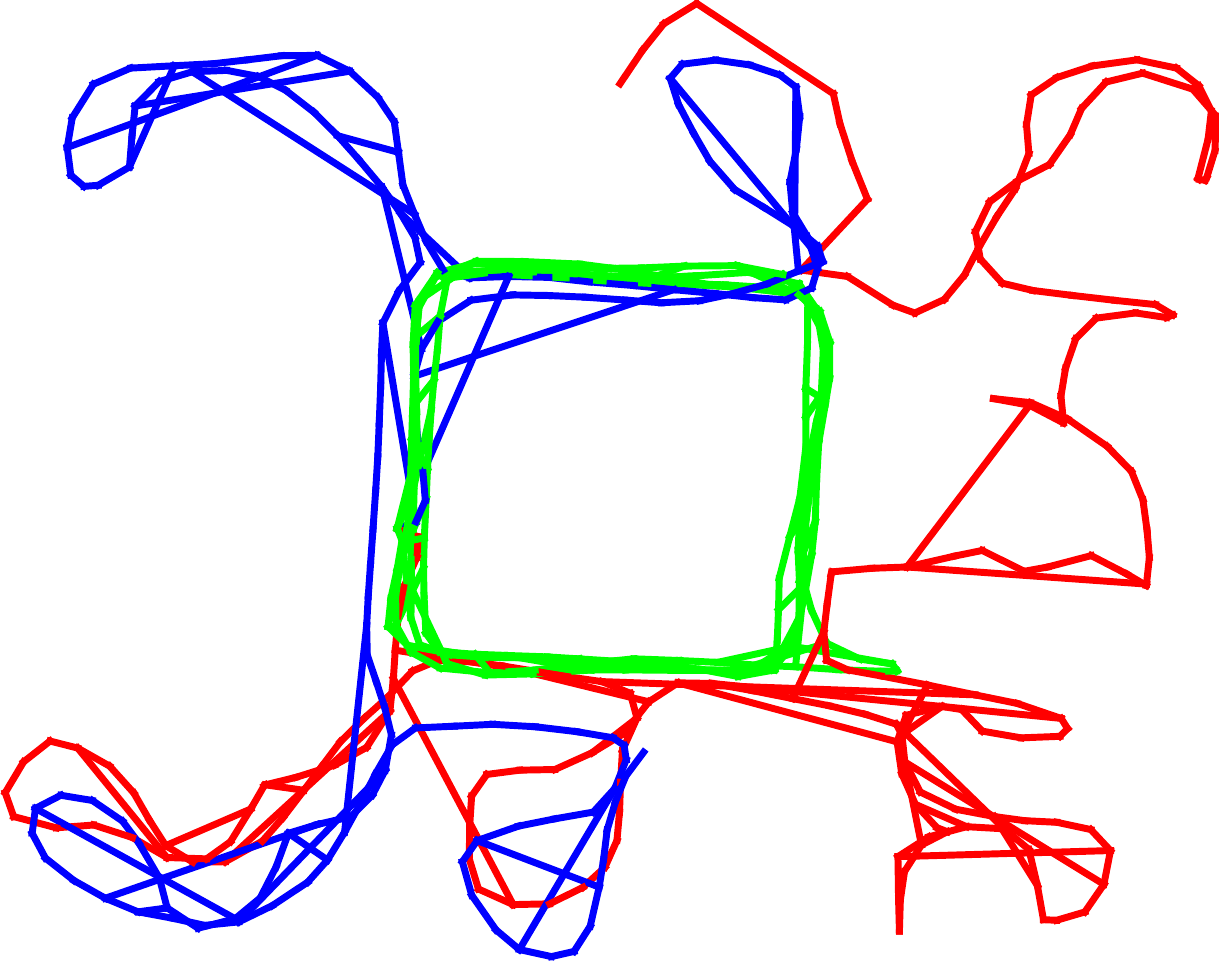} 
\end{minipage}
&
\begin{minipage}{0.5\columnwidth}%
\centering%
\includegraphics[width=\columnwidth, trim=0cm 0cm 0cm 0cm,clip]{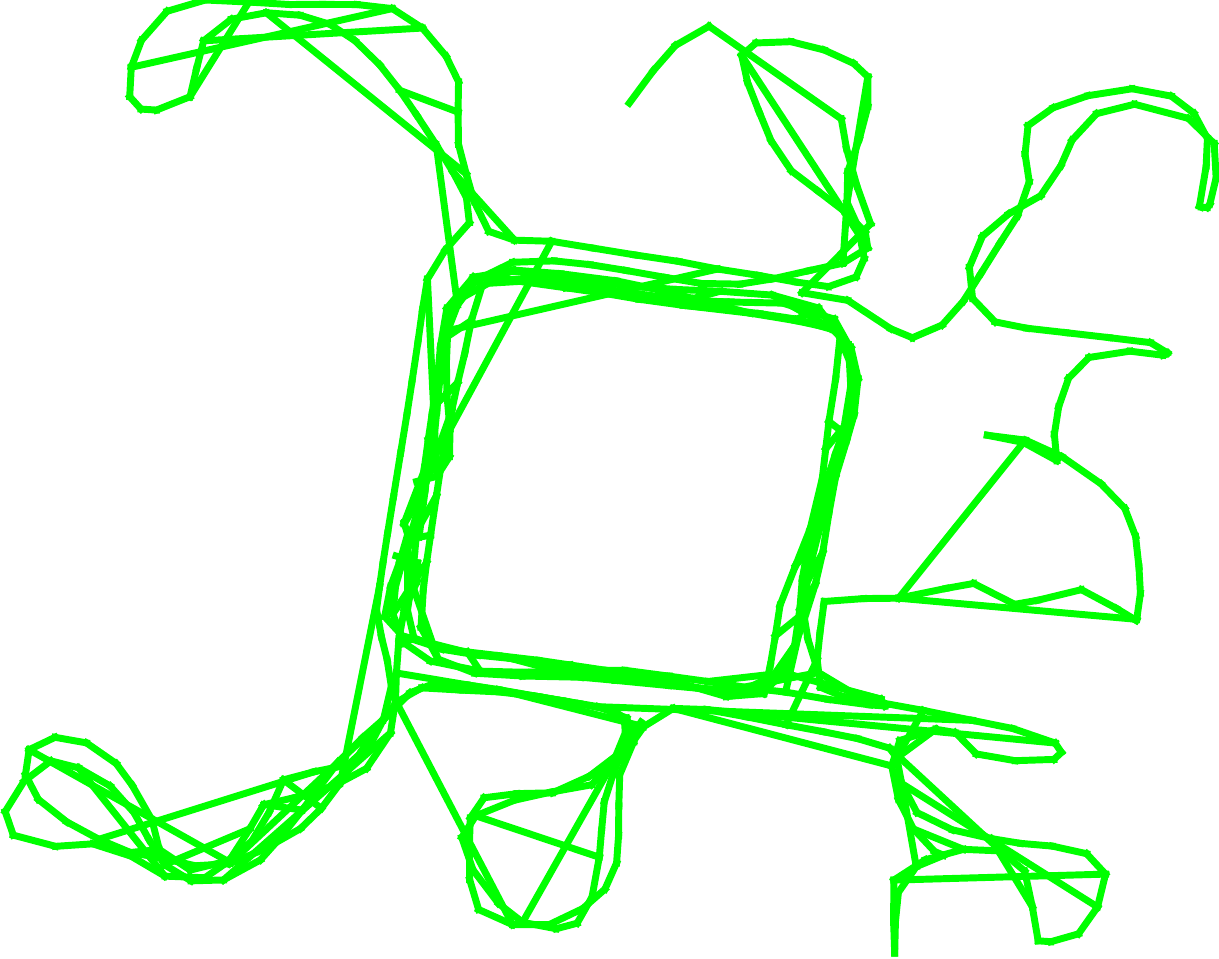} 
\end{minipage}
&
\begin{minipage}{0.5\columnwidth}%
\centering%
\includegraphics[width=\columnwidth, trim= 3.5cm 0cm 4.5cm 0cm, clip]{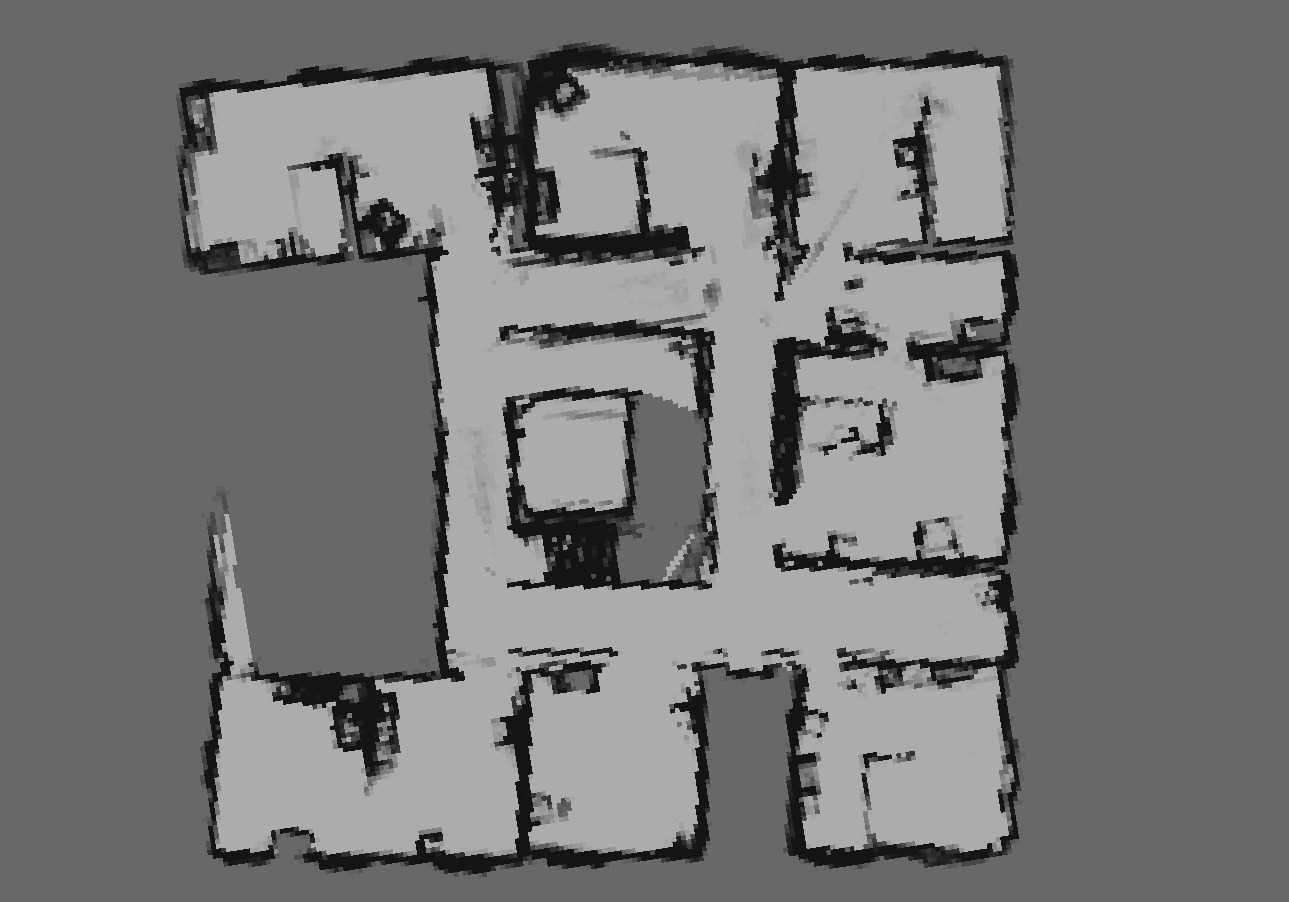} \end{minipage}
\vspace{0.2cm}

\end{tabular}
  \end{minipage}
  \caption{
  Indoor scenarios: 
(Left) aggregated point cloud obtained from the \DGS trajectory estimate. 
(Center) estimated trajectories for \DGS, \GN and \DDFSAM (robots shown 
  in red, blue, green and black for the distributed techniques). 
(Right) overall occupancy grid map obtained from the \DGS trajectory estimate.
  \label{fig:realExperiments2}
  \vspace{-0.5cm}
  }
\end{figure*}


\begin{figure*}[t]
\begin{minipage}{0.7\columnwidth}
\begin{tabular}{c|ccc|c}%

 Point Cloud & \DGS &  DDF-SAM & Centralized  & Occupancy Grid \\
\vspace{0.2cm}
\begin{minipage}{0.5\columnwidth}%
\centering%
\includegraphics[width=\columnwidth, trim=20cm 0cm 20cm 0cm,clip]{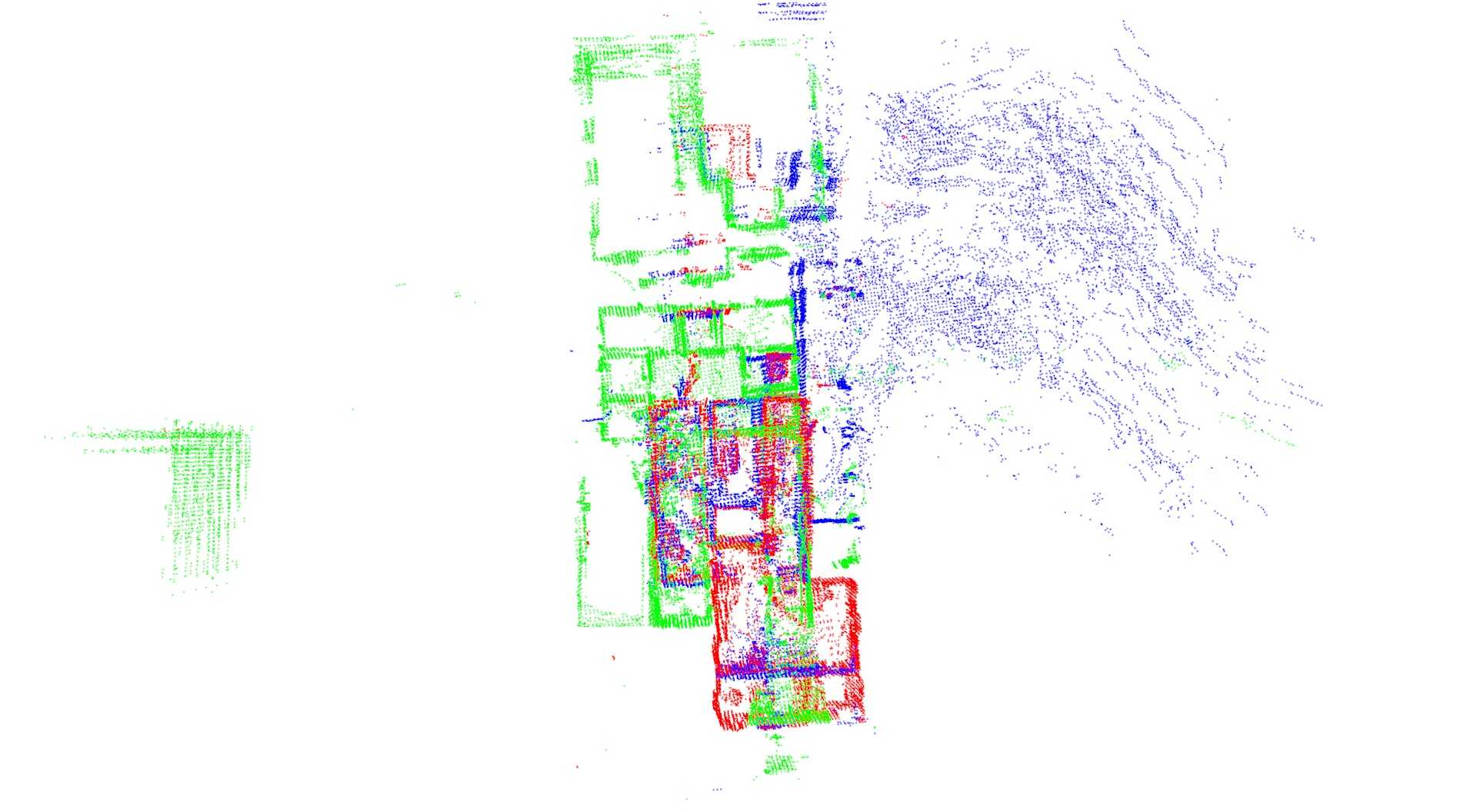} 
\end{minipage}
&
\begin{minipage}{0.5\columnwidth}%
\centering%
\includegraphics[width=\columnwidth, trim=0cm 0cm 0cm 0cm,clip]{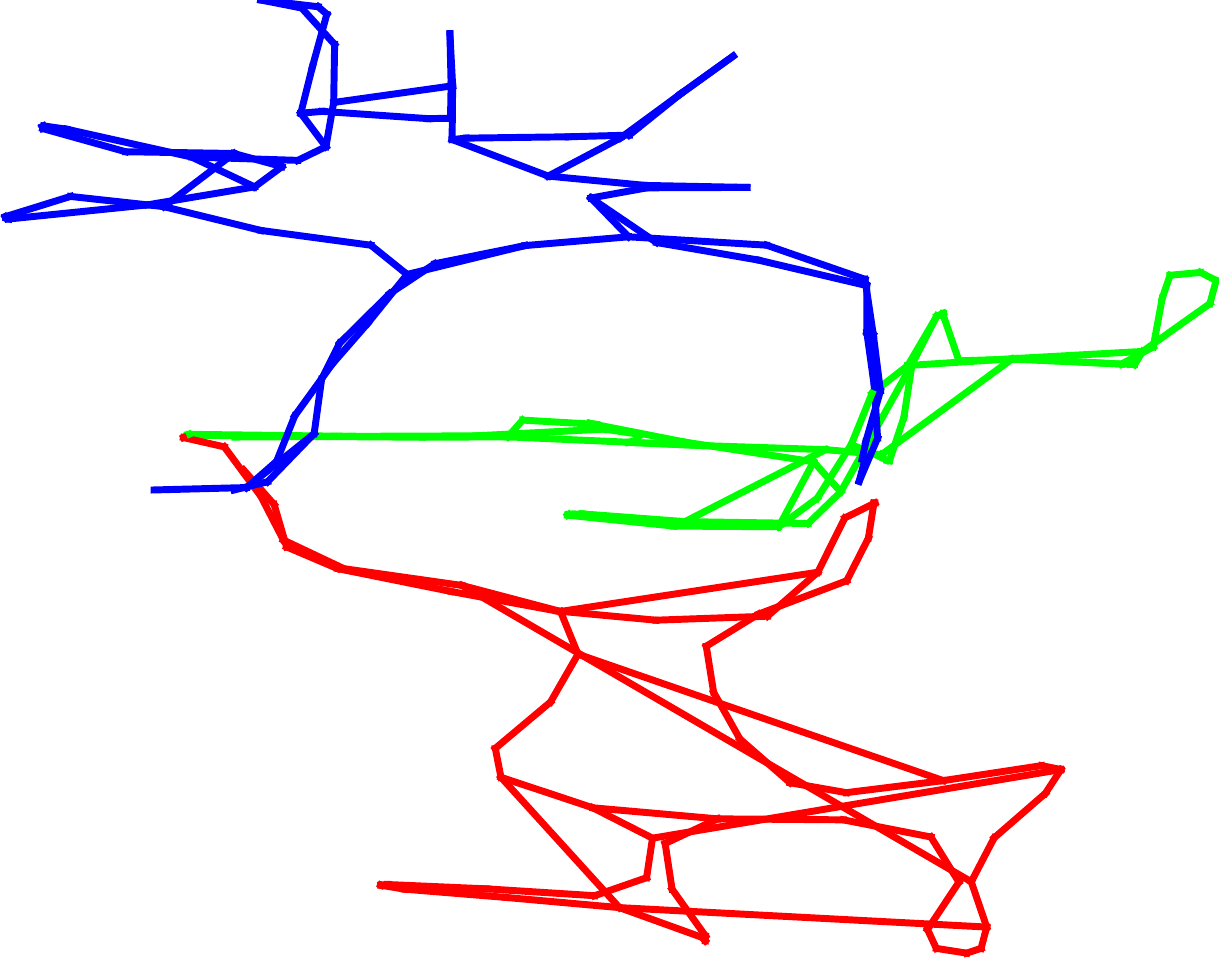} 
\end{minipage}
&
\begin{minipage}{0.5\columnwidth}%
\centering%
\includegraphics[width=\columnwidth, trim=0cm 0cm 0cm 0cm,clip]{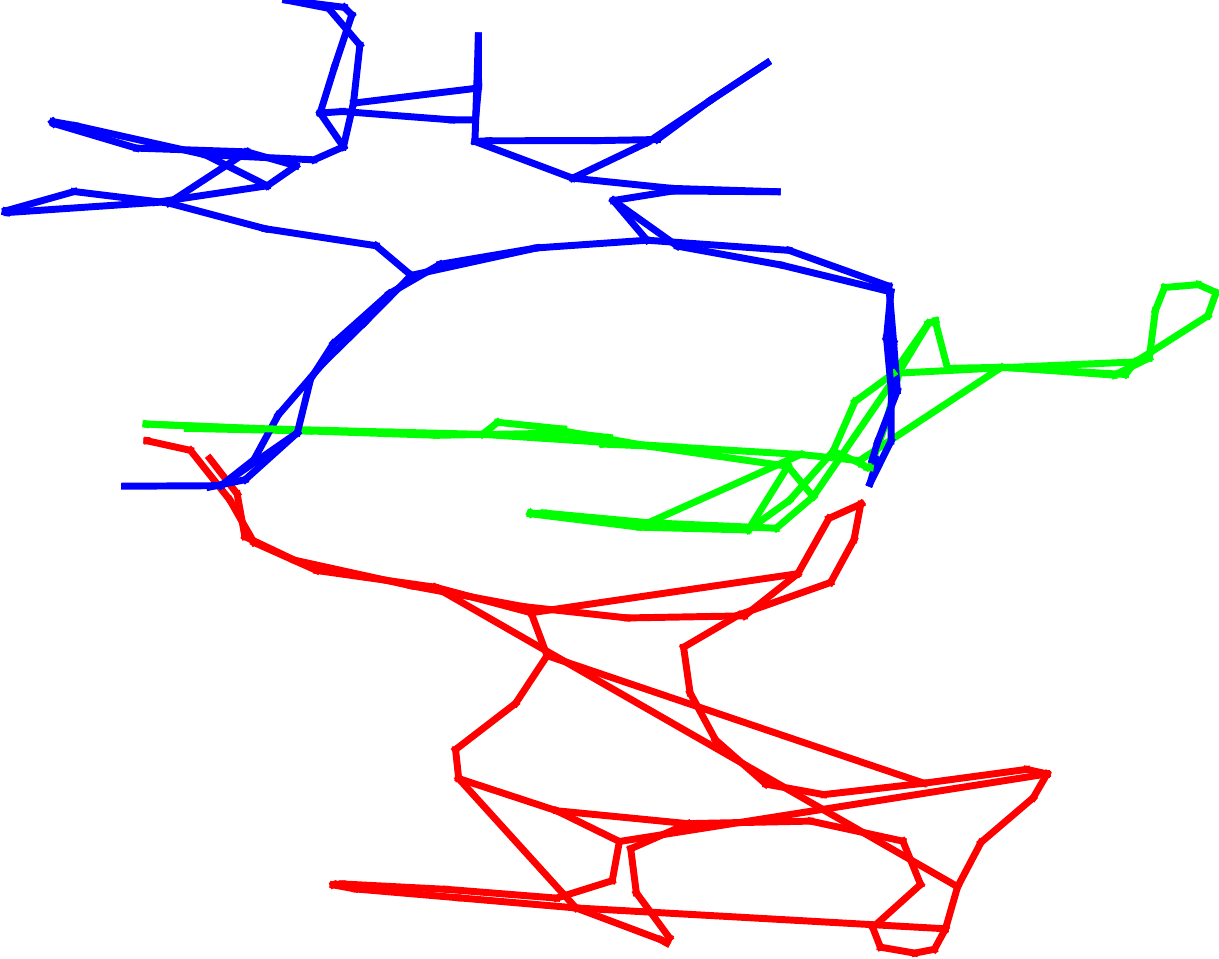} 
\end{minipage}
&
\begin{minipage}{0.5\columnwidth}%
\centering%
\includegraphics[width=\columnwidth, trim=0cm 0cm 0cm 0cm,clip]{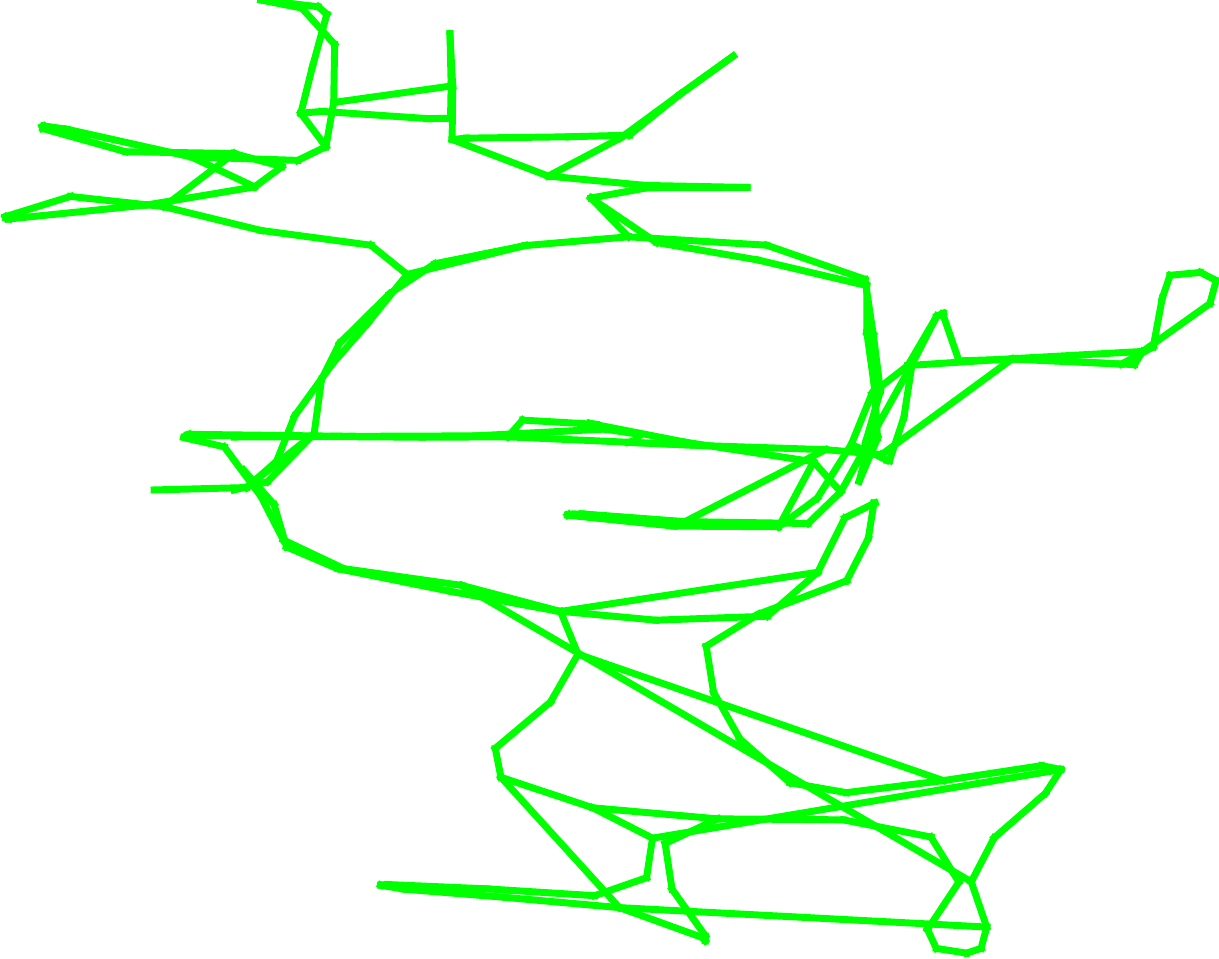} 
\end{minipage}
&
\begin{minipage}{0.5\columnwidth}%
\centering%
\includegraphics[width=\columnwidth, trim= 3.5cm 0cm 4.5cm 0cm, clip, angle=0]{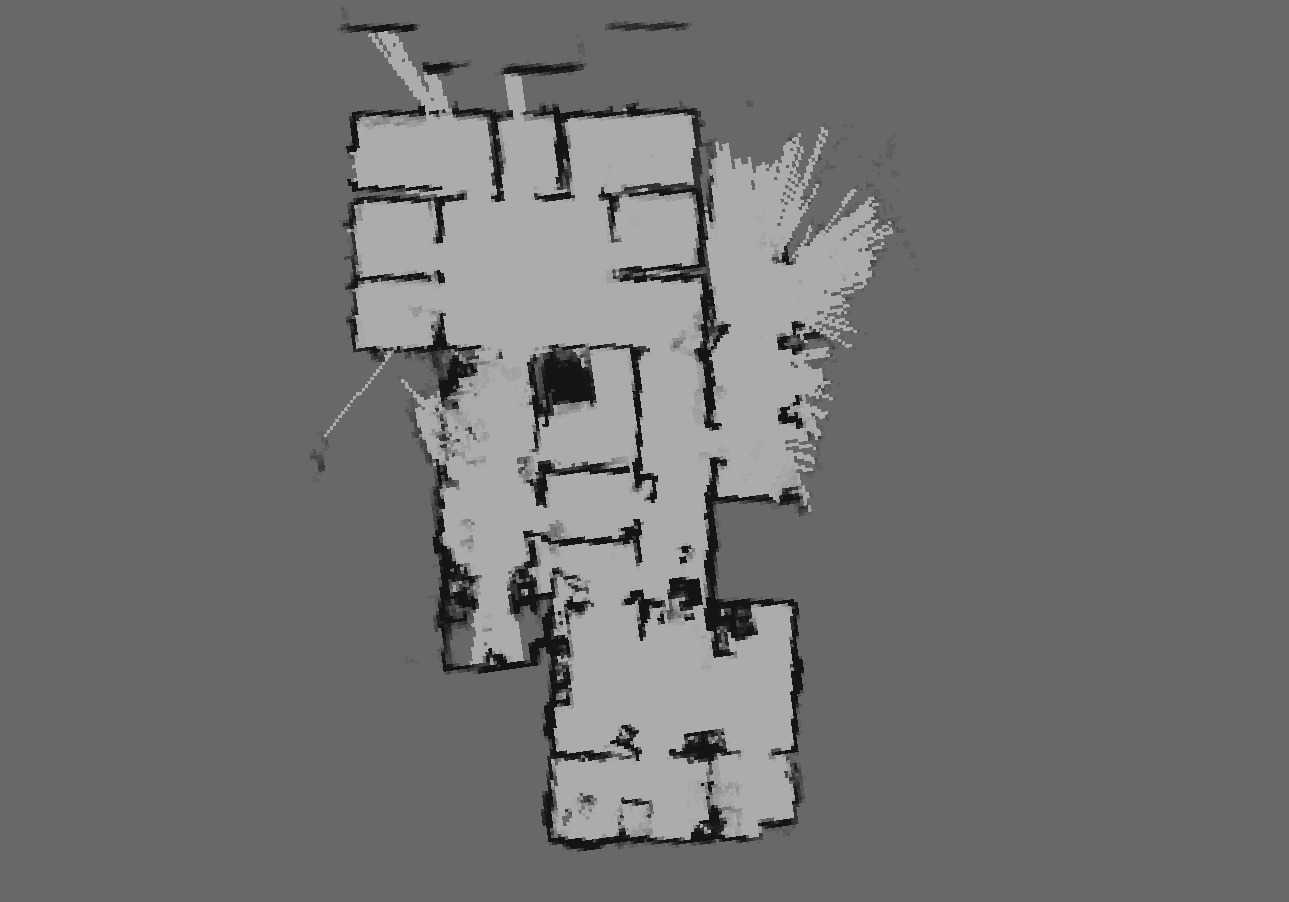} 
\end{minipage}\\
\vspace{0.2cm}

\begin{minipage}{0.5\columnwidth}%
\centering%
\includegraphics[width=\columnwidth, trim=20cm 0cm 25cm 0cm,clip]{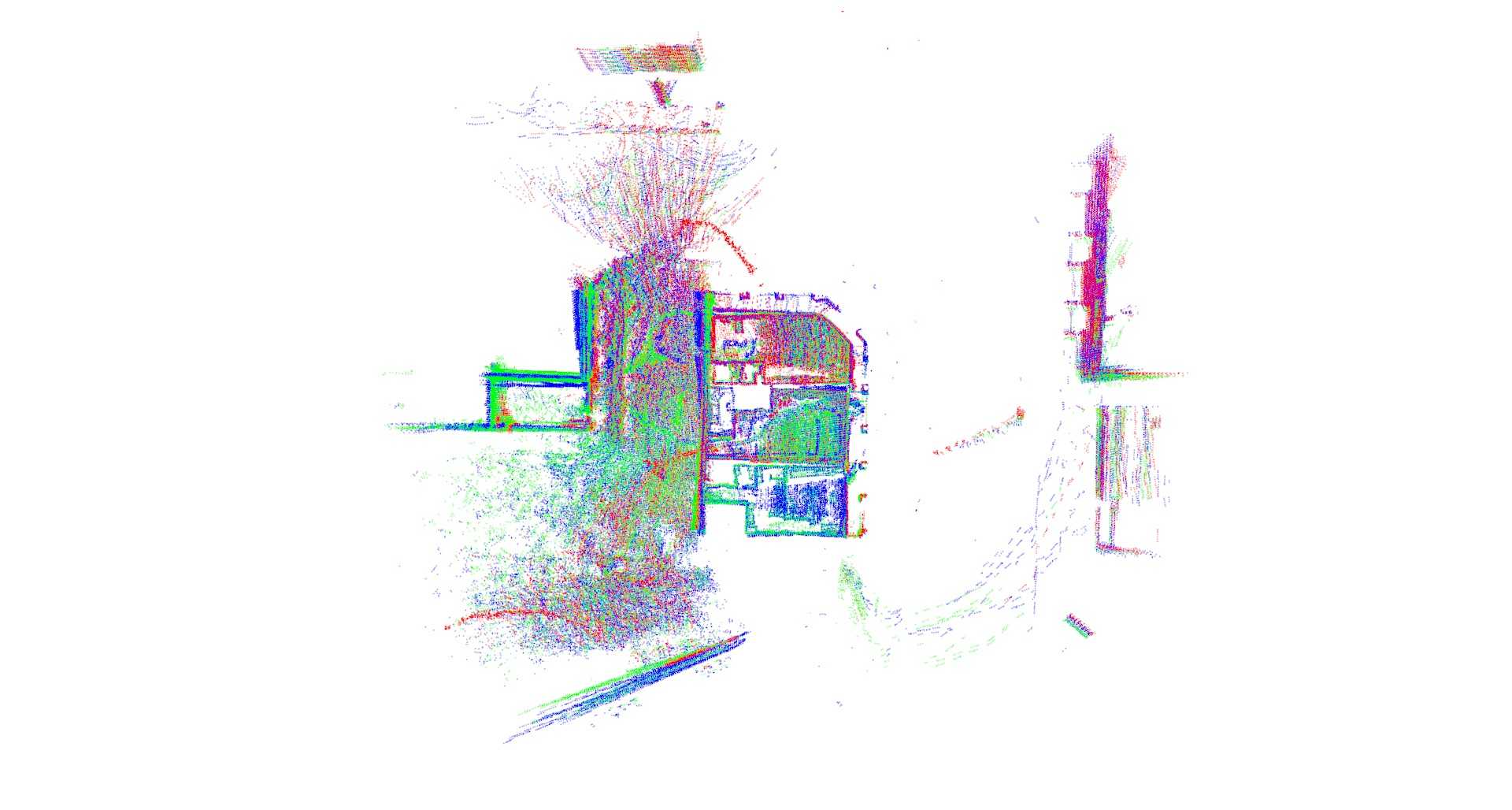} 
\end{minipage}
&
\begin{minipage}{0.5\columnwidth}%
\centering%
\includegraphics[width=\columnwidth, trim=0cm 0cm 0cm 0cm,clip]{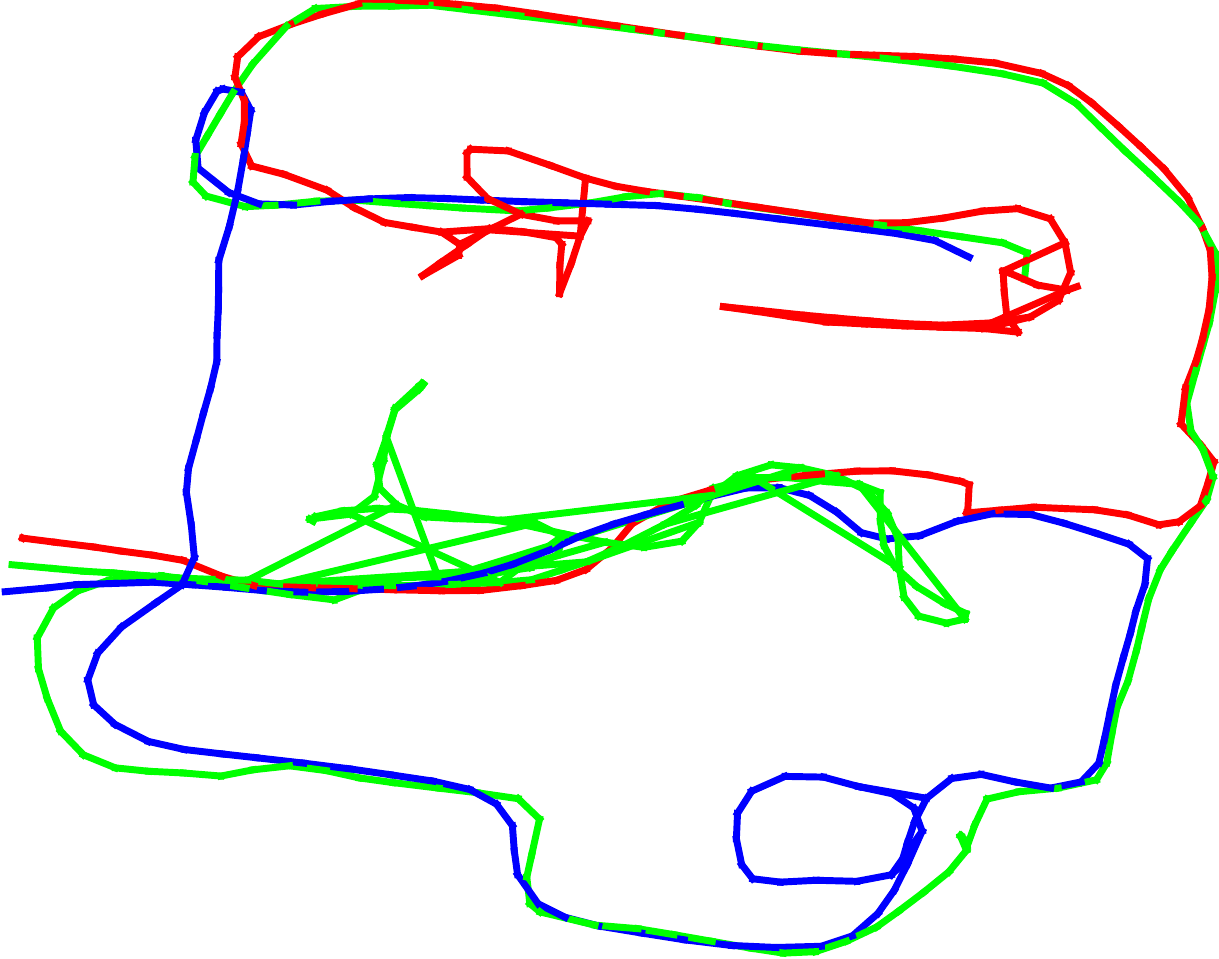} 
\end{minipage}
&
\begin{minipage}{0.5\columnwidth}%
\centering%
\includegraphics[width=\columnwidth, trim=0cm 0cm 0cm 0cm,clip]{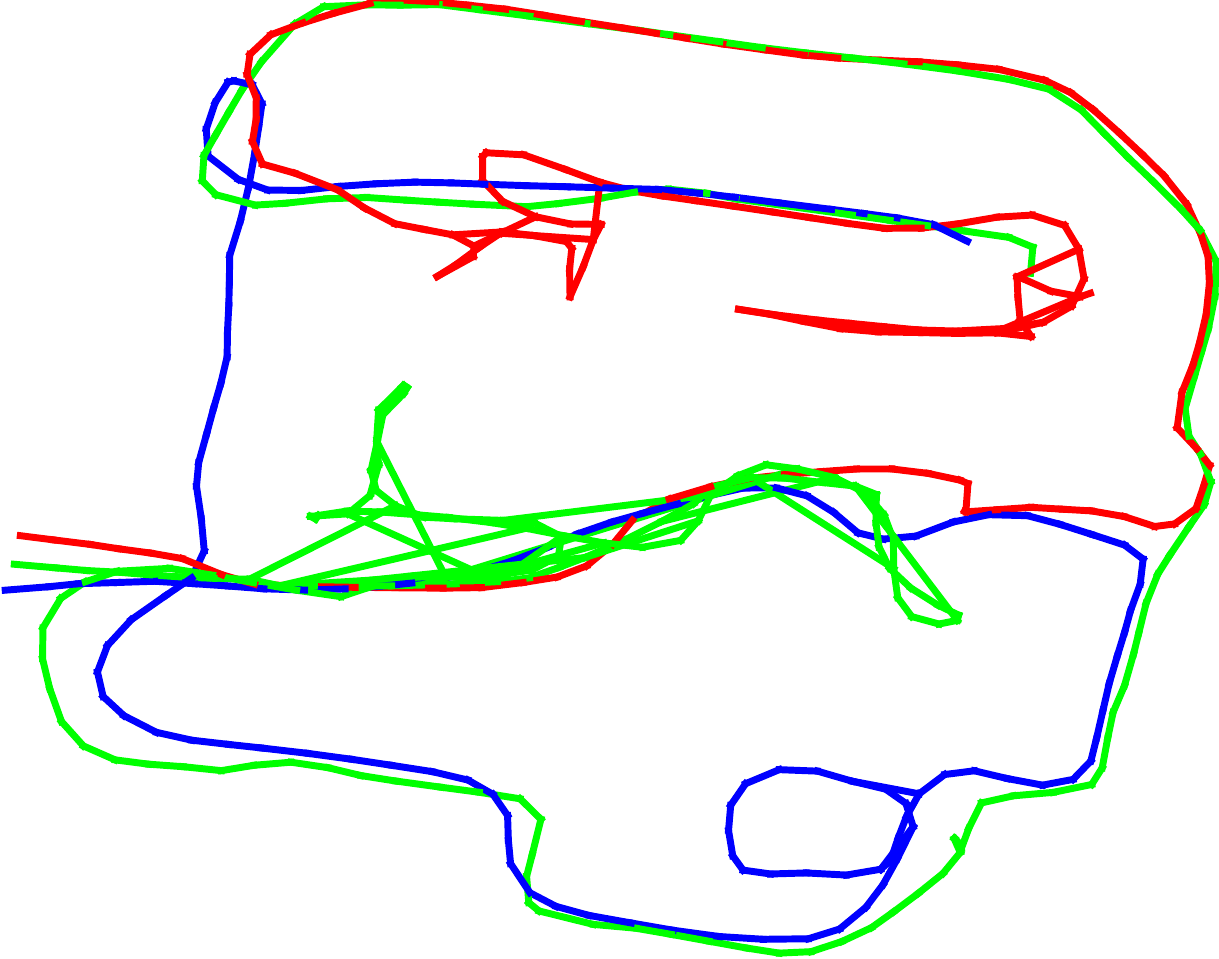} 
\end{minipage}
&
\begin{minipage}{0.5\columnwidth}%
\centering%
\includegraphics[width=\columnwidth, trim=0cm 0cm 0cm 0cm,clip]{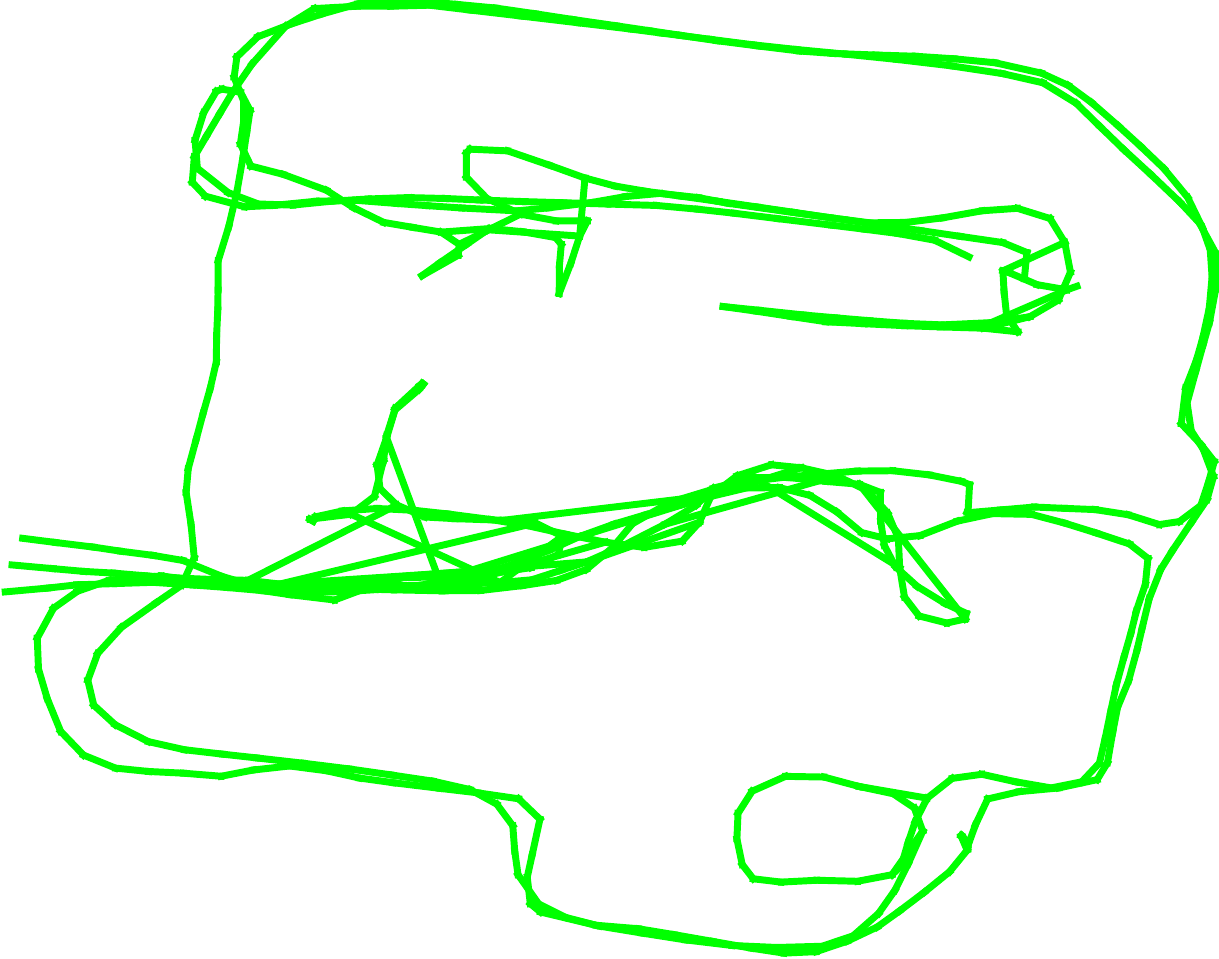} 
\end{minipage}
&
\begin{minipage}{0.5\columnwidth}%
\centering%
\includegraphics[width=\columnwidth, trim= 3.5cm 0cm 4.5cm 0cm, clip, angle=0]{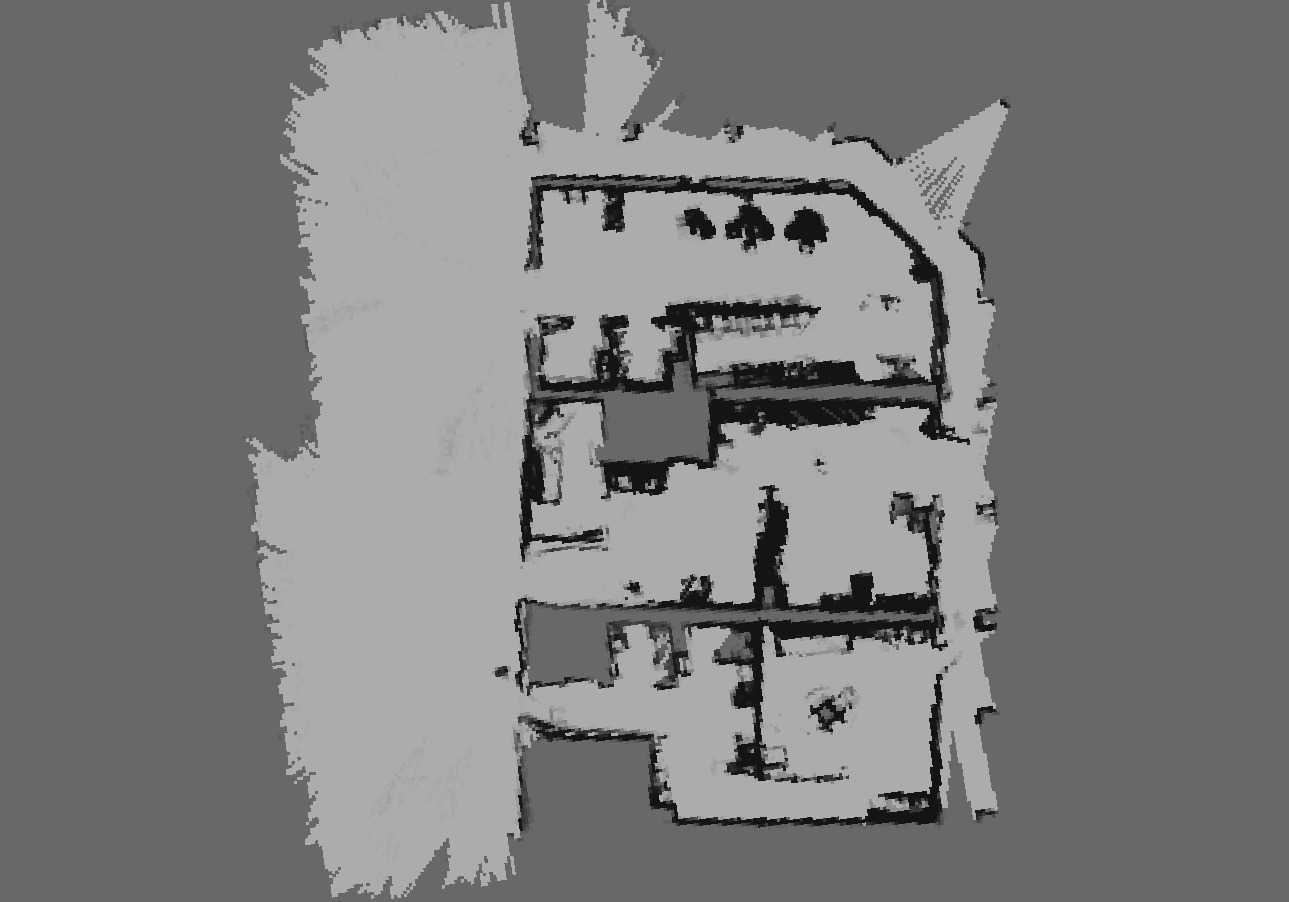} 
\end{minipage}\\
\vspace{0.2cm}

\begin{minipage}{0.5\columnwidth}%
\centering%
\includegraphics[width=\columnwidth, trim=20cm 0cm 25cm 0cm,clip]{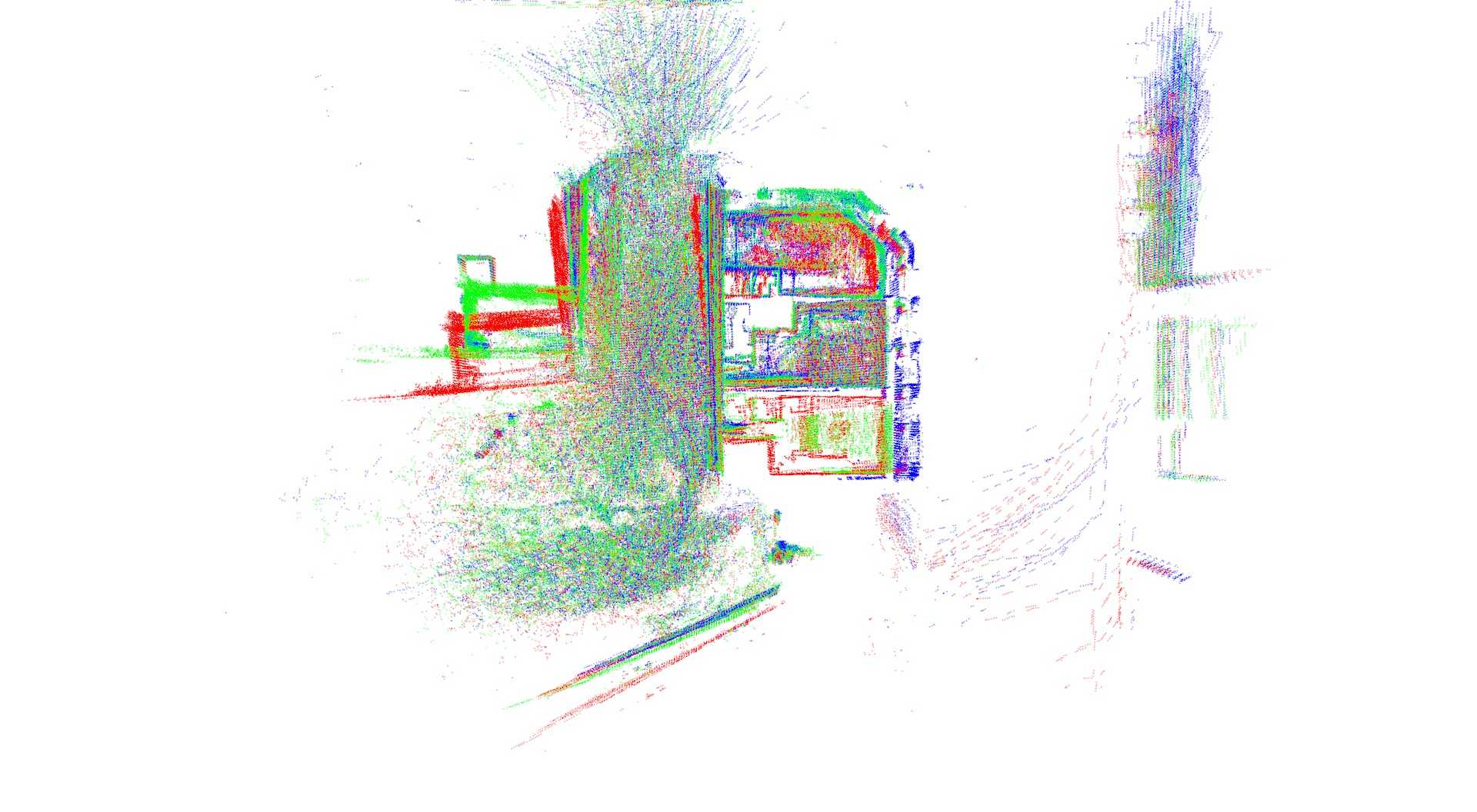} 
\end{minipage}
&
\begin{minipage}{0.5\columnwidth}%
\centering%
\includegraphics[width=\columnwidth, trim=0cm 0cm 0cm 0cm,clip]{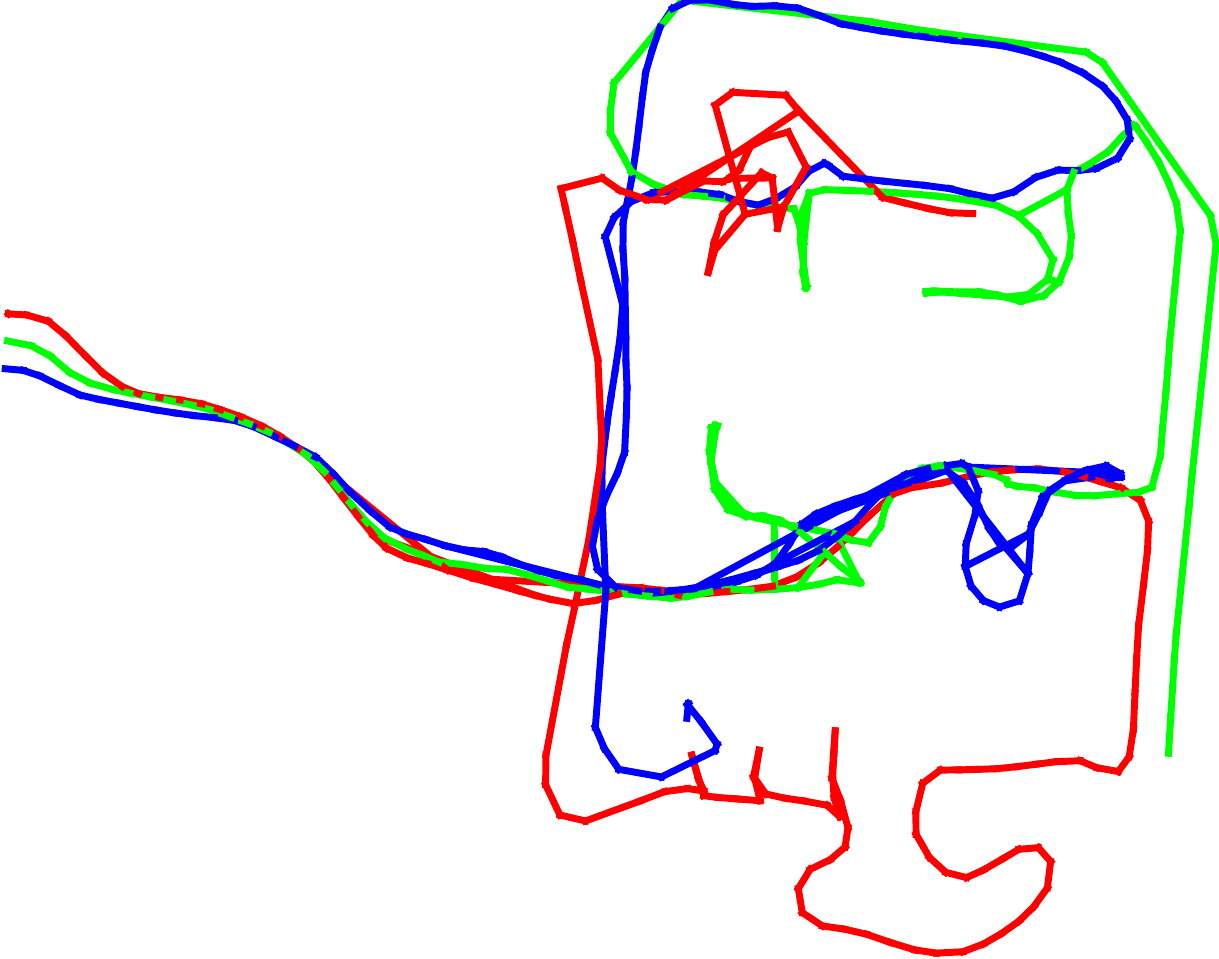} 
\end{minipage}
&
\begin{minipage}{0.5\columnwidth}%
\centering%
\includegraphics[width=\columnwidth, trim=0cm 0cm 0cm 0cm,clip]{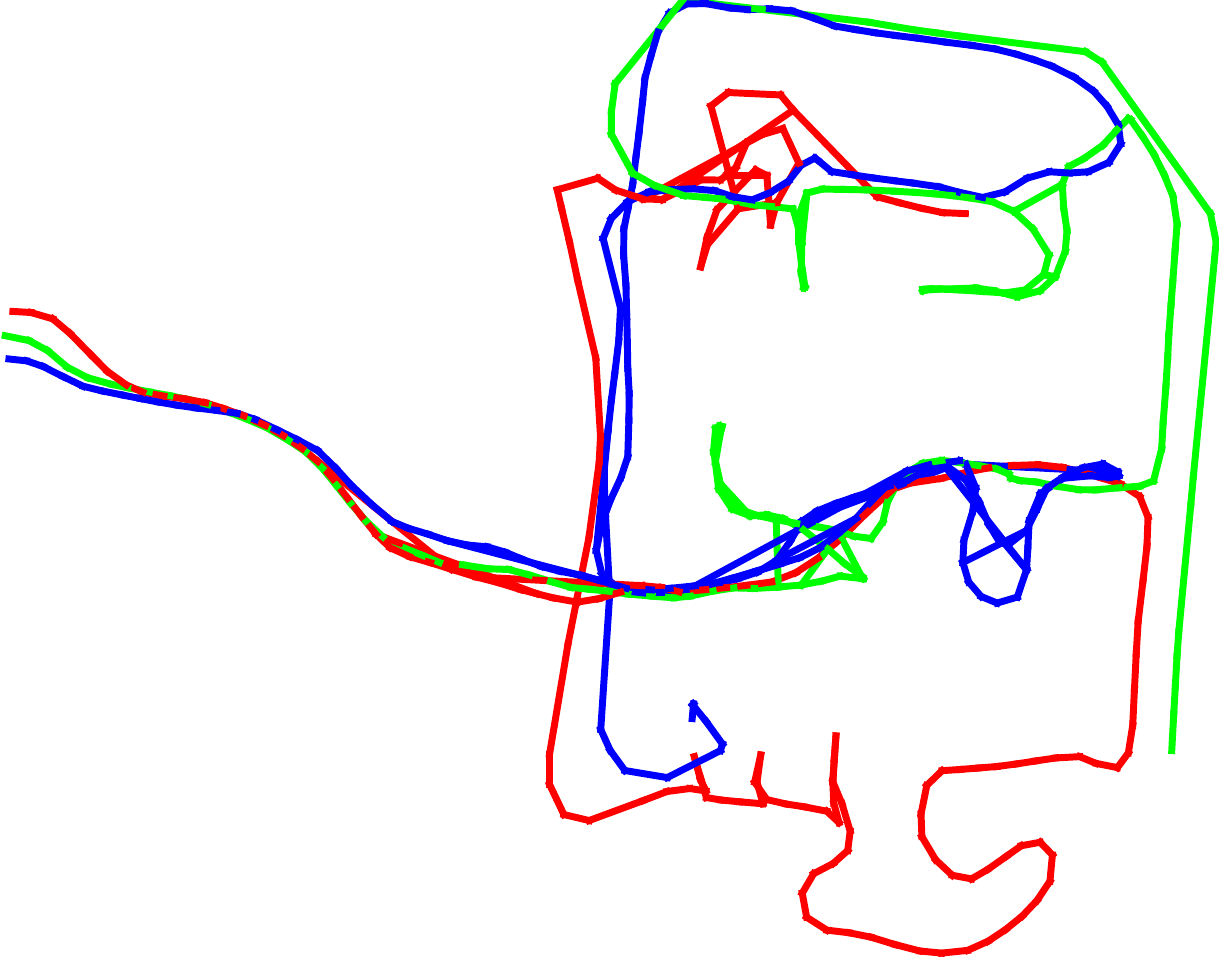} 
\end{minipage}
&
\begin{minipage}{0.5\columnwidth}%
\centering%
\includegraphics[width=\columnwidth, trim=0cm 0cm 0cm 0cm,clip]{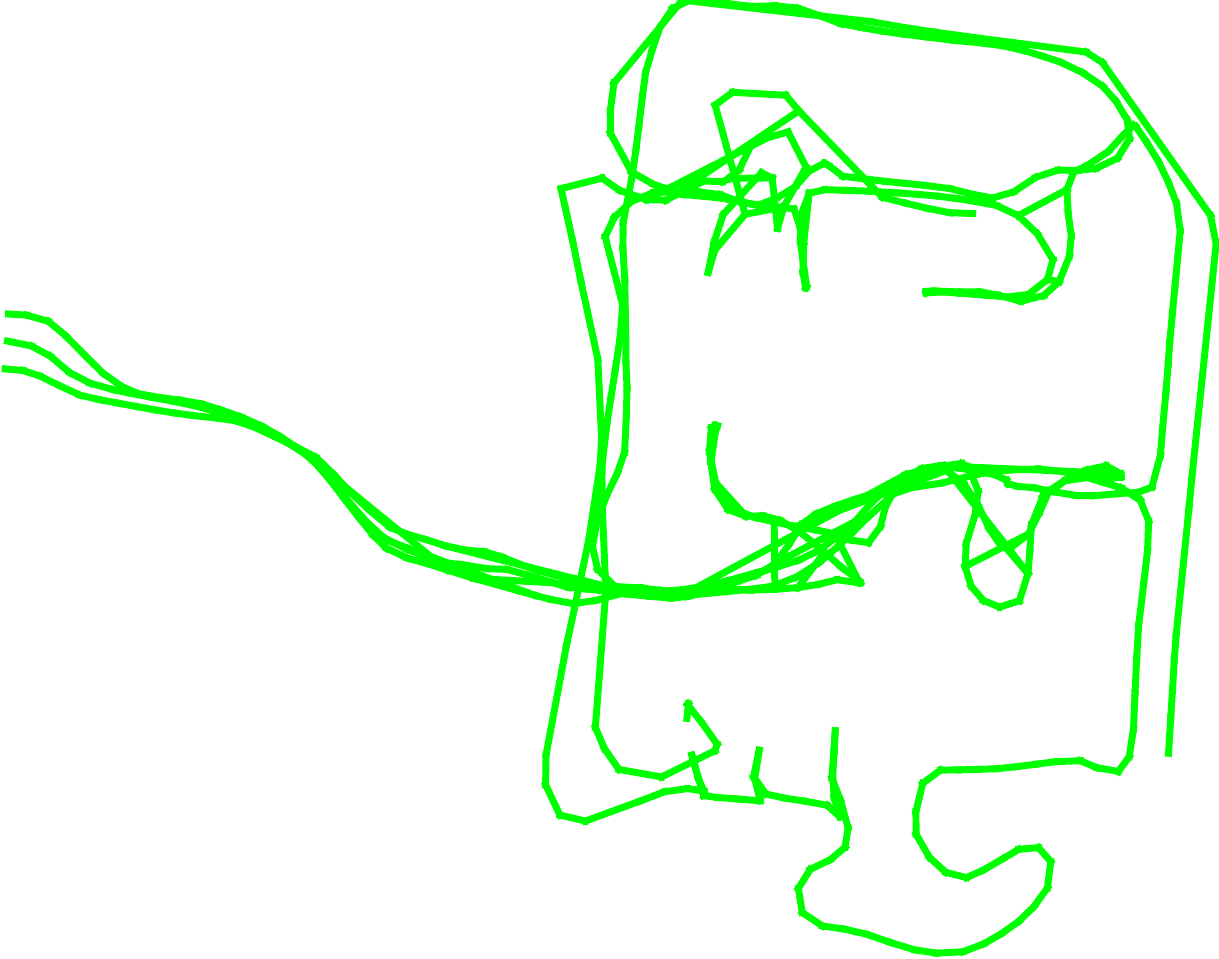} 
\end{minipage}
&
\begin{minipage}{0.5\columnwidth}%
\centering%
\includegraphics[width=\columnwidth, trim= 3.5cm 0cm 4.5cm 0cm, clip, angle=0]{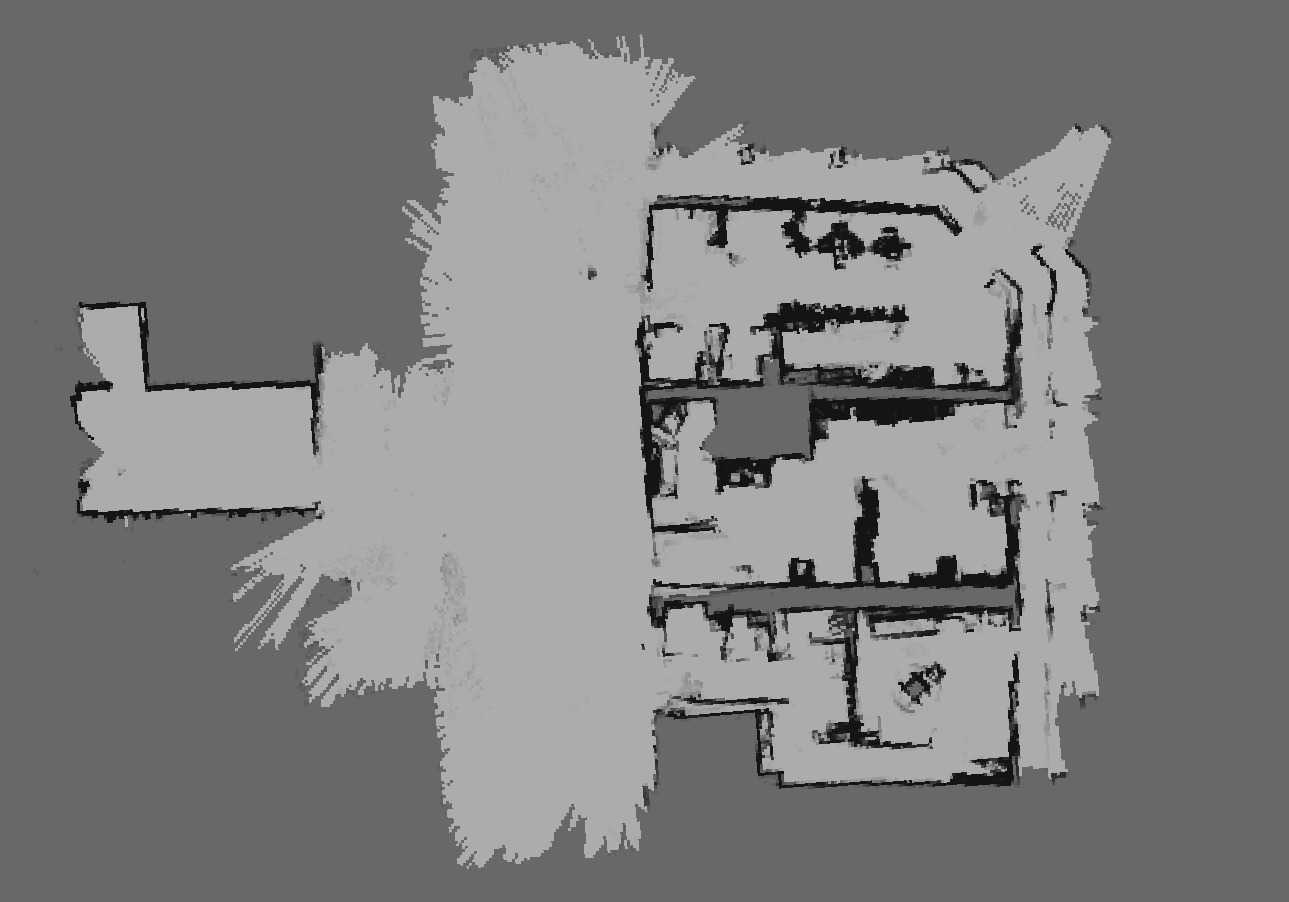} 
\end{minipage}\\
\vspace{0.2cm}

\begin{minipage}{0.5\columnwidth}%
\centering%
\includegraphics[width=\columnwidth, trim=20cm 0cm 10cm 0cm,clip, angle=90]{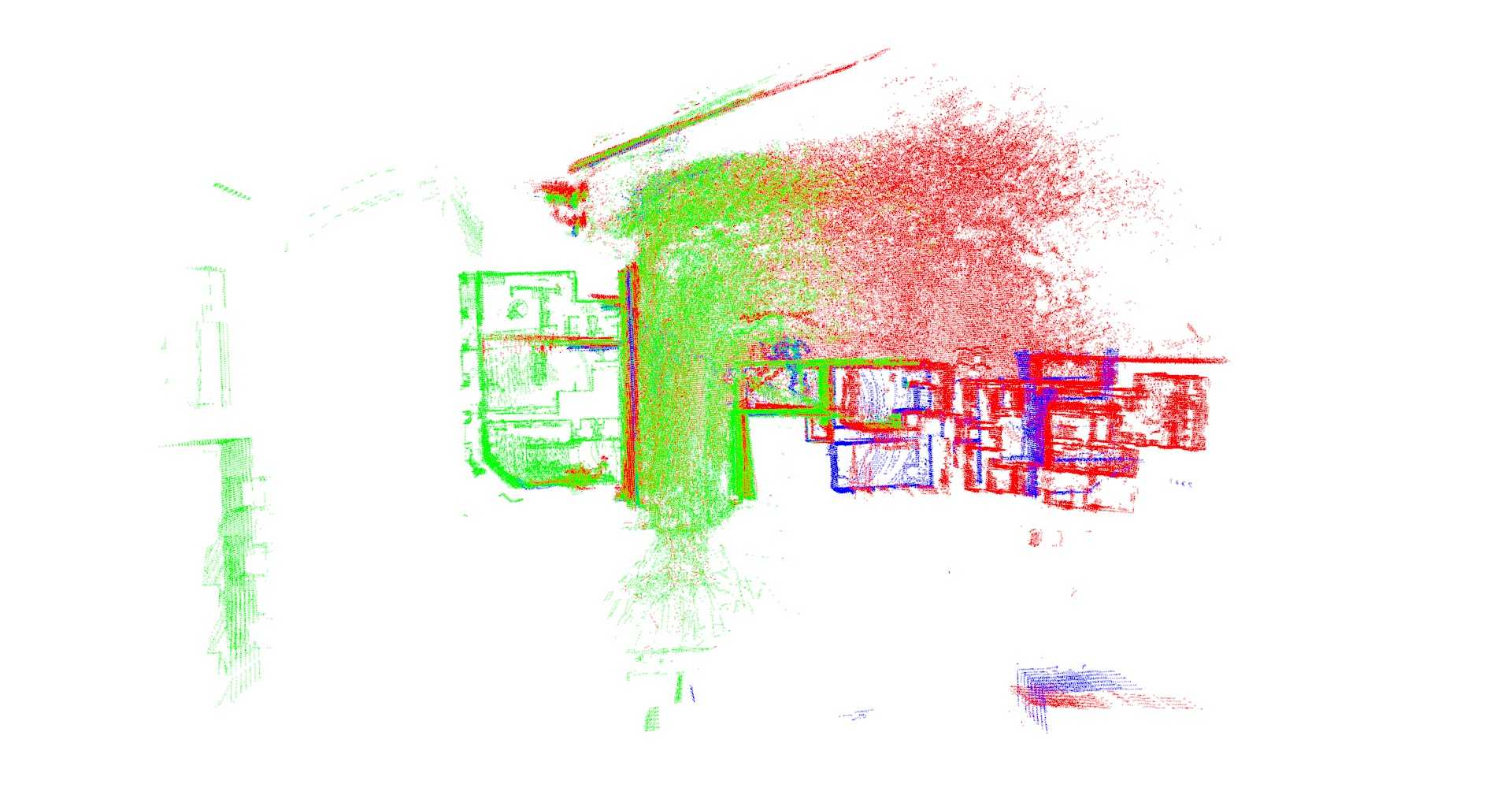} 
\end{minipage}
&
\begin{minipage}{0.5\columnwidth}%
\centering%
\includegraphics[width=\columnwidth, trim=0cm 0cm 0cm 0cm,clip, angle=90]{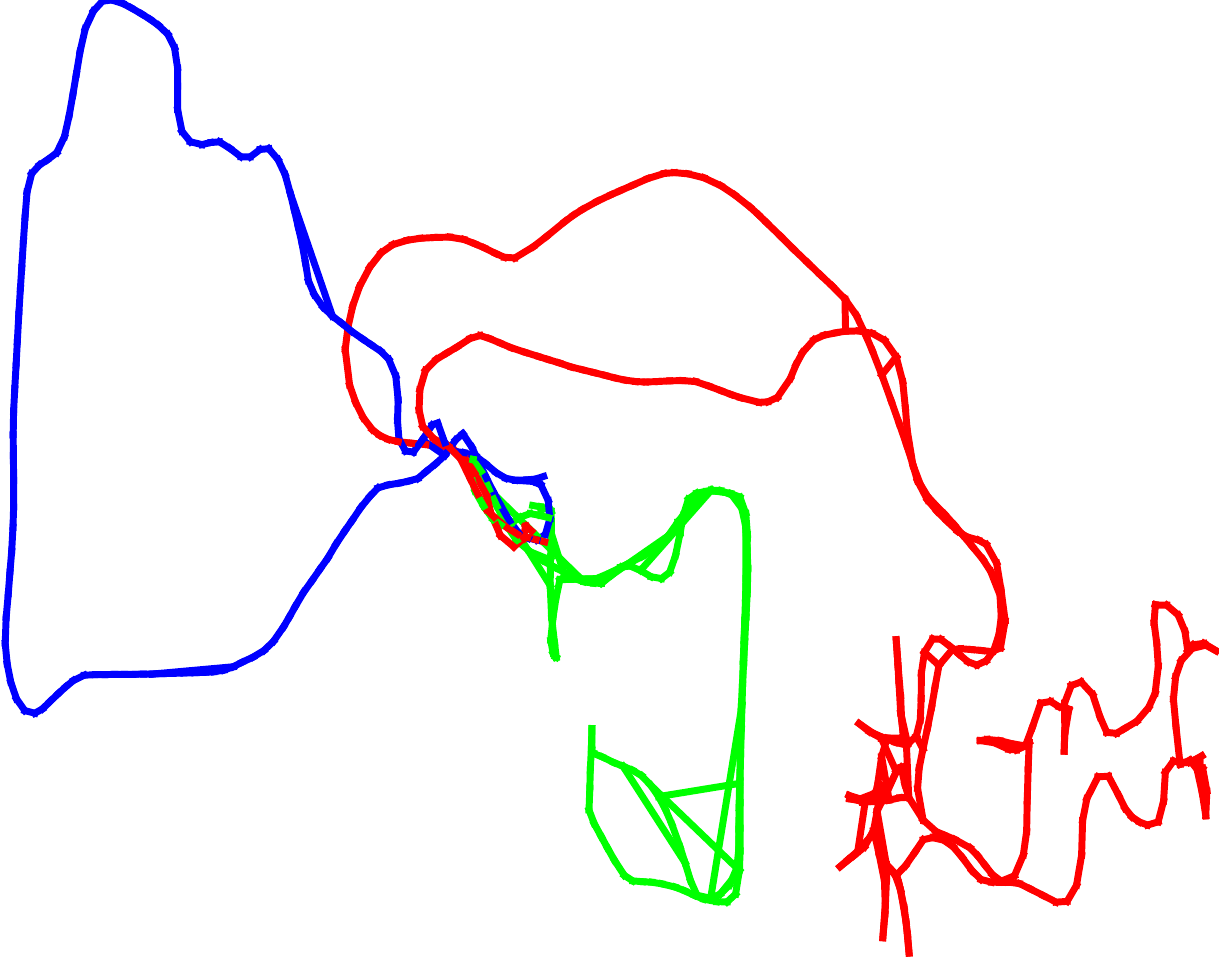} 
\end{minipage}
&
\begin{minipage}{0.5\columnwidth}%
\centering%
\includegraphics[width=\columnwidth, trim=0cm 0cm 0cm 0cm,clip, angle = 90]{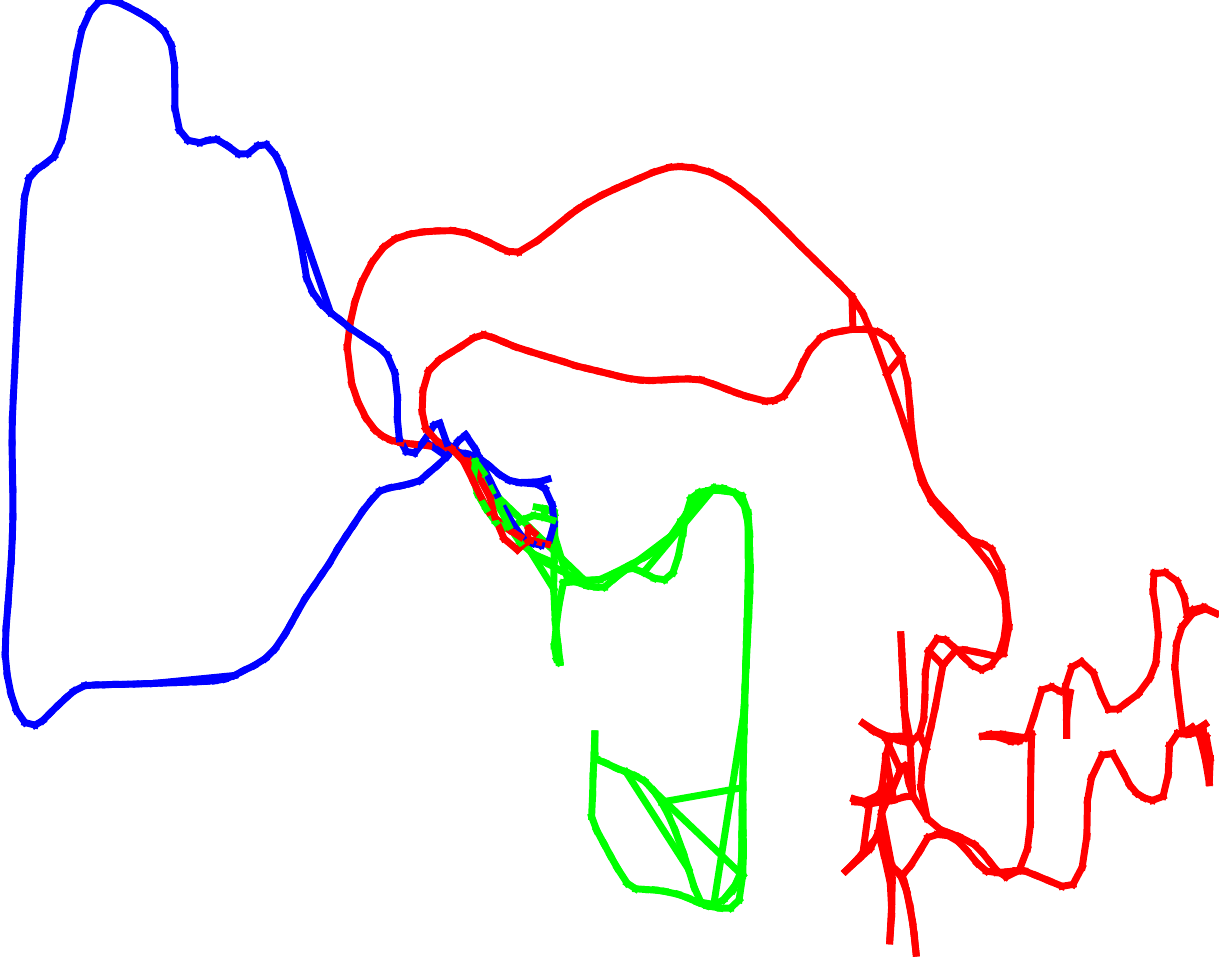} 
\end{minipage}
&
\begin{minipage}{0.5\columnwidth}%
\centering%
\includegraphics[width=\columnwidth, trim=0cm 0cm 0cm 0cm,clip, angle = 90]{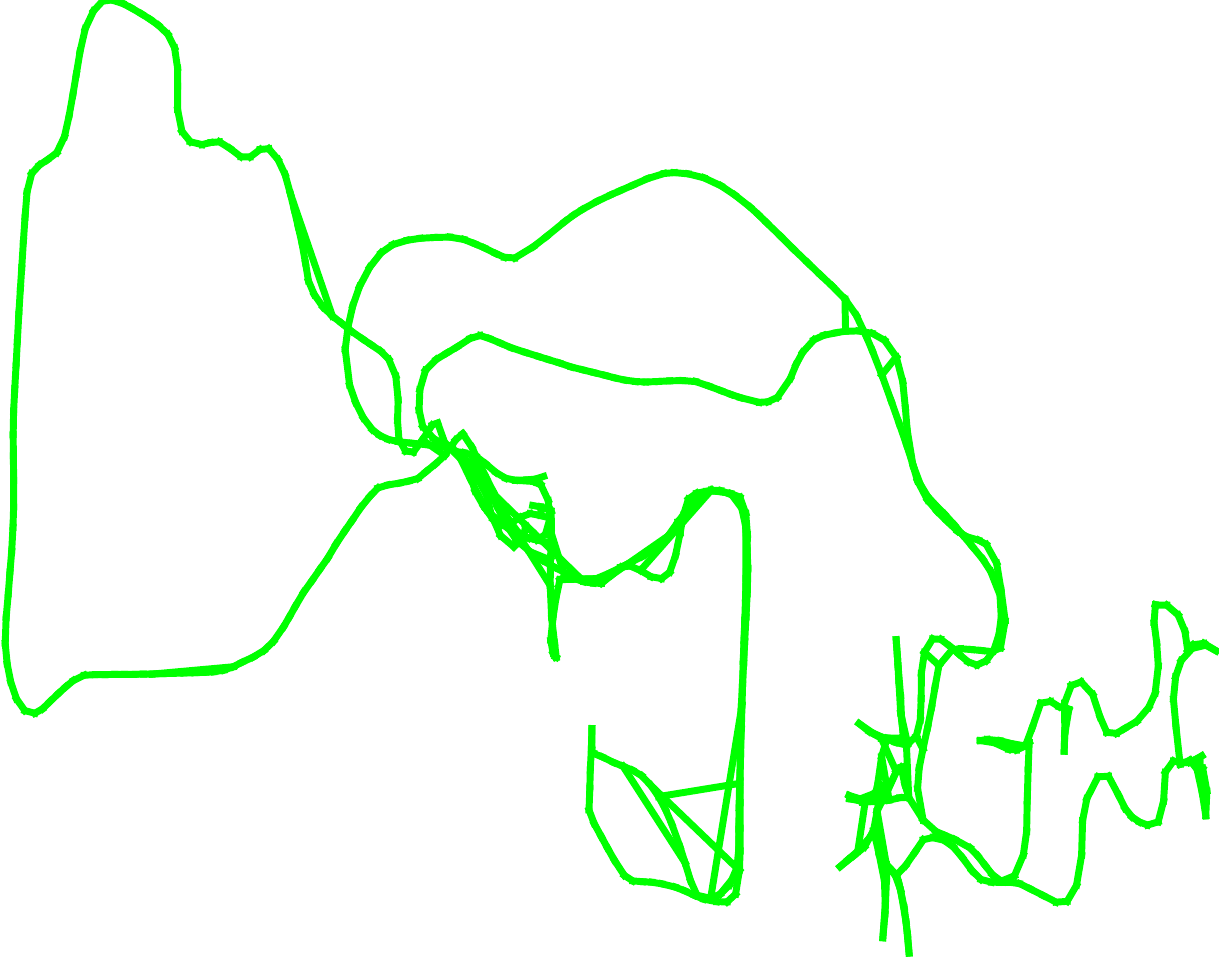} 
\end{minipage}
&
\begin{minipage}{0.5\columnwidth}%
\centering%
\includegraphics[width=\columnwidth, trim= 3.5cm 0cm 4.5cm 0cm, clip, angle=90]{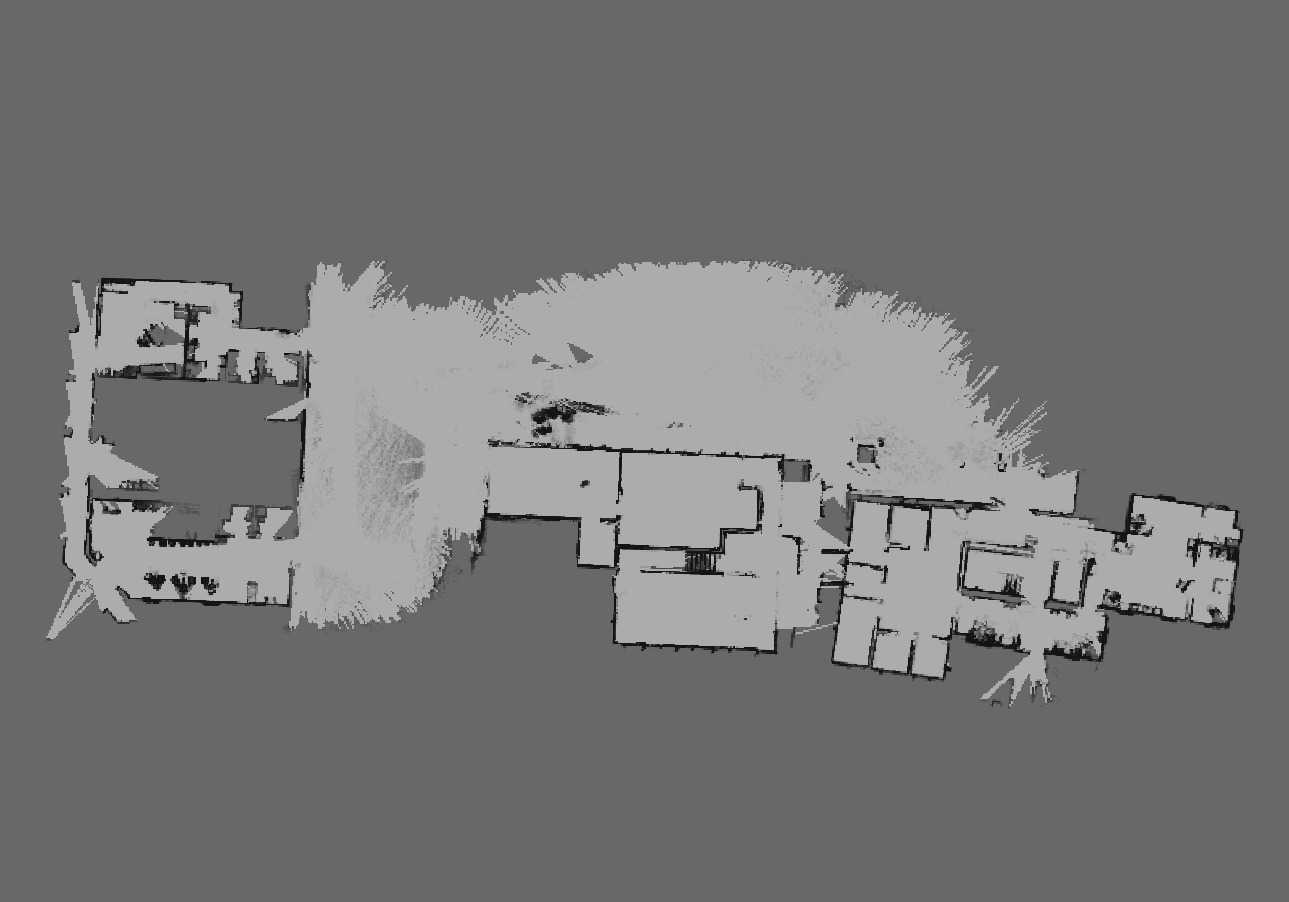} 
\end{minipage}\\
\vspace{0.2cm}

\begin{minipage}{0.5\columnwidth}%
\centering%
\includegraphics[width=\columnwidth, trim=15cm 0cm 20cm 0cm,clip]{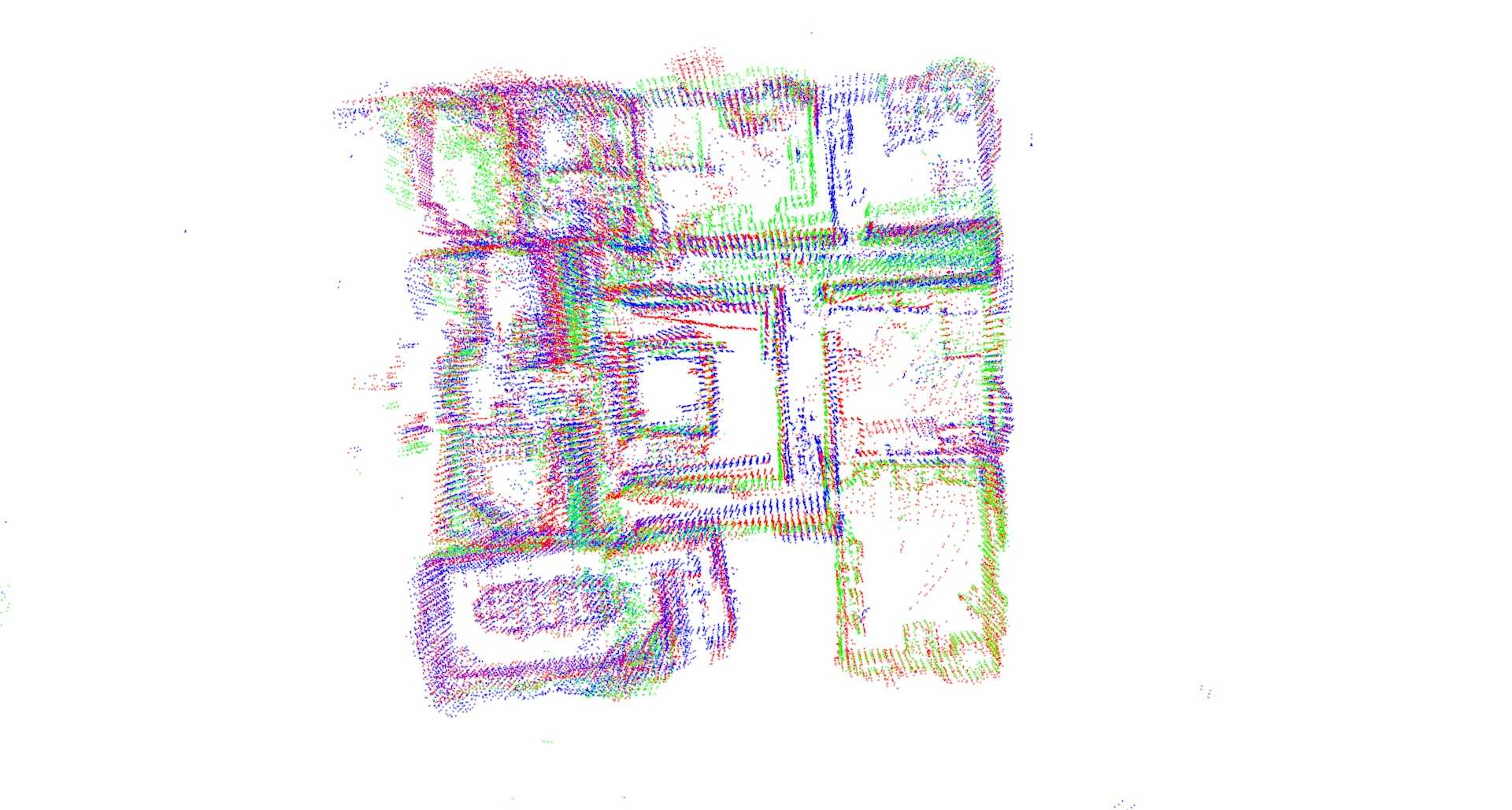} 
\end{minipage}
&
\begin{minipage}{0.5\columnwidth}%
\centering%
\includegraphics[width=\columnwidth, trim=0cm 0cm 0cm 0cm,clip]{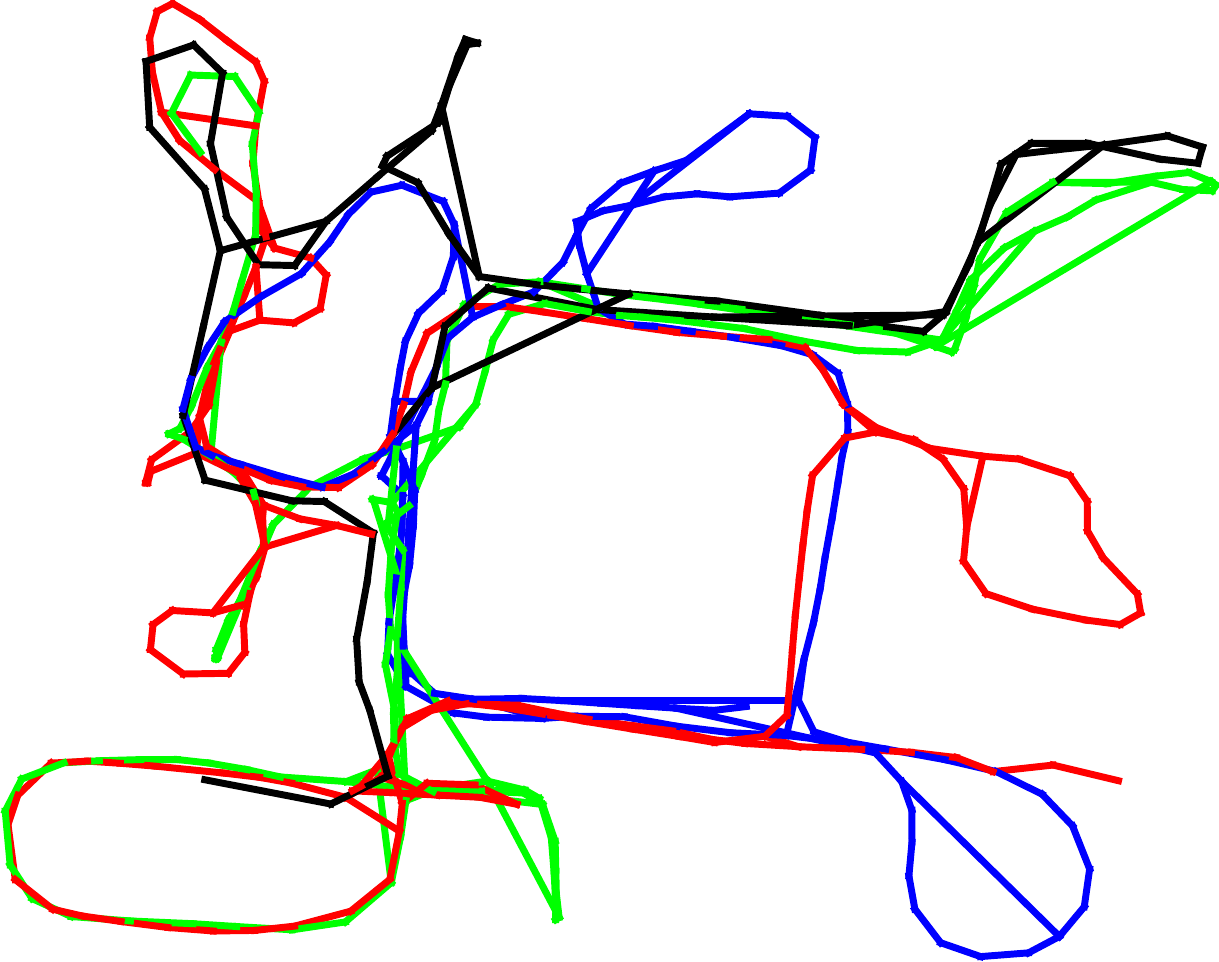} 
\end{minipage}
&
\begin{minipage}{0.5\columnwidth}%
\centering%
\includegraphics[width=\columnwidth, trim=0cm 0cm 0cm 0cm,clip]{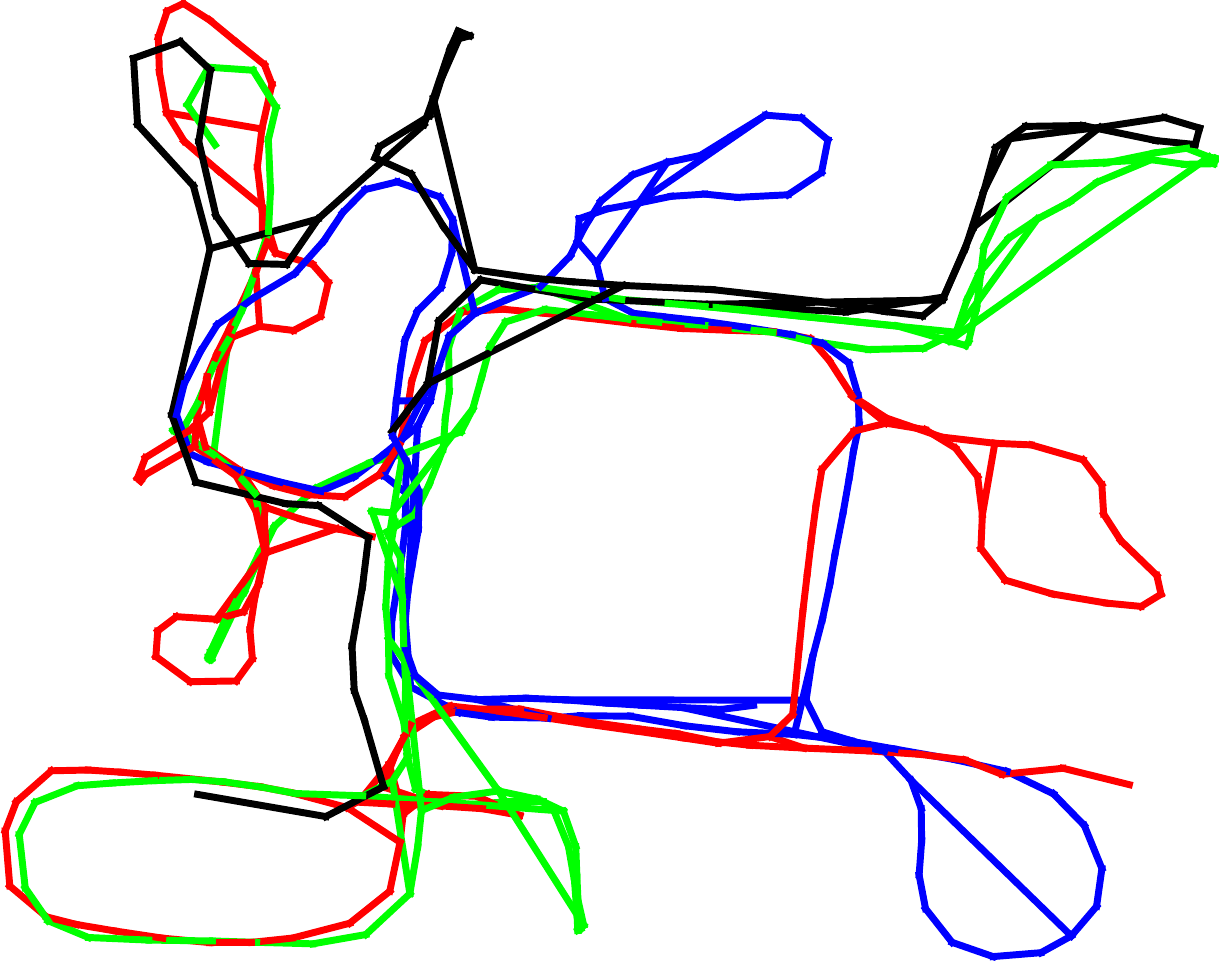} 
\end{minipage}

&
\begin{minipage}{0.5\columnwidth}%
\centering%
\includegraphics[width=\columnwidth, trim=0cm 0cm 0cm 0cm,clip]{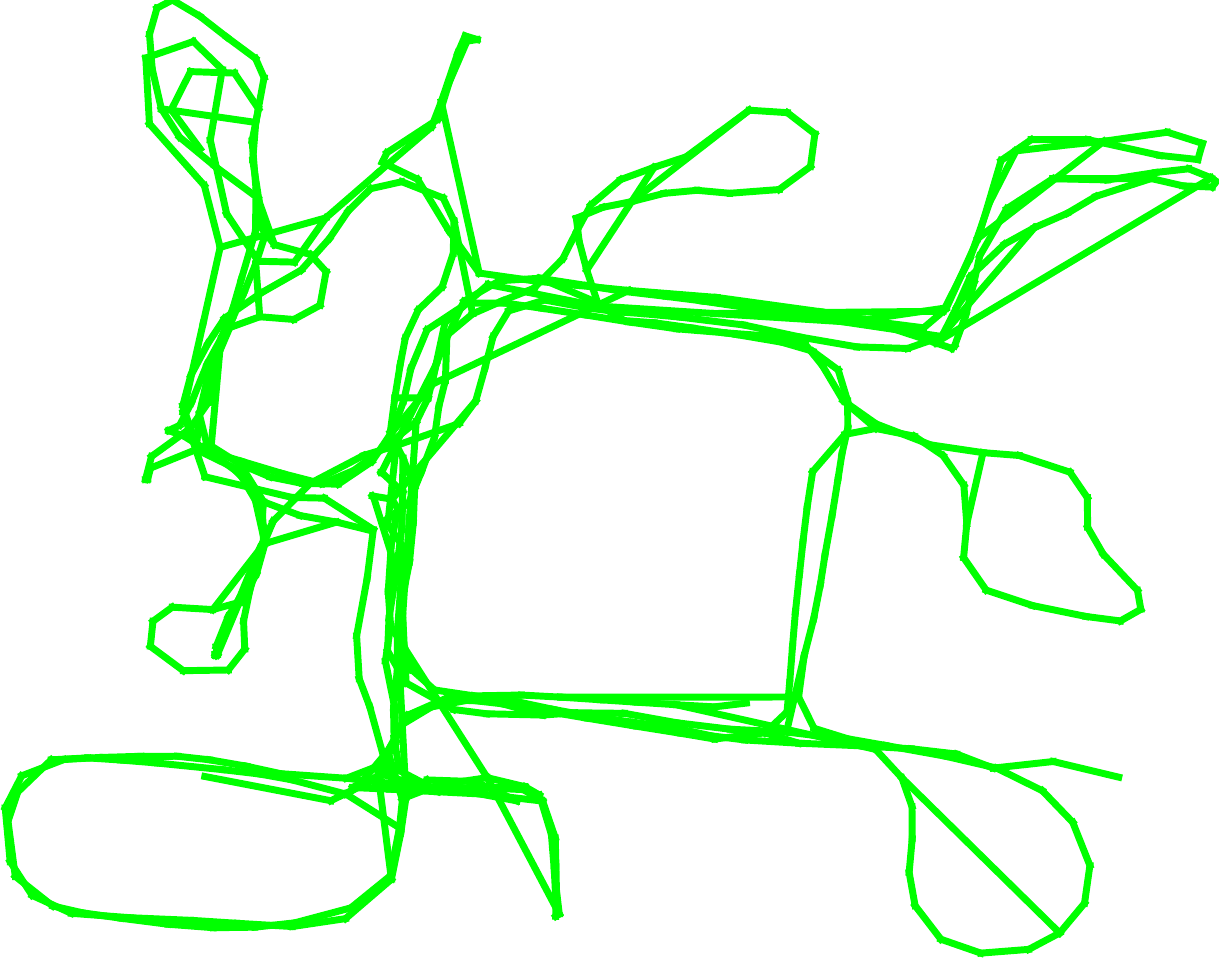} 
\end{minipage}
&
\begin{minipage}{0.5\columnwidth}%
\centering%
\includegraphics[width=\columnwidth, trim= 4.5cm 0cm 3.0cm 0cm, clip]{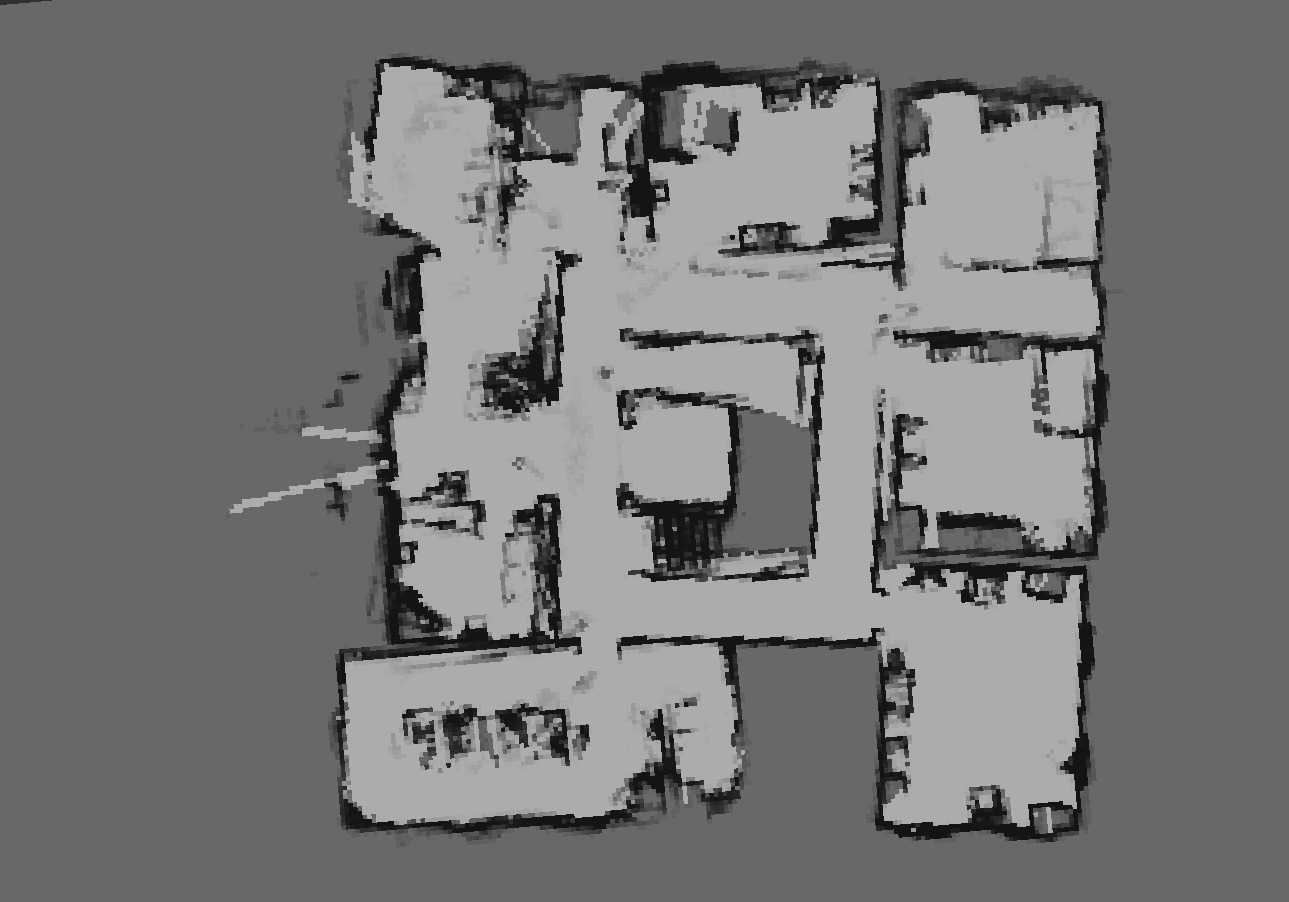} 
\end{minipage}
\vspace{0.2cm}


\end{tabular}
  \end{minipage}
  \caption{
   Mixed indoor-outdoor scenarios: 
(Left) aggregated point cloud obtained from the \DGS trajectory estimate. 
(Center) estimated trajectories for \DGS, \GN and \DDFSAM (robots shown 
  in red, blue, green and black for the distributed techniques). 
(Right) overall occupancy grid map obtained from the \DGS trajectory estimate.
  \label{fig:realExperiments1}
  \vspace{-0.5cm}
  }
\end{figure*}

\begin{figure*}[t]
\begin{minipage}{0.7\columnwidth}
\hspace{2cm}
\begin{tabular}{c|cc|c}%

 Point Cloud & \DGS &  Centralized  & Occupancy Grid \\

\begin{minipage}{0.5\columnwidth}%
\centering%
-
\end{minipage}
&
\begin{minipage}{0.5\columnwidth}%
\centering%
\includegraphics[width=\columnwidth, trim=0cm 0cm 0cm 0cm,clip]{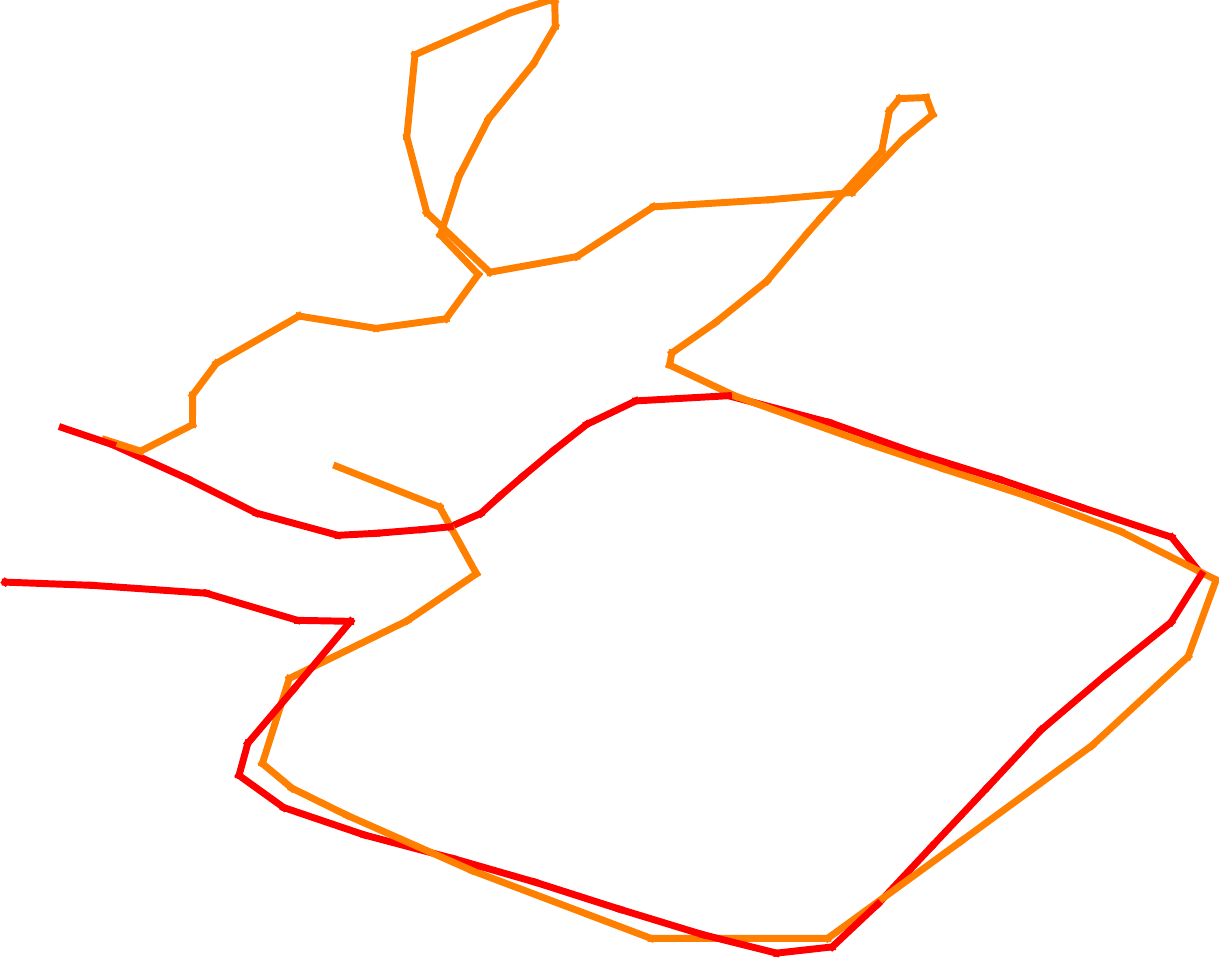} 
\end{minipage}
&
\begin{minipage}{0.5\columnwidth}%
\centering%
\includegraphics[width=\columnwidth, trim=0cm 0cm 0cm 0cm,clip]{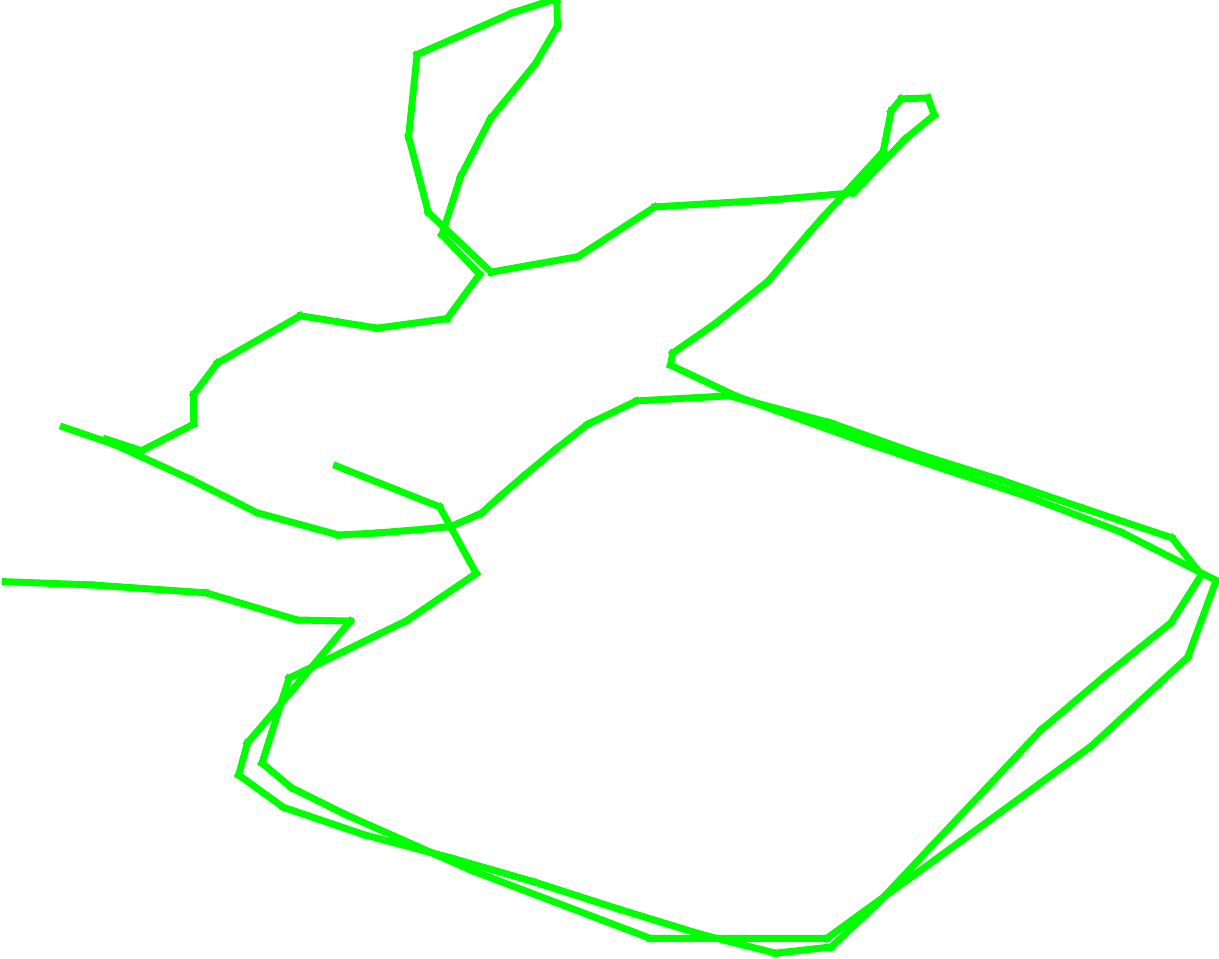} 
\end{minipage}
&
\begin{minipage}{0.5\columnwidth}%
\centering%
-
\end{minipage}\\
\vspace{0.2cm}

\begin{minipage}{0.5\columnwidth}%
\centering%
\includegraphics[width=\columnwidth, trim=0cm 0cm 0cm 0cm,clip]{figures/realTests/1} 
\end{minipage}
&
\begin{minipage}{0.5\columnwidth}%
\centering%
\includegraphics[width=\columnwidth, trim=0cm 0cm 0cm 0cm,clip]{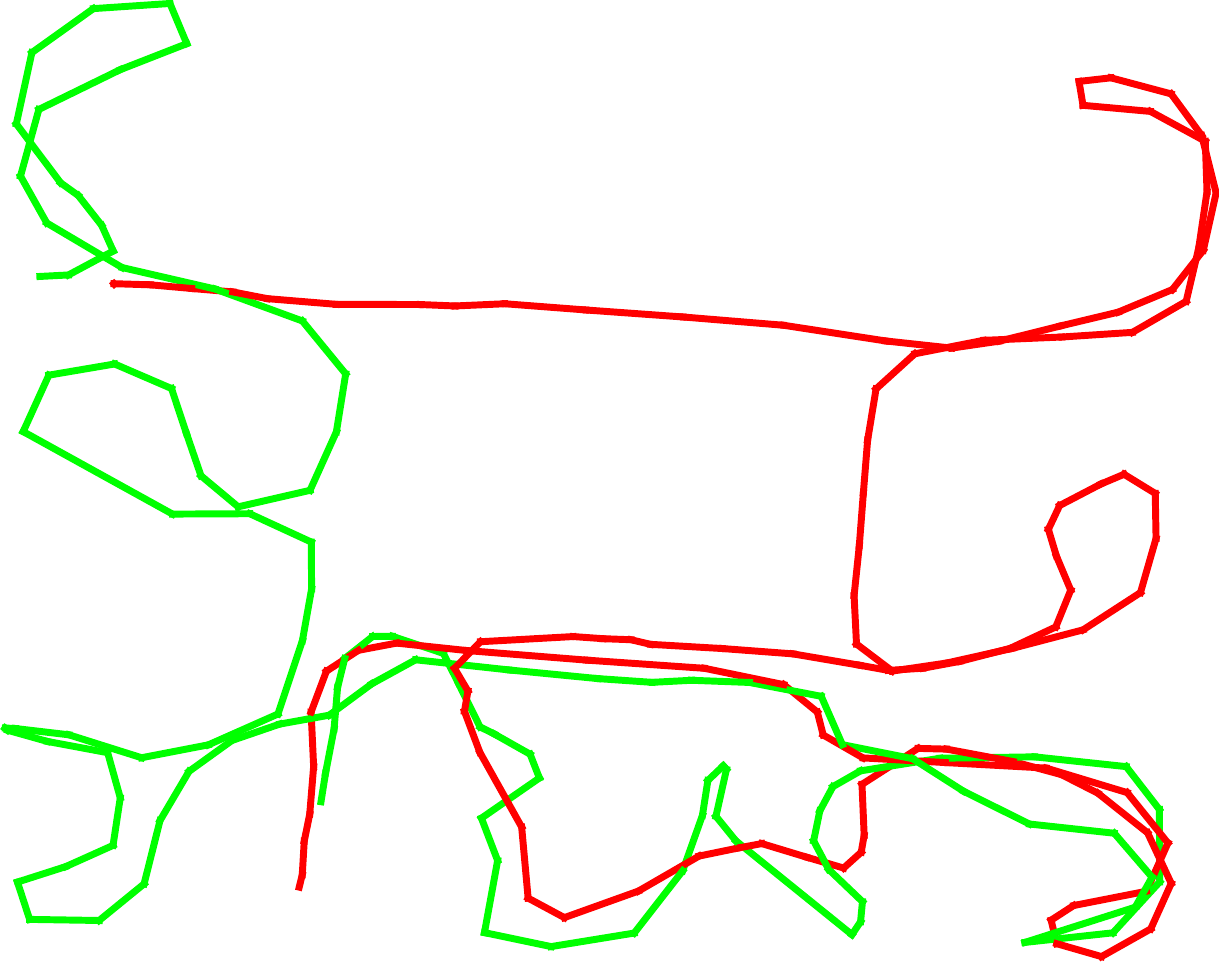} 
\end{minipage}
&
\begin{minipage}{0.5\columnwidth}%
\centering%
\includegraphics[width=\columnwidth, trim=0cm 0cm 0cm 0cm,clip]{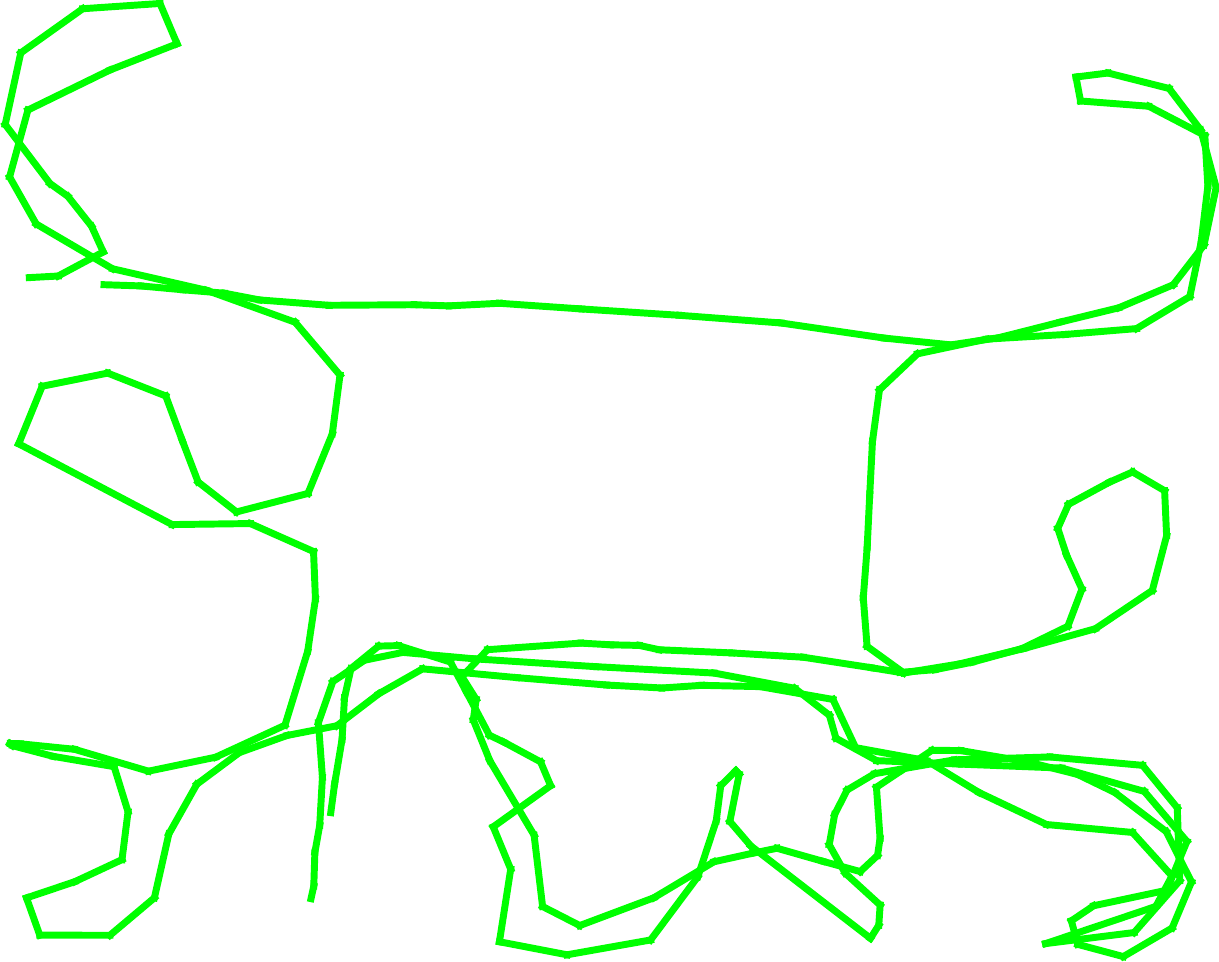} 
\end{minipage}
&
\begin{minipage}{0.5\columnwidth}%
\centering%
\includegraphics[width=\columnwidth, trim= 3.5cm 0cm 4.5cm 0cm, clip, angle=-90]{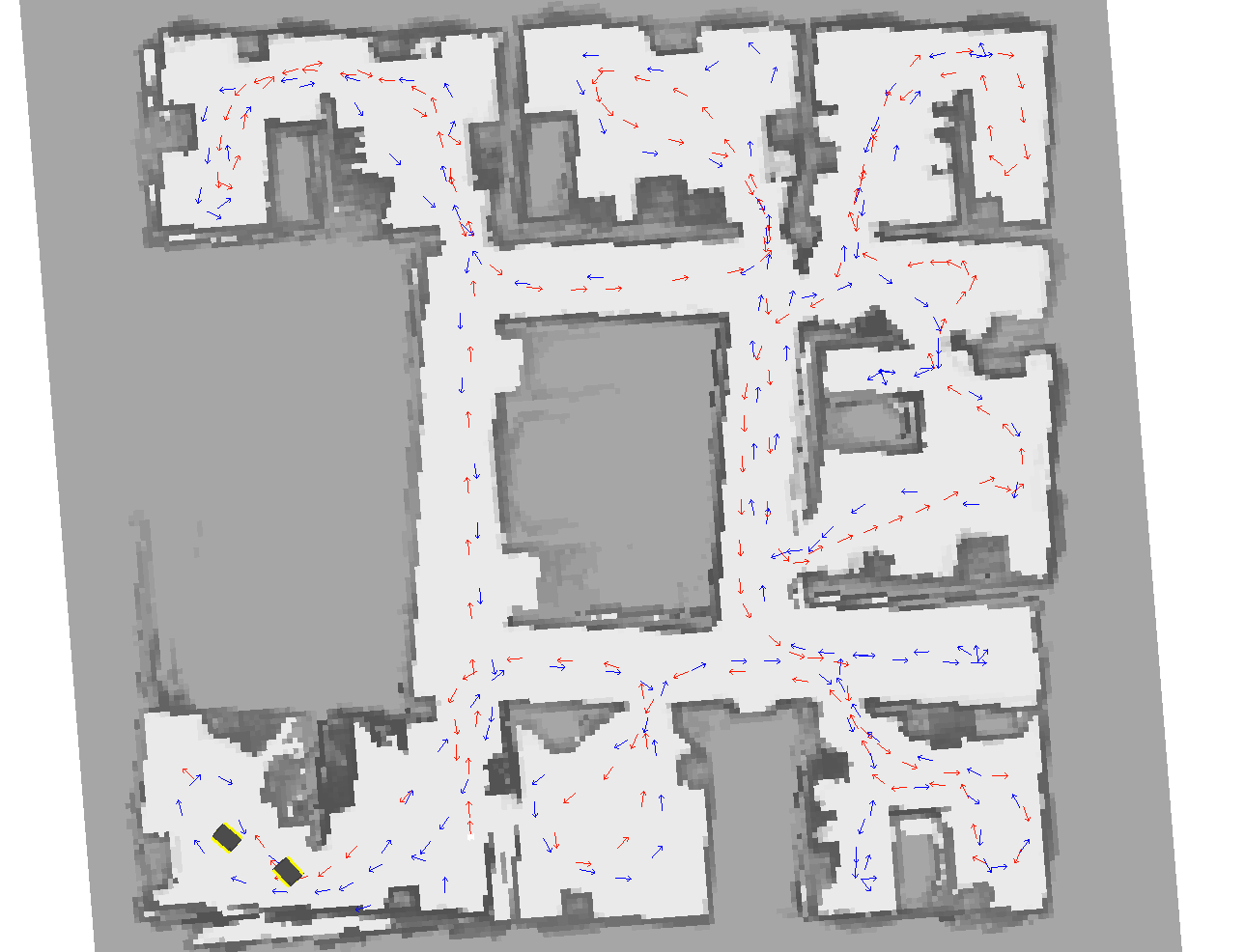} 
\end{minipage}\\
\vspace{0.2cm}

\begin{minipage}{0.5\columnwidth}%
\centering%
\includegraphics[width=\columnwidth, trim=0cm 0cm 0cm 0cm,clip]{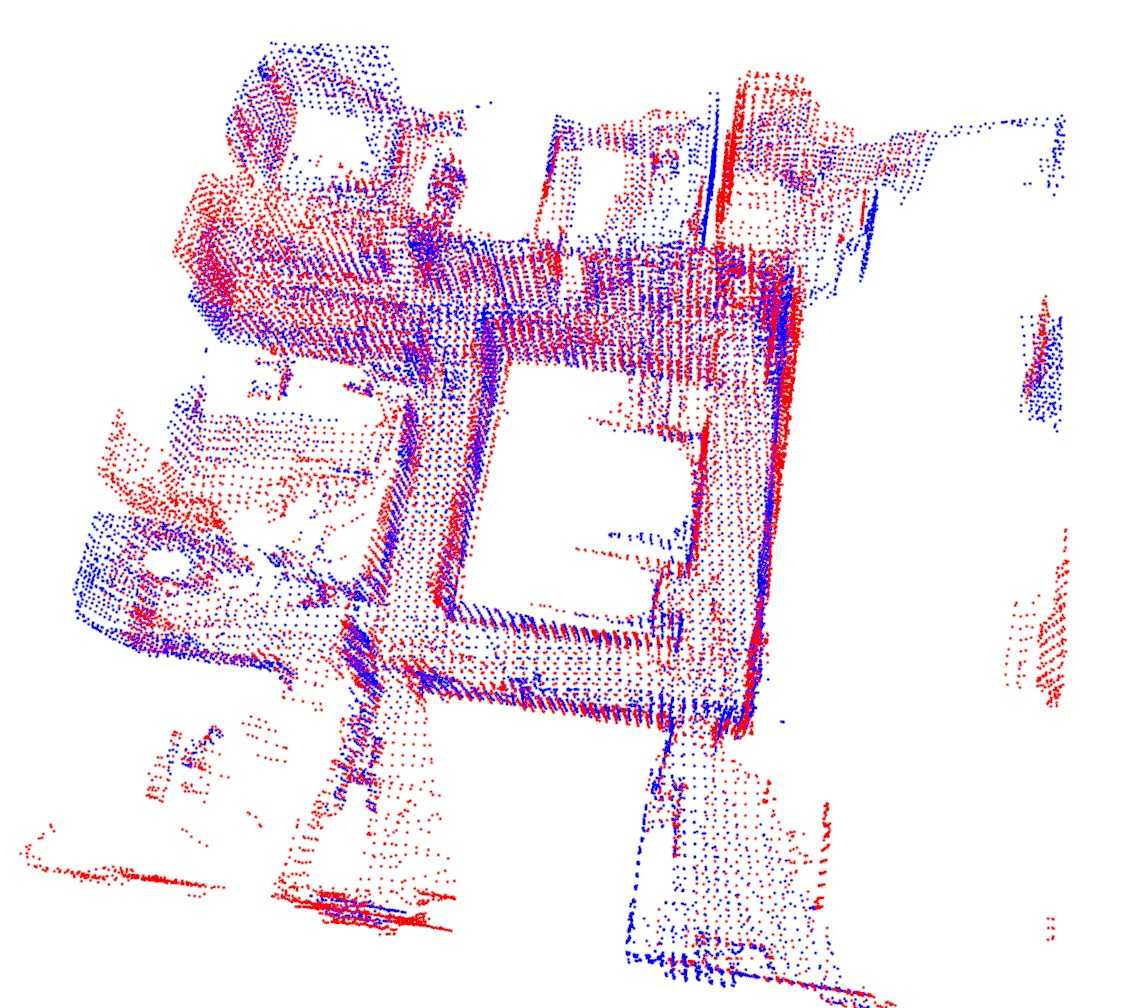} 
\end{minipage}
&
\begin{minipage}{0.5\columnwidth}%
\centering%
\includegraphics[width=\columnwidth, trim=0cm 0cm 0cm 0cm,clip]{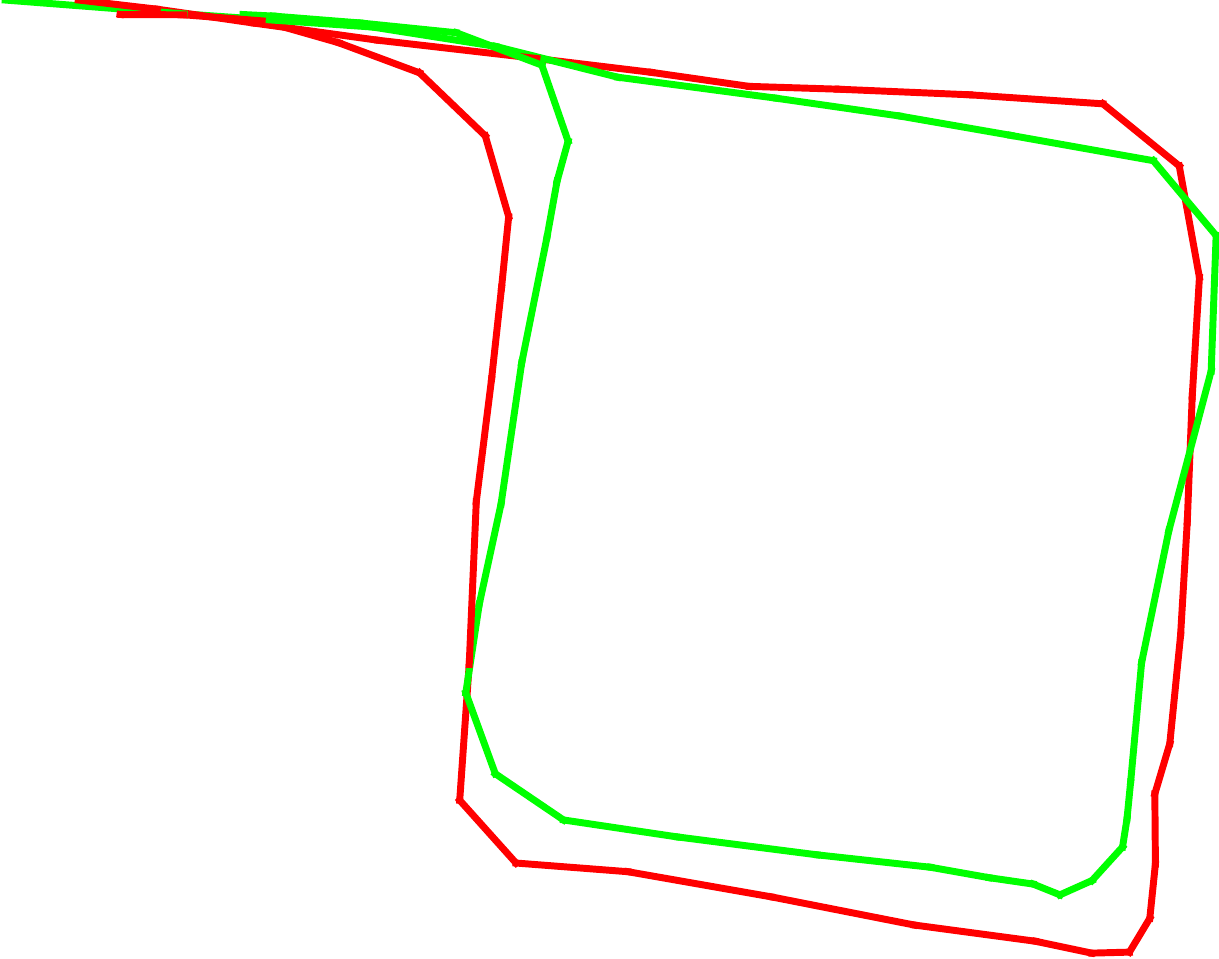} 
\end{minipage}
&
\begin{minipage}{0.5\columnwidth}%
\centering%
\includegraphics[width=\columnwidth, trim=0cm 0cm 0cm 0cm,clip]{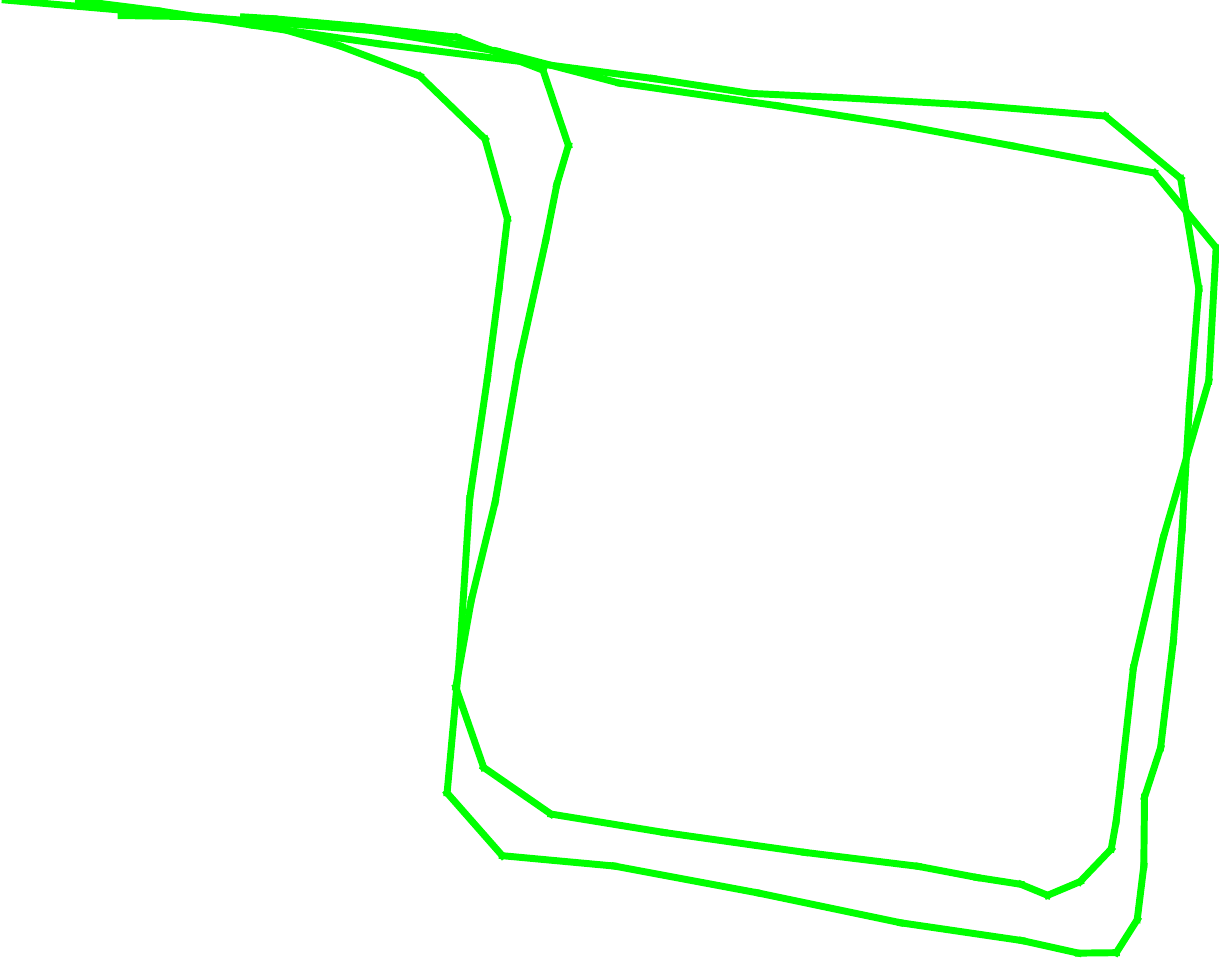} 
\end{minipage}
&
\begin{minipage}{0.5\columnwidth}%
\centering%
-
\end{minipage}\\
\vspace{0.2cm}

\begin{minipage}{0.5\columnwidth}%
\centering%
\includegraphics[width=\columnwidth, trim=0cm 0cm 0cm 0cm,clip]{figures/realTests/2} 
\end{minipage}
&
\begin{minipage}{0.5\columnwidth}%
\centering%
\includegraphics[width=\columnwidth, trim=0cm 0cm 0cm 0cm,clip]{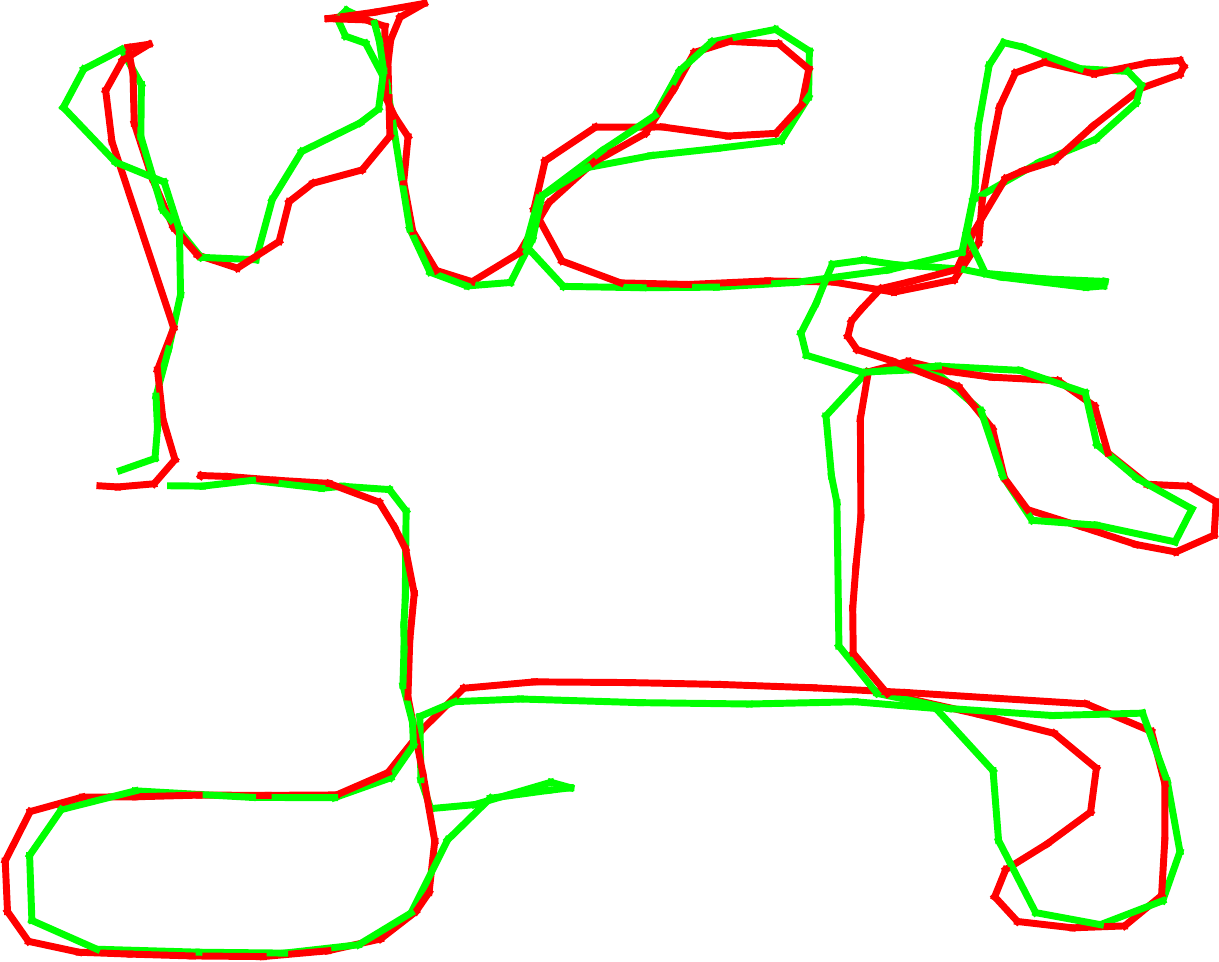} 
\end{minipage}
&
\begin{minipage}{0.5\columnwidth}%
\centering%
\includegraphics[width=\columnwidth, trim=0cm 0cm 0cm 0cm,clip]{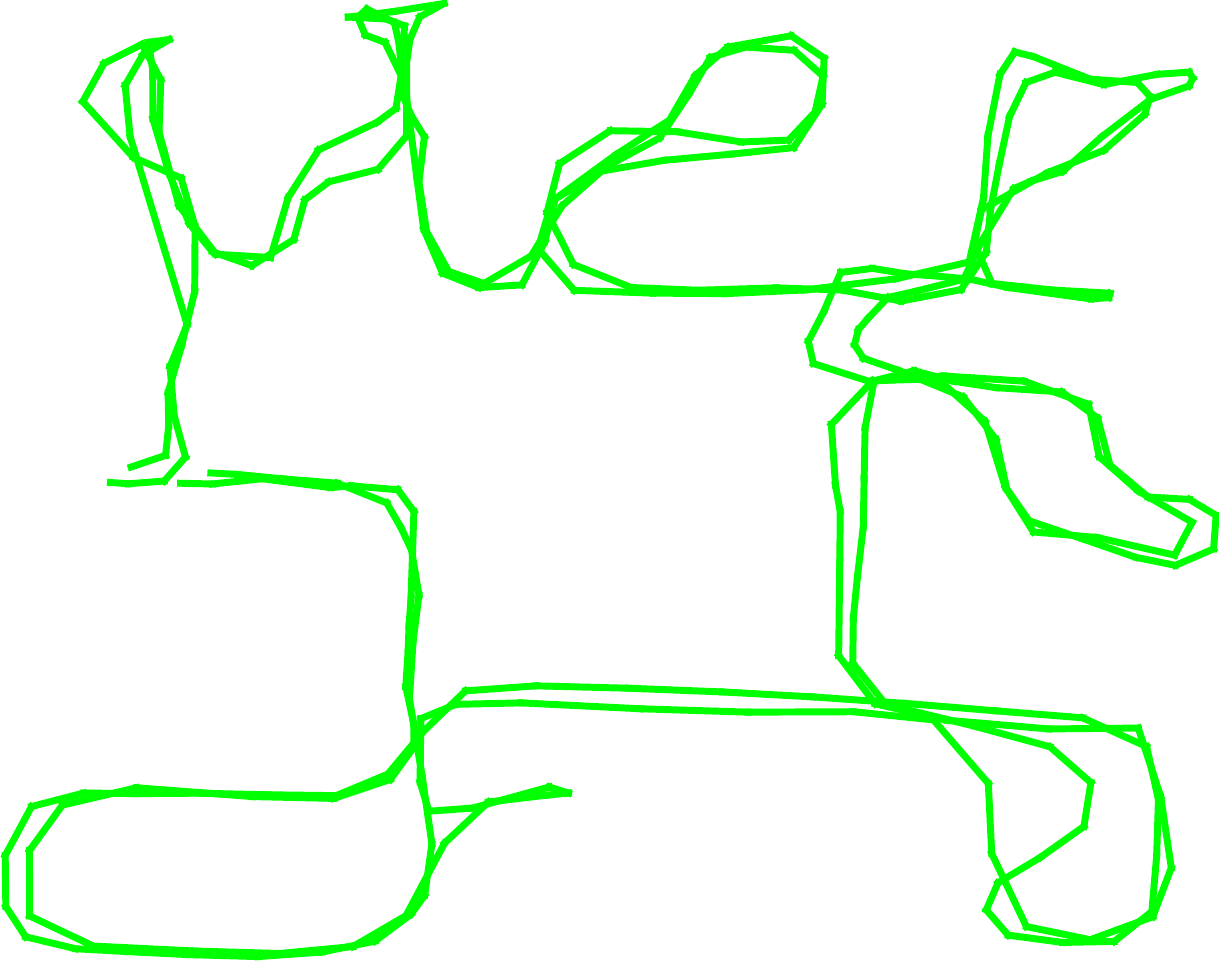} 
\end{minipage}
&
\begin{minipage}{0.5\columnwidth}%
\centering%
\includegraphics[width=\columnwidth, trim= 3.5cm 0cm 4.5cm 0cm, clip, angle=-90]{figures/realTests/hotel-1-03/ortho_map} 
\end{minipage}
\vspace{0.2cm}


\end{tabular}
\end{minipage}
\caption{
Early tests with 2 robots: 
(Left) aggregated point cloud obtained from the \DGS trajectory estimate. 
(Center) estimated trajectories for \DGS and \GN. 
(Right) overall occupancy grid map obtained from the \DGS trajectory estimate.
  \label{fig:realExperiments3}
  \vspace{1cm}
  }
\end{figure*}

\subsection{Field Experiments: Multi Robot Pose Graph Optimization}
\label{sec:real}

We tested the \DGS approach on field data collected by two to four Jackal robots (\Fig\ref{fig:jackalRobot}), 
moving in a MOUT (\emph{Military Operations in Urban Terrain}) test facility.  
Each robot collects 3D scans using Velodyne 32E, and uses IMU and wheel odometry 
to measure its ego-motion. 
3D scans are used to compute inter-robot measurements (via ICP) during rendezvous.
We evaluated our approach in multiple buildings in the MOUT test facility. 

\Figs\ref{fig:realExperiments1}, \ref{fig:realExperiments2}, \ref{fig:realExperiments3} show the 
aggregated 3D point clouds (left), the estimates trajectories (center), and the aggregated 
occupancy grid map (right) over multiple runs.
The central part of the figures
 compares the \DGS estimate against the \DDFSAM estimate and the corresponding centralized estimate.
Note that the test scenarios cover a broad set of operating conditions. 
For instance \Fig\ref{fig:realExperiments2} corresponds to experiments with 4 robots
moving in indoor environment, while \Fig\ref{fig:realExperiments1} corresponds to tests performed 
in a mixed indoor-outdoor scenario (with robots moving on gravel when outdoor, Fig. \ref{fig:jackalRobotOnGravel}).
The four tests of \Fig\ref{fig:realExperiments3} correspond to early results with 2 robots for which we 
do not have a comparison against \DDFSAM. 

Quantitative results are given in Table \ref{tab:realExperiments}, 
which reports the cost attained by the \DGS algorithm as compared to the centralized \GN cost and \DDFSAM. Number of iterations, ATE* and ARE* are also shown.
Each line of the table shows statistics for each of the 15 field tests in the MOUT facility. The first four rows 
(tests 0 to 3) correspond to tests performed in a mixed
indoor-outdoor scenario (\Fig\ref{fig:realExperiments1}). The next seven rows (tests 4 to 10) correspond to tests performed with 4 robots in an indoor environment. The last four rows (tests 11 to 14)
correspond to early results with 2 robots. Higher ATE* and ARE* in the first few rows is due to the fact that the robots move on gravel in outdoors which introduces larger odometric errors. Consistently with what we observed in the 
previous sections, larger measurement errors may induce the \DGS algorithm to perform 
more iterations to reach consensus (e.g., test 3).
The columns ``\#vertices'' and ``\#edges'' describe the size of the overall factor graph (including all robots), 
while the column ``\#links'' reports the total number of rendezvous events. 
In all the tests \DDFSAM performed worse than \DGS which is shown by higher cost attained by \DDFSAM as compared to \DGS. 
This is because \DDFSAM requires good linearlization points to generate condensed graphs and
instead bad linearization points during communication can introduce linearlization errors resulting in higher cost. 


\begin{table*}[h!]
\centering

{
\renewcommand{\arraystretch}{1.8}%
 \setlength\tabcolsep{3.5pt}

\begin{tabular}{|c|c|c|c|c|c|c|c|c|c|c|c|c|c|c|}

\hline 
 \multirow{3}{*}{\#Test} & \multirow{3}{*}{\#vertices} &   \multirow{3}{*}{\#edges} &  \multirow{3}{*}{\#links} & \multicolumn{8}{c|}{Distributed Gauss-Seidel} & \multicolumn{2}{c|}{Centralized} & \multirow{3}{*}{DDF-SAM}\\
\cline{5-14}
& & & & \multicolumn{4}{c|}{$\eta_r =\eta_p = 10^{-1}$} & \multicolumn{4}{c|}{$\eta_r =\eta_p = 10^{-2}$} & Two-Stage & \GN & \\
\cline{5-12}
& & &  & \#Iter & Cost & ATE* & ARE*  & \#Iter & Cost & ATE* & ARE* & Cost & Cost & \\
\hline 
\hline 
0 & 194 & 253 & 16 & 12 & 1.42 & 0.21 & 1.63 & 197 & 1.40 & 0.07 & 0.67 & 1.40 & 1.40 & 4.86 \\ 
 \hline 
1 & 511 & 804 & 134 & 10 & 0.95 & 1.22 & 6.64 & 431 & 0.91 & 1.18 & 6.37 & 0.89 & 0.89 & 6.88 \\ 
 \hline 
2 & 547 & 890 & 155 & 21 & 1.99 & 1.03 & 4.74 & 426 & 1.95 & 1.08 & 4.79 & 1.93 & 1.93 & 12.54 \\ 
 \hline 
3 & 657 & 798 & 47 & 176 & 0.32 & 0.68 & 2.40 & 446 & 0.32 & 0.69 & 2.06 & 0.32 & 0.32 & 2.39 \\ 
 \hline 
4 & 510 & 915 & 179 & 23 & 10.89 & 1.10 & 6.69 & 782 & 10.57 & 0.71 & 4.53 & 10.51 & 10.50 & 37.94 \\ 
 \hline 
5 & 418 & 782 & 151 & 13 & 3.02 & 0.51 & 5.75 & 475 & 2.92 & 0.39 & 4.32 & 2.91 & 2.90 & 18.31 \\ 
 \hline 
6 & 439 & 720 & 108 & 26 & 9.28 & 0.68 & 5.08 & 704 & 9.12 & 0.31 & 2.39 & 9.10 & 9.07 & 72.76 \\ 
 \hline 
7 & 582 & 1152 & 228 & 10 & 3.91 & 0.31 & 3.40 & 579 & 3.78 & 0.26 & 2.43 & 3.78 & 3.78 & 16.38 \\ 
 \hline 
8 & 404 & 824 & 183 & 11 & 1.92 & 0.13 & 1.78 & 410 & 1.89 & 0.12 & 1.25 & 1.89 & 1.89 & 6.82 \\ 
 \hline 
9 & 496 & 732 & 86 & 41 & 4.30 & 0.54 & 4.20 & 504 & 4.29 & 0.45 & 3.04 & 4.28 & 4.27 & 21.53 \\ 
 \hline 
10 & 525 & 923 & 147 & 15 & 5.56 & 0.39 & 3.93 & 577 & 5.43 & 0.23 & 2.04 & 5.43 & 5.40 & 19.59 \\ 
 \hline 
11 & 103 & 107 & 3 & 71 & 0.85 & 1.58 & 14.44 & 328 & 0.84 & 0.27 & 2.18 & 0.84 & 0.84 & - \\ 
 \hline 
12 & 227 & 325 & 50 & 16 & 0.79 & 1.11 & 10.44 & 511 & 0.71 & 0.80 & 7.02 & 0.68 & 0.68 & - \\ 
 \hline 
13 & 77 & 127 & 26 & 10 & 0.33 & 0.34 & 2.23 & 78 & 0.26 & 0.21 & 1.25 & 0.26 & 0.26 & - \\ 
 \hline 
14 & 322 & 490 & 85 & 28 & 1.42 & 0.83 & 5.05 & 606 & 1.07 & 0.47 & 2.10 & 1.04 & 1.04 & - \\ 
 \hline 
  \end{tabular}} 
  \vspace{0.2cm}
\caption{\label{tab:realExperiments} 
Performance of \DGS on field data as compared to the centralized \GN method and \DDFSAM. Number of iterations, ATE* and ARE* with respect to centralized Gauss-Newton estimate are also shown. 
}

\end{table*}



\begin{figure}[t]
\centering
\includegraphics[width=0.9\columnwidth, trim= 0cm 0cm 0cm 0cm, clip]{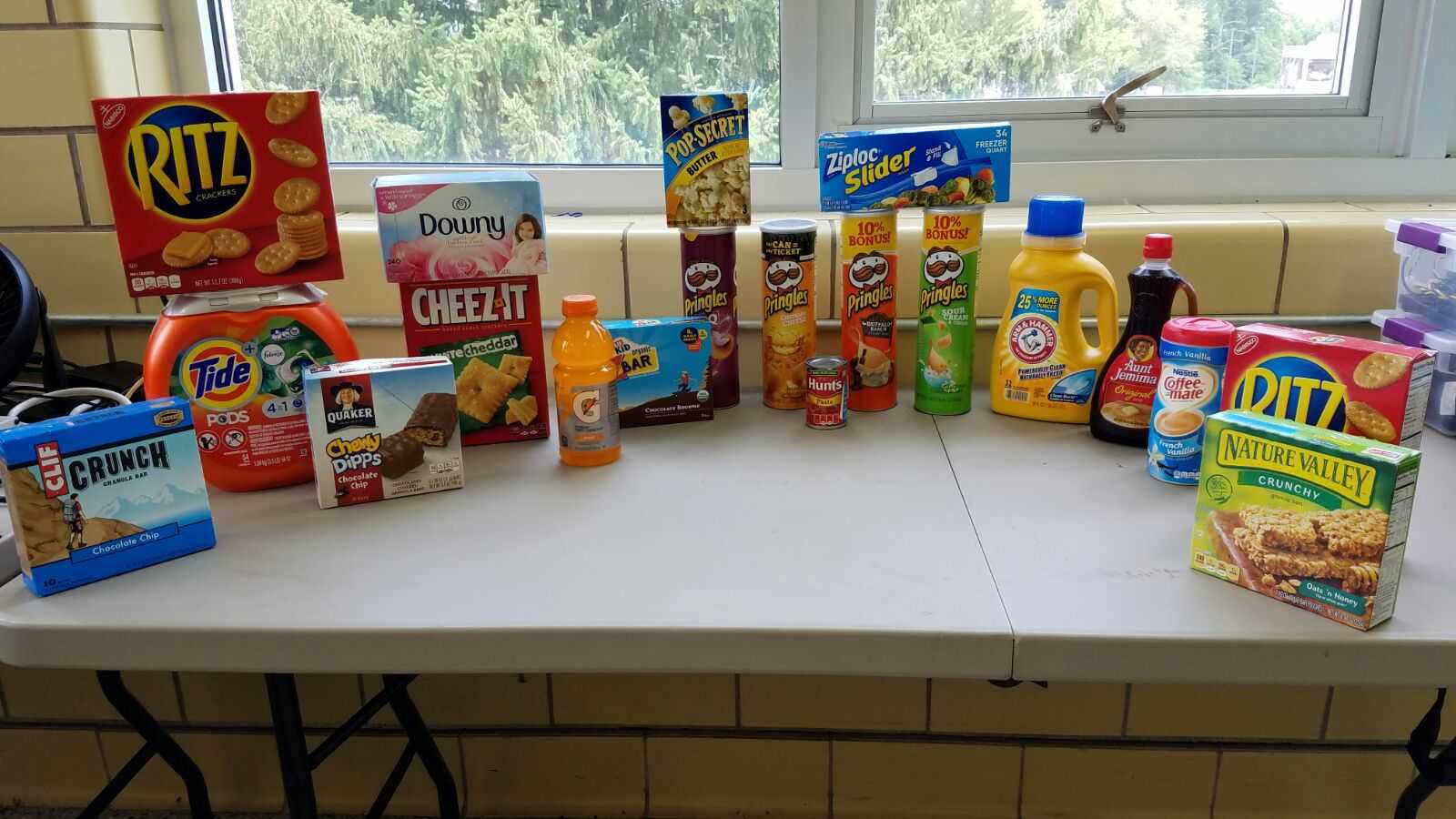}  \\

\caption{
Objects from the BigBird dataset used in the field experiments 
of Section~\ref{sec:real_objectSLAM}.\label{fig:figObjects}
}
\end{figure}



\begin{figure}[t]
\includegraphics[width=0.48\columnwidth, trim= 0cm 0cm 0cm 0cm, clip]{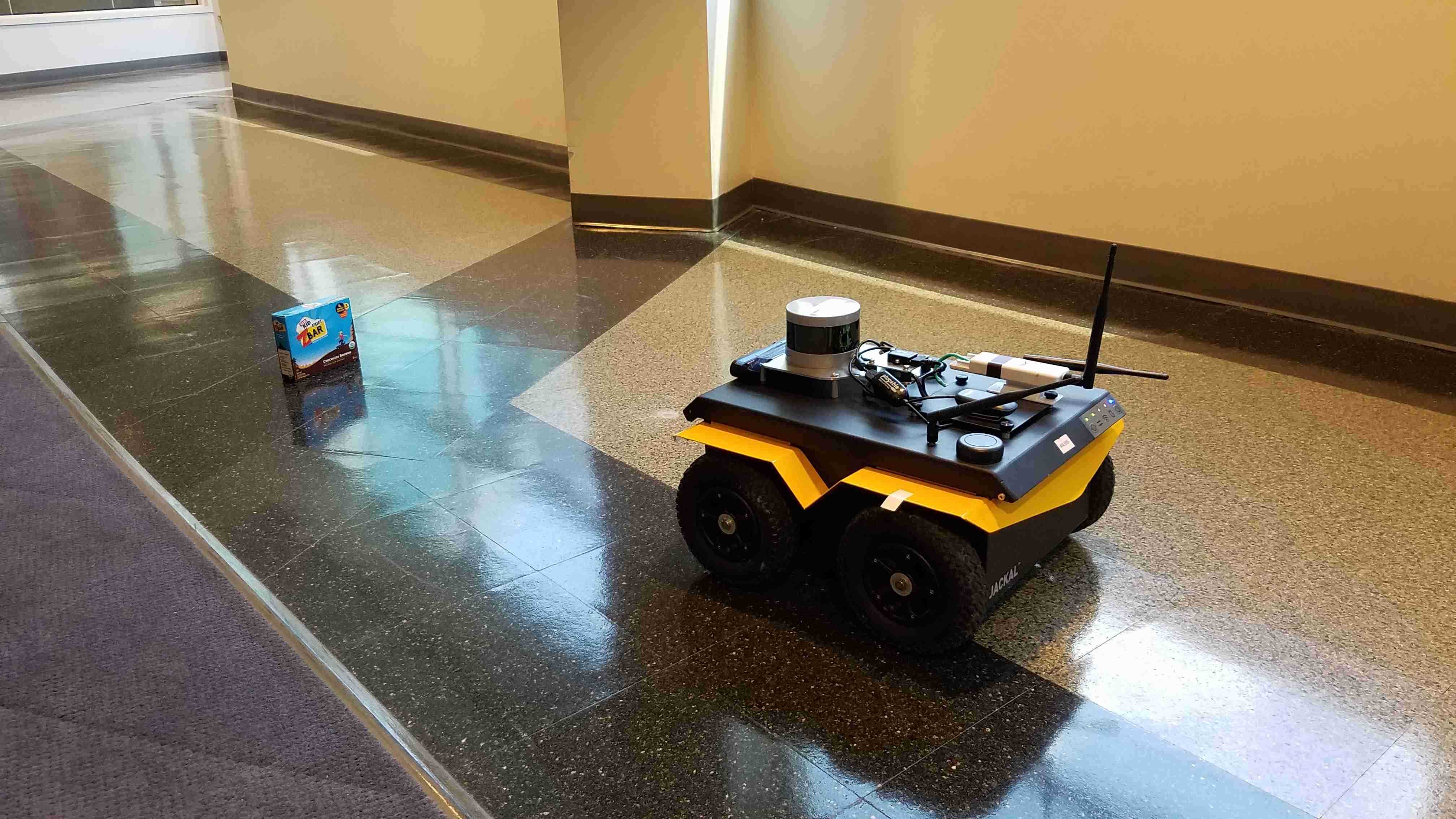}
\includegraphics[width=0.48\columnwidth, trim= 0cm 0cm 0cm 0cm, clip]{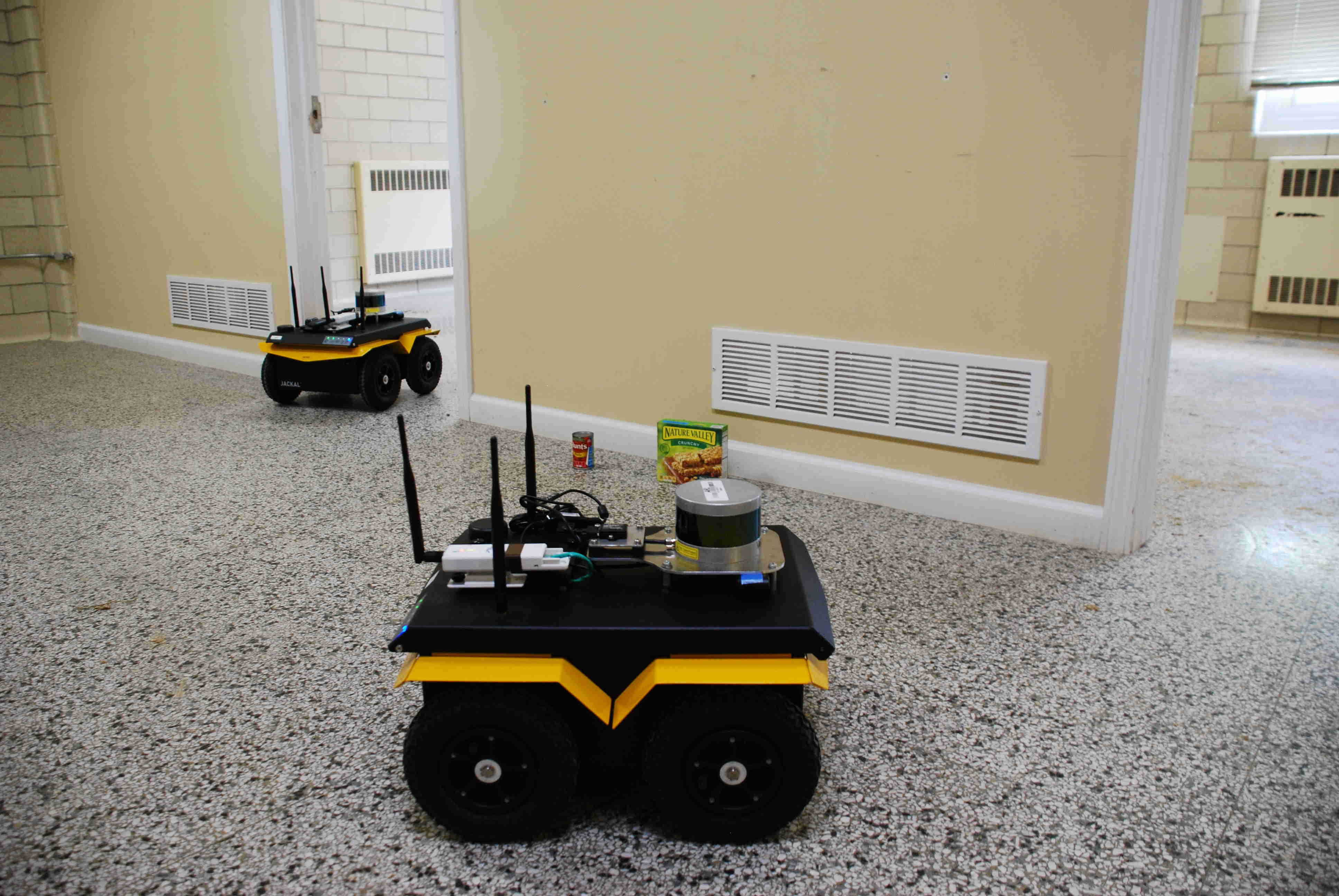}    \\
\caption{
Snapshot of the 
test facility, the two Clearpath Jackal robots, and the objects 
used for object-based SLAM for the tests of  Section~\ref{sec:real_objectSLAM}.\label{fig:jackalRobot_objectSLAM}
}
\end{figure}


\subsection{Field Experiments: Multi Robot Object-based SLAM}
\label{sec:real_objectSLAM}

We test the combination of the \DGS method and our object-based model on
 field data collected by two Jackal robots (\Fig\ref{fig:jackalRobot_objectSLAM}) 
 moving in a MOUT facility. 
We scattered the environment with a set of 
objects (\Fig\ref{fig:figObjects}) from the BigBird dataset (\cite{Singh14icra}).
Each robot is equipped with an Asus Xtion RGB-D sensor and uses wheel odometry to measure its ego-motion. 
We use the RGB-D sensor for object detection and object pose estimation.

We evaluated our approach in two different scenarios, the \stadium and the \housea. 
We did two runs inside the \stadium (\stadiuma \& \stadiumb) and one run in the \housea with objects randomly spread along the robot trajectories. The \stadium datasets were collected in an indoor basketball stadium with the robot trajectories bounded in a roughly rectangular area. The \housea dataset was collected in the living room and kitchen area of a house.


\begin{figure}[t]
\centering
\includegraphics[width=0.45\columnwidth, trim= 0cm 0cm 0cm 0cm, clip]{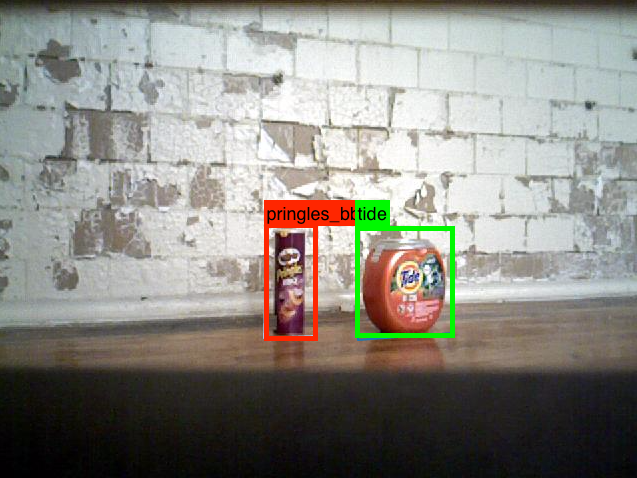}
\includegraphics[width=0.45\columnwidth, trim= 0cm 0cm 0cm 0cm, clip]{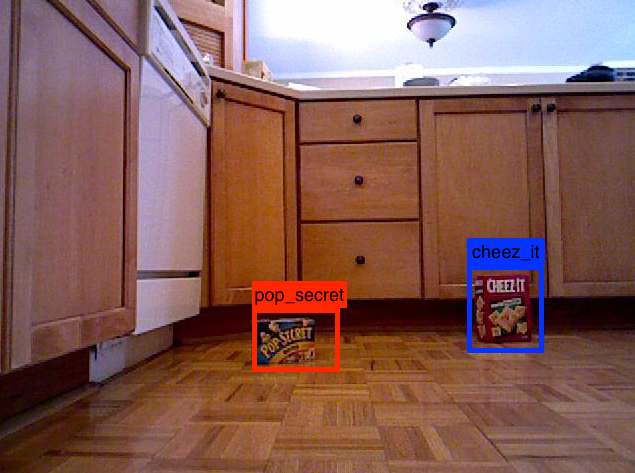}  \\
\caption{Snapshot of the YOLO object detection in two difference scenes: (left) \stadium dataset,
(right) \housea dataset. \label{fig:objectDetectionSnapshot} 
}

\end{figure}


\myparagraph{Object detection}
We used 12 objects from the BigBird dataset in all three runs. 
The two-stage process of object detection (semantic verification) followed by pose estimation  (geometric verification) ensured that we do not add false positive detections.  
Similarly to the standard \GN method, our current distributed optimization technique (\DGS) is not robust to outliers (more comments in Section~\ref{sec:conclusion}). 
 The detection thresholds can be further relaxed 
 when using robust pose graph optimization techniques.

In the first run (\stadiuma), 6 objects were added to the map out of the 12 objects present in the environment. Similarly, 5 objects were detected in \stadiumb and \housea. \Fig\ref{fig:objectDetectionSnapshot} shows 
a snapshot of the bounding box of the detected object in three different scenes. 
Videos showing YOLO object detection results on other datasets are available at \url{https://youtu.be/urZiIJK2IYk} and \url{https://youtu.be/-F6JpVmOrc0}.


\begin{table}[h!]
\centering

{
\renewcommand{\arraystretch}{1.2}%
 \setlength\tabcolsep{3.9pt} 
\begin{tabular}{|c|c|c|c|c|c|c|c|}

\hline 
 \multirow{3}{*}{Scenario}  & \multicolumn{2}{c|}{Avg. Per-Robot} & \multicolumn{2}{c|}{Avg. Comm.}\\
&  \multicolumn{2}{c|}{ Memory Req. (MB)} & \multicolumn{2}{c|}{Bandwidth Req. (MB)}\\
\cline{2-5}
&  PCD &  Obj & PCD &  Obj \\
\hline
\hline
Stadium-1 & 1.2e+03 & 1.9e+00 & 1.9e+01 & 1.5e-05\\
 \hline
Stadium-2 & 1.4e+03 & 1.9e+00 & 1.4e+01 & 1.1e-05\\
 \hline
House  & 2.1e+03 & 1.9e+00 & 1.6e+01 & 1.3e-05\\
 \hline
 
 \end{tabular}}
\caption{\label{tab:fieldExperimentsMemory} 
 Memory and communication requirements for our object based approach (Obj) as compared to Point cloud based approach (PCD) on field data.
}
\end{table}

\myparagraph{Memory Requirements}
Table \ref{tab:fieldExperimentsMemory} compares the average memory requirement per robot to store a dense point cloud map (PCD) with respect to storing a object-based map (Obj) in our real tests. 

Per-robot memory requirement in the case of dense point cloud is computed as $n_fKC$ where $n_f$ is the number of frames, $K$ is the number of points per frame and $C$ is the memory required to store each point. In the case of object level map, it is computed as $n_oPC$ where $n_o$ is the number of object models and $P$ is the average number of points in each object model. Table \ref{tab:fieldExperimentsMemory} shows that, 
as expected, the per-robot memory requirement is orders of magnitude smaller 
with our object-based map as compared to point-cloud-based maps. 

\begin{figure*}[t!]
\begin{minipage}{\columnwidth}
{\renewcommand{\arraystretch}{2.0}%
\begin{tabular}{cc|cc|cc}%

 Centralized & Distributed  &  Centralized & Distributed  &  Centralized & Distributed \\


\begin{minipage}{0.30\columnwidth}%
\centering%
\includegraphics[width=\columnwidth, trim=0cm 0cm 0cm 0cm,clip]{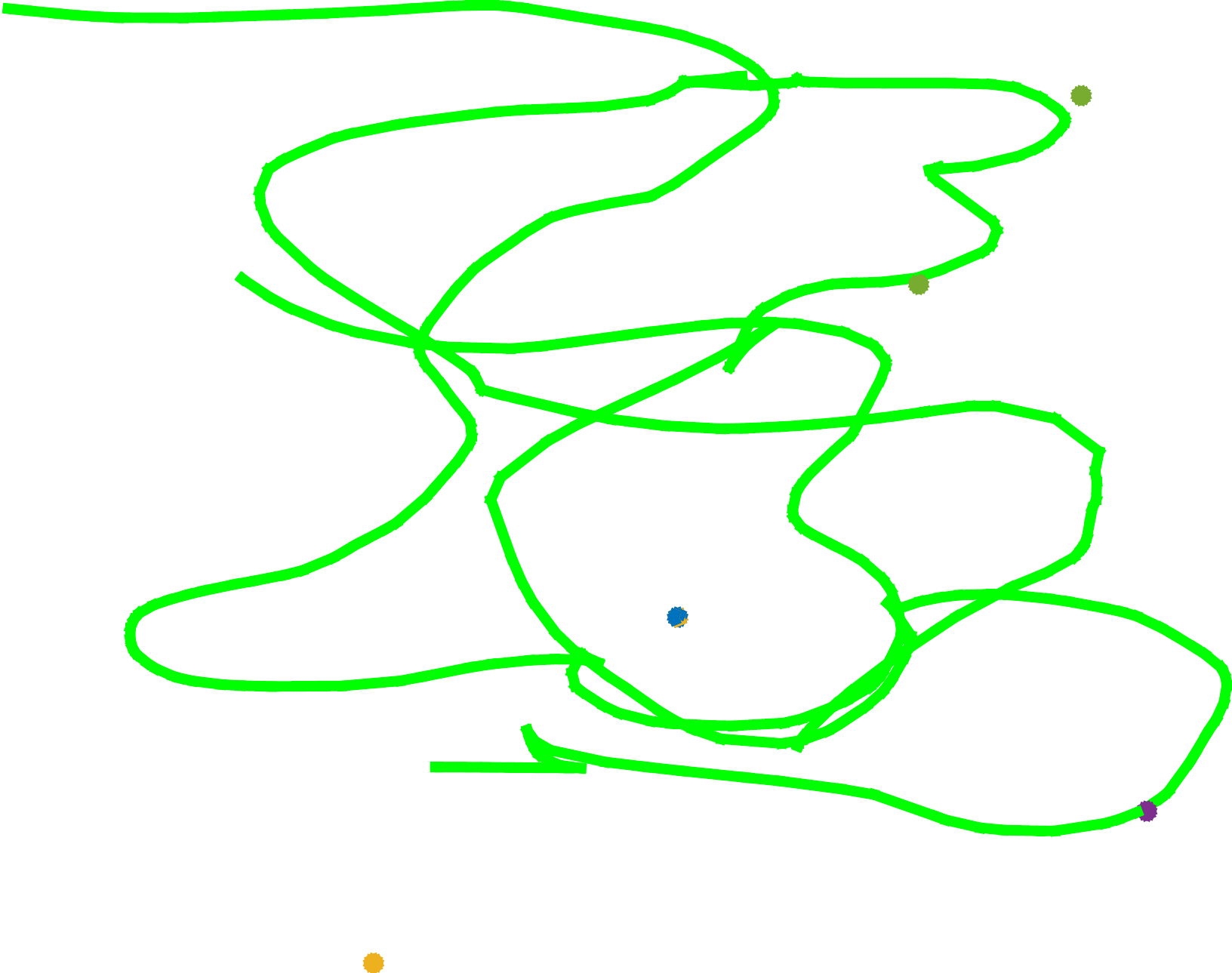} 
\end{minipage}
&
\begin{minipage}{0.30\columnwidth}%
\centering%
\includegraphics[width=\columnwidth, trim= 0cm 0cm 0cm 0cm, clip]{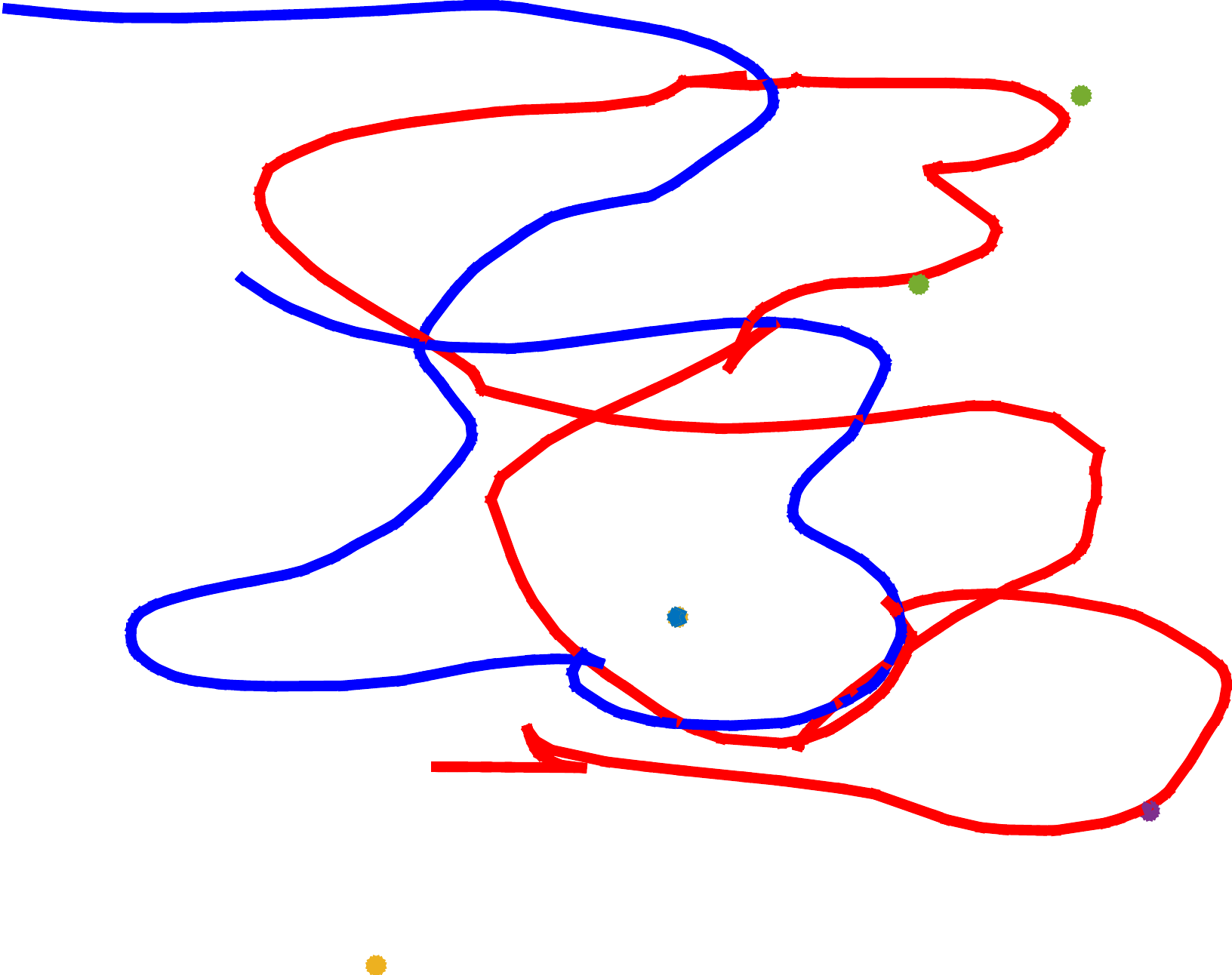} 
\end{minipage}
&
\begin{minipage}{0.30\columnwidth}%
\centering%
\includegraphics[width=\columnwidth, trim=0cm 0cm 0cm 0cm,clip]{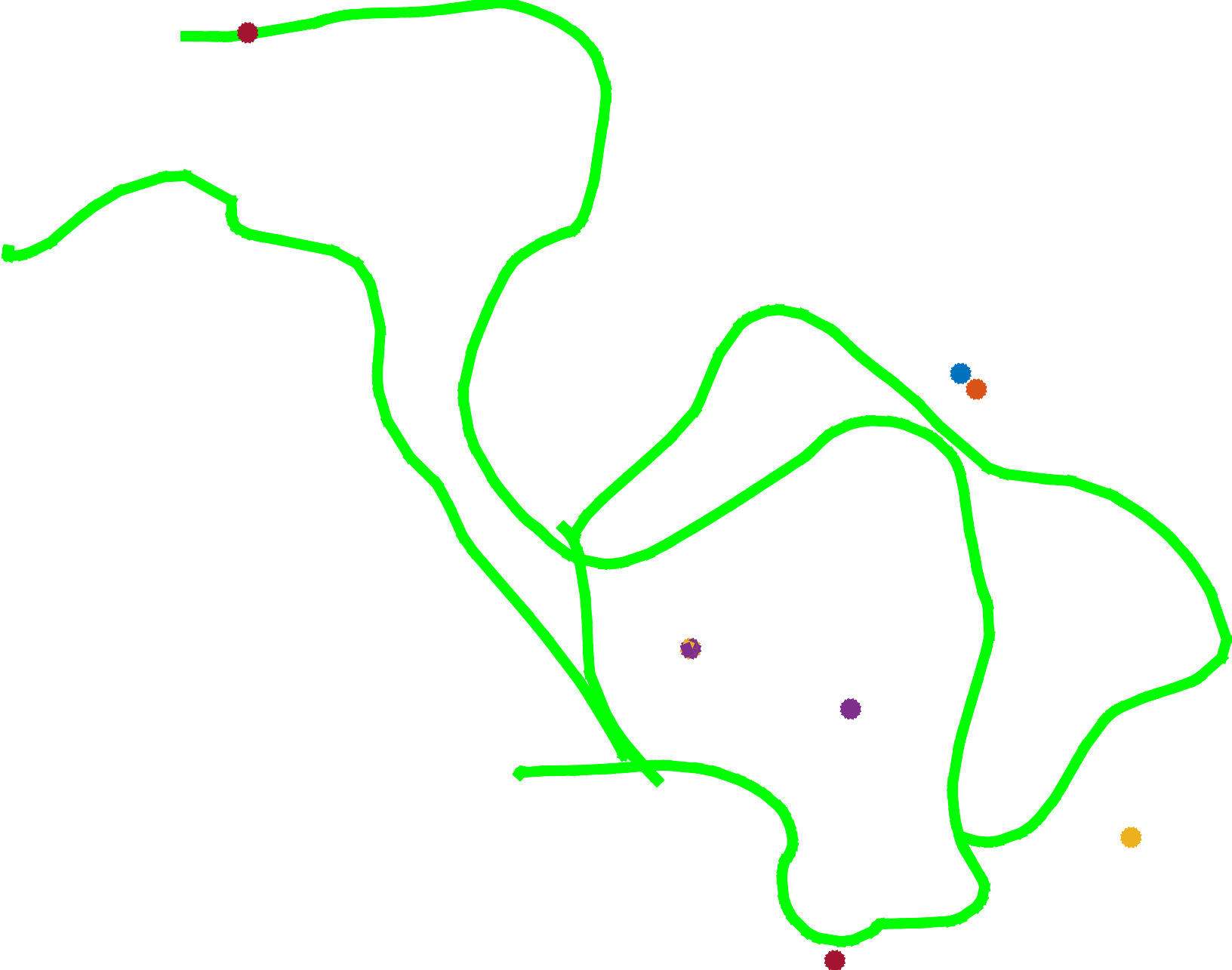} 
\end{minipage}
&
\begin{minipage}{0.30\columnwidth}%
\centering%
\includegraphics[width=\columnwidth, trim= 0cm 0cm 0cm 0cm, clip]{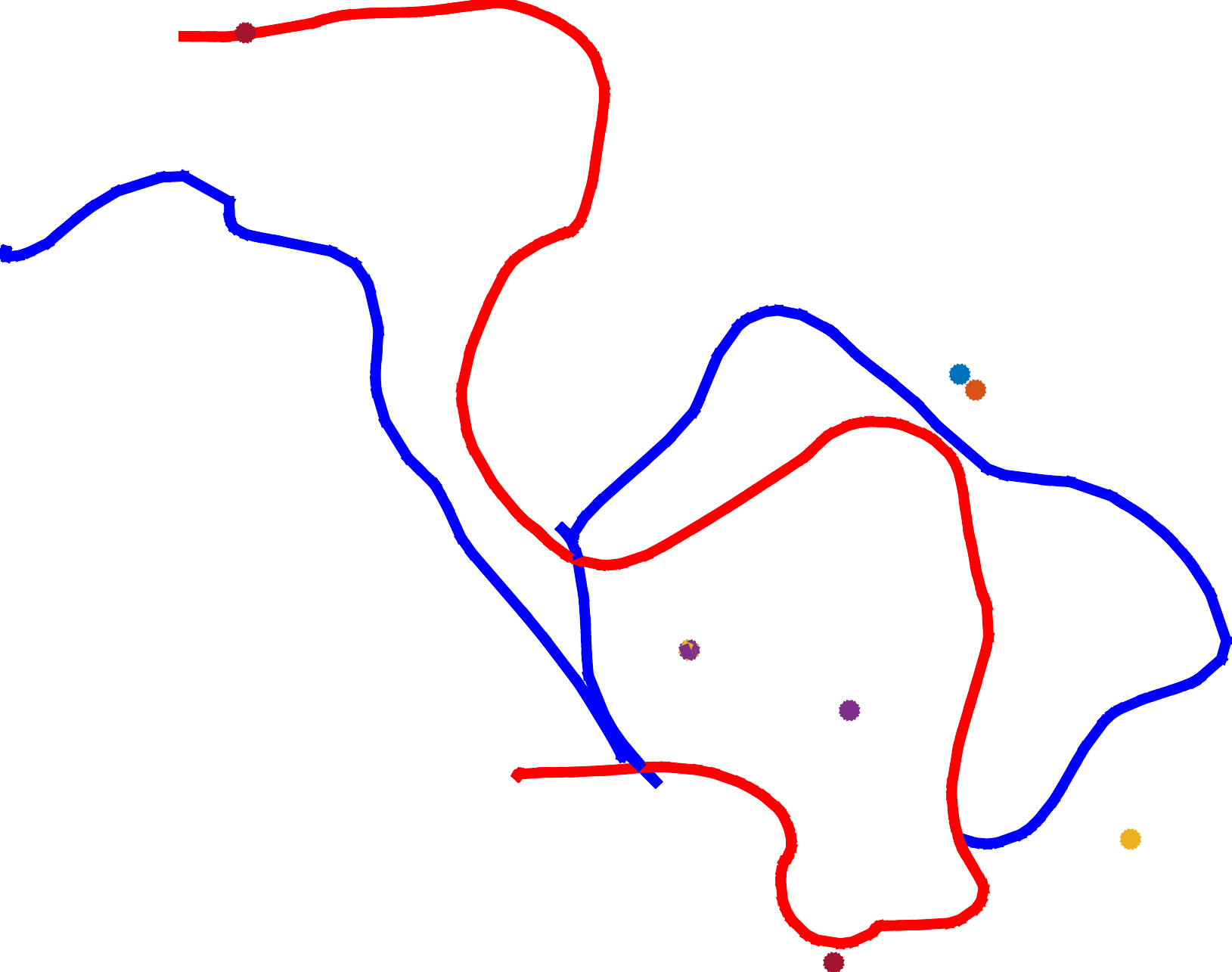} \end{minipage}
&
\begin{minipage}{0.30\columnwidth}%
\centering%
\includegraphics[width=\columnwidth, trim=0cm 0cm 0cm 0cm,clip]{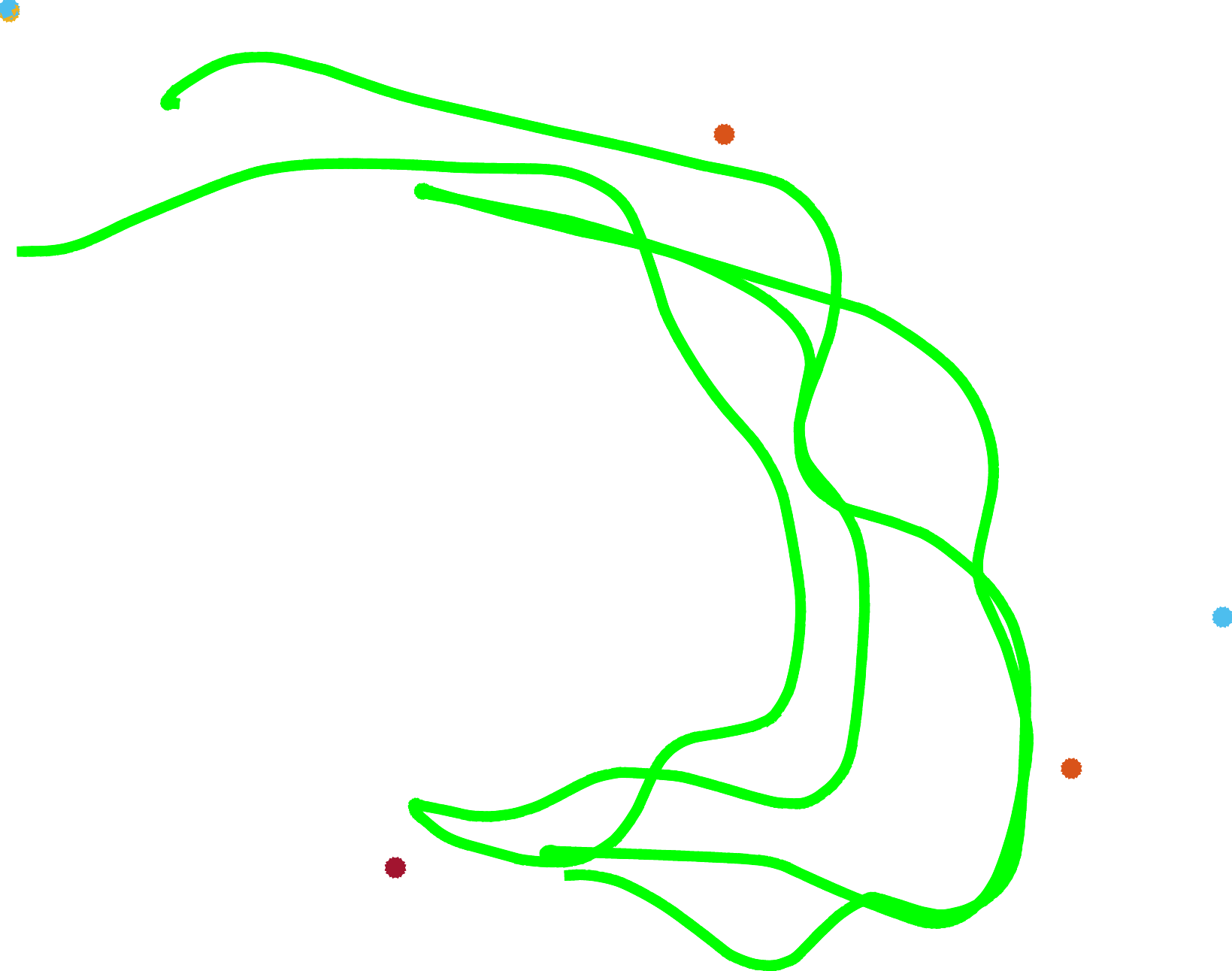} 
\end{minipage}
&
\begin{minipage}{0.30\columnwidth}%
\centering%
\includegraphics[width=\columnwidth, trim= 0cm 0cm 0cm 0cm, clip]{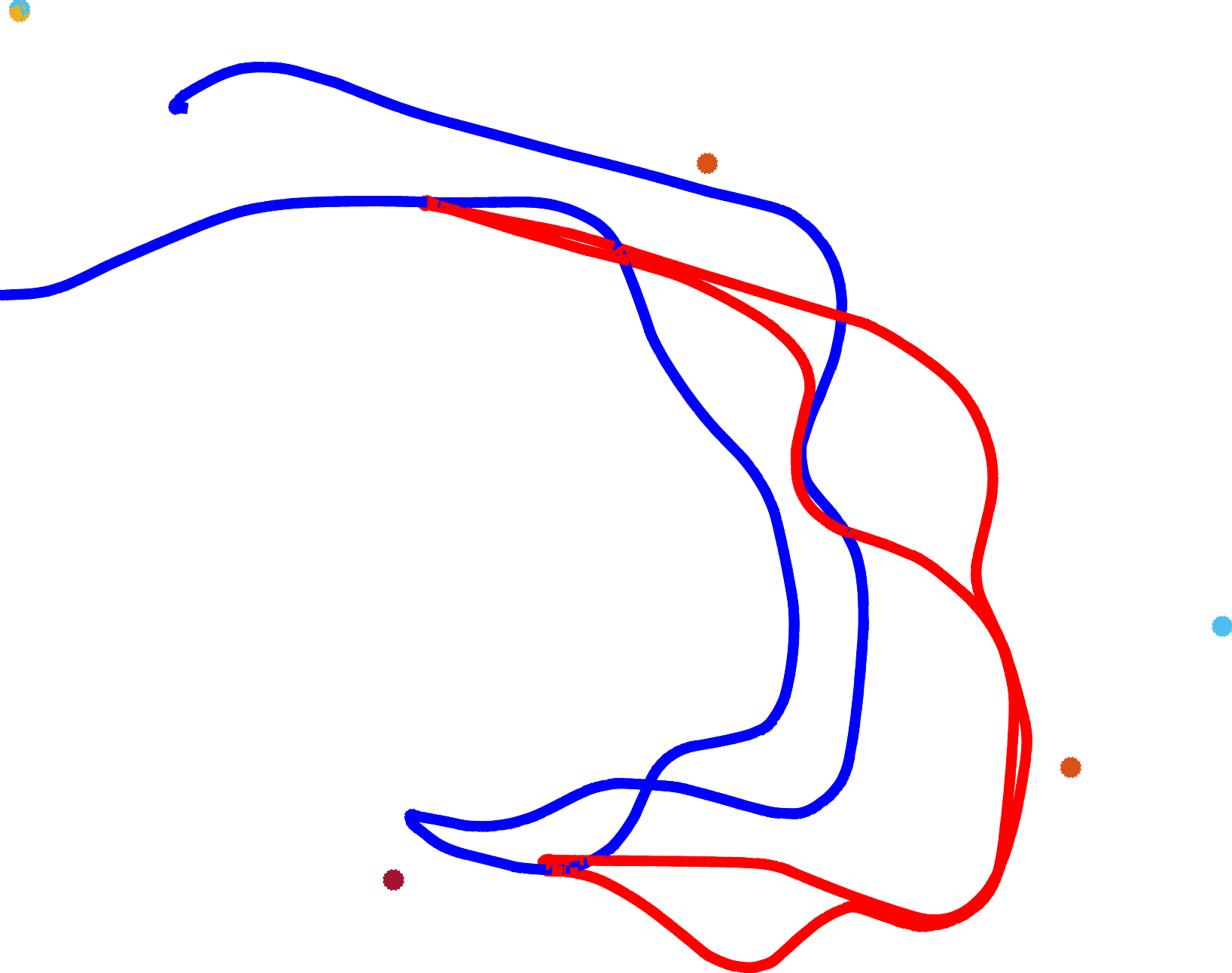} \end{minipage}\\

\multicolumn{2}{c}{\stadiuma}&\multicolumn{2}{c}{\stadiumb}&\multicolumn{2}{c}{\housea}\\
\end{tabular}}
  \end{minipage}
  \caption{\label{fig:fieldExperiments} Real tests: estimated trajectories and object poses for our approach and for the centralized \GN method. Trajectories of the
two robots are shown as red and blue lines, while objects are shown as colored dots.
  }
\end{figure*}

\begin{table*}
\centering
{\renewcommand{\arraystretch}{1.5}%
\begin{tabular}{|c|c|c|c|c|c|c|c|c|c|c|}
\hline 
\multirow{3}{*}{Scenario} & Initial &  \multicolumn{4}{c|}{Distributed Gauss-Seidel} & Centralized & \multicolumn{2}{c|}{ATE* (m)} & \multicolumn{2}{c|}{ARE* (deg)}\\
\cline{3-11}
& & \multicolumn{2}{c|}{$\eta_r\!=\!\eta_p\!=\!10^{-1}$} & \multicolumn{2}{c|}{$\eta_r\!=\!\eta_p\!=\!10^{-2}$} & GN & \multirow{2}{*}{Poses} & \multirow{2}{*}{Lmrks.} & \multirow{2}{*}{Poses} & \multirow{2}{*}{Lmrks.} \\
\cline{3-6}
& Cost & \#Iter & Cost & \#Iter & Cost & Cost  & & & &\\
\hline 
\hline 

Stadium-1& 120.73 & 5.0 & 1.1e-09 & 5.0 & 1.1e-09 & 1.6e-10 & 1.9e-10 & 1.9e-10 & 1.4e-03 & 1.2e-04\\
 \hline
Stadium-2 & 310.24 & 5.0 & 4.5e-12 & 8.0 & 4.4e-12 & 3.5e-13 & 2.1e-03 & 2.2e-03 & 1.2e-02 & 1.4e-02\\
 \hline
House & 43.59 & 5.0 & 1.1e-03 & 6.0 & 1.0e-03 & 8.4e-04 & 4.4e-02 & 6.2e-02 & 4.3e-01 & 4.9e-01\\
 \hline
 \end{tabular}}
\caption{\label{tab:fieldExperimentsAccuracy} 
Number of iterations, cost, ATE* and ARE* of our approach as compared to centralized Gauss-Newton method for Field data
}
\end{table*}

\myparagraph{Communication Bandwidth Requirements}
Table \ref{tab:fieldExperimentsMemory} also compares the 
average communication requirements in the case of transmission of dense point clouds and object-based models. 
When using point clouds, 
the robots are required sending at least one RGB-D frame at every rendezvous to estimate their relative pose. So the average communication for dense point cloud map is computed as $n_c K C$ where $n_c$ is the number of rendezvous, $K$ is the number of points per frame and $C$ is the memory required to send each point. Communication in the case of our object-based map requires sending object category and object pose; a upper bound can be computed as $n_oL$ where $n_o$ is the number of objects and $L$ is the memory required to store category label and pose of an object. Table \ref{tab:fieldExperimentsMemory} confirms that our approach provides a remarkable advantage in terms of communication burden as it requires transmitting 6 orders of magnitude less than a point-cloud-based approach. 

\myparagraph{Accuracy}
\Fig\ref{fig:fieldExperiments} shows the trajectories of the two robots in three runs and the object pose estimates. The figure compares our approach and the corresponding centralized \GN estimate. Quantitative results are given in 
Table \ref{tab:fieldExperimentsAccuracy},
which reports the cost attained by  our approach, the number of iterations, ATE*,  ARE* as compared to the centralized
\GN approach. The table confirms that our distributed approach converges in few iterations and is practically as accurate as the centralized \GN method; in particular the mismatch between the \DGS and the \GN estimates is in the order of millimeters for the position estimates and tenth of degrees for the rotation estimates.
Note that for these indoor experiments the wheel odometry is fairly accurate, 
since the robot moves on wooden or tiled floor.
This results in better performance for the 
proposed technique and in small costs in \GN.
The initial cost, instead, is large mostly because of the error in the initial alignment between the two robots. 

\section{Conclusions and Future Work}\label{sec:conclusion}
We investigate distributed algorithms to estimate the 3D trajectories of multiple 
cooperative robots from relative pose measurements. 
Our first contribution is the design of a 2-stage approach for distributed pose 
estimation and propose a number of algorithmic variants.
One of these algorithms, the Distributed Gauss-Seidel (\DGS) method, is shown to have 
excellent performance in practice: (i) its communication burden scale linearly in 
the number of separators and respect agents' privacy, (ii) 
it is robust to noise and 
the resulting estimates are sufficiently accurate after few communication rounds, 
(iii) the approach is simple to implement and scales well to large teams.  
We demonstrated the effectiveness of the \DGS approach in extensive simulations and field tests.

Our second contribution is to extend the \DGS approach to use objects as landmarks for multi robot mapping. 
We show that using object-based abstractions in a multi robot setup further reduces 
the memory requirement and the information exchange among teammates. 
We demonstrate our multi robot object-based mapping approach in Gazebo simulations and in field tests performed in a 
MOUT (\emph{Military Operations in Urban Terrain}) test facility.

We are currently extending the approach proposed in this paper in several directions. 
First, our current approach for object-based mapping assumes 
that a  model of each
observed objects is known in advance. However it can be challenging to store a large number of object models, 
and to account for intra-class variations.
 As a future work, we plan to extend our approach to the case where object models are not previously known 
 (at an instance level) and instead object shapes are jointly optimized within our SLAM framework. 

Second, our current approach is based on a nonlinear least squares formulation which is not robust to gross 
outliers. Therefore future work will focus on designing more general algorithms that are robust to spurious measurements.
 
 Third, we plan to extend our experimental evaluation to flying robots. 
While we demonstrated the effectiveness of our approach in large teams of ground robots, we believe that 
the next grand challenge is to enable coordination and distributed mapping in swarms of agile micro aerial vehicles 
with limited communication and computation resources.

\bibliographystyle{SageH}
\bibliography{references/refs,iser,iros}

\begin{thebibliography}{85}
\providecommand{\natexlab}[1]{#1}
\providecommand{\url}[1]{\texttt{#1}}
\providecommand{\urlprefix}{URL }
\expandafter\ifx\csname urlstyle\endcsname\relax
  \providecommand{\doi}[1]{DOI:\discretionary{}{}{}#1}\else
  \providecommand{\doi}{DOI:\discretionary{}{}{}\begingroup
  \urlstyle{rm}\Url}\fi

\bibitem[{Anderson et~al.(2010)Anderson, Shames, Mao and
  Fidan}]{Anderson10siam}
Anderson B, Shames I, Mao G and Fidan B (2010) Formal theory of noisy sensor
  network localization.
\newblock \emph{SIAM Journal on Discrete Mathematics} 24(2): 684--698.

\bibitem[{Andersson and Nygards(2008)}]{Andersson08icra}
Andersson L and Nygards J (2008) C-{SAM} : Multi-robot {SLAM} using square root
  information smoothing.
\newblock In: \emph{IEEE Intl. Conf. on Robotics and Automation (ICRA)}.

\bibitem[{Aragues et~al.(2011)Aragues, Carlone, Calafiore and
  Sagues}]{Aragues11icra}
Aragues R, Carlone L, Calafiore G and Sagues C (2011) Multi-agent localization
  from noisy relative pose measurements.
\newblock In: \emph{IEEE Intl. Conf. on Robotics and Automation (ICRA)}. pp.
  364--369.

\bibitem[{Aragues et~al.(2012{\natexlab{a}})Aragues, Carlone, Calafiore and
  Sagues}]{Aragues11scl}
Aragues R, Carlone L, Calafiore G and Sagues C (2012{\natexlab{a}}) Distributed
  centroid estimation from noisy relative measurements.
\newblock \emph{Systems \& Control Letters} 61(7): 773--779.

\bibitem[{Aragues et~al.(2012{\natexlab{b}})Aragues, Cortes and
  Sagues}]{Aragues12tro}
Aragues R, Cortes J and Sagues C (2012{\natexlab{b}}) Distributed consensus on
  robot networks for dynamically merging feature-based maps.
\newblock \emph{IEEE Transactions on Robotics} .

\bibitem[{Bahr et~al.(2009)Bahr, Walter and Leonard}]{Bahr09icra}
Bahr A, Walter M and Leonard J (2009) Consistent cooperative localization.
\newblock In: \emph{IEEE Intl. Conf. on Robotics and Automation (ICRA)}. pp.
  3415--3422.

\bibitem[{Bailey et~al.(2011)Bailey, Bryson, Mu, Vial, McCalman and
  Durrant-Whyte}]{Bailey11icra}
Bailey T, Bryson M, Mu H, Vial J, McCalman L and Durrant-Whyte H (2011)
  Decentralised cooperative localisation for heterogeneous teams of mobile
  robots.
\newblock In: \emph{IEEE Intl. Conf. on Robotics and Automation (ICRA)}.

\bibitem[{Bao et~al.(2012)Bao, Bagra, Chao and Savarese}]{Bao12cvpr}
Bao SYZ, Bagra M, Chao YW and Savarese S (2012) Semantic structure from motion
  with points, regions, and objects.
\newblock In: \emph{CVPR}. pp. 2703--2710.

\bibitem[{Barooah and Hespanha(2005)}]{Barooah05icisip}
Barooah P and Hespanha J (2005) Semantic structure from motion.
\newblock In: \emph{Intl. Conf. on Intelligent Sensing and Information
  Processing}. pp. 226--231.

\bibitem[{Barooah and Hespanha(2007)}]{Barooah07csm}
Barooah P and Hespanha J (2007) Estimation on graphs from relative
  measurements.
\newblock \emph{Control System Magazine} 27(4): 57--74.

\bibitem[{Bertsekas and Tsitsiklis(1989)}]{Bertsekas89book}
Bertsekas D and Tsitsiklis J (1989) \emph{Parallel and Distributed Computation:
  Numerical Methods}.
\newblock Englewood Cliffs, NJ: Prentice-Hall.

\bibitem[{Calafiore et~al.(2012)Calafiore, Carlone and Wei}]{Calafiore12tsmca}
Calafiore G, Carlone L and Wei M (2012) A distributed technique for
  localization of agent formations from relative range measurements.
\newblock \emph{IEEE Trans. on Systems, Man, and Cybernetics, Part A} 42(5):
  1083--4427.

\bibitem[{Carlone et~al.(2011)Carlone, Ng, Du, Bona and Indri}]{Carlone11jirs}
Carlone L, Ng MK, Du J, Bona B and Indri M (2011) Simultaneous localization and
  mapping using {Rao-Blackwellized} particle filters in multi robot systems.
\newblock \emph{J. of Intelligent and Robotic Systems} 63(2): 283--307.

\bibitem[{Carlone et~al.(2015{\natexlab{a}})Carlone, Rosen, Calafiore, Leonard
  and Dellaert}]{Carlone15iros-duality3D}
Carlone L, Rosen D, Calafiore G, Leonard J and Dellaert F (2015{\natexlab{a}})
  Lagrangian duality in {3D SLAM}: Verification techniques and optimal
  solutions.
\newblock In: \emph{IEEE/RSJ Intl. Conf. on Intelligent Robots and Systems
  (IROS)}.

\bibitem[{Carlone et~al.(2015{\natexlab{b}})Carlone, Tron, Daniilidis and
  Dellaert}]{Carlone15icra-init3D}
Carlone L, Tron R, Daniilidis K and Dellaert F (2015{\natexlab{b}})
  Initialization techniques for {3D SLAM}: a survey on rotation estimation and
  its use in pose graph optimization.
\newblock In: \emph{IEEE Intl. Conf. on Robotics and Automation (ICRA)}. pp.
  4597--4604.

\bibitem[{Carron et~al.(2014)Carron, Todescato, Carli and
  Schenato}]{Carron14tcns}
Carron A, Todescato M, Carli R and Schenato L (2014) An asynchronous
  consensus-based algorithm for estimation from noisy relative measurements.
\newblock \emph{IEEE Transactions on Control of Network Systems} 1(3):
  2325--5870.

\bibitem[{Choudhary et~al.(2014)Choudhary, Trevor, Christensen and
  Dellaert}]{Choudhary14iros}
Choudhary S, Trevor AJB, Christensen HI and Dellaert F (2014) {SLAM} with
  object discovery, modeling and mapping.
\newblock In: \emph{2014 {IEEE/RSJ} International Conference on Intelligent
  Robots and Systems, Chicago, IL, USA, September 14-18, 2014}. pp. 1018--1025.
\newblock \doi{10.1109/IROS.2014.6942683}.
\newblock \urlprefix\url{http://dx.doi.org/10.1109/IROS.2014.6942683}.

\bibitem[{Civera et~al.(2011)Civera, G{\'{a}}lvez{-}L{\'{o}}pez, Riazuelo,
  Tard{\'{o}}s and Montiel}]{Civera11iros}
Civera J, G{\'{a}}lvez{-}L{\'{o}}pez D, Riazuelo L, Tard{\'{o}}s JD and Montiel
  JMM (2011) Towards semantic {SLAM} using a monocular camera.
\newblock In: \emph{2011 {IEEE/RSJ} International Conference on Intelligent
  Robots and Systems, {IROS} 2011, San Francisco, CA, USA, September 25-30,
  2011}. pp. 1277--1284.
\newblock \doi{10.1109/IROS.2011.6094648}.
\newblock \urlprefix\url{http://dx.doi.org/10.1109/IROS.2011.6094648}.

\bibitem[{Cunningham et~al.(2013)Cunningham, Indelman and
  Dellaert}]{Cunningham13icra}
Cunningham A, Indelman V and Dellaert F (2013) {DDF-SAM} 2.0: Consistent
  distributed smoothing and mapping.
\newblock In: \emph{IEEE Intl. Conf. on Robotics and Automation (ICRA)}.
  Karlsruhe, Germany.

\bibitem[{Cunningham et~al.(2010)Cunningham, Paluri and
  Dellaert}]{Cunningham10iros}
Cunningham A, Paluri M and Dellaert F (2010) {DDF-SAM}: Fully distributed slam
  using constrained factor graphs.
\newblock In: \emph{IEEE/RSJ Intl. Conf. on Intelligent Robots and Systems
  (IROS)}.

\bibitem[{Davison et~al.(2007)Davison, Reid, Molton and
  Stasse}]{Davison07tpami}
Davison AJ, Reid ID, Molton ND and Stasse O (2007) Monoslam: Real-time single
  camera slam.
\newblock \emph{IEEE Transactions on Pattern Analysis and Machine Intelligence}
  29(6): 1052--1067.
\newblock \doi{http://doi.ieeecomputersociety.org/10.1109/TPAMI.2007.1049}.

\bibitem[{Dellaert(2005)}]{Dellaert05rss}
Dellaert F (2005) Square {Root} {SAM}: Simultaneous location and mapping via
  square root information smoothing.
\newblock In: \emph{Robotics: Science and Systems (RSS)}.

\bibitem[{Dellaert(2012)}]{Dellaert12tr}
Dellaert F (2012) Factor graphs and {GTSAM}: A hands-on introduction.
\newblock Technical Report GT-RIM-CP\&R-2012-002, Georgia Institute of
  Technology.

\bibitem[{Dong et~al.(2015)Dong, Nelson, Indelman, Michael and
  Dellaert}]{Dong15icra}
Dong J, Nelson E, Indelman V, Michael N and Dellaert F (2015) Distributed
  real-time cooperative localization and mapping using an uncertainty-aware
  expectation maximization approach.
\newblock In: \emph{IEEE Intl. Conf. on Robotics and Automation (ICRA)}.

\bibitem[{Estrada et~al.(2005)Estrada, Neira and Tardos}]{Estrada05tro}
Estrada C, Neira J and Tardos J (2005) Hierarchical {SLAM}: Real-time accurate
  mapping of large environments.
\newblock \emph{{IEEE} Trans. Robotics} 21(4): 588--596.

\bibitem[{Finman et~al.(2013)Finman, Whelan, Kaess and Leonard}]{Finman13ecmr}
Finman R, Whelan T, Kaess M and Leonard JJ (2013) Toward lifelong object
  segmentation from change detection in dense {RGB-D} maps.
\newblock In: \emph{2013 European Conference on Mobile Robots, Barcelona,
  Catalonia, Spain, September 25-27, 2013}. pp. 178--185.
\newblock \doi{10.1109/ECMR.2013.6698839}.
\newblock \urlprefix\url{http://dx.doi.org/10.1109/ECMR.2013.6698839}.

\bibitem[{Franceschelli and Gasparri(2010)}]{Franceschelli10icra}
Franceschelli M and Gasparri A (2010) On agreement problems with {Gossip}
  algorithms in absence of common reference frames.
\newblock In: \emph{IEEE Intl. Conf. on Robotics and Automation (ICRA)}, volume
  337. pp. 4481--4486.

\bibitem[{Freris and Zouzias(2015)}]{Freris15cdc}
Freris N and Zouzias A (2015) Fast distributed smoothing of relative
  measurements.
\newblock In: \emph{IEEE Conf. on Decision and Control}. pp. 1411--1416.

\bibitem[{Frese(2006)}]{Frese06ar2}
Frese U (2006) Treemap: An {$O(\log n)$} algorithm for indoor simultaneous
  localization and mapping.
\newblock \emph{Autonomous Robots} 21(2): 103--122.

\bibitem[{Frese et~al.(2005)Frese, Larsson and Duckett}]{Frese05tro}
Frese U, Larsson P and Duckett T (2005) A multilevel relaxation algorithm for
  simultaneous localisation and mapping.
\newblock \emph{{IEEE} Trans. Robotics} 21(2): 196--207.

\bibitem[{G{\'{a}}lvez{-}L{\'{o}}pez et~al.(2016)G{\'{a}}lvez{-}L{\'{o}}pez,
  Salas, Tard{\'{o}}s and Montiel}]{Lopez16ras}
G{\'{a}}lvez{-}L{\'{o}}pez D, Salas M, Tard{\'{o}}s JD and Montiel JMM (2016)
  Real-time monocular object {SLAM}.
\newblock \emph{Robotics and Autonomous Systems} 75: 435--449.
\newblock \doi{10.1016/j.robot.2015.08.009}.
\newblock \urlprefix\url{http://dx.doi.org/10.1016/j.robot.2015.08.009}.

\bibitem[{Grisetti et~al.(2010)Grisetti, Kuemmerle, Stachniss, Frese and
  Hertzberg}]{Grisetti10icra}
Grisetti G, Kuemmerle R, Stachniss C, Frese U and Hertzberg C (2010)
  Hierarchical optimization on manifolds for online 2{D} and 3{D} mapping.
\newblock In: \emph{IEEE Intl. Conf. on Robotics and Automation (ICRA)}.
  Anchorage, Alaska.

\bibitem[{Grisetti et~al.(2012)Grisetti, K{\"u}mmerle and Ni}]{Grisetti12iros}
Grisetti G, K{\"u}mmerle R and Ni K (2012) Robust optimization of factor graphs
  by using condensed measurements.
\newblock In: \emph{IEEE/RSJ Intl. Conf. on Intelligent Robots and Systems
  (IROS)}.

\bibitem[{Hatanaka et~al.(2010)Hatanaka, Fujita and Bullo}]{Hatanaka10cdc}
Hatanaka T, Fujita M and Bullo F (2010) Vision-based cooperative estimation via
  multi-agent optimization.
\newblock In: \emph{IEEE Conf. on Decision and Control}.

\bibitem[{Howard(2006)}]{Howard06ijrr}
Howard A (2006) Multi-robot simultaneous localization and mapping using
  particle filters.
\newblock \emph{Intl. J. of Robotics Research} 25(12): 1243--1256.
\newblock
  \urlprefix\url{http://cres.usc.edu/cgi-bin/print_pub_details.pl?pubid=514}.

\bibitem[{Indelman et~al.(2012)Indelman, Gurfil, Rivlin and
  Rotstein}]{Indelman12ijrr}
Indelman V, Gurfil P, Rivlin E and Rotstein H (2012) Graph-based distributed
  cooperative navigation for a general multi-robot measurement model.
\newblock \emph{Intl. J. of Robotics Research} 31(9).

\bibitem[{Indelman et~al.(2014)Indelman, Nelson, Michael and
  Dellaert}]{Indelman14icra}
Indelman V, Nelson E, Michael N and Dellaert F (2014) Multi-robot pose graph
  localization and data association from unknown initial relative poses via
  expectation maximization.
\newblock In: \emph{IEEE Intl. Conf. on Robotics and Automation (ICRA)}.

\bibitem[{Kim et~al.(2010)Kim, Kaess, Fletcher, Leonard, Bachrach, Roy and
  Teller}]{Kim10icra}
Kim B, Kaess M, Fletcher L, Leonard J, Bachrach A, Roy N and Teller S (2010)
  Multiple relative pose graphs for robust cooperative mapping.
\newblock In: \emph{IEEE Intl. Conf. on Robotics and Automation (ICRA)}.
  Anchorage, Alaska, pp. 3185--3192.

\bibitem[{Kim et~al.(2012)Kim, Mitra, Yan and Guibas}]{Kim12siggraphAsia}
Kim YM, Mitra NJ, Yan DM and Guibas LJ (2012) Acquiring 3d indoor environments
  with variability and repetition.
\newblock \emph{ACM Trans. Graph.} 31(6): 138.

\bibitem[{Knuth and Barooah(2013)}]{Knuth13icra}
Knuth J and Barooah P (2013) Collaborative localization with heterogeneous
  inter-robot measurements by {R}iemannian optimization.
\newblock In: \emph{IEEE Intl. Conf. on Robotics and Automation (ICRA)}. pp.
  1534--1539.

\bibitem[{Koppula et~al.(2011)Koppula, Anand, Joachims and
  Saxena}]{Koppula11nips}
Koppula H, Anand A, Joachims T and Saxena A (2011) Semantic labeling of 3d
  point clouds for indoor scenes.
\newblock In: \emph{Advances in Neural Information Processing Systems (NIPS)}.

\bibitem[{Kuipers(2000)}]{Kuipers2000ai}
Kuipers B (2000) The spatial semantic hierarchy.
\newblock \emph{Artificial Intelligence} 119: 191 -- 233.
\newblock \doi{http://dx.doi.org/10.1016/S0004-3702(00)00017-5}.
\newblock
  \urlprefix\url{http://www.sciencedirect.com/science/article/pii/S0004370200000175}.

\bibitem[{Kundu et~al.(2014)Kundu, Li, Dellaert, Li and Rehg}]{Kundu14eccv}
Kundu A, Li Y, Dellaert F, Li F and Rehg JM (2014) \emph{Joint Semantic
  Segmentation and 3D Reconstruction from Monocular Video}.
\newblock Cham: Springer International Publishing.
\newblock ISBN 978-3-319-10599-4, pp. 703--718.
\newblock \doi{10.1007/978-3-319-10599-4_45}.
\newblock \urlprefix\url{http://dx.doi.org/10.1007/978-3-319-10599-4_45}.

\bibitem[{Lazaro et~al.(2011)Lazaro, Paz, Pinies, Castellanos and
  Grisetti}]{Lazaro11icra}
Lazaro M, Paz L, Pinies P, Castellanos J and Grisetti G (2011) Multi-robot
  {SLAM} using condensed measurements.
\newblock In: \emph{IEEE Intl. Conf. on Robotics and Automation (ICRA)}. pp.
  1069--1076.

\bibitem[{Leonard and Feder(2001)}]{Leonard01joe}
Leonard J and Feder H (2001) Decoupled stochastic mapping.
\newblock \emph{IEEE Journal of Oceanic Engineering} : 561--571.

\bibitem[{Leonard and Newman(2003)}]{Leonard03ijcai}
Leonard J and Newman P (2003) Consistent, convergent, and constant-time {SLAM}.
\newblock In: \emph{Intl. Joint Conf. on AI (IJCAI)}.

\bibitem[{Martinec and Pajdla(2007)}]{Martinec07cvpr}
Martinec D and Pajdla T (2007) Robust rotation and translation estimation in
  multiview reconstruction.
\newblock In: \emph{IEEE Conf. on Computer Vision and Pattern Recognition
  (CVPR)}. pp. 1--8.

\bibitem[{McCormac et~al.(2016)McCormac, Handa, Davison and
  Leutenegger}]{McCormac16arxiv}
McCormac J, Handa A, Davison AJ and Leutenegger S (2016) Semanticfusion: Dense
  3d semantic mapping with convolutional neural networks.
\newblock \emph{CoRR} abs/1609.05130.
\newblock \urlprefix\url{http://arxiv.org/abs/1609.05130}.

\bibitem[{Nerurkar et~al.(2009)Nerurkar, Roumeliotis and
  Martinelli}]{Nerurkar09icra}
Nerurkar E, Roumeliotis S and Martinelli A (2009) Distributed maximum a
  posteriori estimation for multi-robot cooperative localization.
\newblock In: \emph{IEEE Intl. Conf. on Robotics and Automation (ICRA)}. pp.
  1402--1409.

\bibitem[{Ni and Dellaert(2010)}]{Ni10iros}
Ni K and Dellaert F (2010) Multi-level submap based slam using nested
  dissection.
\newblock In: \emph{IEEE/RSJ Intl. Conf. on Intelligent Robots and Systems
  (IROS)}.
\newblock \urlprefix\url{http://frank.dellaert.com/pubs/Ni10iros.pdf}.

\bibitem[{Ni et~al.(2007)Ni, Steedly and Dellaert}]{Ni07icra}
Ni K, Steedly D and Dellaert F (2007) Tectonic {SAM}: Exact; out-of-core;
  submap-based {SLAM}.
\newblock In: \emph{IEEE Intl. Conf. on Robotics and Automation (ICRA)}. Rome;
  Italy.
\newblock \urlprefix\url{http://www.cc.gatech.edu/~dellaert/pubs/Ni07icra.pdf}.

\bibitem[{N\"{u}chter and Hertzberg(2008)}]{Nuchter08ras}
N\"{u}chter A and Hertzberg J (2008) Towards semantic maps for mobile robots.
\newblock \emph{Robot. Auton. Syst.} 56(11): 915--926.
\newblock \doi{10.1016/j.robot.2008.08.001}.
\newblock \urlprefix\url{http://dx.doi.org/10.1016/j.robot.2008.08.001}.

\bibitem[{Olfati-Saber(2006)}]{Olfati-Saber06cdc}
Olfati-Saber R (2006) Swarms on sphere: A programmable swarm with synchronous
  behaviors like oscillator networks.
\newblock In: \emph{IEEE Conf. on Decision and Control}. pp. 5060--5066.

\bibitem[{Paull et~al.(2015)Paull, Huang, Seto and Leonard}]{Paull15icra}
Paull L, Huang G, Seto M and Leonard J (2015) Communication-constrained
  multi-{AUV} cooperative {SLAM}.
\newblock In: \emph{IEEE Intl. Conf. on Robotics and Automation (ICRA)}.

\bibitem[{Pillai and Leonard(2015)}]{Pillai15rss}
Pillai S and Leonard J (2015) Monocular slam supported object recognition.
\newblock In: \emph{Proceedings of Robotics: Science and Systems (RSS)}. Rome,
  Italy.

\bibitem[{Piovan et~al.(2013)Piovan, Shames, Fidan, Bullo and
  Anderson}]{Piovan13automatica}
Piovan G, Shames I, Fidan B, Bullo F and Anderson B (2013) On frame and
  orientation localization for relative sensing networks.
\newblock \emph{Automatica} 49(1): 206--213.

\bibitem[{Pronobis and Jensfelt(2012)}]{Pronobis12icra}
Pronobis A and Jensfelt P (2012) Large-scale semantic mapping and reasoning
  with heterogeneous modalities.
\newblock In: \emph{IEEE International Conference on Robotics and Automation
  (ICRA)}.
\newblock \doi{10.1109/ICRA.2012.6224637}.

\bibitem[{Ranganathan and Dellaert(2007)}]{Ranganathan07rss}
Ranganathan A and Dellaert F (2007) {Semantic Modeling of Places using
  Objects}.
\newblock In: \emph{Robotics: Science and Systems (RSS)}. Atlanta; USA.

\bibitem[{Redmon et~al.(2015)Redmon, Divvala, Girshick and
  Farhadi}]{Redmon15arxiv}
Redmon J, Divvala SK, Girshick RB and Farhadi A (2015) You only look once:
  Unified, real-time object detection.
\newblock \emph{CoRR} abs/1506.02640.
\newblock \urlprefix\url{http://arxiv.org/abs/1506.02640}.

\bibitem[{Rogers et~al.(2011)Rogers, Trevor, Nieto-Granda and
  Christensen}]{Rogers11iros}
Rogers JG, Trevor AJB, Nieto-Granda C and Christensen HI (2011) Simultaneous
  localization and mapping with learned object recognition and semantic data
  association.
\newblock In: \emph{2011 IEEE/RSJ International Conference on Intelligent
  Robots and Systems}. pp. 1264--1270.
\newblock \doi{10.1109/IROS.2011.6095152}.

\bibitem[{Rosen et~al.(2016)Rosen, Carlone, Bandeira and Leonard}]{Rosen16wafr}
Rosen D, Carlone L, Bandeira A and Leonard J (2016) {SE-Sync}: A certifiably
  correct algorithm for synchronization over the special euclidean group.

\bibitem[{Roumeliotis and Bekey(2002)}]{Roumeliotis02tra}
Roumeliotis S and Bekey G (2002) Distributed multi-robot localization.
\newblock \emph{{IEEE} Trans. Robot. Automat.} .

\bibitem[{Russell et~al.(2011)Russell, Klein and Hespanha}]{Russell11tsp}
Russell W, Klein D and Hespanha J (2011) Optimal estimation on the graph cycle
  space.
\newblock \emph{{IEEE} Trans. Signal Processing} 59(6): 2834--2846.

\bibitem[{Rusu(2009)}]{RusuDoctoralDissertation}
Rusu RB (2009) \emph{{Semantic 3D Object Maps for Everyday Manipulation in
  Human Living Environments}}.
\newblock PhD Thesis, Technische Universit\"at M\"unchen.

\bibitem[{Rusu et~al.(2008)Rusu, Marton, Blodow, Dolha and Beetz}]{Rusu08iros}
Rusu RB, Marton ZC, Blodow N, Dolha ME and Beetz M (2008) Functional object
  mapping of kitchen environments.
\newblock In: \emph{IEEE/RSJ Intl. Conf. on Intelligent Robots and Systems
  (IROS)}.

\bibitem[{Salas-Moreno et~al.(2013)Salas-Moreno, Newcombe, Strasdat, Kelly and
  Davison}]{Moreno13cvpr}
Salas-Moreno RF, Newcombe RA, Strasdat H, Kelly PH and Davison AJ (2013)
  {SLAM++}: Simultaneous localisation and mapping at the level of objects.
\newblock In: \emph{The IEEE Conference on Computer Vision and Pattern
  Recognition (CVPR)}.

\bibitem[{Sarlette and Sepulchre(2009)}]{Sarlette09sicon}
Sarlette A and Sepulchre R (2009) Consensus optimization on manifolds.
\newblock \emph{SIAM J. Control and Optimization} 48(1): 56--76.

\bibitem[{Segal et~al.(2009)Segal, Haehnel and Thrun}]{Segal09rss}
Segal A, Haehnel D and Thrun S (2009) Generalized-icp.
\newblock In: \emph{Proceedings of Robotics: Science and Systems}. Seattle,
  USA.
\newblock \doi{10.15607/RSS.2009.V.021}.

\bibitem[{Simonetto and Leus(2014)}]{Simonetto14tsp}
Simonetto A and Leus G (2014) Distributed maximum likelihood sensor network
  localization.
\newblock \emph{{IEEE} Trans. Signal Processing} 52(6).

\bibitem[{Singh et~al.(2014)Singh, Sha, Narayan, Achim and
  Abbeel}]{Singh14icra}
Singh A, Sha J, Narayan KS, Achim T and Abbeel P (2014) Bigbird: A large-scale
  3d database of object instances.
\newblock In: \emph{2014 IEEE International Conference on Robotics and
  Automation (ICRA)}. pp. 509--516.
\newblock \doi{10.1109/ICRA.2014.6906903}.

\bibitem[{Sturm et~al.(2012)Sturm, Engelhard, Endres, Burgard and
  Cremers}]{Sturm12iros}
Sturm J, Engelhard N, Endres F, Burgard W and Cremers D (2012) A benchmark for
  the evaluation of rgb-d slam systems.
\newblock In: \emph{Proc. of the International Conference on Intelligent Robot
  Systems (IROS)}.

\bibitem[{Suger et~al.(2014)Suger, Tipaldi, Spinello and Burgard}]{Suger14icra}
Suger B, Tipaldi G, Spinello L and Burgard W (2014) {An Approach to Solving
  Large-Scale SLAM Problems with a Small Memory Footprint}.
\newblock In: \emph{IEEE Intl. Conf. on Robotics and Automation (ICRA)}.

\bibitem[{Thrun and Liu(2003)}]{Thrun03isrr}
Thrun S and Liu Y (2003) Multi-robot {SLAM} with sparse extended information
  filters.
\newblock In: \emph{Proceedings of the 11th International Symposium of Robotics
  Research (ISRR'03)}. Sienna, Italy: Springer.

\bibitem[{Thunberg et~al.(2011)Thunberg, Montijano and Hu}]{Thunberg11cdc}
Thunberg J, Montijano E and Hu X (2011) Distributed attitude synchronization
  control.
\newblock In: \emph{IEEE Conf. on Decision and Control}.

\bibitem[{Todescato et~al.(2015)Todescato, Carron, Carli and
  Schenato}]{Todescato15ecc}
Todescato M, Carron A, Carli R and Schenato L (2015) Distributed localization
  from relative noisy measurements: a robust gradient based approach.
\newblock In: \emph{European Control Conference}.

\bibitem[{Trevor et~al.(2013)Trevor, Gedikli, Rusu and
  Christensen}]{Trevor13spme}
Trevor AJB, Gedikli S, Rusu RB and Christensen HI (2013) Efficient organized
  point cloud segmentation with connected components.
\newblock In: \emph{Semantic Perception Mapping and Exploration (SPME)}.

\bibitem[{Trevor et~al.(2012)Trevor, {Rogers III} and
  Christensen}]{Trevor12icra}
Trevor AJB, {Rogers III} JG and Christensen HI (2012) Planar surface slam with
  3d and 2d sensors.
\newblock In: \emph{IEEE Intl. Conf. on Robotics and Automation (ICRA)}. St.
  Paul, MN: IEEE.

\bibitem[{Tron et~al.(2012{\natexlab{a}})Tron, Afsari and Vidal}]{Tron12cdc}
Tron R, Afsari B and Vidal R (2012{\natexlab{a}}) Intrinsic consensus on
  {SO(3)} with almost global convergence.
\newblock In: \emph{IEEE Conf. on Decision and Control}.

\bibitem[{Tron et~al.(2012{\natexlab{b}})Tron, Afsari and Vidal}]{Tron12tac}
Tron R, Afsari B and Vidal R (2012{\natexlab{b}}) Riemannian consensus for
  manifolds with bounded curvature.
\newblock \emph{IEEE Trans. on Automatic Control} .

\bibitem[{Tron and Vidal(2009)}]{Tron09cdc}
Tron R and Vidal R (2009) Distributed image-based {3-D} localization in camera
  networks.
\newblock In: \emph{IEEE Conf. on Decision and Control}.

\bibitem[{Valentin et~al.(2015)Valentin, Vineet, Cheng, Kim, Shotton, Kohli,
  Niessner, Criminisi, Izadi and Torr}]{Valentin15tog}
Valentin J, Vineet V, Cheng MM, Kim D, Shotton J, Kohli P, Niessner M,
  Criminisi A, Izadi S and Torr P (2015) Semanticpaint: Interactive 3d labeling
  and learning at your fingertips.
\newblock \emph{ACM Trans. Graph.} 34(5): 154:1--154:17.
\newblock \urlprefix\url{http://doi.acm.org/10.1145/2751556}.

\bibitem[{Vineet et~al.(2015)Vineet, Miksik, Lidegaard, Nießner, Golodetz,
  Prisacariu, Kähler, Murray, Izadi, Pérez and Torr}]{Vineet15icra}
Vineet V, Miksik O, Lidegaard M, Nießner M, Golodetz S, Prisacariu VA, Kähler
  O, Murray DW, Izadi S, Pérez P and Torr PHS (2015) Incremental dense
  semantic stereo fusion for large-scale semantic scene reconstruction.
\newblock In: \emph{2015 IEEE International Conference on Robotics and
  Automation (ICRA)}. pp. 75--82.
\newblock \doi{10.1109/ICRA.2015.7138983}.

\bibitem[{Wei et~al.(2015)Wei, Aragues, Sagues and Calafiore}]{Wei15sensors}
Wei M, Aragues R, Sagues C and Calafiore G (2015) Noisy range network
  localization based on distributed multidimensional scaling.
\newblock \emph{IEEE Sensors Journals} 15(3): 854--874.

\bibitem[{Zhao et~al.(2013)Zhao, Huang and Dissanayake}]{Zhao13iros}
Zhao L, Huang S and Dissanayake G (2013) Linear {SLAM}: A linear solution to
  the feature-based and pose graph {SLAM} based on submap joining.
\newblock In: \emph{IEEE/RSJ Intl. Conf. on Intelligent Robots and Systems
  (IROS)}.

\bibitem[{Zhou and Roumeliotis(2006)}]{Zhou06iros}
Zhou X and Roumeliotis S (2006) Multi-robot {SLAM} with unknown initial
  correspondence: The robot rendezvous case.
\newblock In: \emph{IEEE/RSJ Intl. Conf. on Intelligent Robots and Systems
  (IROS)}. IEEE, pp. 1785--1792.

\end{thebibliography}


\end{document}